\documentclass[]{article}

\usepackage[utf8]{inputenc}
\usepackage{amsmath,amsthm,amssymb,amsfonts,dsfont,stmaryrd,mathtools}
\usepackage{graphicx,color,stmaryrd,url,xspace,afterpage,tikz,rotating}
\usepackage{subfig,booktabs,multirow,nicefrac,bbm}
\usepackage{authblk}
\usetikzlibrary{calc}
\newcommand{\twopartdef}[4]
{
  \left\{
    \begin{array}{ll}
      #1 & \mbox{if } #2 \\
      #3 & \mbox{if } #4
    \end{array}
  \right.
}

\begin{document}


\title{A survey of exemplar-based texture synthesis}

\author[1]{Lara Raad\thanks{lara.raad@upf.edu}}
\author[2]{Axel Davy\thanks{axel.davy@cmla.ens-cachan.fr}}
\author[2]{Agnès Desolneux\thanks{desolneux@cmla.ens-cachan.fr}}
\author[2]{Jean-Michel Morel\thanks{morel@cmla.ens-cachan.fr}}

\affil[1]{{\footnotesize Dept. of Information \& Communications Technologies,
          Universitat Pompeu Fabra}}
\affil[2]{{\footnotesize CMLA, \'Ecole normale sup\'erieure Paris-Saclay}}

\maketitle

\begin{abstract}
Exemplar-based texture synthesis is the process of generating, from an input sample, new texture images of arbitrary size and which are perceptually equivalent to the sample. The two main approaches are statistics-based methods and patch re-arrangement methods. In the first class, a texture is characterized by a statistical signature; then, a random sampling conditioned to this signature produces genuinely different texture images. The second class boils down to a clever ``copy-paste'' procedure, which stitches together large regions of the sample. Hybrid methods try to combine ideas from both approaches to avoid their hurdles. The  recent approaches using convolutional neural networks fit to this classification, some being statistical and others performing patch re-arrangement in the feature space. They produce impressive synthesis on various kinds of textures. Nevertheless, we found that most real textures are organized at multiple scales, with global structures revealed at coarse scales and highly varying  details  at finer ones. Thus, when confronted with large natural images of textures the results of state-of-the-art methods degrade rapidly, and the problem of modeling them remains wide open.
\end{abstract}

\let\thefootnote\relax\footnote{The images in this document are lossy compressed. To compare the zoomed-in images, please refer to the uncompressed pdf, which can be found at \url{http://desolneux.perso.math.cnrs.fr/papers/exemplar-based-texture-synthesis-v2-full-res.pdf}}

\section{Introduction}
This paper proposes a review of {\it exemplar-based texture theory},  a topic that occupied David Mumford at the end of the last century \cite{ZhuMumford,zhu98}, and again in his book on pattern theory \cite{mumford-desolneux}. Textures are ubiquitous in our visual environment. In the  past fifty years their definition has occupied psychophysicists, mathematicians and computer scientists who have built increasingly sophisticated models.  The main progress on the elusive topic of defining textures has come from computer graphics with the problem of reproducing other examples of the \textit{same} texture given a sample. There is so far no complete mathematical theory that would, first, give a formal axiomatic of texture, and then prove that some texture synthesis algorithm matches this  definition. Rather, each exemplar-based  texture  method formulates its own definition of texture and sometimes (but rarely) convergence or consistency proofs.  The method to work on texture modeling  still relies on a visual  exploration of synthesized  textures, their defects and successes  being linked  to some improvement or shortcoming of the mathematical model. All the more, texture modeling remains a valid challenge for mathematicians, as textures represent arguably the vaster and most common class of observable functions. They indeed cover a majority of the area of most  digital images.  This  article accounts for the very rapid and impressive recent apparition of  new  texture synthesis methods with striking results. We shall retrace their theoretical roots. By performing objective experiments and not hiding the failures of each method, this paper will uncover some flaws in the current  definition of \textit{exemplar-based texture  modeling}. This will lead us to propose a slightly different definition of the problem that seems to address better its challenges.

The Oxford Dictionary of English defines texture as the feel, appearance, or consistency of a surface or a substance. Focusing on visual appearance, texture is analog to color, a perceived quality of a surface, where the RGB bands are replaced by the output of a specific bank of filters~\cite[p.215]{mumford-desolneux}. Julesz defined  textures as classes of pictures that cannot be discriminated in preattentive vision and advanced two statistical hypotheses to characterize them~\cite{julesz,Julesz62,julesz81theory}. Grenander proposed to use the term ``texture'' for strictly stationary stochastic processes~\cite[p.398]{grenander}. Giving a precise definition of textures is a slippery task; in a sense, each model implicitly proposes one and as we will see the jury is still out.

\emph{Exemplar-based texture synthesis} is the process of generating, from an input texture sample, new texture images of arbitrary size and which are perceptually equivalent to the input. It is common to classify them under the two classical statistical estimation categories: \emph{parametric methods} and \emph{non-parametric methods}. The parametric methods aim at characterizing a given texture sample by estimating a set of statistics which will define an underlying stochastic process. The new images will then be samples of this stochastic process, i.e. they will have the same statistics as the input sample. The question here is: what would be the appropriate set of statistics to yield a correct synthesis for the wide variety of texture images? The results of these methods are satisfying but only on a small group of textures, and often fail when important structures are visible in the input. The non-parametric methods reorganize local neighbourhoods from the input sample in a consistent way to create new texture images. These methods return impressive visual results. Nevertheless, they often yield verbatim copies of large parts of the input sample. Furthermore, they can diverge, starting to reproduce iteratively one part of the input sample and neglecting the rest of it, thus growing what experts call ``garbage''. Because ``non-parametric'' methods are not completely parameter-free, and ``parametric'' methods can have a reduced set of parameters, in this paper we will denote by \emph{patch re-arrangement methods} the former and by \emph{statistics-based methods} the latter.

{What constitutes a texture? The answer depends of course on human perception. But a mathematical formulation can be used to characterize patterns that are perceived as textures.} The statistical characterization of texture images was initiated by B\'ela Julesz \cite{Julesz62,julesz73}. Julesz was the first to point out that texture images could be reliably organized according to their N-th order statistics into groups of textures that are preattentively indistinguishable by humans \cite{Julesz62}. {(Focusing on pre-attentive vision helps to reduce the subjective impact of high level processing.)} Julesz~\cite{julesz73} demonstrated that many texture pairs sharing the same second-order statistics would not be discerned by human preattentive vision. This hypothesis constitutes the first Julesz axiom for texture perception. One consequence of this axiom is that two textures sharing the same Fourier modulus but with different phase should be perceptually equivalent. Indeed, the square Fourier modulus of an image corresponds to its spatial auto-correlation, thus the second-order statistics. This motivates a class of algorithms (the random phase methods) aiming at creating textures with a given second-order statistic. An example of such algorithms is \cite{Wijk_spot_noise_texture_synthesis_1991}. In  a more recent extension \cite{GalerneRPN},  a texture is generated by randomizing the Fourier phase while maintaining the Fourier modulus. The Random Phase Noise method in \cite{GalerneRPN} correctly synthesizes textures with no salient details, namely microtexture, which adapt well to a Gaussian distribution, but it fails for more structured ones, macrotextures, as can be experimented in the executable paper \cite{GalerneRPNIpol}. Indeed, textures may share the same second and even third order statistics while being visually different~\cite{julesz78,caelli78}. This led Julesz~\cite{julesz81theory, julesz81textons} to propose a second theory to explain texture preattentive discrimination by introducing the notion of \textit{textons}. {Textons are local conspicuous features like bars or corners. Giving a mathematical definition for textons is far from trivial and was studied in for example~\cite{zhu-textons,desolneux2012}.} Julesz' second theory states that only the first order statistics of these textons are relevant for texture perception: images having the same texton densities (thus, just  a first order statistic) could  not be discriminated. Texton theory proposes the main axiom that texture perception is invariant to random shifts of the textons \cite{julesz81theory}. This axiom is extensively used in the stochastic dead leaves models~\cite{matheron68,serra82,bordenave2006}.

{Several models of the early visual processing in mammals are based on a multiscale analysis with Gabor kernels, and are used in particular for modeling the perception of texture~\cite{bergen-adelson-1986,turner1986,malik-perona1990}. Wavelet analysis provided a natural frame for these models and resulted in effective methods for texture classification and segmentation~\cite{chang1993,laine1993,unser1995,manjunathi1996}.} Heeger and Bergen~\cite{HeegerBergen} extended Julesz' approach to multiscale statistics. They characterized a texture sample by the histograms of its wavelet coefficients. By enforcing the same histograms on a white noise image they obtained a new multiscale exemplar-based texture synthesis method. Yet this method only measures marginal statistics. It misses important correlations between pixels across scales and orientations which are crucial to generate edges and conspicuous patterns. We refer to the on-line execution of this method~\cite{HeegerBergenIpol} where some successes but many failures are evident, as is also the case for RPN~\cite{GalerneRPNIpol}. Within a similar range of results, the De~Bonet~\cite{DeBonet} method randomizes the initial texture image and preserves only a few statistics, namely the dependencies across scales of a multi-resolution filter bench response. 
Other methods are also based on statistics of wavelet coefficients or more involved multiscale image representations~\cite{PS, Peyre_texture_synthesis_grouplets_2010, Rabin_Peyre_Delon_Bernot_wasserstein_barycenter_texture_mixing_proc_2010}. The Heeger-Bergen method was extended by Portilla and Simoncelli~\cite{PS} who proposed to  evaluate on the sample some 700 cross-correlations, autocorrelations and statistical moments of the wavelet coefficients. Enforcing the same statistics on synthetic images, starting from white noise, achieves striking results for a  wide range of texture examples. This method, which for a decade represented the state-of-the-art for psychophysically and statistically founded algorithms is nevertheless computationally heavy, and its convergence is not guaranteed. Its results, though generally indiscernible from the original samples in a pre-attentive examination, often present blur and phantoms. Earlier, Zhu, Wu and Mumford~\cite{zhu98} proposed to model texture images by inferring a probability distribution on a set of images with the same texture appearances and then to sample from it. To infer this probability distribution, the set of images is filtered by a pre-selected set of filters (which capture the important features of a given texture image) and their histograms are extracted. These are estimates of the marginals of the probability distribution sought for. Then the maximum entropy probability distribution is constructed matching the previous marginals. To sample from this probability distribution the Gibbs sampler is adopted, thus generating  new texture images. The resulting model is a Markov random field. The limitation of this method is its practical aspect. Inferring the probability distribution and sampling from it are complex tasks. {More recent work by Zhu~et~al.~\cite{zhu-liu-wu-2000,wu-zhu-liu-2000} advanced the \emph{Julesz ensembles} texture model based on a common set of statistics; they proved that this model is equivalent to FRAME in the limit of an infinite image grid. An efficient MCMC sampling method was also proposed.}
These two texture generators have been recently revisited with neural networks. Gatys' texture generator \cite{gatys} and DeepFrame \cite{deepframe} can be seen respectively as extended versions of \cite{PS} and \cite{zhu98}, and get significantly better results. Some new neural network methods, based on generative neural networks, also get notable results \cite{jetchev2016texture}. {All these recent methods show that the Julesz program of seeking the right statistics to characterize a texture is still well alive.}

It is worth mentioning that texture models can be used to complete missing parts of an image or texture inpainting. These methods rely on the definition of texture images as the realization of a random field. For inpainting this boils down to the estimation of a random texture model on the masked input image (a set of valid pixels of the image) from which a new image is sampled conditioned to some of the known values of input image. The method presented in \cite{galerne2016microtexture,galerne_gaussian_inpainting_2017,galerne_gaussian_inpainting_ipol_2017} is particularly well adapted for micro-textures. A Gaussian model is estimated on the masked input image; then the result is generated by a conditional sampling from the estimated model using the kriging estimation framework.

Patch re-arrangement methods constitute a totally different category of texture synthesis algorithms. {This category started by pixel re-arrangement using square patches as context.} The initial Efros and Leung \cite{EfrosLeung} method was inspired by Shannon's Markov random field model for the English language \cite{shannon}. In analogy with Shannon's algorithm for synthesizing sentences, the texture is constructed pixel by pixel. For each new pixel in the reconstructed image, a patch centered in the pixel is compared to all the patches of the input sample. The patches in the sample that are similar help predict the pixel value in the synthetic image. Several optimizations have been proposed to accelerate this algorithm. Among them Wei and Levoy \cite{WeiLevoy} managed to fix the shape and size of the learning patch, and Ashikhmin \cite{ashikhmin} proposed to extend existing patches whenever possible instead of searching in the entire sample texture. Yet, as already pointed out in the original paper \cite{EfrosLeung}, an iterative procedure may fail by producing ``garbage'' when the  neighborhood's size is too small. On the other hand, it can lead to a trivial verbatim reproduction of big pieces of the sample when the neighborhood is too large. This can be experimented in the online executable paper \cite{EfrosLeungIpol}. Many extensions of \cite{EfrosLeung} have been proposed that manage to accelerate the procedure and reduce the ``garbage'' problem by stitching entire patches instead of pixels. {Among the first methods proposing to re-arrange whole patches, Xu~et~al.~\cite{chaos} proposed to synthesize a texture by picking random patches from the sample texture and placing them randomly in the output texture image. A blending step is applied across the overlapping blocks to avoid edge artifacts. 
} In \cite{liang2001} the authors proposed to synthesize the image by quilting together patches that were taken from the input image among those who best match the patch under construction. A blending step was also added to overcome some edge artifacts. Efros and Freeman \cite{EfrosFreeman} proposed an extension of the latter introducing the quilting method (a blending step) that computes a path with minimal contrast across overlapping patches, thus mitigating the transition effect from patch to patch.

Kwatra et al.~\cite{kwatra2003} extended \cite{EfrosFreeman} by using a graph-cut algorithm to define the edges of the patch to quilt in the synthesis image. Another extension of \cite{EfrosLeung} was proposed by Kwatra et al.~\cite{kwatra2005} where to synthesize a texture image they improve the quality of the synthesis image sequentially by minimizing a patch-based energy function. In the same spirit as \cite{kwatra2005}, where texture optimization is performed, the authors in \cite{lefebvre2005parallel} proposed to synthesize textures in a multiscale framework using the coordinate maps of the sample texture at different scales. They introduced spatial randomness by applying a jitter function to the coordinates at each level, combined to a correction step inspired by \cite{ashikhmin}. One of the key strengths of the method is that it is a parallel synthesis algorithm which makes it extremely fast. These patch-based approaches often present satisfactory visual results. In particular they have the ability to reproduce highly structured textures (macrotextures). However, the risk remains of copying even several times verbatim large parts of the input sample. For practical applications this may result in the appearance of repeated patterns in the synthesized image. Furthermore, a fidelity to the global statistics of the initial sample is not guaranteed, in particular when the texture sample is not stationary. We refer to \cite{WeiLefebvre} for an extensive overview of the different patch re-arrangement methods.

Recent research tries to revisit the use of previous methods. Using neural networks has seen some success, as well as combining patch re-arrangement and statistics-based methods to overcome the drawbacks mentioned previously~\cite{peyre,tartavel}. These approaches will be called \emph{hybrid methods}. Peyr\'e~\cite{peyre} proposed to use a patch-based approach where all the patches of the synthesized image are created from a sparse dictionary learnt on the input sample. Tartavel~et~al.~\cite{tartavel} extended~\cite{peyre} by minimizing an energy that involves a sparse dictionary of patches combined to constraints on the Fourier spectrum of the input sample in a multiscale framework. Raad~et~al.~\cite{raad2016} proposed to model the self-similarities of a given input texture with conditional multivariate Gaussian distributions in the patch space in a multiscale framework. A new image is generated patch by patch, where for each given patch a multivariate Gaussian model is inferred from its nearest neighbours in the patch space of the input sample, and hereafter sampled from this model.

The academic literature shows that current methods are able to produce impressive texture synthesis on various kinds of textures. Our experiments will illustrate this, and the opposite. Indeed, this literature is still working, in a sense, on toy examples. Most textures are defined by texture samples of relatively small size and the structures are present in a small range of scales. When confronting the methods with more challenging data, the quality of the results degrades rapidly. This can be seen for most natural images of textures, which are non-stationary, due for example to the presence of illumination changes and perspective. As a matter of fact large  photographs of textures are non-stationary because even  homogeneous material always shows an internal variation of structure.  Thus the classic exemplar-based texture synthesis problem can be seen in this light as an almost impossible Fourier spectrum extrapolation, given a very small texture example. Hence our exploration not only of the solutions, but of the  problem itself will illustrate the limitations of the current question, and introduce a more general question: how  to emulate the real, non-stationary textures, for which we dispose of large samples?  Then the question is no longer to ``extend'' a small patch into a larger texture of the same kind, but rather to be able  to fabricate other examples of a given large and complex texture, given only  one sample of it.

{This survey concentrates on the problem of texture synthesis on flat 2D domains. There are several interesting extensions and applications of the basic problem which are not discussed here. These include \emph{surface texture synthesis}, in which a texture is to be placed onto a curved surface, \emph{dynamic texture synthesis}, when the goal is to generate textures whose appearance evolves over time such as for videos of time-variant materials, or \emph{solid texture synthesis}, where the aim is to generate the color content of 3D blocks of synthesized materials from which, for example, computer graphics objects can be carved out. Other related problems include \emph{image completion} and \emph{resolution enhancement} by texture synthesis. Also, the computational cost in real-time applications (e.g. games) or when the data volume is large (e.g. film production) impose further restrictions leading to particular algorithms. For a discussion of these topics, we refer the reader to Wei~et~al.~\cite{WeiLefebvre} and the references therein.}

We now sketch our plan.  We shall present the main trends in exemplar-based texture synthesis by describing in detail several methods illustrating the three main families. In each case, the strength and limitations will be commented as well as some relevant variations. Section~\ref{sec:parametric} introduces the statistics-based methods which perform statistical optimization and describes several algorithms. Then Section~\ref{sec:non-parametric} focuses on patch re-arrangement methods, presenting three works. The third main class of hybrid methods is discussed in Section~\ref{sec:hybrid}. The experimental Section~\ref{sec:exp} first compares the main families of algorithms in a varied set of textures; then, the limitations of all current methods are revealed with high-resolution and non-stationary examples. Finally, Section~\ref{sec:conclusion} concludes the paper. {All the results displayed were generated for this paper, with the original code published with the methods \cite{HeegerBergenIpol, GalerneRPNIpol, codePS, codegatys, codeDF, codesgan, EfrosLeungIpol, EfrosFreemanIpol, codecnnmrf} or with the modifications mentioned in this paper.}

\section{Statistics-based methods}\label{sec:parametric}

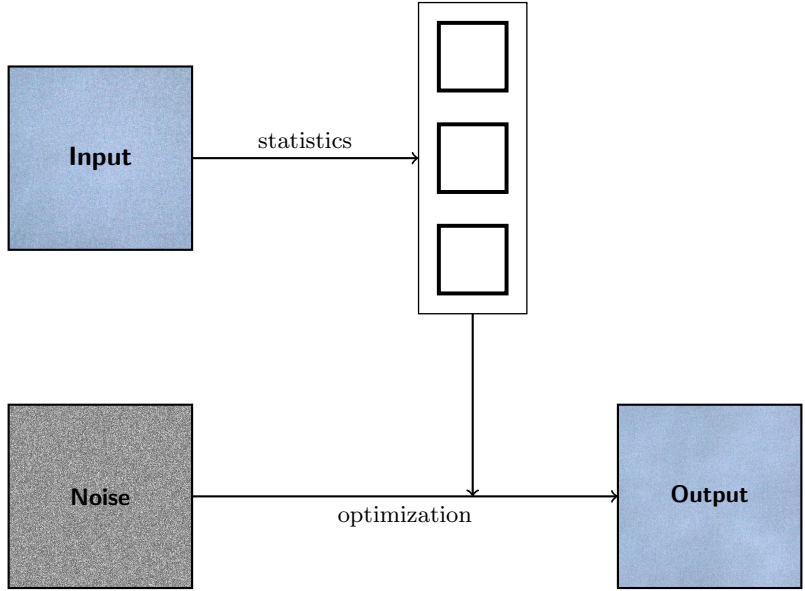
\begin{figure}[t]
  \centering
  \begin{tikzpicture}[scale=0.9]
    \centering

    \node[anchor=center, inner sep=0] (input) at (0,0) {\includegraphics[width=0.2\textwidth]{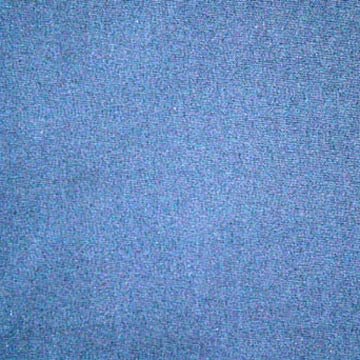}};
    \draw[thick, fill=white, fill opacity=.4] (input.south west) rectangle (input.north east) node[midway, opacity=1, font=\sffamily\bfseries] {Input};

    \draw[line width=1.5pt] (5, 1) rectangle (6,2);
    \draw[line width=1.5pt] (5, -0.5) rectangle (6,0.5);
    \draw[line width=1.5pt] (5, -2) rectangle (6,-1);
    \draw[line width=0.5pt] (4.7, -2.3) rectangle (6.3,2.3);

    \draw[thick,->] (input.east) -- node[midway,above] {\small{statistics}} (4.7,0);

    \node[anchor=center, inner sep=0] (noisy) at (0,-5) {\includegraphics[width=0.2\textwidth]{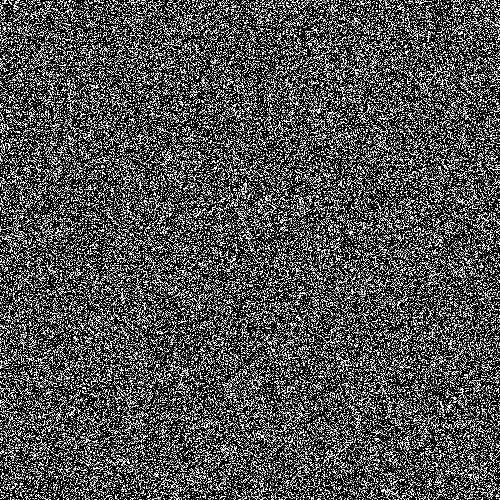}};
    \draw[thick, fill=white, fill opacity=.4] (noisy.south west) rectangle (noisy.north east) node[midway, opacity=1, font=\sffamily\bfseries] {\small{Noise}};

    \node[anchor=center, inner sep=0] (output) at (9,-5) {\includegraphics[width=0.2\textwidth]{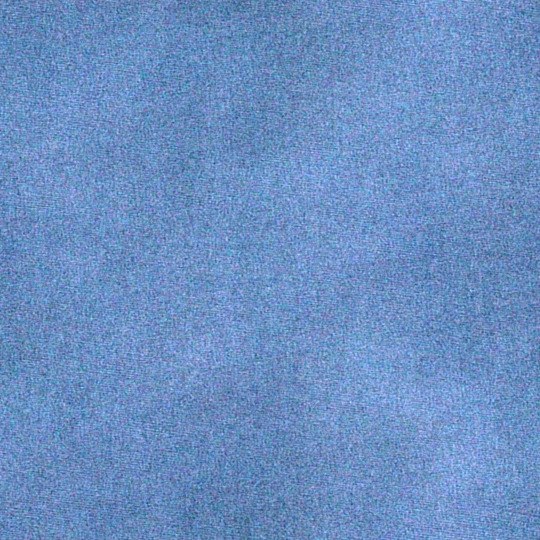}};
    \draw[thick, fill=white, fill opacity=.4] (output.south west) rectangle (output.north east) node[midway, opacity=1, font=\sffamily\bfseries] {\small{Output}};

    \draw[thick,->] (noisy.east) -- node[midway,below] {\small{optimization}} (output.west);
    \draw[thick,->] (5.5,-2.3) -- (5.5,-5);

  \end{tikzpicture}
  \caption{Statistics-based methods. A set of statistics is extracted from an input sample (analysis step). Then, starting with a noise image, an optimization procedure is applied  to enforce these statistics on the output image  (synthesis step).}
  \label{fig:statistics-based}
\end{figure}

Statistics-based texture synthesis methods follow the general approach proposed by Julesz, illustrated in Figure~\ref{fig:statistics-based}. The synthesis is performed in two steps: first, a set of statistics is estimated from the sample texture; second, a random image is generated, subject to these statistical constraints. Methods in this class differ in the set of statistics considered and in the optimization method used to impose them on a random image. We will describe several algorithms of this class with increasing sophistication. It will appear that the number of statistics enforced plays a key role in the success.

\subsection{Micro-texture synthesis by phase randomization}\label{sec:rpn}

The Random Phase Noise (RPN) method synthesizes a new texture from a rectangular sample by simply randomizing the phase of the Fourier coefficients of the input sample. The results are very satisfying for textures that are characterized by their Fourier modulus, a class called \textit{micro-texture} by some authors. This method is also able to create a random texture from any input image, not necessarily a texture sample. It is in spirit quite close to the noise generators from computer graphics \cite{perlin,Wijk_spot_noise_texture_synthesis_1991}. The rest of this section describes the main ideas of this method and we refer the reader to \cite{GalerneRPNIpol} for more details and a catalog of several synthesis examples.

\begin{figure}[t]
  \centering
  \begin{tikzpicture}
    \node[anchor=south, inner sep=0] (input1) at (0,0) {\includegraphics[width = .15\textwidth]{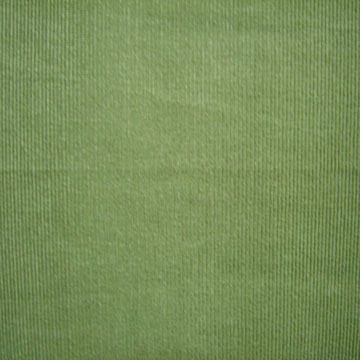}};
    \node [anchor=north east, rounded corners, fill=white, opacity=.5, text opacity=1] at (input1.north east) {\footnotesize input};
    \node[anchor=south, inner sep=0] (output1) at (3,0) {\includegraphics[width = .3\textwidth]{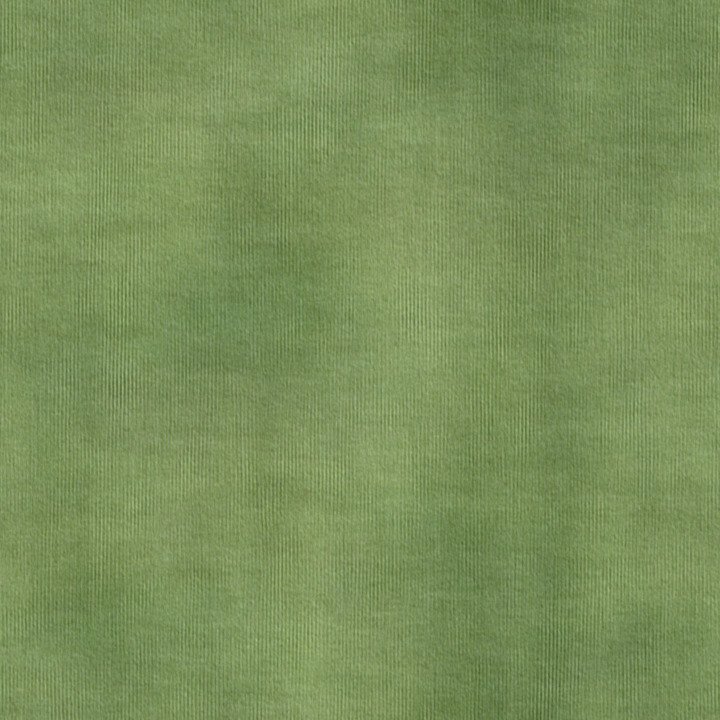}};
    \node [anchor=north east, rounded corners, fill=white, opacity=.5, text opacity=1] at (output1.north east) {\footnotesize output};

    \node[anchor=south, inner sep=0] (input2) at (6.5,0) {\includegraphics[width = .15\textwidth]{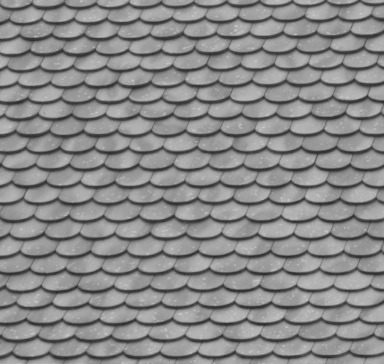}};
    \node [anchor=north east, rounded corners, fill=white, opacity=.5, text opacity=1] at (input2.north east) {\footnotesize input};
    \node[anchor=south, inner sep=0] (output2) at (9.5,0) {\includegraphics[width = .3\textwidth]{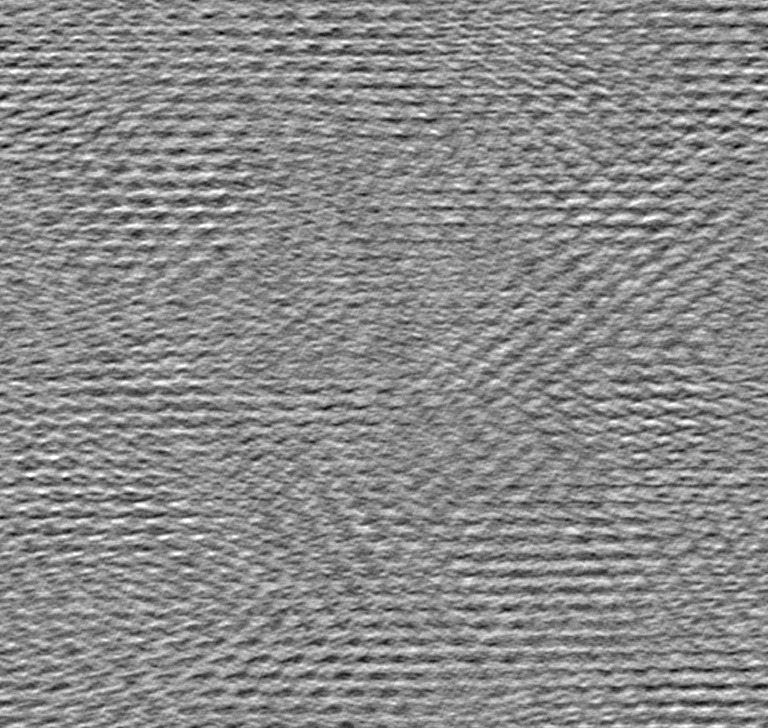}};
    \node [anchor=north east, rounded corners, fill=white, opacity=.5, text opacity=1] at (output2.north east) {\footnotesize output};
  \end{tikzpicture}
  \caption{Synthesis results of the RPN method \cite{Wijk_spot_noise_texture_synthesis_1991,GalerneRPN}. This method works extremely well for micro-textures including tissues and granular textures with no geometric structures \cite{GalerneRPNIpol}. For more structured texture images it fails. Two examples are shown: a successful synthesis on the left and a failure case on the right.}
  \label{fig:RPN-examples}
\end{figure}

The RPN of an image $u$ defined on a domain $\Omega$ is obtained by adding a random phase $\theta$ to the Fourier phase of the input sample image. The random phase is a white noise image uniformly distributed over $(-\pi,\pi]$ and is constrained to be symmetric. In the case of an RGB color image $u = (u_R, u_G, u_B)$, the RPN image is obtained by adding the same random phase to the Fourier transform of each color channel. Adding the same random phase to the original phases of each color channel preserves the phase displacements between channels. This is important as it permits to create new textures without creating false colors~\cite{GalerneRPN}.

More precisely, a \emph{uniform random phase} is defined as a random image $\theta \in \mathbb{R}^{M\times N}$ satisfying the following conditions:
\begin{itemize}
\item $\theta$ is odd: $\forall m,n \in \Omega, \theta(-m,-n)=-\theta(m,n)$;
\item $\theta(m,n)$ is uniform on the interval $(-\pi,\pi]$ for $(m,n) \not\in \{(0,0), (M/2,0),(0,N/2), (M/2,N/2)\}$;
\item $\theta(m,n)$ is uniform on the set $\{0,\pi\}$ for $(m,n) \in \{(0,0), (M/2,0),(0,N/2), (M/2,N/2)\}$;
\item for every subset $\mathcal{S}$ of the Fourier domain which does not contain distinct symmetric points, the family of random variables $\{\theta(m,n) \vert (m,n)\in\mathcal{S}\}$ is independent.
\end{itemize}
The RPN of an image $u\in \mathbb{R}^{M\times N}$ is defined as the random image $X$ where there exists a uniform random phase $\theta$ such that
\begin{equation}
  \hat X(\xi,\eta) = \hat u(\xi,\eta)e^{i\theta(\xi,\eta)}, (\xi,\eta) \in \Omega,
  \label{eq:def-rpn-1}
\end{equation}
where $\hat u$ denotes the Fourier transform of $u$. An equivalent definition is
\begin{equation}
  \hat X(\xi,\eta) = \vert \hat u(\xi,\eta)\vert e^{i\theta(\xi,\eta)},
  \label{eq:def-rpn-2}
\end{equation}
where $\theta$ is a uniform random phase. Given the phase $\phi$ of a real-valued image and a uniform random phase $\theta$, the random image $(\theta + \phi) \mod 2\pi$ is also a uniform random phase, which proves this equivalence. The first definition~\eqref{eq:def-rpn-1} leads to a natural extension of RPN to color images~\cite{GalerneRPN}, while the second definition~\eqref{eq:def-rpn-2} highlights the fact that the RPN depends only on the Fourier modulus of the sample image $u$.

Similarly, an Asymptotic Discrete Spot Noise (ADSN) associated with an image $u$ is defined as the convolution of a normalized zero-mean copy of $u$ with a Gaussian white noise. A Gaussian white noise image has a uniform random phase and its Fourier modulus is a white Rayleigh noise; the phase and modulus are independent. Thus, the phase of the ADSN is a uniform random phase whereas its Fourier modulus is the pointwise multiplication of the Fourier modulus of $u$ by a Rayleigh noise~\cite{GalerneRPN}. Both ADSN and RPN have uniform random phases, but the modulus distributions are different.  RPN keeps the Fourier modulus of the original image, while for ADSN the Fourier modulus is degraded by a pointwise multiplication by a white Rayleigh noise. Regardless of their theoretical differences, ADSN and RPN produce results that are perceptually very similar~\cite{GalerneRPN}.

The RPN method is the fastest method presented in this review since it basically needs the computation of two FFTs. Nevertheless, this method is limited to micro-textures and it will fail synthesizing structured textures, namely macro-textures. In Figure~\ref{fig:RPN-examples} two synthesis examples are shown. The first synthesis (left example in Figure~\ref{fig:RPN-examples}) shows outstanding results. This micro-texture is indeed well represented by its Fourier modulus. However this is not at all the case for the second texture synthesis (right example in Figure \ref{fig:RPN-examples}). Clearly, the knowledge of the modulus of the Fourier coefficients of this texture is not sufficient to recover the strong contrast of the input.

\subsection{The Heeger and Bergen pyramid based texture synthesis}\label{sec:hb}

Heeger and Bergen~\cite{HeegerBergen} proposed to characterize a texture by the first order statistics of both its color and its responses to multiscale and multi-orientation filters organized in a steerable pyramid~\cite{steerable-pyramid}. This proposition, motivated by the study of human texture perception, focuses on the synthesis of microtextures defined as images that don't have conspicuous patterns (e.g., granite, bark, sand).

 Let us describe the input texture image $u$ and the synthesized texture $v$ using the Heeger and Bergen method. First the image $u$ is filtered using a steerable pyramid decomposition~\cite{steerable-pyramid,simoncelli_steerable} with $S$ scales and $Q$ orientations at each scale. The steerable pyramid is a linear multiscale and multi-orientation image decomposition. Given an input image, it is first filtered to provide a high frequency image and a low frequency image. Band-pass oriented filters are then sequentially applied to the low frequency image which is also down-sampled. These band-pass oriented filters are applied $S$ times to the corresponding low frequency image. This decomposition yields images of different sizes corresponding to the different scales and orientations on which the gray level histograms are extracted as well as the gray level histogram of $u$. These histograms define the set of statistics that characterize $u$. 

The second step consists in generating the output image $v$, which is initialized with a noise image. Its pixel values are iteratively modified to match the histograms of $u$ and of its steerable decomposition. These histogram matchings are performed on $v$ alternately in the image domain and in the multiscale transform domain, until all the output histograms match the ones of $u$. A third parameter is introduced here and it is the number of iterations used to achieve a stabilization of the histogram matching.

To the best of our knowledge, no theorem guarantees that this iteration will end with an image respecting all statistics; there is of course one solution to it, namely the  example image. But the goal is to create an image different from the example. Hence the random initialization, which is supposed to lead always to different samples of the same texture. This remark applies to all texture synthesis methods we will consider: their success will mainly be judged visually and experimentally.

To treat RGB color images, instead of applying the method to each color channel of the input image which are highly correlated, the authors proposed to change the color space RGB to a more adapted color space. This new color space is obtained by principal component analysis of the RGB values of the input image $u$. In~\cite{HeegerBergenIpol} a detailed explanation of the original method of Heeger and Bergen~\cite{HeegerBergen} is provided with a complete analysis of the steerable pyramid decomposition and the histogram matching step. The authors also provide in~\cite{HeegerBergenIpol} a minor improvement in the edge handling of the convolutions as well as an experimental section illustrating the influence of the parameters, namely the number of iterations, the number of scales and the number of orientations. As we said, there is no theoretical proof of convergence of the method but an experimental study shows that the results tend to stabilize after five to ten iterations~\cite{HeegerBergenIpol}. Increasing the number of orientations changes the results slightly, but four orientations are enough in general. The number of scales is very important. Taking the highest number permits to take into account all the scales of the texture. When the input texture has no evident structure then this parameter has less influence in the result.

\begin{figure}[t]
  \centering
  \begin{tikzpicture}
    \node[anchor=south, inner sep=0] (input1) at (0,0) {\includegraphics[width = .18\textwidth]{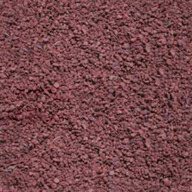}};
    \node [anchor=north east, rounded corners, fill=white, opacity=.5, text opacity=1] at (input1.north east) {\footnotesize input};
    \node[anchor=south, inner sep=0] (output1) at (3,0) {\includegraphics[width = .27\textwidth]{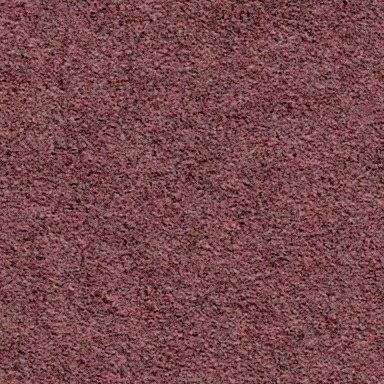}};
    \node [anchor=north east, rounded corners, fill=white, opacity=.5, text opacity=1] at (output1.north east) {\footnotesize output};

    \node[anchor=south, inner sep=0] (input2) at (6.5,0) {\includegraphics[width = .18\textwidth]{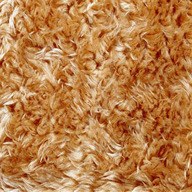}};
    \node [anchor=north east, rounded corners, fill=white, opacity=.5, text opacity=1] at (input2.north east) {\footnotesize input};
    \node[anchor=south, inner sep=0] (output2) at (9.5,0) {\includegraphics[width = .27\textwidth]{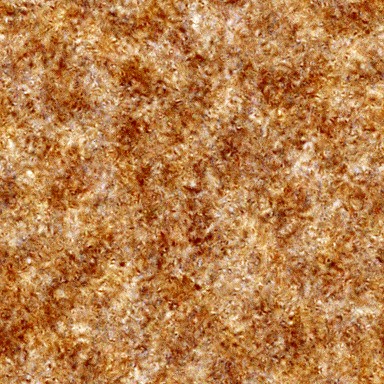}};
    \node [anchor=north east, rounded corners, fill=white, opacity=.5, text opacity=1] at (output2.north east) {\footnotesize output};
  \end{tikzpicture}
  \caption{Synthesis results of the Heeger and Bergen method \cite{HeegerBergen}. This method works for microtextures. For more structured texture images it fails. Two examples are shown: a successful synthesis on the left and a failure case on the right.}
  \label{fig:HB-examples}
\end{figure}
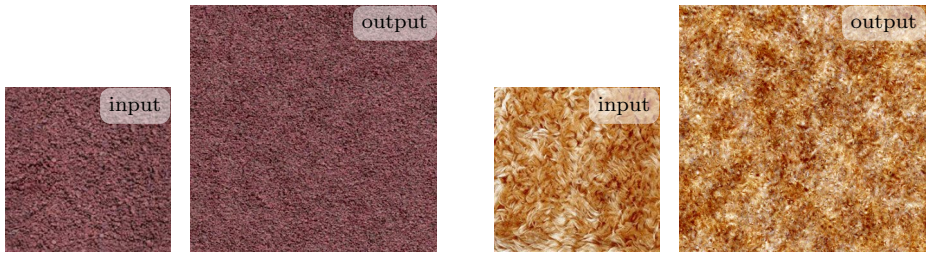

As our experiments here will show, the results yielded by this approach are convincing for some stochastic textures, but the method fails for most complex texture images. In particular it generally fails (visually) for quasi-periodic textures, random mosaic textures, textures having more than one dominant orientation, and textures having correlations of high frequency content over large distances. This demonstrates experimentally that all the spatial information characterizing a texture is not captured by the first order statistics of a set of linear filter outputs. In Figure~\ref{fig:HB-examples} two synthesis examples are shown: a successful synthesis and a failure case.

{
\subsection{FRAME: a mathematical model for textures}\label{sec:frame}

FRAME, which stands for Filters, Random fields And Maximum Entropy, is a mathematical model of textures developed by Zhu, Wu and Mumford in~\cite{ZhuMumford} and~\cite{zhu98}. It is the most mathematical solid work among the stream of work on texture modeling during that period. It puts the graphics method of Heeger and Bergen~\cite{HeegerBergen} in a mathematical sound setting, i.e., it has a formal statistical model, and it can match the marginal statistics. It also answers the Julesz quest by pursuing the minimum statistical constraint that are necessary.

The FRAME model is based on the maximum entropy principle. It starts with a set of filters that are selected from a general filter bank to capture features of the texture. These filters are applied to observed texture images, and the histograms of the filtered images are extracted. Then, the maximum entropy principle is employed to derive a distribution $f$, which has in expectation the same filter responses as the original image, while being of maximum entropy. More precisely, let $u$ be an observed texture image and let $F^k$, $k=1,\ldots, K$ be a set of filters. Let $H_u^k$ be the (discrete) histograms of the filter responses $F^k \ast u$, and for any image $v$, let  $H_v^k$ be the histograms of the filter responses $F^k \ast v$. Zhu, Wu and Mumford seek for a distribution $f(v)$ on images $v$ such that
\begin{equation}
  \mathbb{E}_f(H_v^k) =  H_u^k,
\label{eq:expect}
\end{equation}
while being of maximum entropy (i.e. while being ``as random as possible''). By Lagrange Multipliers (maximization under constraints), the solution has the form
$$
f(v ; \lambda) = \frac{1}{Z(\lambda)} \exp\left(- \sum_{i=1}^L \sum_{k=1}^K \lambda_i^k H_u^k(i) \right),
$$
where $L$ is the number of bins of the discrete histograms. To find the value of the parameters $\lambda$ satisfying Equation~\eqref{eq:expect}, Zhu, Wu and Mumford use a gradient descent to find the right $\lambda$ and the Gibbs sampler algorithm to sample random images $v$ from a distribution $f(\cdot ; \lambda)$. The distribution $f(\cdot ; \lambda)$ defines a Markov Random Field (MRF).

Finally, a stepwise algorithm is proposed to choose the filters from a general filter bank. This ``filter pursuit'' step is achieved thanks to the minimax entropy principle: find the set of filters such that the associated distribution $f$ is of minimum entropy, since it is equivalent to be of minimal Kullback-Leibler divergence from the ``true'' underlying distribution. A detailed explanation of this fact can be found in~\cite{zhu98}.

The FRAME model was later extended, in particular with non linear filters, using the output of some layers of a neural network. It is then called DeepFrame~\cite{deepframe}, and we will talk again about it in Section~\ref{sec:cnn}.
}

\subsection{The Portilla and Simoncelli algorithm}\label{sec:ps}

{The key issue in FRAME and in the method of Heeger and Bergen is to choose the ``right'' filters and the statistics that will be matched.} In~\cite{PS}, Portilla and Simoncelli proposed an important improvement on Heeger and Bergen’s method~\cite{HeegerBergen}. The texture is again synthesized starting from a noise image and coercing it to have the same statistics as the input image. As we have seen, marginal statistics are not enough to capture the relations across scales and orientations. Portilla and Simoncelli proposed to match a set of joint statistics measurements of the coefficients of the steerable pyramid decomposition of the input texture. The statistics used to characterize the input texture are the autocorrelation and cross-correlation coefficients (inner and intra scales), as well as the statistical moments of order one, two, three and four of the input sample's values. To enforce these statistics on the result, the image under construction is projected iteratively into the subspace of constraints using a gradient projection approach until stabilization. The final output image may not have exactly the same statistics as the input sample. It merely represents a local minimum. Again there is no proof of a convergence of the method anyway.

Portilla and Simoncelli's technique is based on the theories of human visual perception, in particular Julesz' hypothesis stating that two images are perceptually equivalent if and only if they agree on a set of statistic measurements. The goal is to establish the minimal set of measurements in a way that all types of textures are correctly synthesized using that set of measurements. In the same way as Heeger and Bergen’s method, the input texture sample is decomposed with a multiscale oriented linear basis: the steerable pyramid~\cite{steerable-pyramid,simoncelli_steerable}. For each pair of coefficients at nearby positions, orientations and scales, the average value of their product, of their magnitude product and their relative phase is measured. In addition to these parameters, some marginal statistics on the input image pixels distribution are kept: the mean, the variance, the skewness, the kurtosis and the range. The  number of parameters will depend on the number of sub-band images and on the size of the neighbourhood considered to estimate the statistical constraints of the example texture.

The second part of the algorithm is the synthesis step coercing to a random noise image, the measurements previously computed. The synthesized image is initialized with a Gaussian white noise image and then iteratively the algorithm alternates between: 1) constructing the steerable pyramid and enforcing the sample statistics of each sub-band image matching those of the corresponding sub-bands of the target image; 2) reconstructing an image from the pyramid and then forcing it to have the same marginal statistics as the input texture.

A texture is defined as a two-dimensional stationary random field $X(m,n)$ on a finite lattice $(m,n) \in \Omega \subset \mathbb{Z}^2$. Julesz' hypothesis is the basis to connect this statistical definition to perception: there exists a set of constraint functions $\lbrace \phi_k,k = 1,\dots,K\rbrace$ such as if two random fields, $X$ and $Y$, are identical in expectation over this set of functions then any two samples drawn from $X$ and $Y$ will be perceptually equivalent under some fixed comparison conditions. The importance of human perception as a fundamental criterion of equivalence between textures can be seen through this hypothesis, as well as the existence of such a set of statistical measurements capable of capturing this equivalence. To choose the set of constraint functions Portilla and Simoncelli proceeded as follows:

\begin{enumerate}
  \item Set an initial set of constraints and synthesize a large library of texture examples;
  \item Group the synthesis failures classifying them according to visual features that distinguish them from their original texture examples and keep the group with the poorest results;
  \item Add a new statistical constraint to the set capturing the missing feature of the failure group;
  \item Re-synthesize the failure group and verify the wanted feature is captured; otherwise go back to the previous point;
  \item Verify that the original constraints are still needed; for each constraint, find a texture example that fails when the constraint is removed from the set;
  \item Delete the unnecessary constraint, re-synthesize the library and go back to the second point.
\end{enumerate}

Following this strategy, the constraint set is adapted to a reference set of textures and not just to one texture, and it is driven by perceptual criteria. The set of constraints is composed of:

\begin{enumerate}
  \item Marginal statistics formed by: skewness and kurtosis of the low-pass images of each level of the pyramid, variance of the high-pass image of the pyramid, skewness, kurtosis, variance, mean and range of the image. The marginal statistics set the general degree of pixel intensity and their distribution. This is why they cannot be discarded from the statistics set~\cite{PS}.
  \item Autocorrelation of the low-band coefficients. This allows to capture the periodic structures of a texture as well as long-range correlation. Omitting this constraint from the original set yields unsatisfying results for textures having periodic or long-range correlation patterns~\cite{PS}.
  \item Autocorrelation and cross-correlation of the magnitude of the sub-bands. These statistics appear to be relevant because observation reveals that oriented bands have a particular behaviour concerning certain pattern and their periodicity whatever the orientation~\cite{PS}. The cross-correlations kept are of each sub-band image with others of the same scale (inner cross-correlation) and of each sub-band with sub-bands at the coarser scale (intra cross-correlation).
  \item Cross-correlation of the real part of the sub-bands with the real and imaginary parts of the coefficients' phase of the coarser scale. This statistic is important to capture the strong illumination effects present in some texture images. In particular, the synthesized image looses its three-dimensional effect and the shadows structure if they are not considered~\cite{PS}.
\end{enumerate}

The set of statistics is summarized in Table~\ref{table:ps_statistics}. As mentioned previously, the number of parameters used depends on the number of scales $S$ and orientations $Q$ of the steerable decomposition as well as the size of the neighbourhood $N_a$ used to compute the auto-correlations. The total number of parameters is $6 + 1 +2(S+1) + (S+1)(N_a^2+1)/2 + SQ(N_a^2+1)/2 + SQ(Q-1)/2 + (S-1)Q^2 + 2(S-1)Q^2$, where the terms correspond (from left to right) to: the marginal statistics of $u$, the variance of high pass image, the skewness and kurtosis of the low band images, the auto-correlation of the low band images, the auto-correlation of the sub band images, the inner cross-correlation of the sub band images, the intra cross-correlation of the low band images and the cross correlation of the real part of the sub band images with the real and imaginary part of the phase sub band images. In general $S=4$, $Q=4$ and $N_a=7$ are used, leading to a total of $710$ parameters.

\begin{table}[p]
  \centering
  \begin{tabular}{|@{\hspace*{2mm}}p{.30\textwidth}@{\hspace*{2mm}} p{.70\textwidth}@{\hspace*{2mm}}|}
    \hline

    \rule{0pt}{3ex}
    \emph{range of $u$} & $\max(u)$ and $\min(u)$ \\
    \rule{0pt}{3ex}
    \emph{mean of $u$} & $\mu_1(u)$\\
    \rule{0pt}{3ex}
    \emph{variance of $u$} & $\mu_2(u)$\\
    \rule{0pt}{3ex}
    \emph{skewness of $u$} & $\mu_3(u)/(\mu_2(u))^{1.5}$\\
    \rule{0pt}{3ex}
    \emph{kurtosis of $u$} & $\mu_4(u)/(\mu_2(u))^{2}$\\
    \rule{0pt}{3ex}
    \emph{lowband's skewness} & $\mu_3(l_s)/(\mu_2(l_s))^{1.5},~1\leq s \leq S+1$\\
    \rule{0pt}{3ex}
    \emph{lowband's kurtosis} & $\mu_4(l_s)/(\mu_2(l_s))^{2},~1\leq s \leq S+1$\\
    \rule{0pt}{3ex}
    \emph{highband's variance} & $\mu_2(h)$\\
    \rule{0pt}{3ex}
    \emph{$\Re\{l_s\}$ auto-correlation} & $\Gamma_{\Re\{l_s\}}\left(m,n\right),~1\leq s \leq S+1$\\
    \rule{0pt}{3ex}
    \emph{$\vert u^{s,q}\vert$ auto-correlation} & $\Gamma_{\vert u^{s,q}\vert}\left(m,n\right),~1\leq s \leq S, \hspace{1mm} 0\leq q \leq Q-1$\\
    \rule{0pt}{3ex}
    \emph{inner cross-correlation} & $\mathcal{C}\left( \left\vert u^{s,q} \right\vert , \left\vert u^{s,q'} \right\vert \right),~1\leq s \leq S,~0\leq q,q' \leq Q-1$\\
    \rule{0pt}{4ex}
    \emph{intra cross-correlation} & $\mathcal{C}\left( \left\vert u^{s,q} \right\vert , \left\vert u^{s+1,q'} \right\vert \right),~1\leq s \leq S-1,~0\leq q,q' \leq Q-1$\\
    \rule{0pt}{4ex}
    \emph{cross-correlation with the real part of the phase} & $\mathcal{C}\left( \Re \left\lbrace u^{s,q} \right\rbrace, \frac{\Re \left\lbrace u^{s+1,q'} \right\rbrace}{\left\vert u^{s+1,q'} \right\vert} \right),~1\leq s \leq S-1,~0\leq q,q' \leq Q-1$\\
    \rule{0pt}{4ex}
    \emph{cross-correlation with the imaginary part of the phase} & $\mathcal{C}\left( \Re \left\lbrace u^{s,q} \right\rbrace, \frac{\Im \left\lbrace u^{s+1,q'} \right\rbrace}{\left\vert u^{s+1,q'} \right\vert} \right), ~ 1\leq s \leq S-1,~0\leq q,q' \leq Q-1$\\

    \hline
    \rule{0pt}{5ex}
    \emph{Central sample moment} & $ \mu_{n}(u) = \twopartdef{\frac{1}{MN}\sum_{i=0}^{M-1}{\sum_{j=0}^{N-1}{u(i,j)}}}{n=1}{\frac{1}{MN}\sum_{i=0}^{M-1}{\sum_{j=0}^{N-1}{\left( u(i,j) - \mu_{1}\left( u\right) \right)^{n}}}}{n>1} $\\

    \rule{0pt}{4ex}
    \emph{Translation operator} & $\tau_{x,y}\left( u\right) : u(m,n)\mapsto u(\lfloor m-x\rfloor_{M},\lfloor n-y\rfloor_{N})$ \hspace{3mm} \mbox{$0\leq m\leq M-1$}, \hspace{2mm} \mbox{$0\leq n\leq N-1$}, \hspace{2mm} \mbox{$(x,y)\in \Omega $}\\

    \rule{0pt}{4ex}
    \emph{Correlation} & $ \mathcal{C}(u,v) = \frac{1}{MN}\sum_{i=0}^{M-1}{\sum_{j=0}^{N-1}{\left( u(i,j)-m(u)\right)\left( v(i,j)-m(v)\right)^{*}}} $\\

    \rule{0pt}{4ex}
    \emph{Auto-correlation} & $ \Gamma_{u}\left( x,y\right) = \mathcal{C}\left( u, \tau_{x,y}\left( u\right)\right) $\\

    \hline
  \end{tabular}

  \caption{Summary of the set of statistical constraints for the Portilla-Simoncelli method.}
  \label{table:ps_statistics}
\end{table}

After setting the set of statistical constraints, a sample verifying them has to be generated. Let ${c_k}$ be the corresponding estimated values of the constraint functions for a particular texture image. Portilla and Simoncelli~\cite{PS} ``samples'' an image from the set of images that yield the same estimated constraints values $A_{\vec{\phi},\vec{c}} = \{u:\phi_k(u) = c_k,~\forall k\}$. To pick at random from this set the authors proposed to select at random a sample $u_0$ from $\mathbb{R}^{\vert \Omega\vert}$ and then project it sequentially onto subsets of $A_{\vec{\phi},\vec{c}}$. To emulate this the authors proposed a gradient projection. That is moving in the direction of the gradient of the constraint $\phi_k(v)$:
\begin{equation*}
  v' = v + \lambda_k\overrightarrow{\nabla}\phi_k(v)
 \label{eq:gradient-direction}
\end{equation*}
choosing $\lambda_k$ such that
\begin{equation}
  \phi_k(v') = c_k.
  \label{eq:verifiy-constraints}
\end{equation}
Computing $\overrightarrow{\nabla}\phi_k(v)$ is usually simple, and it  remains to find the $\lambda_k$ that solves~\eqref{eq:verifiy-constraints}. When there are multiple solutions for $\lambda_k$, the one with smaller amplitude is chosen, modifying as little as possible the image. In that way, we stay as close as possible to the already projected set. When there is no solution, the $\lambda_k$ is the one that comes closest to satisfying~\eqref{eq:verifiy-constraints}. Finally this method can be extended to the adjustment of a subset of constraints. Once the set of statistical measurements is defined and a method to sample from the Julesz' ensemble of textures, the synthesis can be performed as explained previously.

\begin{figure}[t]
  \centering
  \begin{tikzpicture}
    \node[anchor=south, inner sep=0] (input1) at (0,0) {\includegraphics[width = .15\textwidth]{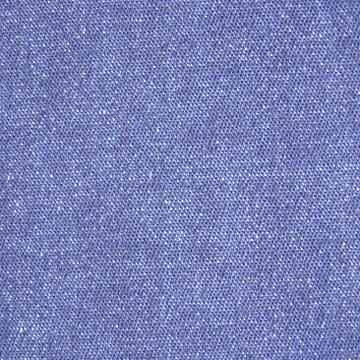}};
    \node [anchor=north east, rounded corners, fill=white, opacity=.5, text opacity=1] at (input1.north east) {\footnotesize input};
    \node[anchor=south, inner sep=0] (output1) at (3,0) {\includegraphics[width = .3\textwidth]{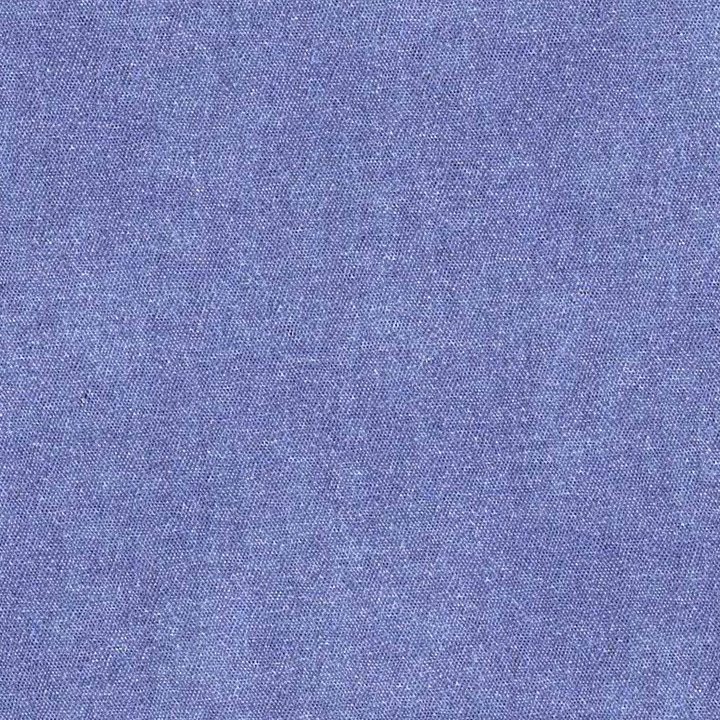}};
    \node [anchor=north east, rounded corners, fill=white, opacity=.5, text opacity=1] at (output1.north east) {\footnotesize output};

    \node[anchor=south, inner sep=0] (input2) at (6.5,0) {\includegraphics[width = .15\textwidth]{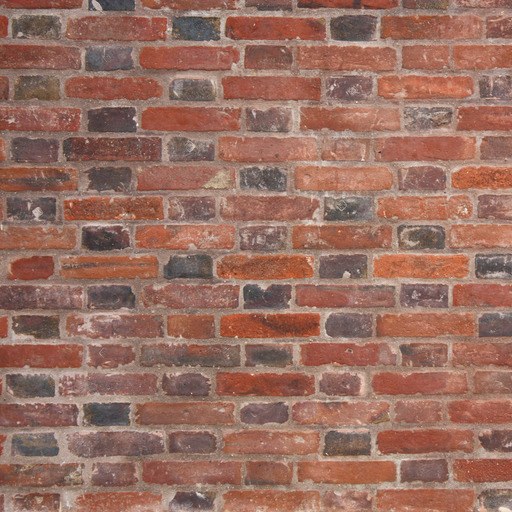}};
    \node [anchor=north east, rounded corners, fill=white, opacity=.5, text opacity=1] at (input2.north east) {\footnotesize input};
    \node[anchor=south, inner sep=0] (output2) at (9.5,0) {\includegraphics[width = .3\textwidth]{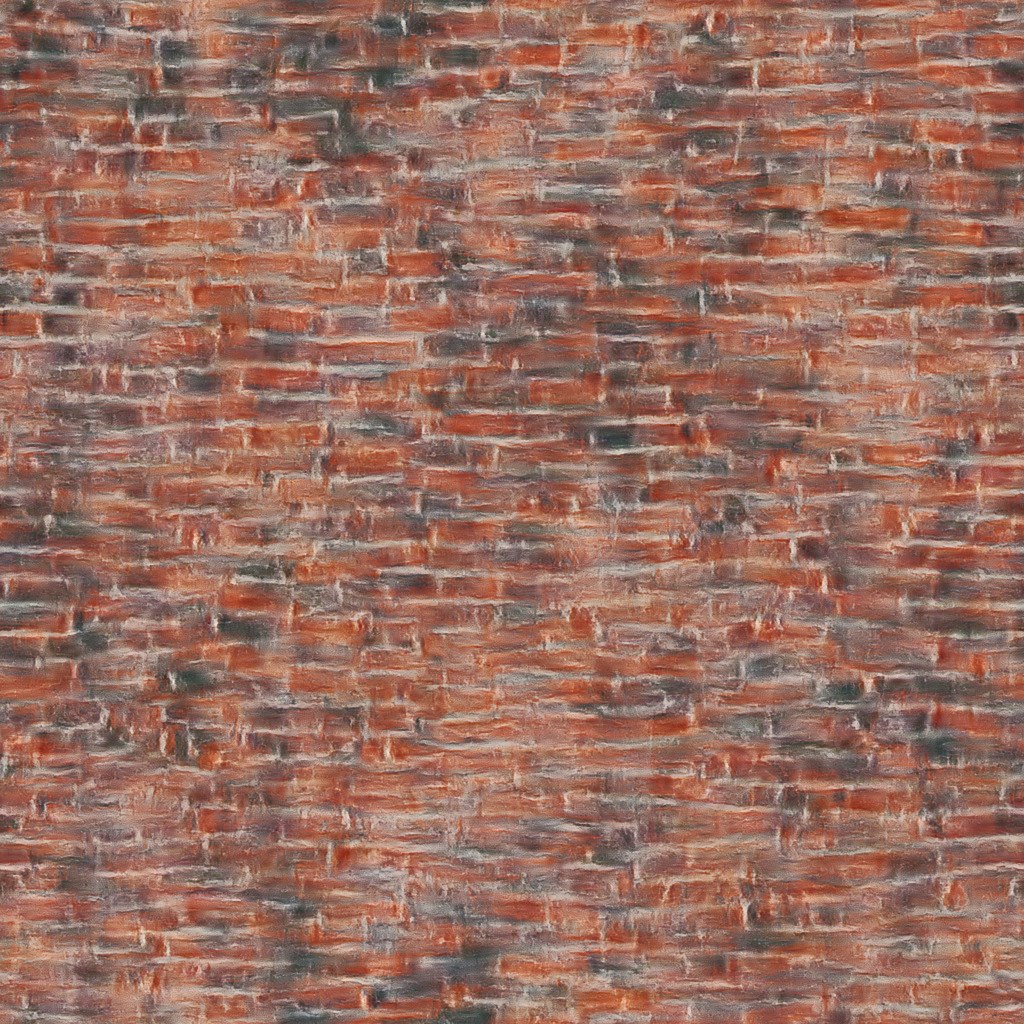}};
    \node [anchor=north east, rounded corners, fill=white, opacity=.5, text opacity=1] at (output2.north east) {\footnotesize output};
  \end{tikzpicture}
  \caption{Synthesis results of the Portilla and Simoncelli method \cite{PS}. It is satisfactory for many small grain textures (left) but may miss the global structure (right).}
  \label{fig:PS-examples}
\end{figure}
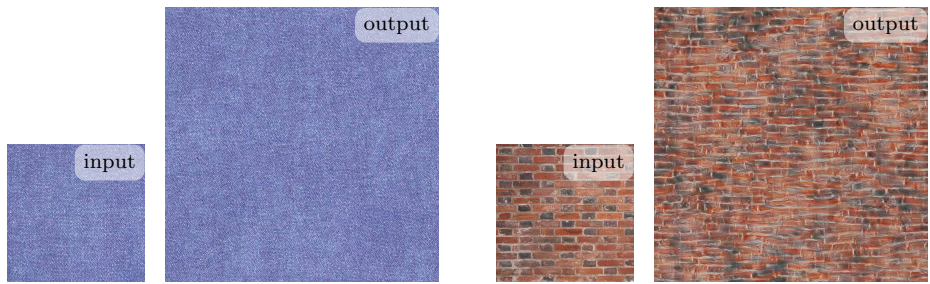

In a pre-attentive examination, the results are in general indistinct from the original  texture samples. Nevertheless, on attentive examination the synthesis of structured textures often present blurry and jammed results. Long range structures are missed and the method tends to homogenize the output texture. Figure~\ref{fig:PS-examples} shows two synthesis results. The first example (left in Figure~\ref{fig:PS-examples}) represents a quasi-periodic image where the method yields excellent results although it contains some global structures. In the second example (right in Figure~\ref{fig:PS-examples}), even though we recognize the nature of the input sample, one can observe that strong structures are missing. It is impossible to recover the lined up tiles.

Increasing the number of orientations $Q$ will improve the results since more information is captured. However for $Q>4$ the improvement is hardly noticeable. The number of levels $S$ of the steerable pyramid is the most influential parameter. Depending on the nature of the texture, it will need to be increased to capture the details at all  scales. Once again, for microtextures this parameter is less influential. Finally, the size of the neighborhood $N_a$ used to compute the autocorrelation is important whenever the texture has periodic information.

As we will see in Section~\ref{sec:exp}, even though imperfect, the results are very impressive, as they succeed modeling most textures using a moderately large set of global statistics. This brings us to the following two questions. Is the set of  statistics considered enough to describe any kind of textures? Is the optimization step enough to enforce these statistics? Fifteen years later, Gatys~et~al.~\cite{gatys} proposed a texture synthesis method based on Convolutional Neural Networks (CNN) which can be seen as an extension of Portilla and Simoncelli's work, where the set of statistics used is much larger and unknown; also, the optimization is performed by the backpropagation method.

\subsection{Texture synthesis using CNN}\label{sec:cnn}

It is hard to define metrics to determine if two textures are similar or not according to human taste. Julesz' conjecture that humans cannot distinguish two textures with same second order statistics was invalidated. Yet this does not rule out a more general hypothesis, according to which  there is a set of low-level filters such that if two textures respect the same statistics for these filters, they are indistinguishable. Portilla and Simoncelli's approach \cite{PS} and Zhu, Wu, and Mumford's FRAME (Filters, Random field, And Maximum Entropy) \cite{ZhuMumford,zhu98} can be seen as fixing a set of hand-picked filters and synthesing new textures by enforcing the response to the filters to have similar statistics. The set of filters is chosen to match human expectations about textures. However determining the exact set of filters equivalent to human vision is very hard, and both approaches use only a subset of them. Portilla and Simoncelli achieve similar statistics by iterating specific projections, starting from white noise, while FRAME achieves that with a Gibbs Sampler and some simplifications (quantizing the image intensities, etc). Recently, Convolutional Neural Networks (CNNs) have given a breath of fresh air to these approaches. CNNs are compositions of layers of convolutions, non-linearities and pooling. In the past few years, CNNs have been successfully applied in a wide variety of domains, in particular in image related tasks. Arguably, the win by a large margin of CNNs \cite{krizhevsky2012imagenet} in the 2012 ILSVRC challenge \cite{ILSVRC15}, an image classification challenge, helped spark interest of the global community to these methods.  We refer the reader to the corresponding literature for more details on the working of CNNs.

By taking a fully trained CNN on some visual classification task, and restricting to lower layers, one gets a set of low level filters which can directly be used for synthesizing texture, as shown in several works. The topic is quite active recently, and the question ``how to best synthesize a texture with the help of neural networks" is far from being solved. In the following, we will focus on two different approaches: Gatys' texture generator \cite{gatys} and DeepFrame \cite{deepframe}. Gatys' approach is to minimize the distance between the Gram matrices defined by the local filter responses of the network layers, while DeepFrame generates textures by sampling from an exponential model. The use of CNNs by these new approaches solves the issues of their ancestors: first the filters do not need to be handpicked anymore, they are encoded directly by the CNN. A pre-trained CNN successful on some image-related tasks can be selected for the texture generation. The choice of the CNN and whether it is pre-trained or the weights are random, affect the result. Second, the architecture of Neural Networks eases the generation process. The statistics of all the filters can be handled at the same time, via backpropagation for example. DeepFrame needs no quantization, unlike its predecessor, and synthesizes textures at a faster speed. Because the filter responses at a given Neural Network layer also encode the image content, texture transfer -- also named style transfer -- can be achieved by applying the statistics of the filter responses of a source image to a target image while keeping overall the filter responses similar \cite{gatys2016image}. While initially both Gatys' texture generator and DeepFrame used the VGG network \cite{simonyan2014very} trained on ImageNet \cite{deng2009imagenet}, more recent work obtained good results with networks with random weights \cite{gatysrecent} or by integrating the network training with the generation process \cite{deepframemultilayer}. {The success of VGG for texture generation seems to stem from its training on an object classification task. This implies that its trained features are valuable ``textons'' able to discriminate shape and object  features. One could imagine using a network trained to distinguish textures directly, instead of VGG.  But, to the best of our knowledge, no network has been trained on an ImageNet equivalent to textures.}

We now take a closer look at Gatys' texture generator and at DeepFrame. Gatys' texture model is a generalization of Julesz' model. It postulates that textures are described by the correlations between the neural network activations (features). Thus, by starting from random noise and imposing the correlations between the features to be the same as for a given input texture, one should get a new sample of this texture.

More precisely, Gatys's texture generator seeks to minimize the cost
$$
  E = \sum_l w_l ||G^l - T^l||_F^2
$$
where $||.||_F$ is the Frobenius norm, $w_l$ are weights and $G^l, T^l$ are the Gram matrices, respectively for the image and the target texture, of the feature maps of a pretrained neural network at a layer $l$. In \cite{gatys}, a custom 19-layer VGG network was used where max pooling was replaced by average pooling and the network weights were rescaled. Let $N_l$ be the number of feature maps at layer $l$ (this usually corresponds to the number of ``channels"), and $M_l$ the size of each feature map at layer $l$ ($M_l \times N_l$ is the number of outputs of layer $l$). If we denote by $F_{ij}^l, i\in\{1\cdots, N_l\}, j\in \{1\cdots, M_l\},$ the $j$-th output with the $i$-th feature map at layer $l$, then
$$
   \left(G^l\right)_{ij} = \frac{1}{M_l} \sum_{k=0 }^{M_l} F_{ik}^l F_{jk}^l.
$$
The texture generator minimizes the cost via backpropagation in the network, and thus falls into a local minimum. Starting from white noise, several thousand iterations can be needed to reach visual convergence. While in \cite{gatys} the features were extracted from VGG \cite{simonyan2014very}, a Deep Convolutional Neural Network trained on image classification tasks, in \cite{gatysrecent} it is noted that taking a pre-trained network is not necessary and a network with random weights can give satisfying results. The minimization of $E$ is done with L-BFGS-B \cite{zhu1994lbfgs} and the bounds are set to the minima and maxima of the source texture. After convergence, the histogram of the source is enforced.

\begin{figure}[t]
  \centering
  \begin{tikzpicture}
    \node[anchor=south, inner sep=0] (input1) at (0,0) {\includegraphics[width = .23\textwidth]{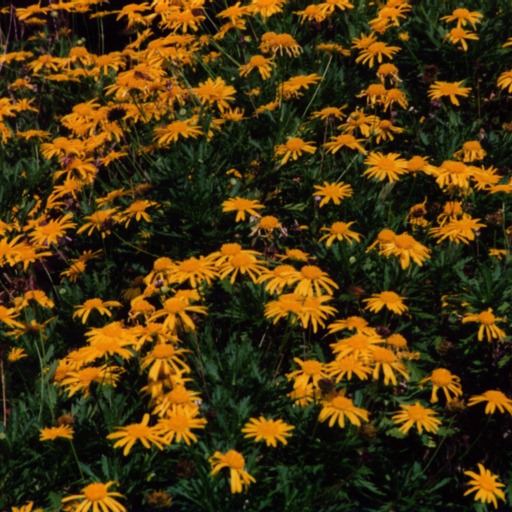}};
    \node [anchor=south] at (input1.north) {\footnotesize input};
    \node[anchor=south, inner sep=0] (output1) at (3.2,0) {\includegraphics[width = .23\textwidth]{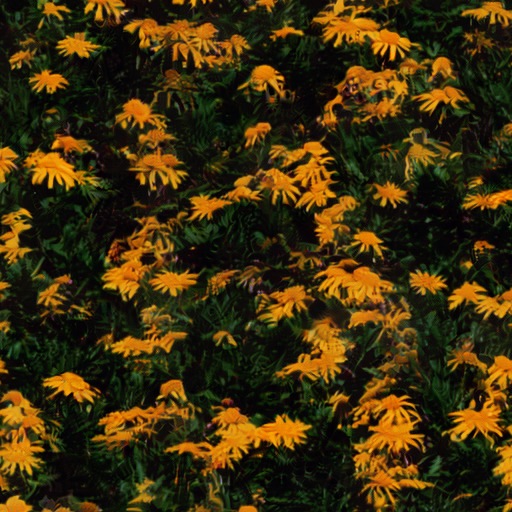}};
    \node [anchor=south] at (output1.north) {\footnotesize no padding + crop};
    \node[anchor=south, inner sep=0] (output2) at (6.4,0) {\includegraphics[width = .23\textwidth]{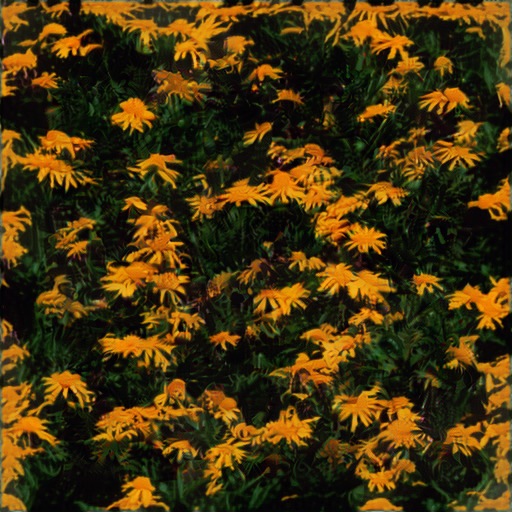}};
    \node [anchor=south] at (output2.north) {\footnotesize no padding};
    \node[anchor=south, inner sep=0] (output3) at (9.6,0) {\includegraphics[width = .23\textwidth]{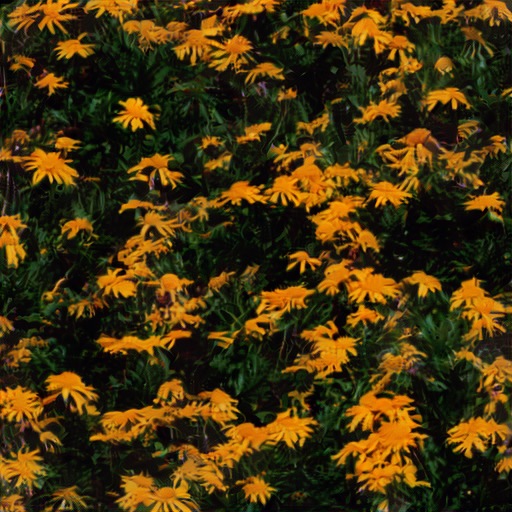}};
    \node [anchor=south] at (output3.north) {\footnotesize padding};
  \end{tikzpicture}
  \caption{This figure shows the impact of the padding in the neural network. The second image shows the result of a $1024\times1024$ generated texture without the network padding, cropped to $512\times512$, while the figures on the right show $512\times512$ sized generated results without or with padding. The same random initialization was used for all three results (and cropped for the last two results). The differences are particularly visible on the border of the pictures, since it is where each variant imposes different statistics.}
  \label{fig:Gatys-examples}
\end{figure}
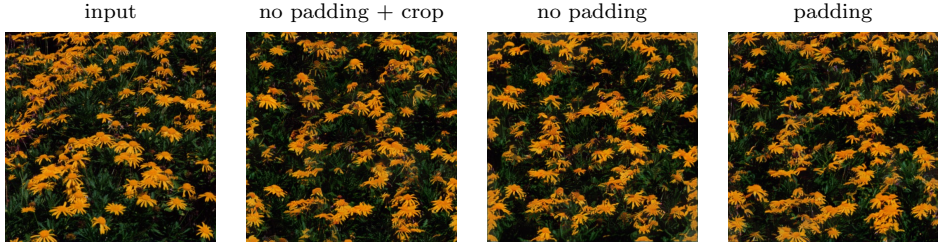

To generate the results in this article, we made a few changes compared to \cite{gatys}. The 19-layer VGG network used in \cite{gatys} pads the outputs at every convolution layer with zeros on each layer (to have the layer outputs be the same size as the layer inputs). That, plus the fact that pixels on the border are ``seen" by fewer features than the pixels in the center, means that all pixels on the image are not imposed the same distribution. If we take the same layers than in \cite{gatys} (\verb!conv1_1!, \verb!pool1!, \verb!pool2!, \verb!pool3!, \verb!pool4!) the top layer's outputs (\verb!pool4!) depend each on a $124 \times 124$ area of the source. Thus $123$ pixels should be removed on each border in order to have all remaining pixels seen by the same number of features. Removing 123 pixels on each border is not sufficient however to get the same constraints on the border and the center since the neighbouring pixels affect the features, and those neighbouring pixels are not affected by the same features. Thus to generate the results in this article, we decided to both remove the padding and generate bigger images -- 256 pixels more on each border -- which we then crop. The impact of this change can be seen on Figure~\ref{fig:Gatys-examples}. Other than that, we took the same parameters. In  \cite{aittala2016reflectance} the method solves the same problem by removing the network padding and enforcing periodicity. With the default network and parameters of Gatys' texture generator, except for the boundaries, a pixel is seen by $37504$ filters. In Gatys' method, textures are only described by the Gram matrices. The number of elements in the Gram matrices totals 352256, 176640 if we remove the redundant values (the matrices are symmetric). This number of parameters doesn't depend on the image size, and once the Gram matrices of the source computed, the output texture can be any size.

To fix some of the shortcomings of Gatys' texture generator \cite{gatys}, several works complete the objective function. The method in \cite{gatysamelioregousseau} incorporates spectrum constraints to significantly improve the generation of textures with low frequency patterns. In  \cite{berger2016incorporating} the proposed method considers  spatial co-occurences of features to help handling long-range consistency constraints. In \cite{wilmot2017stable} it is noticed that the Gram matrices have several particularities that decrease the quality of the texture obtained in several cases with instabilities, particularly visible when generating a texture with a   size different from the source. In our experiments we didn't notice such an instability problem, although we observed some instabilities (see for example the fourth column of figure \ref{fig:algComp1-1} and the first column of figure \ref{fig:algComp2-1}). It is possible  that the instabilities are affected by the parameter choice. To solve the instability problem, the authors added to the objective function a term to force the feature maps histograms to be the same as for the source. The authors of \cite{novak2016improving} also discussed some insufficiencies of Gram matrices in the case of style transfer, and in particular proposed to shift the activations to avoid sparsity. To accelerate the speed of the texture generation, the method of \cite{ulyanov2016texture} trains for a given texture a new CNN, which outputs new samples of the texture. The CNN is trained with the same objective function as for Gatys' texture generator. Once the CNN is trained, generation is fast.

DeepFrame's texture generator samples from an exponential model. The model is defined by the probability density function
$$
  f(u; w) = \frac{1}{Z(w)} \exp\left[\sum_{k=1}^K \sum_{x\in \Omega} w_k F_k(u)(x) \right] g(u),
$$
where $F_k$ corresponds to a filter map extracted from a CNN, $\Omega$ is the image domain of $u$ the image, $Z(w)$ is a normalizing constant and $g(u)$ is a reference distribution, like
$$
  g(u) = \frac{1}{(2\pi \sigma^2)^{|\Omega|/2}} \exp\left[ -\frac{1}{2\sigma^2} ||u||^2 \right].
$$
In contrast, the FRAME model defined the probability density function
$$
{ 
   f(u; \lambda ) = \frac{1}{Z(\lambda)} \exp\left[\sum_{k=1}^K \sum_{x\in \Omega} \lambda_k [F_k * u(x)] \right]
}
$$
where the $(F_k)_{k=1..K}$ were kernels, such as Gabor filters, or Difference of Gaussian filters, and $\lambda_k$ was a discretization function with finite number of possible outputs.

In a first phase, the DeepFrame parameters $w=(w_k)$ are tuned for the source texture, then in a second phase new samples of the texture are generated via Langevin dynamics. While in \cite{deepframe} a pre-trained network is used, in the method of \cite{deepframemultilayer}  its own network is trained on the source.

While both Gatys' texture generator and DeepFrame have a fixed texture model used to generate new samples, for which they learn parameters, a third successful CNN method to synthesize texture learns directly its model: in \cite{jetchev2016texture} a generative CNN is trained to synthesize new images from one or several samples of a source. The training is based on the adversarial model: a discriminator tries to distinguish the fake generated samples from true ones, while a generator creates new samples. Spatial invariance assumptions are encoded in the networks, but else, the texture model is in some sense learned by the two networks. This method can still be considered as a statistics-based method, because in some sense the discriminator checks the statistics of the texture are correct. To generate samples with this method (``SGAN" for Spatial Generative Adversarial Networks), we took the default network parameters, and applied the source histogram. We stopped after a few hundred epochs. The outputs suffer from a sort of noise pattern, which changes after every epoch. When the noise pattern was too important, we decided to select among the last twenty epochs the generator's result with the less noise. SGAN is a recent method, and there are certainly ways to better select the parameters and reduce this noise, but this goes beyond our goals here. {Recently a new extension called PSGAN (for Periodic Spatial Generative Adversarial Networks)~\cite{bergmann2017learning} was introduced to fix some shortcomings of SGAN, in particular to improve the generation result for textures with periodic patterns.}

{CNNs are also successful in the synthesis of images more general than textures~\cite{oord2016pixel,NIPS2016_6527,salimans2017pixelcnn++,dosovitskiy2016generating}, in particular with methods relying on Generative Adversarial Networks (GAN)~\cite{goodfellow2014generative,denton2015deep,radford2015unsupervised,im2016generating,wang2016generative}, but these methods are out of the scope of this study, which focuses on synthesizing new texture samples based on a single reference sample. These methods generally need a database of images.}

\begin{figure}[t]
  \centering
  \begin{tikzpicture}
    \node[anchor=south, inner sep=0] (input1) at (0,0) {\includegraphics[width = .23\textwidth]{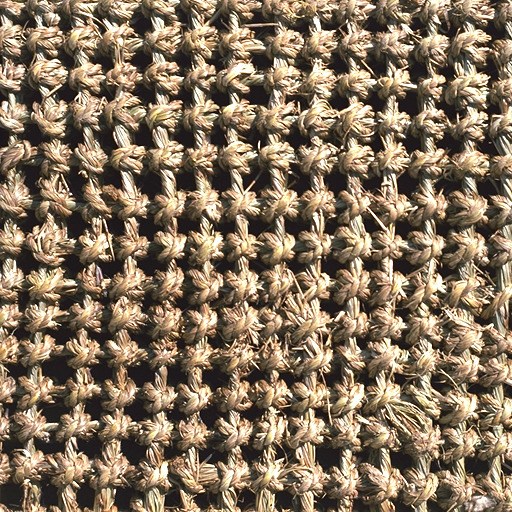}};
    \node [anchor=north east, rounded corners, fill=white, opacity=.5, text opacity=1] at (input1.north east) {\footnotesize input};
    \node[anchor=south, inner sep=0] (output1) at (3.2,0) {\includegraphics[width = .23\textwidth]{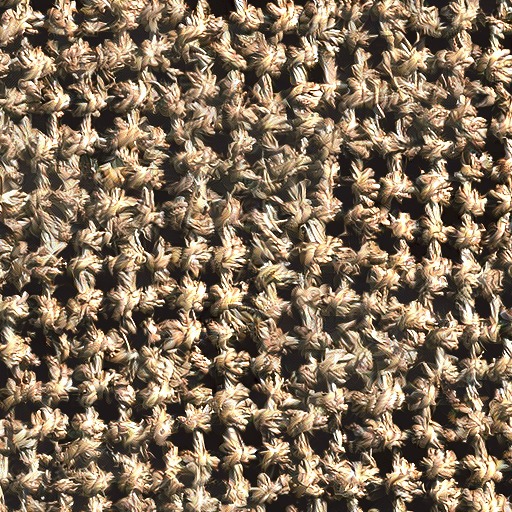}};
    \node [anchor=north east, rounded corners, fill=white, opacity=.5, text opacity=1] at (output1.north east) {\footnotesize Gatys};
    \node[anchor=south, inner sep=0] (output2) at (6.4,0) {\includegraphics[width = .23\textwidth]{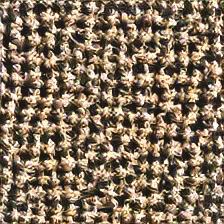}};
    \node [anchor=north east, rounded corners, fill=white, opacity=.5, text opacity=1] at (output2.north east) {\footnotesize DeepFrame};
    \node[anchor=south, inner sep=0] (output3) at (9.6,0) {\includegraphics[width = .23\textwidth]{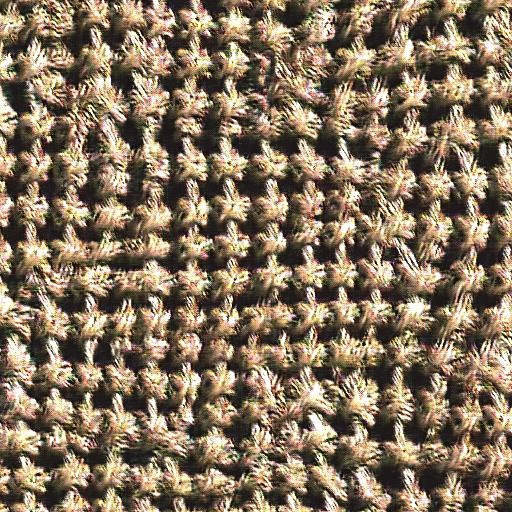}};
    \node [anchor=north east, rounded corners, fill=white, opacity=.5, text opacity=1] at (output3.north east) {\footnotesize SGAN};

    \node[anchor=south, inner sep=0] (input1) at (0,3.6) {\includegraphics[width = .23\textwidth]{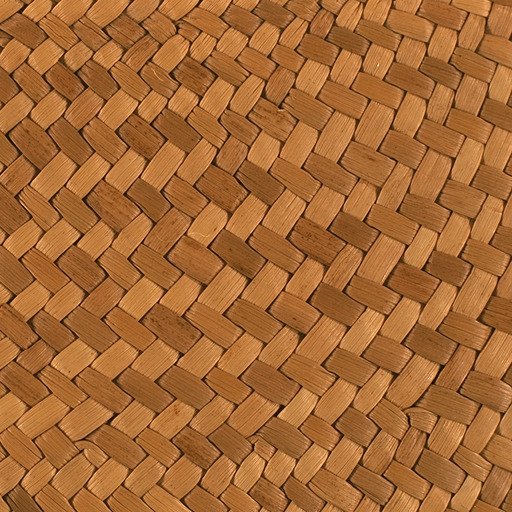}};
    \node [anchor=north east, rounded corners, fill=white, opacity=.5, text opacity=1] at (input1.north east) {\footnotesize input};
    \node[anchor=south, inner sep=0] (output1) at (3.2,3.6) {\includegraphics[width = .23\textwidth]{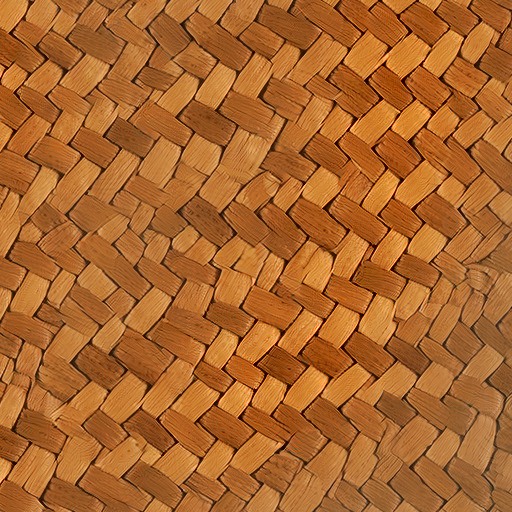}};
    \node [anchor=north east, rounded corners, fill=white, opacity=.5, text opacity=1] at (output1.north east) {\footnotesize Gatys};
    \node[anchor=south, inner sep=0] (output2) at (6.4,3.6) {\includegraphics[width = .23\textwidth]{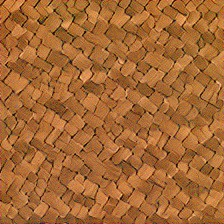}};
    \node [anchor=north east, rounded corners, fill=white, opacity=.5, text opacity=1] at (output2.north east) {\footnotesize DeepFrame};
    \node[anchor=south, inner sep=0] (output3) at (9.6,3.6) {\includegraphics[width = .23\textwidth]{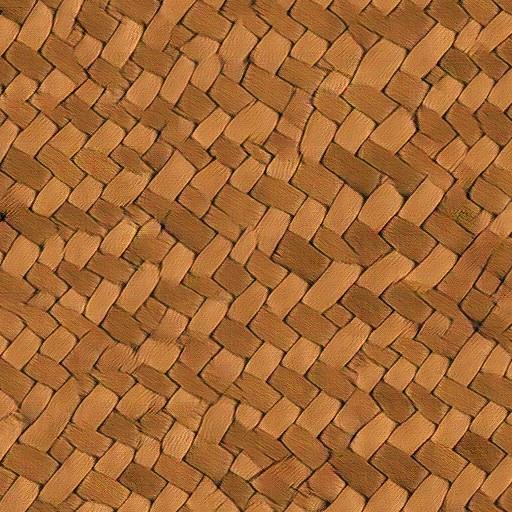}};
    \node [anchor=north east, rounded corners, fill=white, opacity=.5, text opacity=1] at (output3.north east) {\footnotesize SGAN};
  \end{tikzpicture}
  \caption{Comparison between Gatys' texture generator \cite{gatys}, DeepFrame \cite{deepframe} and SGAN \cite{jetchev2016texture}. For all three methods, we used the default parameters, except that in the case of Gatys we used the method we described above where we remove the network padding and crop the result and in case of DeepFrame and SGAN, we specified the result's histogram on the source histogram. Overall, SGAN looks the best when looking from far, but when zoomed in, Gatys seems to respect the best the local structures.}
  \label{fig:CNN-examples}
\end{figure}
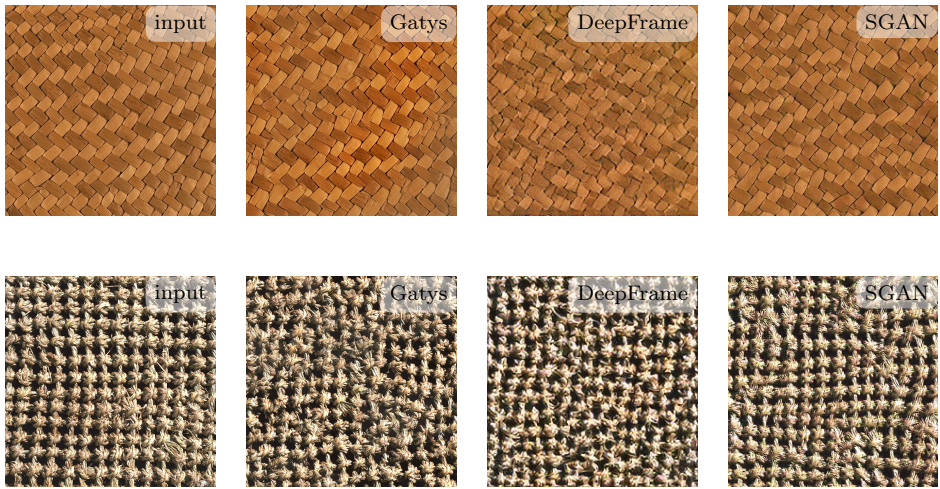

\section{Patch re-arrangement methods}\label{sec:non-parametric}

In contrast to the statistics-based methods, the patch re-arrangement methods do not attempt to characterize textures by a statistical model. Spanning from the groundbreaking work by Efros and Leung~\cite{EfrosLeung}, this family of algorithms consists of clever heuristics to re-arrange parts of the sample texture in a random way in order to create a new texture. By copying directly from the sample image, these methods often are able to keep complex structures from the input. By the same token, the process is frequently limited to copying and the results show little innovation relative to the sample. We will illustrate the family here by the original Efros and Leung~\cite{EfrosLeung} algorithm, a further extension by Efros and Freeman~\cite{EfrosFreeman} which incorporate more recent techniques, and a more recent CNN based method~\cite{li2016combining}.

\subsection{The Efros and Leung algorithm}\label{sec:el}

In his foundational paper of information theory~\cite{shannon}, Claude E. Shannon proposed to approximate the information contents of natural languages by the entropy of generative stochastic processes. He used a Markov chain to generate English text sequentially, letter by letter. Given a piece of already generated text, the next letter is sampled from the probability distribution of English text conditioned to the previous $n$ letters. The following sequence was generated by Shannon using a third-order model:

\begin{quote}
in no ist lat whey cratict froure birs grocid pondenome of demonstures of the reptagin is
regoactiona of cre
\end{quote}
Although very few words are real English words, this simple model produces surprisingly good English ``textures''. Inspired by Shannon's method, Efros and Leung~\cite{EfrosLeung} proposed to adapt the same ideas for image texture synthesis.

\begin{figure}[t]
  \centering
  \begin{tikzpicture}[scale=0.8]

  \centering

  \node[anchor=center, inner sep=0] (input) at (0,0) {\includegraphics[width=0.25\textwidth]{figures/simoncelli_10}};
  \draw[thick, fill=white, fill opacity=.0] (input.south west) rectangle (input.north east);
  \node [anchor=north east, rounded corners, fill=white, opacity=.4, text opacity=1] at (input.north east) {\footnotesize input};

  \draw[line width=0.6pt,dashed] (-0.3, -.55) rectangle (0.8,0.45);
  \draw[line width=0.6pt,dashed] (0, -0.4) rectangle (1,0.6);
  \draw[line width=0.6pt,dashed] (-1.6, 0.45) rectangle (-0.6,1.45);
  \draw[line width=0.6pt,dashed] (0.2, -1.05) rectangle (1.2,-0.05);
  \draw[line width=0.6pt,dashed] (-1.6, -1.7) rectangle (-0.6,-0.7);
  \draw[line width=0.6pt,dashed] (-1.4, 0.65) rectangle (-0.4,1.65);
  \draw[line width=1pt,yellow] (-1.75, -0.35) rectangle (-0.75,0.65);

  \node[anchor=north west, inner sep=0] (patch) at (3.5,0.5) {\includegraphics[width=0.066\textwidth]{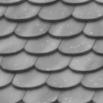}};
  \draw[line width=1pt,yellow, fill=white, fill opacity=.0] (patch.south west) rectangle (patch.north east);

  \coordinate (a) at (4,0);
  \draw (a);
  \fill[yellow,opacity=.6] (a) circle (1pt);

  \draw[thick,->] (input.east) -- node[midway,above] {\normalsize{(1)}} (3.5,0);
  \draw[thick,->] (4.5,0) -- node[midway,above] {\normalsize{(2)}} (6,0);

  \node[anchor=west, inner sep=0] (output) at (6,0) {\includegraphics[width=0.375\textwidth]{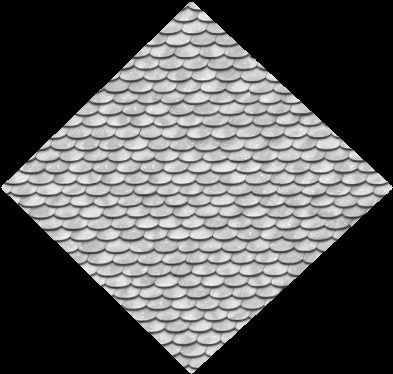}};
  \draw[thick, fill=white, fill opacity=.0] (output.south west) rectangle (output.north east);
  \node [anchor=north east, rounded corners, fill=white, opacity=.4, text opacity=1] at (output.north east) {\footnotesize output};

  \draw[line width=1pt,red] (6.95,0.9) rectangle (7.95,1.9);

  \coordinate (a) at (7.45,1.4);
  \draw (a);
  \fill[red,opacity=.6] (a) circle (1pt);

\end{tikzpicture}
\caption{Overview of the Efros and Leung algorithm \cite{EfrosLeung}. Given a texture image (left) a new image (right) is being synthesized a pixel at a time. For a pixel $(m,n)$ (red point in the output) being synthesized the method finds all neighbourhoods in the left image that match the neighbourhood of $(m,n)$ (dashed squares) and then chooses randomly one of the neighbourhoods (yellow square) and assigns its central pixel value to $(m,n)$.}
\label{fig:EL-overview}
\end{figure}

Efros and Leung in \cite{EfrosLeung} synthesize a new texture image by considering that a pixel value depends on the values of its neighbouring pixels. The method is illustrated in Figure \ref{fig:EL-overview} and works as follows. For a given input texture, a new image is synthesized sequentially, pixel by pixel. For a pixel $(m,n)$ being synthesized, the algorithm finds all the neighbourhoods in the input image that are similar to the neighbourhood of $(m,n)$ up to a patch distance tolerance. Then one of these neighbourhoods is randomly chosen and its central pixel value is affected to the pixel $(m,n)$. The neighbourhood of $(m,n)$ is a square patch but only the known pixels (coming from the seed or already synthetized) of this patch are considered when comparing to the neighbourhoods of the input. Denoting $p_1$ and $p_2$ two patches of size $P\times P$, the comparison is made using a Gaussian-weighted distance defined as
\begin{equation}
  d\big(p_1,p_2\big) = \frac{1}{\sum_{i,j}G_{\sigma}(i,j)}\sum_{i,j}\Big(p_1(i,j)-p_2(i,j)\Big)^2G_\sigma (i,j),
\end{equation}
where $G_\sigma$ is a Gaussian kernel with standard deviation $\sigma$.

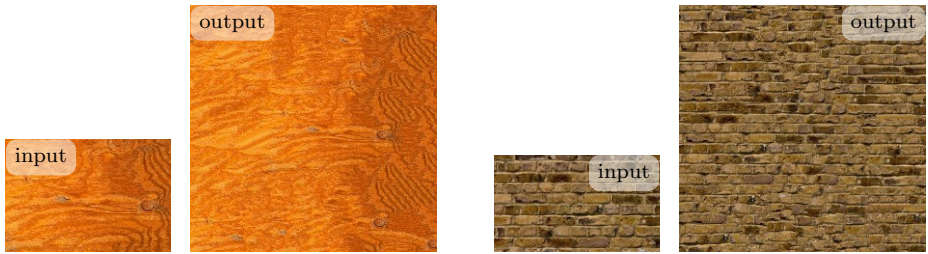
\begin{figure}[t]
  \centering
  \begin{tikzpicture}
    \node[anchor=south, inner sep=0] (input1) at (0,0) {\includegraphics[width = .18\textwidth]{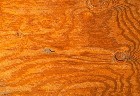}};
    \node [anchor=north west, rounded corners, fill=white, opacity=.5, text opacity=1] at (input1.north west) {\footnotesize input};
    \node[anchor=south, inner sep=0] (output1) at (3,0) {\includegraphics[width = .27\textwidth]{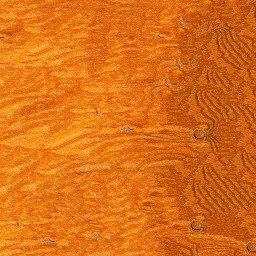}};
    \node [anchor=north west, rounded corners, fill=white, opacity=.5, text opacity=1] at (output1.north west) {\footnotesize output};

    \node[anchor=south, inner sep=0] (input2) at (6.5,0) {\includegraphics[width = .18\textwidth]{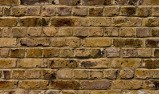}};
    \node [anchor=north east, rounded corners, fill=white, opacity=.5, text opacity=1] at (input2.north east) {\footnotesize input};
    \node[anchor=south, inner sep=0] (output2) at (9.5,0) {\includegraphics[width = .27\textwidth]{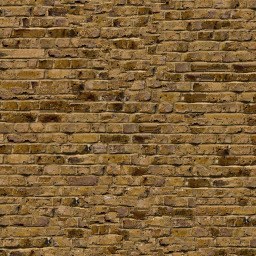}};
    \node [anchor=north east, rounded corners, fill=white, opacity=.5, text opacity=1] at (output2.north east) {\footnotesize output};
  \end{tikzpicture}
  \caption{Synthesis results of the Efros and Leung method \cite{EfrosLeung}. Left: the example shows the garbage growing effect. Right: the example shows the strength of this method to synthesize macrotextures. The patch size used for both synthesis is $P=40$.}
  \label{fig:EL-examples}
\end{figure}

{Levina and Bickel in~\cite{levina-bickel} provided a theoretical justification of Efros and Leung's work. The Efros and Leung algorithm is based on resampling from the random field directly, without constructing an explicit model for the distribution. The authors of \cite{levina-bickel} formalized this algorithm in the framework of resampling from random fields and proved that it provides consistent estimates of the joint distribution of pixels in a window of specified size.}

\begin{figure}[t]
  \centering
  \includegraphics[width=.16\textwidth]{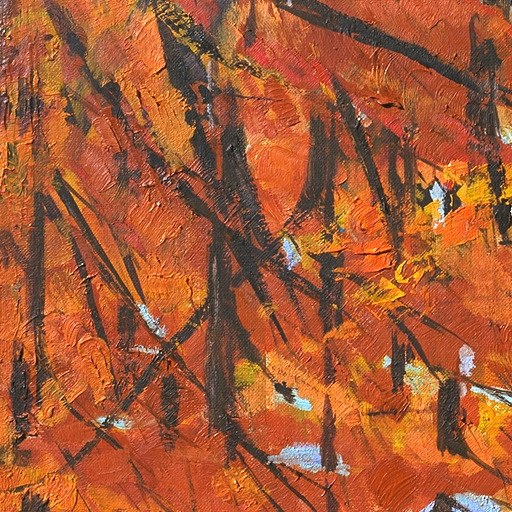}%
  \hfill
  \includegraphics[width=.16\textwidth]{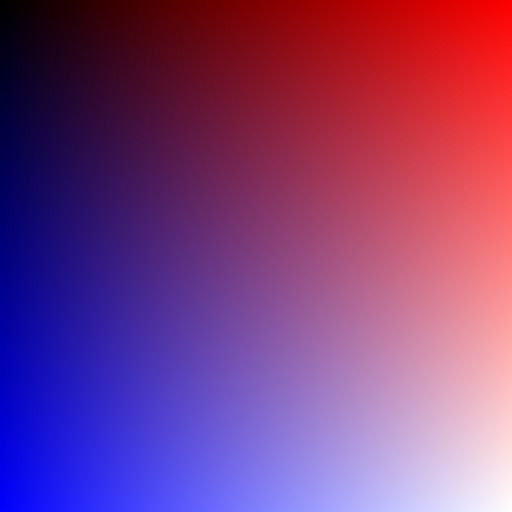}%
  \hfill%
  \includegraphics[width=.32\textwidth]{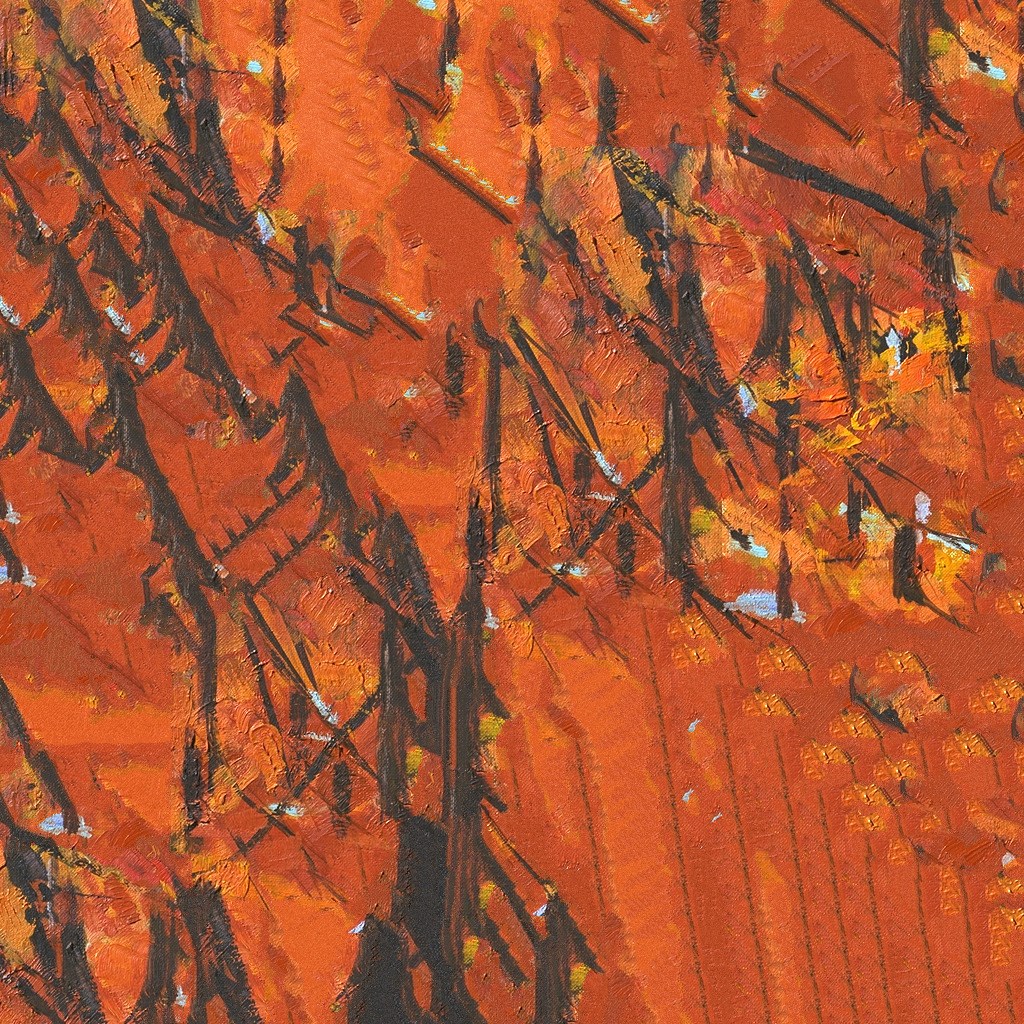}%
  \hfill%
  \includegraphics[width=.32\textwidth]{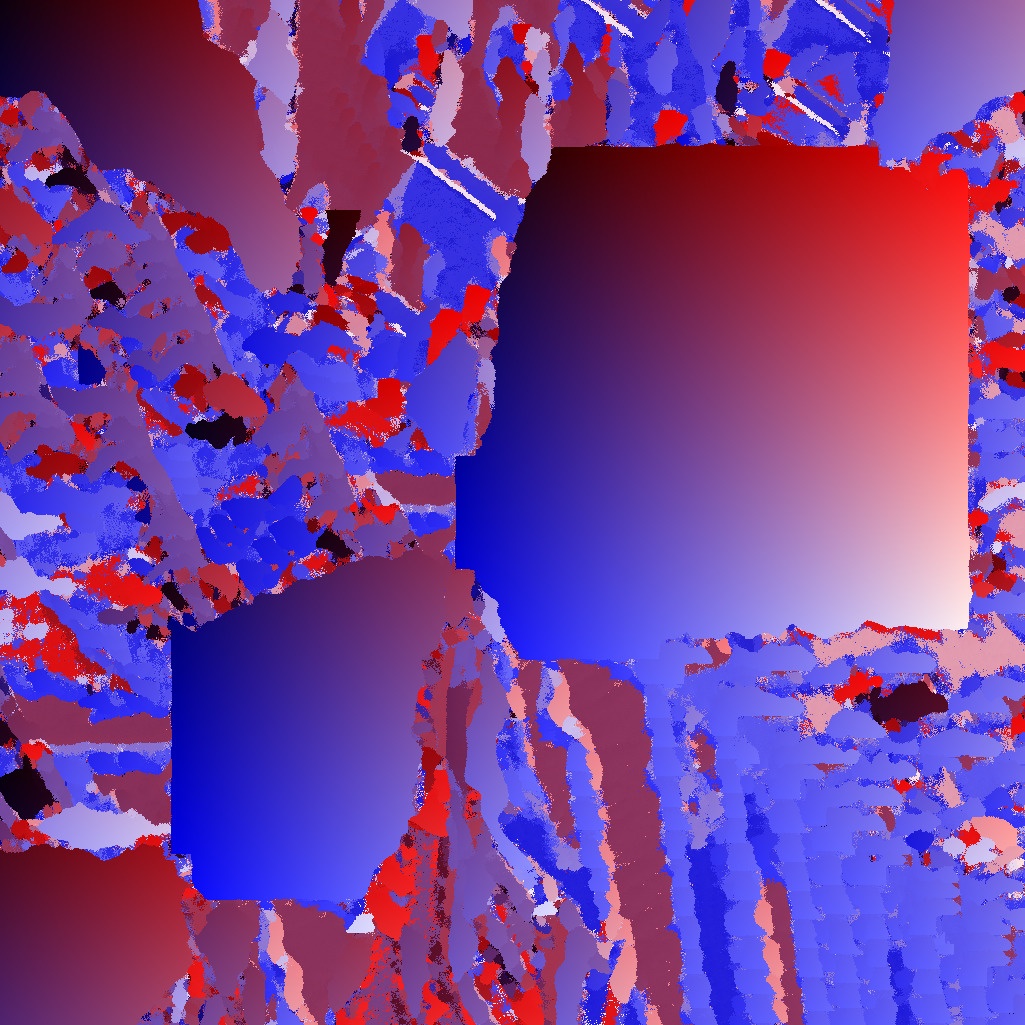}%
  \caption{From left to right: texture sample, position map, synthesized image and synthesis map. The synthesis map shows for each synthesized patch its initial position in the texture sample. It allows then to identify exactly the verbatim copy regions (they correspond to continuous color areas of the map). This method reveals  the verbatim copies of the input in the generated texture and the repetitions (garbage).}
  \label{fig:synthesis_map}
\end{figure}

In general the visual results are very impressive, especially for structured textures. Nevertheless this algorithm suffers from two important drawbacks: verbatim copies of the input and garbage growing (the algorithm starts reproducing iteratively one part of the example and neglects the rest). Figure~\ref{fig:EL-examples} shows two synthesis examples. The first synthesis result illustrates a failure case. In particular one can observe the effect of garbage growing, which reproduces incoherently the right side of the wood sample texture. The second example shows the strength of this method when it comes to synthesize textures with conspicuous patterns as in this case the brick patterns. To illustrate the verbatim-copy regions, \emph{position and synthesis maps} are used to visualize from which regions of the input texture each synthetized pixel comes from. A synthesis and the corresponding map are shown in  Figure~\ref{fig:synthesis_map} (obtained with the online demo \cite{EfrosLeungIpol}). Large continuous zones are identified in the synthesis maps which corresponds to the verbatim copies produced by the method. This representation also shows that the synthesized image is indeed a re-arrangement of pieces of the input sample. 

Increasing the patch size $P$ results in increasing the verbatim copied regions. However if the patch size is too small the local aspect of this method fails in recovering the global configuration of the input texture in particular for macrotextures. A second parameter of the method is the tolerance parameter $\varepsilon$ which is used to select the most similar patches in the input image. Large tolerance values increase the garbage growing effect.

The Efros and Leung method also suffers from its high computational complexity. Several optimizations have been proposed to accelerate this algorithm. Among them Wei and Levoy \cite{WeiLevoy} managed to fix  the shape and  size of the learning patch and Ashikhmin \cite{ashikhmin} proposed to extend existing patches whenever possible instead of searching in the entire sample texture. The following section describes a particularly important extension of the method.

\subsection{The Efros and Freeman algorithm}\label{sec:ef}

\begin{figure}[t]
  \centering
  \begin{minipage}{\linewidth}
    \begin{minipage}{0.32\linewidth}
      \centering
      {\it{Vertical overlap}}
      \includegraphics[width=\textwidth]{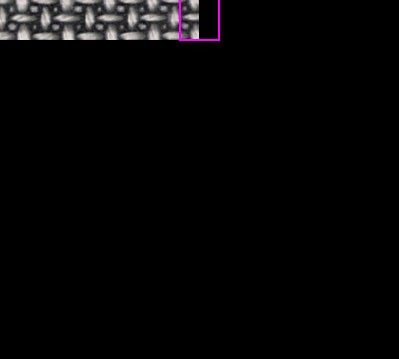}
      \bf{Iteration $\bf10$}
    \end{minipage}%
      \hfill%
    \begin{minipage}{0.32\linewidth}
      \centering
      {\it{Horizontal overlap}}
      \includegraphics[width=\textwidth]{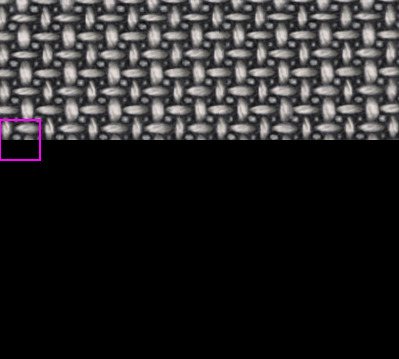}
      \bf{Iteration $\bf115$}
    \end{minipage}%
      \hfill%
    \begin{minipage}{0.32\linewidth}
      \centering
      {\it{L-shape overlap}}
      \includegraphics[width=\textwidth]{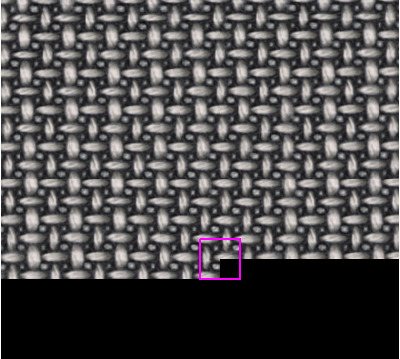}
      \bf{Iteration $\bf239$}
    \end{minipage}
  \end{minipage}
  \caption{Three different iterations of the synthesis process are shown. At each iteration a patch is being synthesized. This patch is represented by the pink square in the three iterations shown. From left to right the three overlap cases are represented: vertical, horizontal and L-shape.}
  \label{fig:intro:patchModel}
\end{figure}

Efros and Freeman's method \cite{EfrosFreeman} is an extension of Efros and Leung's. It is based on the same principle where the pixel values are conditioned to their neighbourhood values. Efros and Freeman proposed to generate a new image sequentially, patch by patch (instead of pixel by pixel) in a raster scan order as illustrated in Figure~\ref{fig:intro:patchModel}. At each step a patch that is only partially defined on a region called \emph{overlap region} is completed. This overlap region is of width $w_o$. This is the patch under construction. To do so a patch of the input image among those who match the patch under construction on its overlap region is randomly selected (\emph{patch selection} step). An optimal boundary cut between the chosen patch ($p_{\text{in}}$) and the one under construction ($p_{\text{old}}$) is then computed across the overlap region (\emph{stitching} step). This optimal boundary cut is used to construct a new patch ($p_{\text{new}}$) by blending the ($p_{\text{in}}$) and ($p_{\text{old}}$) along the cut. There are three possible overlap regions: vertical overlap for the first row, horizontal overlap for the first column, and L-shaped overlap everywhere else (Figure~\ref{fig:intro:patchModel}).

In the \emph{patch selection} step, to select a patch $p_\text{in}$ of an input image $u$ one computes the square distance between the overlap region of the patch $p_\text{old}$ and the corresponding regions of all the patches of $u$. The minimal distance $D_\text{min}$ is determined and $p_\text{in}$ is randomly picked among all patches whose distance to $p_\text{old}$ is lower than $(1+\varepsilon)D_\text{min}$ where $\varepsilon$ is the tolerance parameter. The squared distance image $d$ contains at each position $(m,n)$ the distance between $p_\text{old}$ and the patch from $u$ according to some binary weight $t$ that equals one in the overlap region and zero otherwise. More precisely, one has
\begin{equation}
  d(m,n) = \sum_{i,j} t(i,j)(p_\text{old}(i,j) - u(m+i,n+j))^2.
\label{eq:distance}
\end{equation}

The patch $p_\text{in}$ of $u$ having coordinates $(m,n)$ is similar to the partially defined patch $p_\text{old}$ on their overlap region. To get the final patch $p_{\text{new}}$ one must combine the patches $p_\text{old}$ and $p_\text{in}$. Denoting $t$ the binary weight for the overlap regions as in~\eqref{eq:distance}, then, for any binary image $r$ such that $0\leq r(i,j)\leq t(i,j),~(i,j)\in\{1,\dots,P\}^2$, $P$ can be defined as the combination
\begin{equation*}
p_{\text{new}} = t\, p_{\text{old}} + \left(1-t\right)p_{\text{in}}.
\end{equation*}
The main contribution of Efros and Freeman~\cite{EfrosFreeman} is to look for a binary shape $M$ where the transition between $p_{\text{old}}$ and $p_{\text{new}}$ along the boundary of the shape is minimal. For simplicity, and to be able to use linear programming, the authors do not allow for any shape, but only for the ones whose boundaries are simple forward paths from one end to the other of the overlap region. This results in two pieces of image being  sewn together along some general boundary path, hence the algorithm's name ``quilting''.

\begin{figure}[t]
  \centering
  \begin{tikzpicture}
    \node[anchor=south, inner sep=0] (input1) at (0,0) {\includegraphics[width = .15\textwidth]{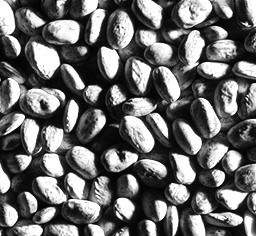}};
    \node [anchor=north east, rounded corners, fill=white, opacity=.5, text opacity=1] at (input1.north east) {\footnotesize input};
    \node[anchor=south, inner sep=0] (output1) at (3,0) {\includegraphics[width = .3\textwidth]{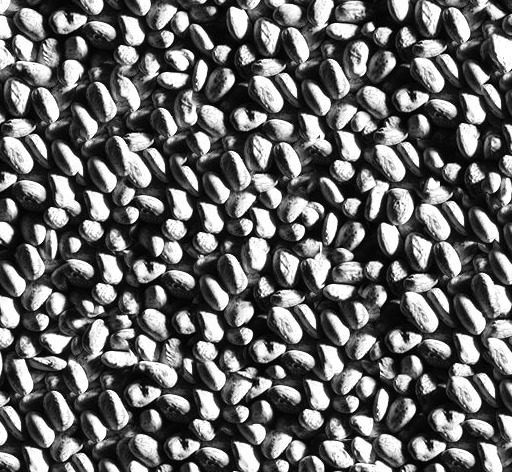}};
    \node [anchor=north east, rounded corners, fill=white, opacity=.5, text opacity=1] at (output1.north east) {\footnotesize output};

    \node[anchor=south, inner sep=0] (input2) at (6.5,0) {\includegraphics[width = .15\textwidth]{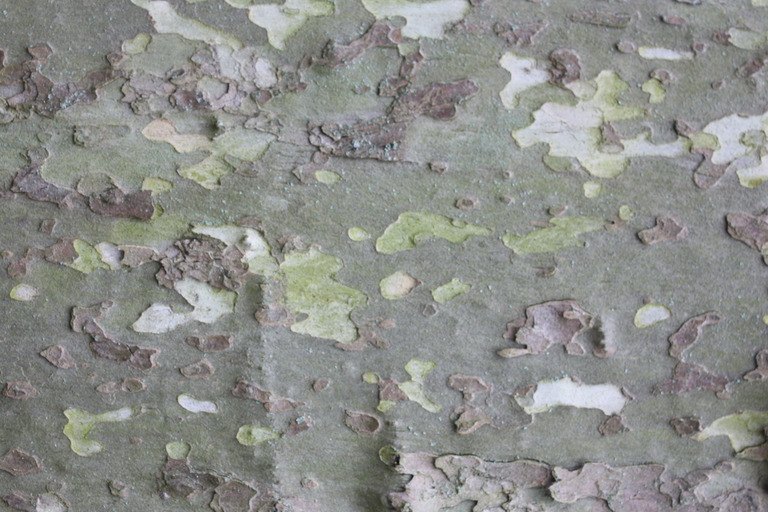}};
    \node [anchor=north east, rounded corners, fill=white, opacity=.5, text opacity=1] at (input2.north east) {\footnotesize input};
    \node[anchor=south, inner sep=0] (output2) at (9.5,0) {\includegraphics[width = .3\textwidth]{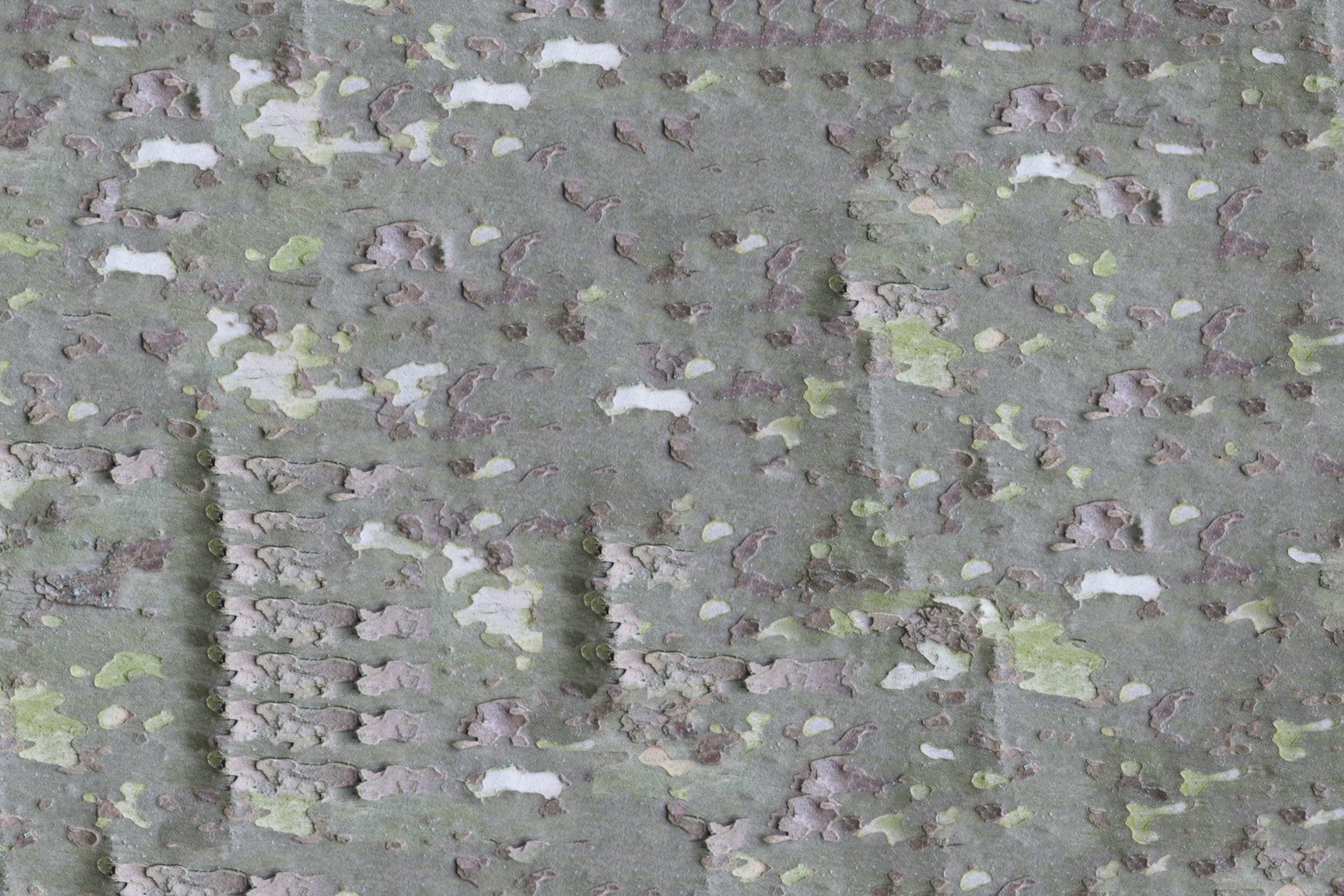}};
    \node [anchor=north east, rounded corners, fill=white, opacity=.5, text opacity=1] at (output2.north east) {\footnotesize output};
  \end{tikzpicture}
  \caption{Synthesis results of the Efros and Freeman method~\cite{EfrosFreeman}. It works for microtextures but risks losing the example's global statistics. It works for macrotextures too, but risks  verbatim copies. Two examples are shown: a success (left) and a failure (right). The parameters used for are $P=80$ and $O=P/4$.}
  \label{fig:EF-examples}
\end{figure}
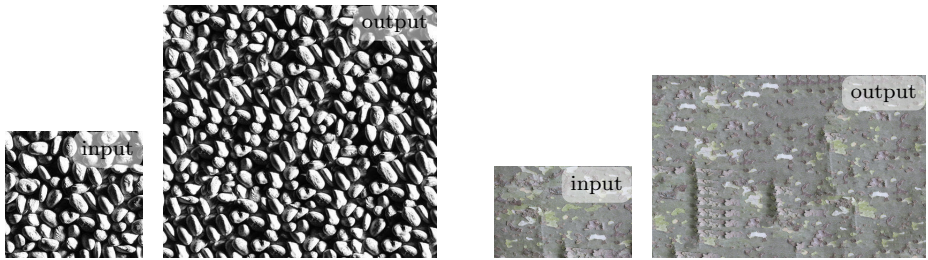

\begin{figure}[t]
  \centering
  \includegraphics[width=.16\textwidth]{figures/bark1004}%
  \hfill
  \includegraphics[width=.16\textwidth]{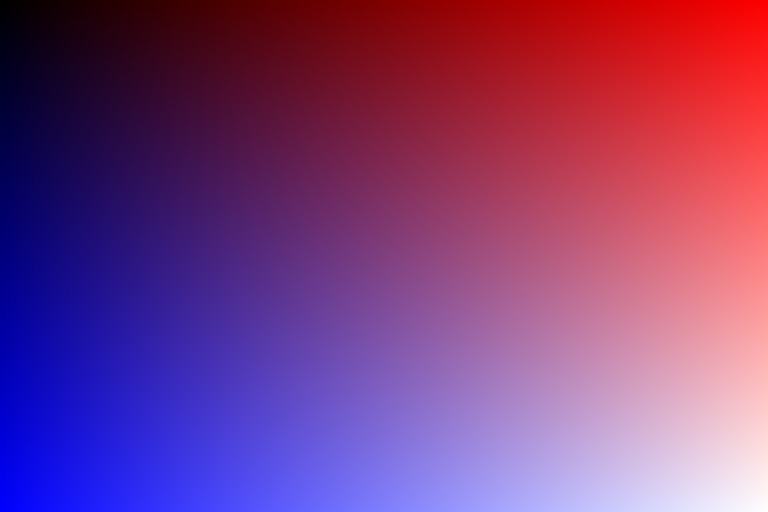}%
  \hfill%
  \includegraphics[width=.32\textwidth]{figures/bark1004_out_w80_r2}%
  \hfill%
  \includegraphics[width=.32\textwidth]{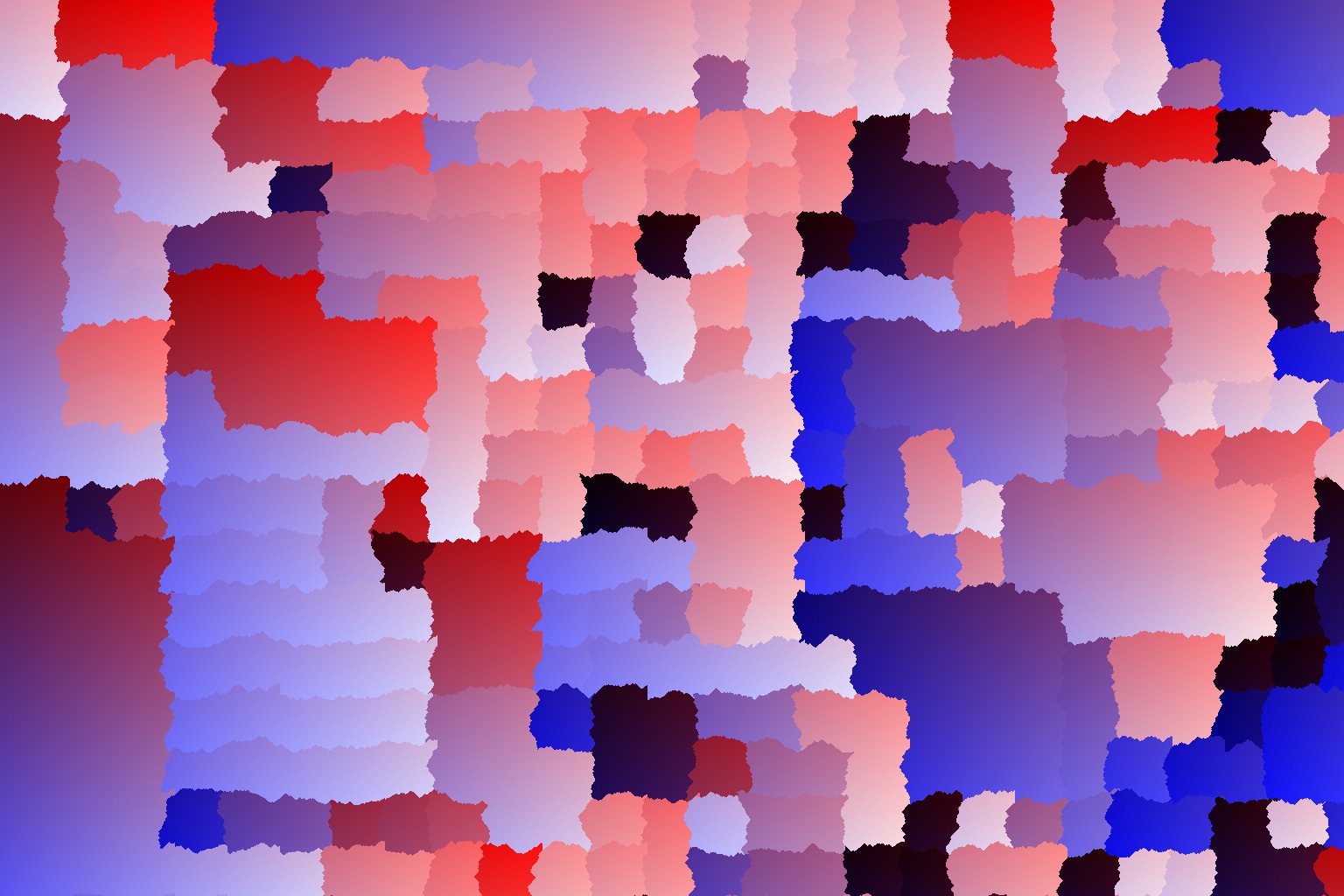}%
  \caption{From left to right: texture sample, position map, synthesized image and synthesis map. The synthesis map shows for each synthesized patch its initial position in the texture sample. It puts in evidence the verbatim copy regions (they correspond to continuous color areas of the map) and the repetitions (corresponding to repeated continuous patches of the same color).  }
  \label{fig:synthesis_map_EF}
\end{figure}

This method yields very impressive visual results, in particular for highly structured textures. In terms of speed the gain is truly significant with respect to the methods which synthesize an image pixel by pixel. The patch size being larger, the risk of garbage growing is reduced compared to the Efros-Leung algorithm. Nevertheless the risk of verbatim copies remains and is even amplified. Moreover, the respect of the global statistics of the input is not guaranteed and this is quite visible when the input texture is not stationary (for example if there is a change of illumination across the image). Figure~\ref{fig:EF-examples} shows two synthesis examples. The first one (left) shows an excellent synthesis result where the strong structures of the input are perfectly recovered. The second one (right) puts in evidence the verbatim copy of parts of the input and the garbage growing effect. To illustrate this the synthesis map of the second example is shown in Figure~\ref{fig:synthesis_map_EF}.

The  parameters $P$ and $\varepsilon$ play the same role as in Efros and Leung's method. A third parameter, the overlap size $O$ is used. Increasing this value tends to increase the verbatim copies of large regions. However if this value is too small then garbage growing increases. The value  $O = 0.25P$ is generally satisfactory.

{
\subsection{High level patch re-arrangement with Convolutional Neural Networks}

The CNNs texture synthesis methods presented in Section~\ref{sec:cnn} typically generated new texture samples by enforcing similar statistics on the feature maps of a pre-trained CNN. At the crossroad of patch re-arrangement methods and CNNs, lies CNNMRF~\cite{li2016combining}. What distinguishes this method to  those of Section~\ref{sec:cnn} is that the texture samples are generated by enforcing similar patches of feature maps on selected upper layers of a pre-trained CNN. The image is then obtained with backpropagation and a smoothness constraint.

More precisely, starting from random noise, an image is generated by minimizing the energy:
$$E = \sum_l \sum_i ||\psi_i(F^l) - \psi_{NN(i)}(F_s^l)||^2 + R$$

\noindent where $l$ goes among the selected layers (\verb!relu3_1! and \verb!relu4_1! of the VGG network~\cite{simonyan2014very}), $i$ goes among all the positions in the layer, $\psi_i(F^l)$ represents the patch at the $i$-th position and $\psi_{NN(i)}(F_s^l)$ is its best matching patch in the source according to the normalized cross-correlation. The default patch size is $3\times3$ times the number of feature maps (often referred as the number of ``channels'' of the layer). $R$ is a regularizer term to impose smoothness of the resulting image.
In~\cite{li2016combining}, the energy also contains a term to enforce the content of the source if doing texture transfer. This term isn't used for texture synthesis.

As noted by the author,  a  natural seamless patch blending is obtained by  performing a  patch re-arrangement on the levels of the CNN instead of doing it directly on the image, like in the other methods of this section.

Similarly to what was done in Section~\ref{sec:cnn}, we removed the padding of the VGG network to generate the results of this method. Indeed if the padding is kept, the spatial invariance assumption is violated. Moreover pixels on the border of the generated images are seen by fewer features, which reinforces the violation of the spatial invariance. Thus in addition to removing the network padding, we generated bigger images and then cropped the result. On figure~\ref{fig:CNNMRFexamples}, the generated texture with the network padding and no border crop kept tends to reproduce exactly significant parts of the input on the borders. This problem doesn't appear on the image with the padding removed and the border cropped. To generate the figures in Section~\ref{sec:exp} which features images of size $1024\times1024$, we couldn't add $256$ pixels more on each border, as was done in Section~\ref{sec:cnn}, due to memory constraints. Instead we generated images of size $1280\times1280$, which were then cropped.

\begin{figure}[t]
  \centering
  \begin{tikzpicture}
    \node[anchor=south, inner sep=0] (input1) at (0,0) {\includegraphics[width = .23\textwidth]{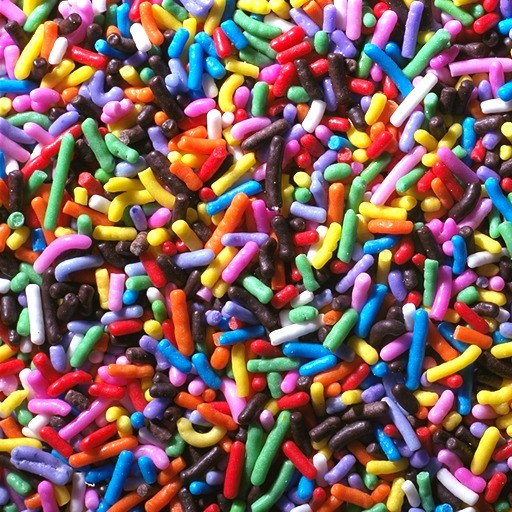}};
    \node [anchor=south] at (input1.north) {\footnotesize input};
    \node[anchor=south, inner sep=0] (output1) at (3.2,0) {\includegraphics[width = .23\textwidth]{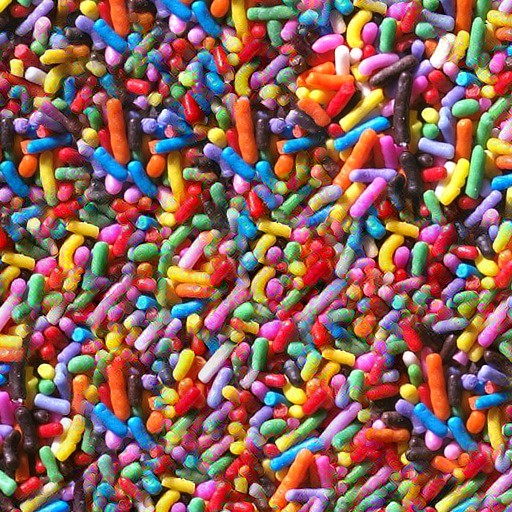}};
    \node [anchor=south] at (output1.north) {\footnotesize no padding + crop};
    \node[anchor=south, inner sep=0] (output2) at (6.4,0) {\includegraphics[width = .23\textwidth]{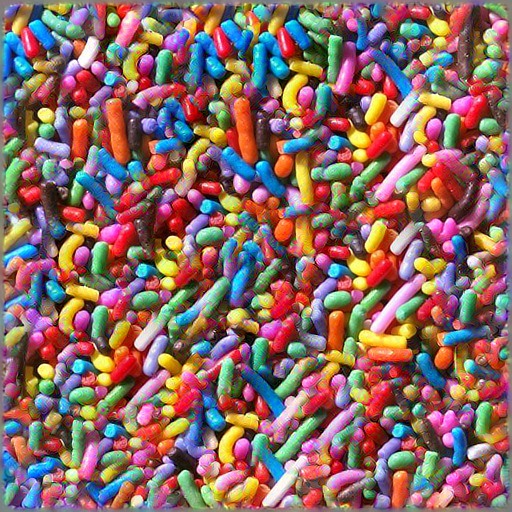}};
    \node [anchor=south] at (output2.north) {\footnotesize no padding};
    \node[anchor=south, inner sep=0] (output3) at (9.6,0) {\includegraphics[width = .23\textwidth]{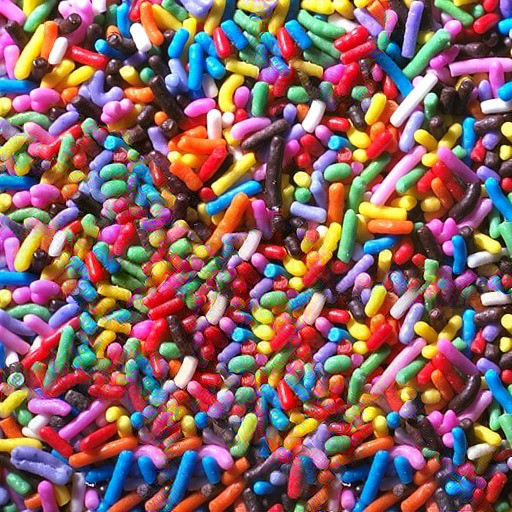}};
    \node [anchor=south] at (output3.north) {\footnotesize padding};
  \end{tikzpicture}
  \caption{This figure shows the impact of the padding in the neural network when generating images with CNNMRF~\cite{li2016combining}. The second image shows the result of a $1024\times1024$ generated texture without the network padding, cropped to $512\times512$, while the figures on the right show $512\times512$ sized generated results without or with padding. The same random initialization was used for all three results (and cropped for the last two results).}
  \label{fig:CNNMRFexamples}
\end{figure}
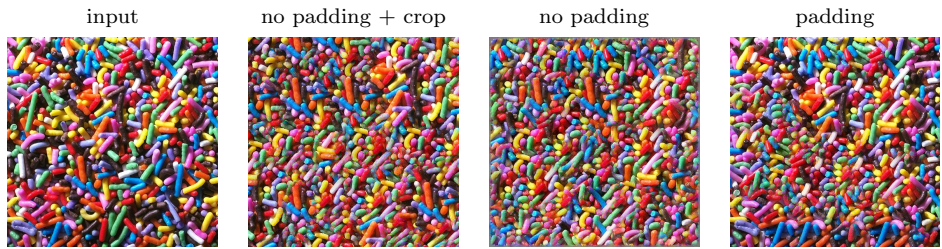

}

\section{Hybrid methods}\label{sec:hybrid}

The two main approaches to texture synthesis are the statistics-based methods and the patch re-arrangement methods. In the first class, a texture is characterized by a statistical signature; then, a random sampling conditioned to this signature produces genuinely different texture images. Nevertheless, these methods often fail for macrotextures. The second class boils down to a clever ``copy-paste'' procedure, which stitches together verbatim copies of large regions of the example. A third kind of \emph{hybrid methods} combines ideas from both approaches, leading to synthesized textures that are everywhere different from the original but with better quality than the purely statistics-based methods. We will describe one such method, its multiscale extension and the explicit combination of complementary algorithms.

\subsection{Local Gaussian models for texture synthesis}\label{sec:lgm}

Raad~et~al.'s method~\cite{raad2016} uses locally Gaussian (LG) texture model in the patch space. Each texture patch is modeled by a multivariate Gaussian distribution learned from its similar patches. Inspired by~\cite{EfrosFreeman}, the idea of searching for patches to stitch together in the original sample is maintained. However, instead of using the exact patch taken in the input texture, the stitched patch is sampled from its Gaussian model. Locally Gaussian patch models have been proved very useful in image denoising~\cite{ipol.2013.16}. This  approach permits to maintain the coherence between patches with respect to the input sample, while creating new patches that do not exist in the sample texture but are still perceptually equivalent to it.

{The multivariate Gaussian models involved are defined by their mean vector $\mu$ and their covariance matrix $\Sigma$. For a given patch $p$, of size $P \times P$ pixels, these parameters are estimated from the set of the $R$ nearest patches $\mathcal{U}_{p}^u$ (nearest neighbours of $p$ taken in $u$) as defined
\begin{equation}
  \begin{array}{lcl}
    \displaystyle \mu &=& \frac{1}{R}\sum_{\rho\in\mathcal{U}_{p}^u}{\rho},\\
    &&\\
    \displaystyle \Sigma &=& \frac{1}{R-1} \sum_{\rho\in\mathcal{U}_{p}^u} \left(\rho - \mu\right) \left(\rho -\mu\right)^t .
  \end{array}
  \label{eq:gaussian_parameters}
\end{equation}
The sampled vector $p'$ is defined as
\begin{equation}
  p' = \frac{1}{\sqrt{R-1}} \sum_{\rho\in\mathcal{U}_p^u} {a_\rho(\rho-\mu)} + \mu , \qquad a_\rho\sim\mathcal{N}(0,1),
  \label{eq:intro:gauss_sample}
\end{equation}
where $a_\rho$ are scalar random variables associated to each patch and following a normal distribution. Note that $p'$ follows the distribution $\mathcal{N}(\mu,\Sigma)$. These models have reasonable variances, confirming that effectively the patches simulated have an acceptable degree of innovation \cite{raad2016}.}

{The new texture image is synthesized by stitching together patches sampled from multivariate Gaussian distributions \eqref{eq:intro:gauss_sample} in the input sample patch space. The method is iterative: the patches are synthesized in a raster-scan order (top to bottom and left to right). The goal of each iteration is to generate a new patch $p_v^{m,n}$ (patch in $v$ placed at $(m,n)$) that is partially defined on a region called the overlap area (see Figure \ref{fig:intro:patchModel}). The known part of the patch defines the set of patches $ \mathcal{U}_{p_v^{m,n}}^{u}$ from which its Gaussian model is inferred. The generated patch $p_{v}^{m,n}$ is then sampled as defined in \eqref{eq:intro:gauss_sample}. The last step consists in stitching the patch into the output texture using the quilting method of \cite{EfrosFreeman}.}

This synthesis algorithm generates a texture that is perceptually equivalent to the sample texture yet not composed of patches existing in the input texture. Thus, this method reduces some of the drawbacks of the statistics-based and the patch-based methods. Indeed the method yields satisfying results for micro- and macro-textures, and reduces the verbatim copies of the input. However, this method remains local and is (like all patch based approaches) not forced to respect the global statistics of the texture sample.

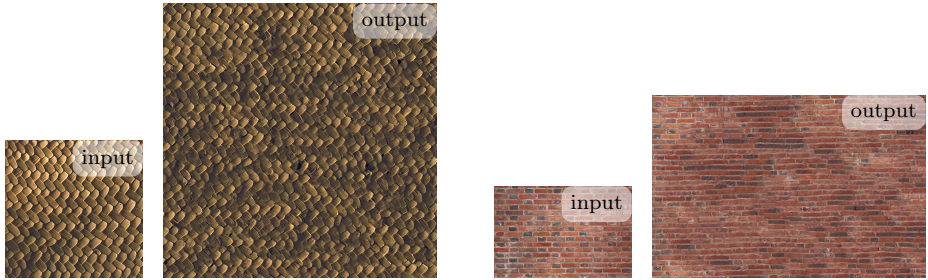
\begin{figure}[t]
  \centering
  \begin{tikzpicture}
    \node[anchor=south, inner sep=0] (input1) at (0,0) {\includegraphics[width = .15\textwidth]{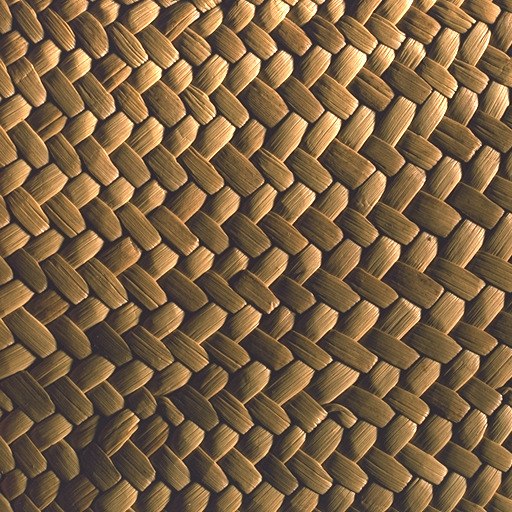}};
    \node [anchor=north east, rounded corners, fill=white, opacity=.5, text opacity=1] at (input1.north east) {\footnotesize input};
    \node[anchor=south, inner sep=0] (output1) at (3,0) {\includegraphics[width = .3\textwidth]{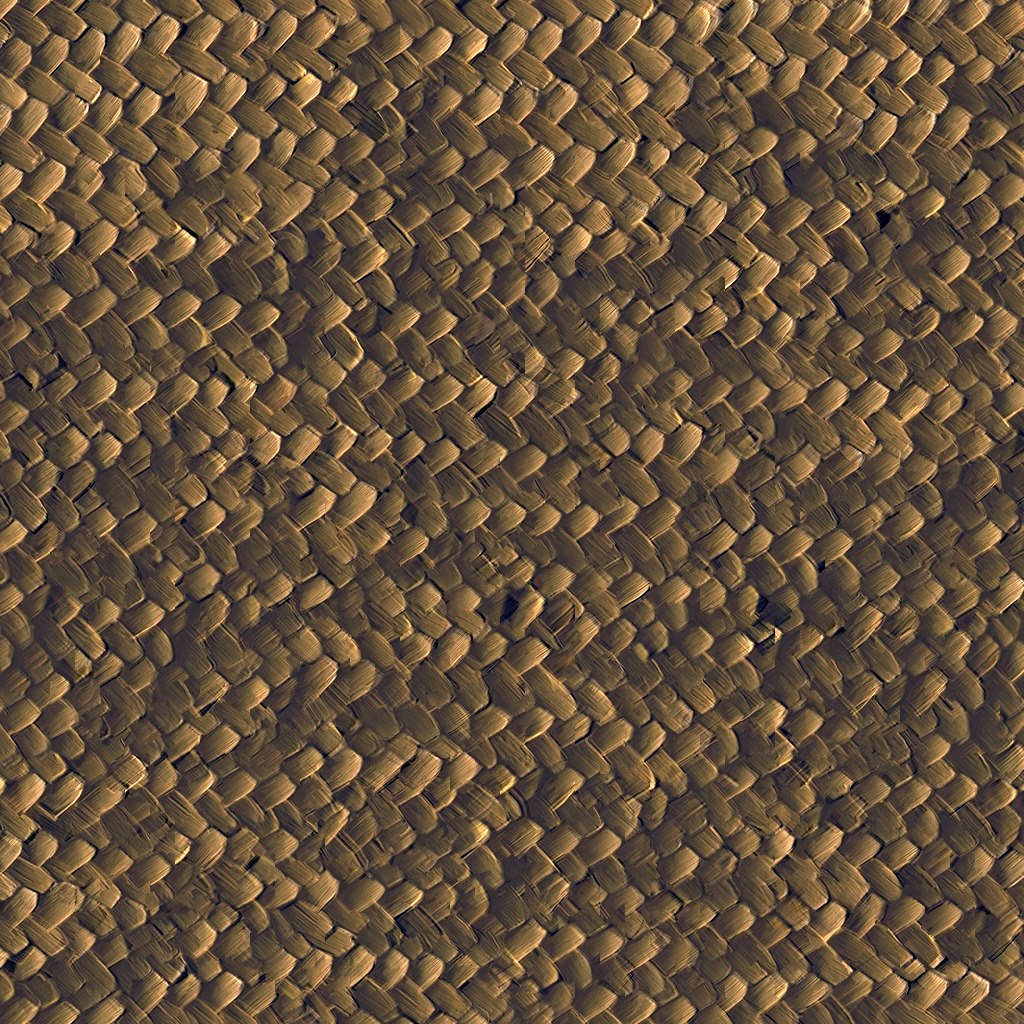}};
    \node [anchor=north east, rounded corners, fill=white, opacity=.5, text opacity=1] at (output1.north east) {\footnotesize output};

    \node[anchor=south, inner sep=0] (input2) at (6.5,0) {\includegraphics[width = .15\textwidth]{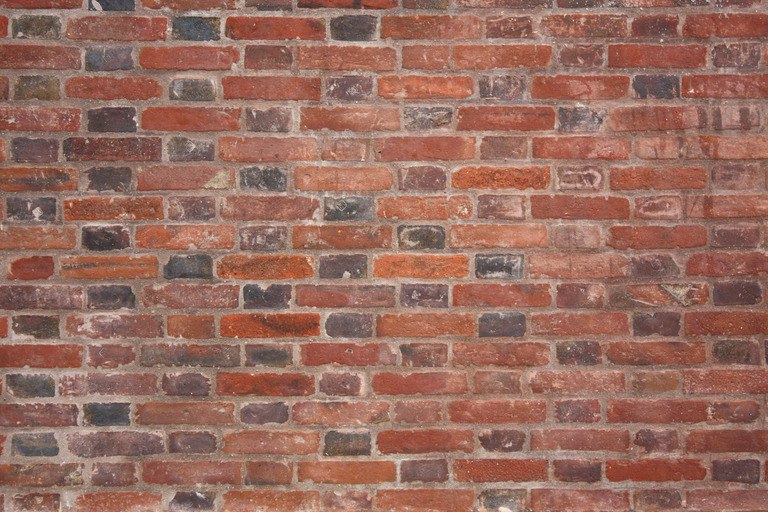}};
    \node [anchor=north east, rounded corners, fill=white, opacity=.5, text opacity=1] at (input2.north east) {\footnotesize input};
    \node[anchor=south, inner sep=0] (output2) at (9.5,0) {\includegraphics[width = .3\textwidth]{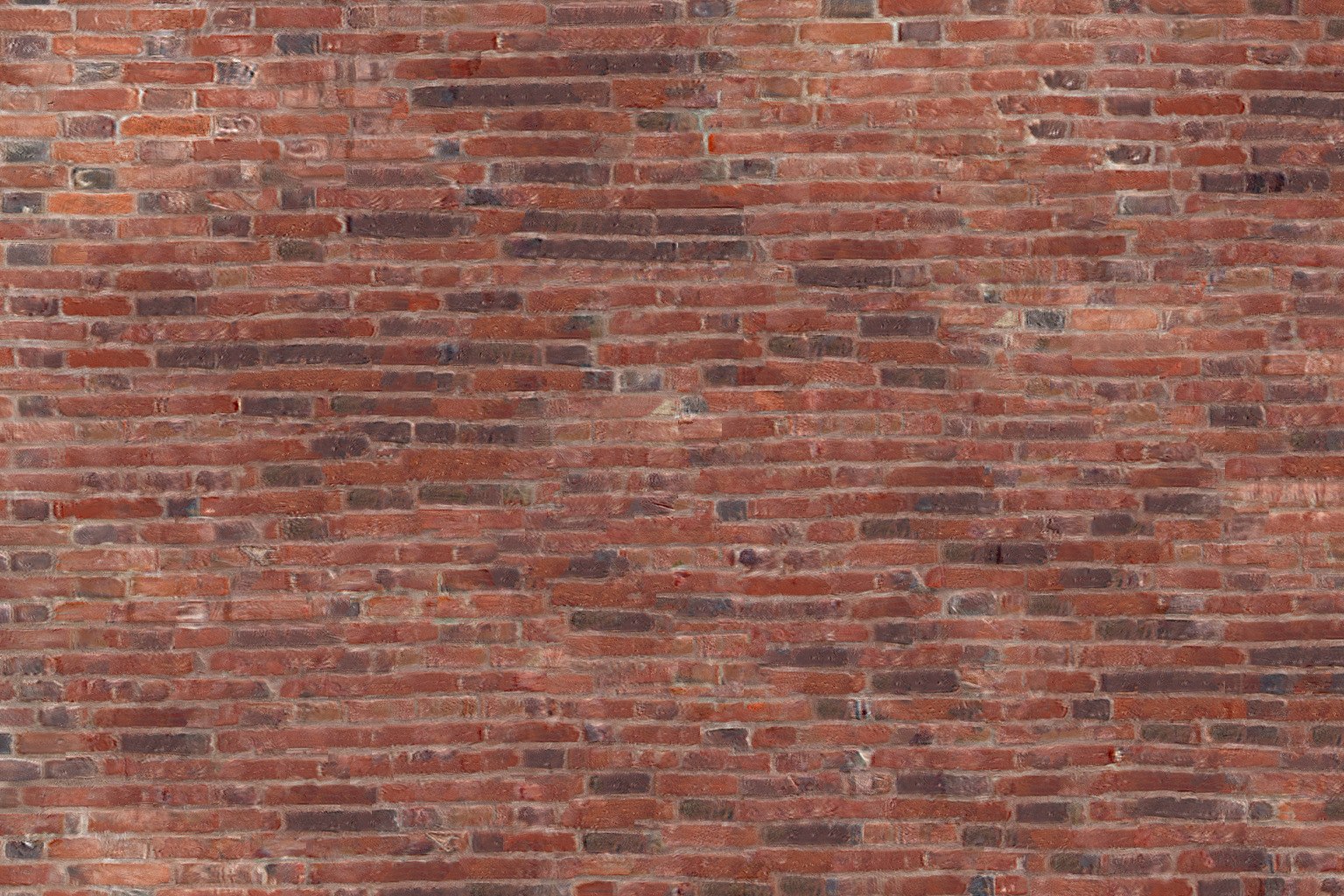}};
    \node [anchor=north east, rounded corners, fill=white, opacity=.5, text opacity=1] at (output2.north east) {\footnotesize output};
  \end{tikzpicture}
  \caption{Synthesis results of the locally Gaussian method \cite{raad2016}. It works well for macrotextures. As one can observe in both examples the result is slightly blurry, a characteristics of the Gaussian model. The parameters used for are $P=40$, $R=30$ and $O=P/2$.}
  \label{fig:lg-results}
\end{figure}

\begin{figure}[p]

  \begin{minipage}{\linewidth}
    \centering
    \begin{minipage}{\linewidth}
      \centering
      \includegraphics[width=.25\textwidth]{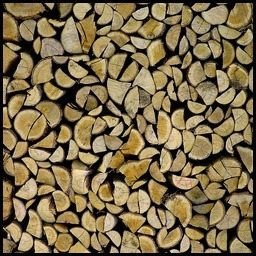}
      \includegraphics[width=.25\textwidth]{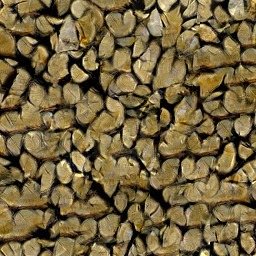}
      \includegraphics[width=.25\textwidth]{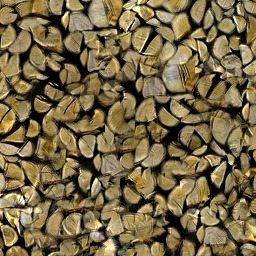}
    \end{minipage}
    \begin{minipage}{\linewidth}\centering
      \centering
      \begin{minipage}{.25\linewidth}
        \centering
        {input}
      \end{minipage}
      \begin{minipage}{.25\linewidth}
        \centering
        ${R=10, P=20}$
      \end{minipage}
      \begin{minipage}{.25\linewidth}
        \centering
        ${R=10, P=30}$
      \end{minipage}
    \end{minipage}
    \begin{minipage}{\linewidth}
      \centering
      \includegraphics[width=.25\textwidth]{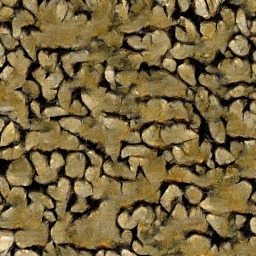}
      \includegraphics[width=.25\textwidth]{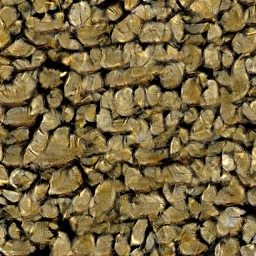}
      \includegraphics[width=.25\textwidth]{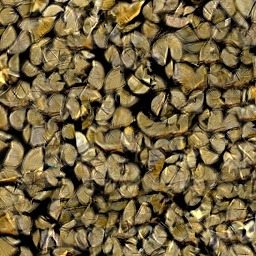}
    \end{minipage}
    \begin{minipage}{\linewidth}
      \centering
      \begin{minipage}{.25\linewidth}
        \centering
        ${R=20, P=10}$
      \end{minipage}
      \begin{minipage}{.25\linewidth}
        \centering
        ${R=20, P=20}$
      \end{minipage}
      \begin{minipage}{.25\linewidth}
        \centering
        ${R=20, P=30}$
      \end{minipage}
    \end{minipage}
    \begin{minipage}{\linewidth}
      \centering
      \includegraphics[width=.25\textwidth]{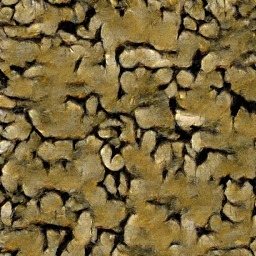}
      \includegraphics[width=.25\textwidth]{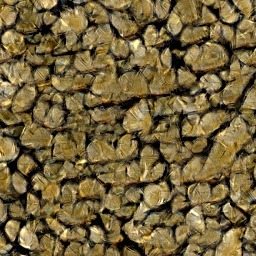}
      \includegraphics[width=.25\textwidth]{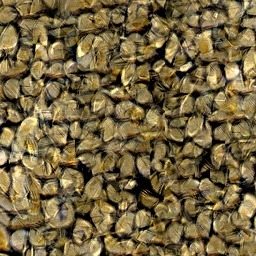}
    \end{minipage}
    \begin{minipage}{\linewidth}
      \centering
      \begin{minipage}{.25\linewidth}
        \centering
        ${R=30, P=10}$
      \end{minipage}
      \begin{minipage}{.25\linewidth}
        \centering
        ${R=30, P=20}$
      \end{minipage}
      \begin{minipage}{.25\linewidth}
        \centering
        ${R=30, P=30}$
      \end{minipage}
    \end{minipage}
  \end{minipage}

  \caption{Texture synthesis result for the left top corner texture image. We show the results obtained for different values of $R$ (the number of similar patches) and $P$ (the side patch size). From left to right $P = 10, 20, 30$. From top to bottom, the number of nearest neighbours is $R=10, 20, 30$. All the results are obtained for an overlap of a half patch size $O = P/2$.}
  \label{fig:intro:synth1}
\end{figure}

Figure~\ref{fig:lg-results} shows two results of the method. The algorithm remains dependent on the choice of the patch size $P$ and of the number of nearest neighbours $R$ as illustrated in Figure~\ref{fig:intro:synth1}. These values may have to be adjusted for each texture sample. As for the overlap size a convenient value is $O = P/2$. If this value is too small then the region used to infer the Gaussian models is not enough. The patches used to infer the model can be very different on a high portion of the patch. The algorithm has a low computational complexity, compared for instance with classic patch-based denoising algorithms~\cite{lebrun2013nonlocal,dabov2007image}. An alternative to reduce the dependency of the method to the patch size is to work in a multiscale approach.

\subsection{Multiscale texture synthesis methods}\label{sec:multiscale}

Most real textures are organized at multiple scales: the global structure is revealed at coarse scales but important detail are present at finer ones. As we have seen, the results of patch-based methods depend strongly on the patch size. Small patch sizes may capture the finer details of the input but the resulting texture will lack global coherence. On the other hand, using large patches will maintain the global structures at the risk of a ``copy-paste'' effect. Furthermore, with large patches it becomes impossible to model the patch variability due to the lack of sufficient samples. This is apparent in the examples of Figure~\ref{fig:lg-results}, where modeling patches as multivariate Gaussian vectors leads to a slightly blurry texture. A natural solution is to use a multiscale approach~\cite{kwatra2005,tartavel,lefebvre2005parallel,hertzmann2001image,raad2016} using several patch sizes for a single texture synthesis, capturing different levels of details.

{This section illustrates the ideas and difficulties of a multiscale extension using as example  the local Gaussian models for texture synthesis presented in the previous section~\cite{raad2016}. The Multi-Scale Locally Gaussian (MSLG) method works at $S$ scales and can be summarized in a few sentences. The synthesis begins at the coarsest scale ($s=S-1$) using the local Gaussian method where the quilting step is replaced by a simple average of the overlapping patches. For the remaining scales ($s=S-2, \dots , 0$), a synthesis is performed by using the result of the previous scale ($s+1$) and the sample image at the corresponding resolution. At each scale the synthesis is done patch by patch in a raster-scan order. Each new patch, added to the synthesized image, overlaps part of the previously synthesized patch and it is the combination of a low resolution patch and a high resolution one sampled from a multivariate Gaussian distribution. The Gaussian distribution of the high frequencies of a given patch is estimated from the high frequencies of its $m$ nearest neighbours in the corresponding scale input image. The synthesis result of the finer scale is the desired output image.

  Let us denote the sample texture by $u$ and $u_s,~s=1,\dots,S-1$ are the zoomed out versions by a factor $2^s, ~ s=1,\dots,S-1$. The synthesis result at each scale is denoted by $v_s,~s=1,\dots,S-1$ and $v$ is the synthesis result returned by the multiscale algorithm. An additional image $\tilde{v}_s$ is needed at each scale, corresponding to a low resolution version of $v_s$ obtained by interpolating $v_{s+1}$. To estimate the parameters of the Gaussian distribution of the patch $p_{v_s}^{m',n'}$ being processed, the set $\mathcal{U}_{p_{v_s}^{m',n'}}^{u_s}$ of $R$ nearest patches in $u_s$ is considered. The $R$ nearest neighbours in $u_{s}$ to the current patch are those minimizing the $L^2$ distance restricted to the overlap area:
\begin{align}
  d(p_{u_{s}}^{m,n},p_{v_{s}}^{m',n'})^2 & =
      \frac{1}{\vert\mathcal{O}\vert}\sum_{(i,j)\in \mathcal{O}}{(u_{s}(m+i,n+j)-v_{s}(m'+i,n'+j) )^2} \nonumber\\
  & + \frac{1}{P^2}\sum_{i,j=0}^{P-1}{(\tilde{u}_s(m+i,n+j)-\tilde{v}_{s}(m'+i,n'+j) )^2},
  \label{eq:distance2}
\end{align}
where $\tilde{u}_s$ denotes the low resolution of the image $u_s$, $\tilde{u}_s = u_s\ast G_{\sigma}$ and $\tilde{v}_{S-1} = u_{S-1}\ast G_{\sigma}$. In~\eqref{eq:distance2}, the overlap area is denoted as $\mathcal{O}$ and the size of patch overlap is fixed to $P/2$. On the set $\mathcal{U}_{p_{v_s}^{m',n'}}^{u_s}$ only the high frequency of the patches $p_{u_s}^{i,j}-p_{\tilde{u}_s}^{i,j}$ is considered to infer the multivariate Gaussian distribution $\mathcal{N}(\mu_H, \Sigma_H)$. The patch $p_{v_s}^{m,n}$ is synthesized as the combination of a low resolution patch $p_{\tilde{v}_s}^{m,n}$ yield from the previous scale with a high resolution one $p \sim \mathcal{N}(\mu_H, \Sigma_H)$, thus $p_{v_s}^{m,n}= p_{\tilde{v}_s}^{m,n} + p$. For more details please refer to~\cite{raad2016}.}

\begin{figure}[t]
  \centering
  \begin{tikzpicture}
    \node[anchor=south, inner sep=0] (input1) at (0,0) {\includegraphics[width = .16\textwidth]{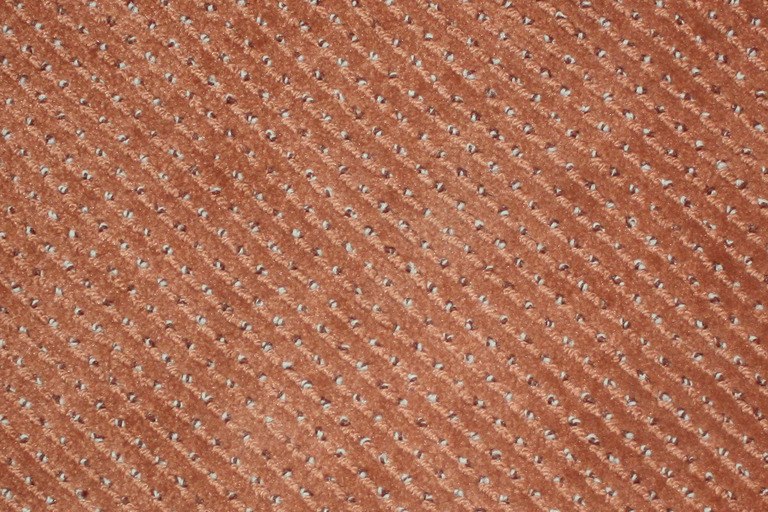}};
    \node [anchor=north east, rounded corners, fill=white, opacity=.5, text opacity=1] at (input1.north east) {\footnotesize input};
    \node[anchor=south, inner sep=0] (output1) at (3.2,0) {\includegraphics[width = .32\textwidth]{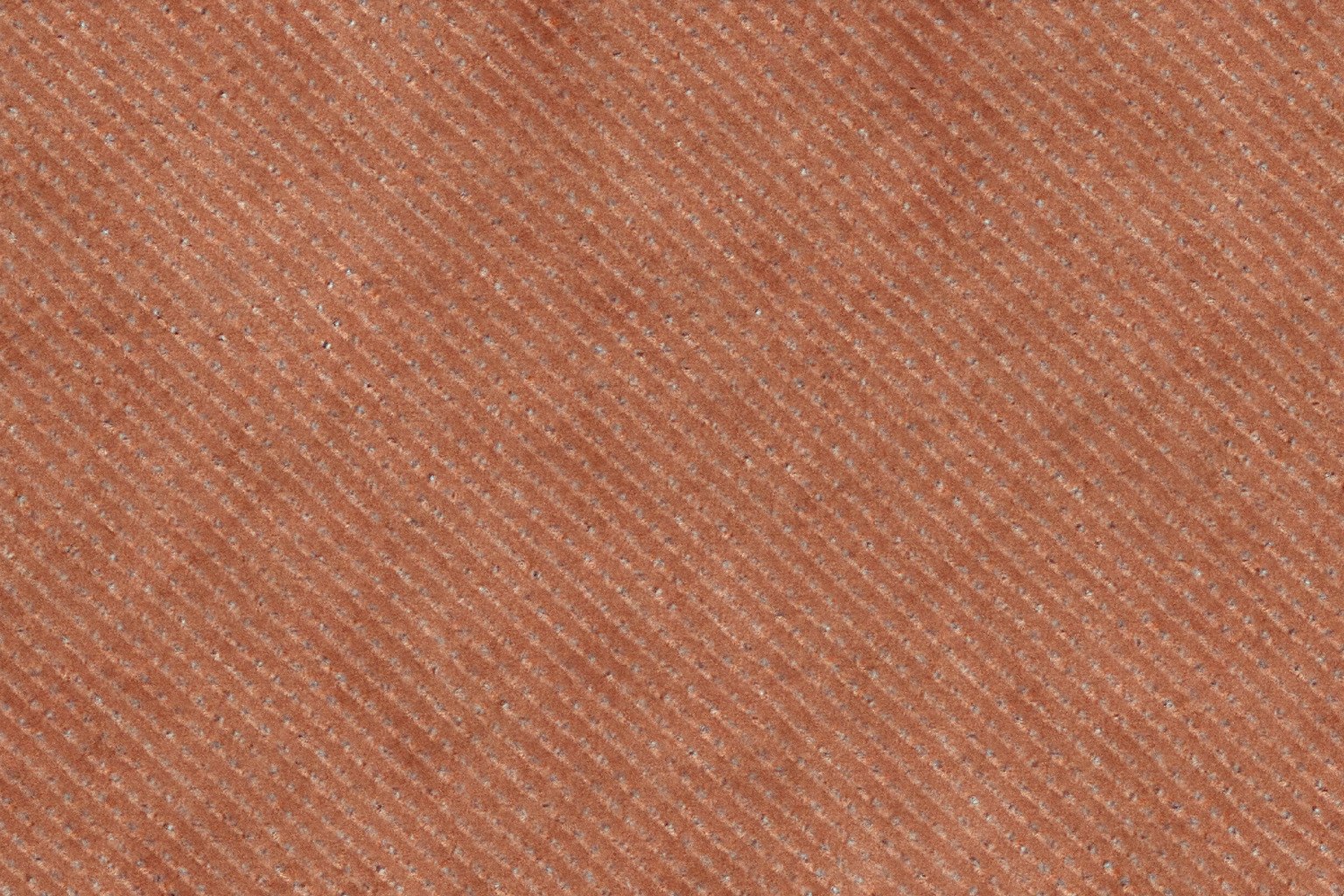}};
    \node [anchor=north east, rounded corners, fill=white, opacity=.5, text opacity=1] at (output1.north east) {\footnotesize output};

    \node[anchor=south, inner sep=0] (input2) at (6.7,0) {\includegraphics[width = .16\textwidth]{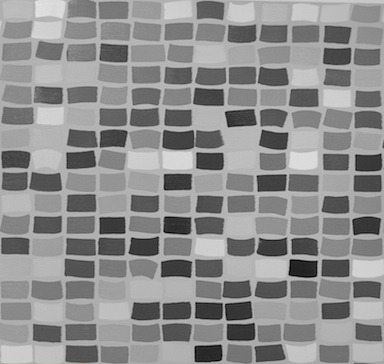}};
    \node [anchor=north east, rounded corners, fill=white, opacity=.5, text opacity=1] at (input2.north east) {\footnotesize input};
    \node[anchor=south, inner sep=0] (output2) at (9.4,0) {\includegraphics[width = .24\textwidth]{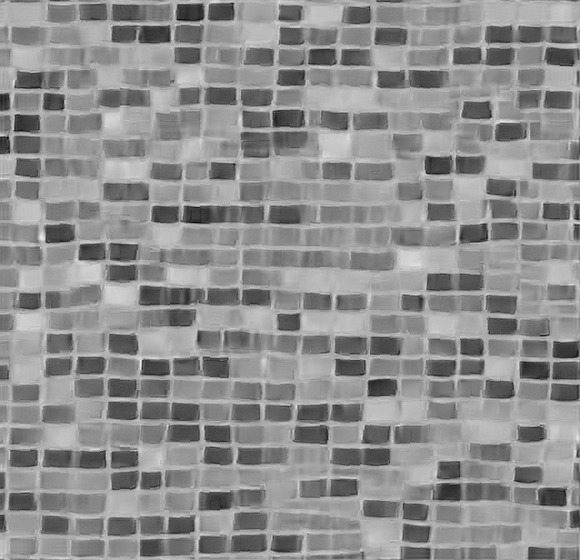}};
    \node [anchor=north east, rounded corners, fill=white, opacity=.5, text opacity=1] at (output2.north east) {\footnotesize output};
  \end{tikzpicture}
  \caption{Synthesis results of the multiscale locally Gaussian method \cite{raad2016}. Both examples show that the details of different scales are correctly synthesized when using a patch size $P=20$. However the results are slightly blurred with respect to the input. The number of scales is $S=3$ for the first example (left) and $S=2$ for the second example (right).}
  \label{fig:mslg-results}
\end{figure}
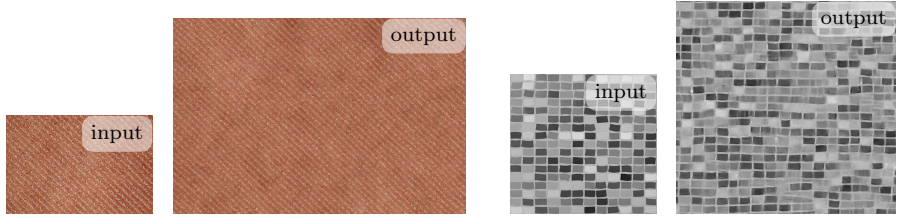

Figure~\ref{fig:mslg-results} shows two synthesis examples. In both cases the result is satisfyingly recovering the details of the different scales for reasonable values of the patch size $P=20$. However one can notice that the results are blurry with respect to the input and this effect is increased with respect to the single scale approach.

\subsection{Combination of methods}\label{sec:combination}

  A smart combination of complementary methods may keep the advantages of each one. We will illustrate the methodology by combining a multiscale approach with three other methods:
\begin{description}
  \item[MSLG+EF] The Multi-Scale Locally Gaussian method combined with the Efros and Freeman method.
  \item[MSLG+PS] The Multi-Scale Locally Gaussian method combined with the Portilla and Simoncelli method.
  \item[MSLG+Gatys] The Multi-Scale Locally Gaussian method combined with the Gatys~et~al. method.
\end{description}

\begin{figure}[t]
  \centering
  \begin{tikzpicture}
    \node[anchor=south, inner sep=0] (input1) at (0,0) {\includegraphics[width = .18\textwidth]{figures/Flowers_0000}};
    \node [anchor=north east, rounded corners, fill=white, opacity=.5, text opacity=1] at (input1.north east) {\footnotesize input};
    \node[anchor=south, inner sep=0] (output1) at (4,0) {\includegraphics[width = .36\textwidth]{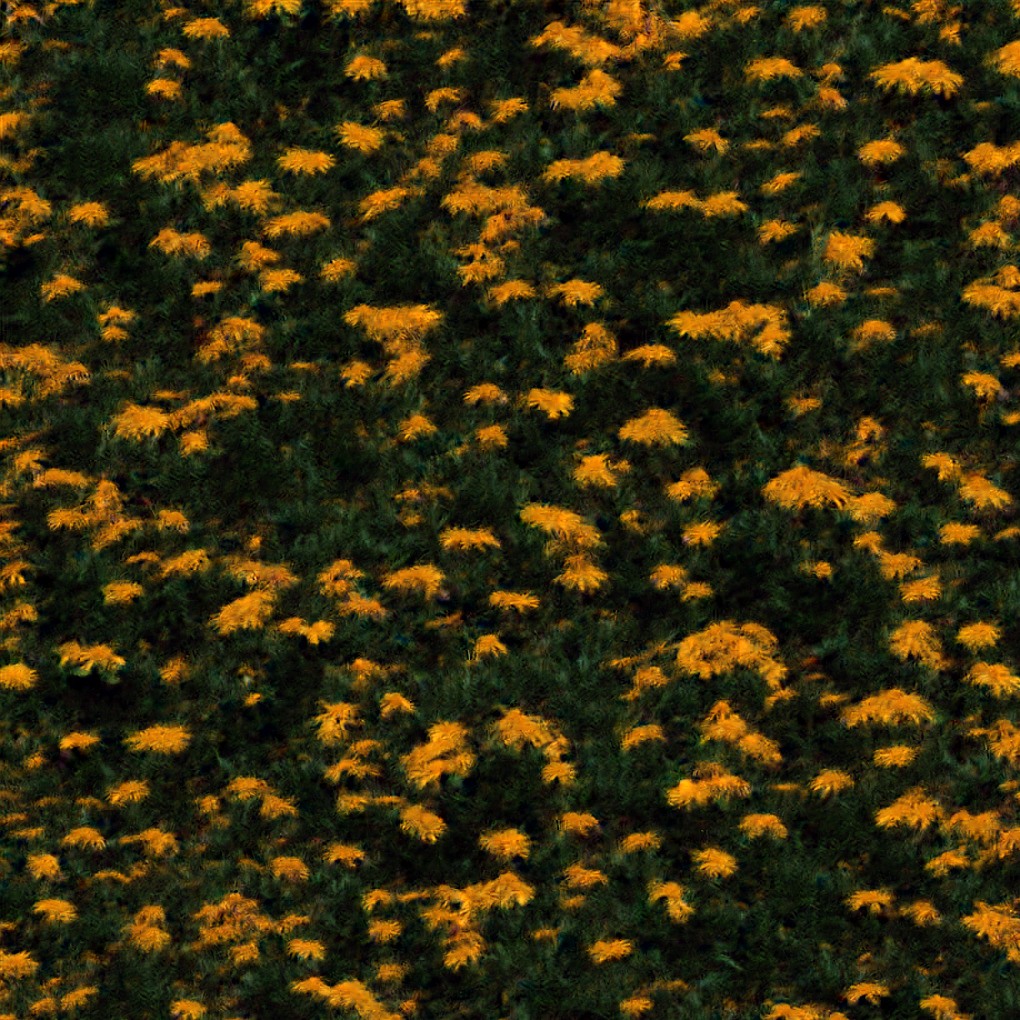}};
    \node [anchor=north east, rounded corners, fill=white, opacity=.5, text opacity=1] at (output1.north east) {\footnotesize MSLG};
    \node[anchor=south, inner sep=0] (output2) at (9,0) {\includegraphics[width = .36\textwidth]{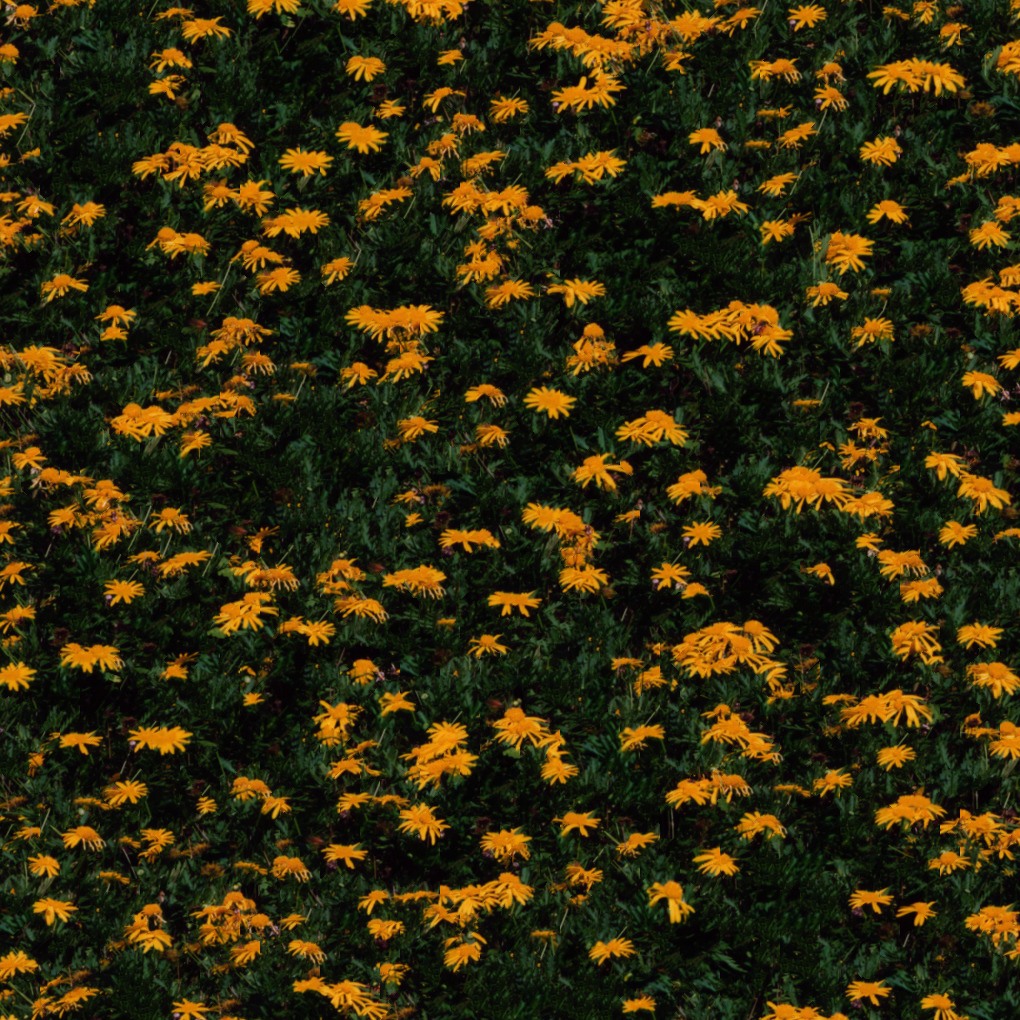}};
    \node [anchor=north east, rounded corners, fill=white, opacity=.5, text opacity=1] at (output2.north east) {\footnotesize MSLG+EF};
  \end{tikzpicture}
  \caption{Synthesis results of the combination of the Multi-Scale Locally Gaussian method with the Efros and Freeman (MSLG+EF).}
  \label{fig:mslg-ef}
\end{figure}

The combination of the Multi-Scale Locally Gaussian method with the Efros and Freeman method (MSLG+EF) consists of two steps. The first step synthesizes the given input $u$ with the Multi-Scale Locally Gaussian method generating a new texture image that we denote $u_\text{mslg}$. The second step consists in applying the Efros and Freeman algorithm to the given input sample, initializing the output image that we denote $u_\text{ef}$ with the image $u_\text{mslg}$. The method is basically the same as the one described in section \ref{sec:ef}. The only step of the algorithm that is modified is the \emph{patch selection} step. In the method described in \cite{EfrosFreeman} at each iteration the added patch was chosen among those (in the input sample) whose overlap region was similar to the one of the patch under construction. When combining the methods, instead of only comparing the overlap areas, the entire patches are compared. Initializing the output with a first synthesis $u_\text{mslg}$ enables the method to use the whole patch under construction to find a candidate in the input sample $u$. The candidate patch taken from $u$ is then quilted in $u_\text{ef}$ at the corresponding position with the same stitching step as in~\cite{EfrosFreeman}. This combination allows to recover the lost resolution of the MSLG synthesis as illustrated in Figure~\ref{fig:mslg-ef}. However it is not capable of masking the garbage growing effects as effectively MSLG+PS combination does.

\begin{figure}[t]
  \centering
  \begin{tikzpicture}
    \node[anchor=south, inner sep=0] (input1) at (0,0) {\includegraphics[width = .18\textwidth]{figures/Fabric_0000}};
    \node [anchor=north east, rounded corners, fill=white, opacity=.5, text opacity=1] at (input1.north east) {\footnotesize input};
    \node[anchor=south, inner sep=0] (output1) at (4,0) {\includegraphics[width = .36\textwidth]{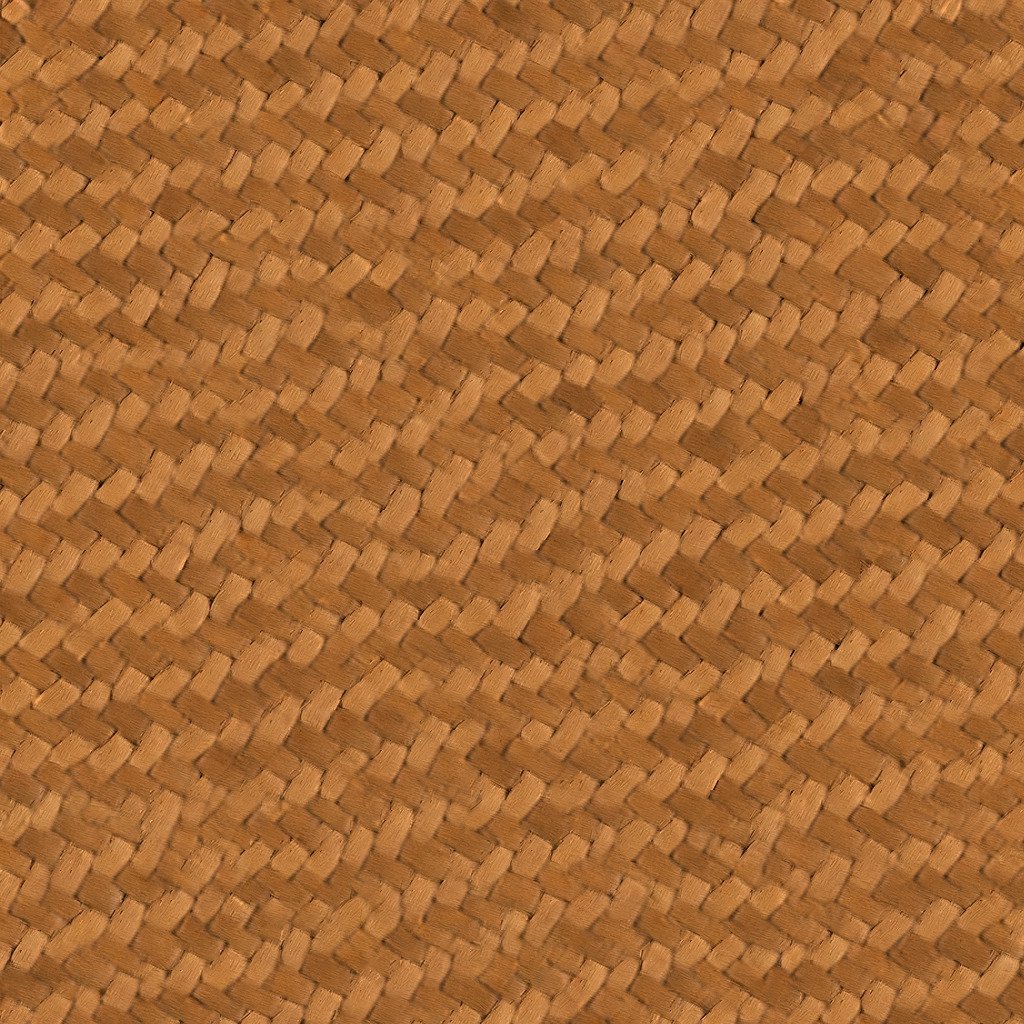}};
    \node [anchor=north east, rounded corners, fill=white, opacity=.5, text opacity=1] at (output1.north east) {\footnotesize MSLG};
    \node[anchor=south, inner sep=0] (output2) at (9,0) {\includegraphics[width = .36\textwidth]{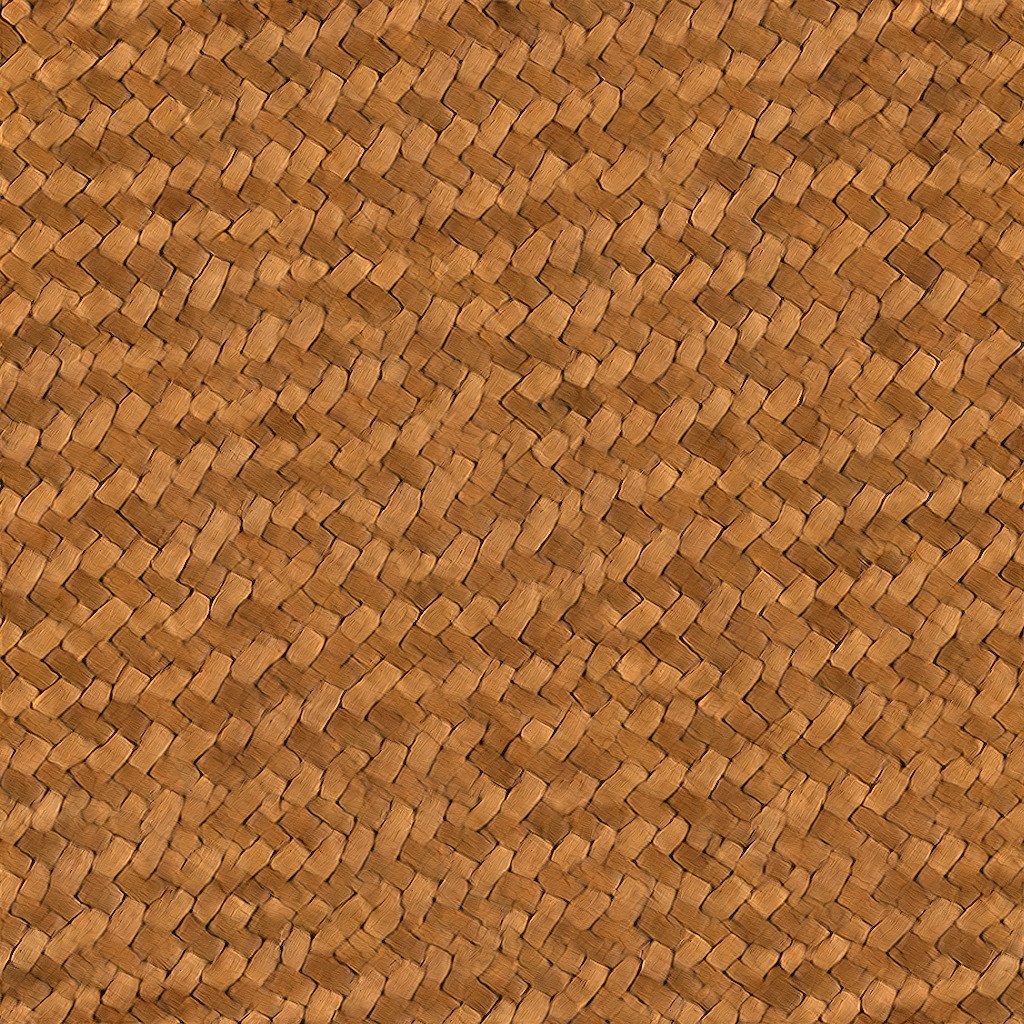}};
    \node [anchor=north east, rounded corners, fill=white, opacity=.5, text opacity=1] at (output2.north east) {\footnotesize MSLG+PS};
  \end{tikzpicture}
  \caption{Synthesis results of the combination of the Multi-Scale Locally Gaussian method with the Portilla and Simoncelli methods (MSLG+PS).}
  \label{fig:mslg-ps}
\end{figure}

The combination of the Multi-Scale Locally Gaussian method with the Portilla and Simoncelli method (MSLG+PS) consists of two steps. In the first step, given the input image $u$, a new texture $u_\text{mslg}$ is generated using MSLG. The second step uses PS where the initialization ``noise image'' is replaced by $u_\text{mslg}$ generating the output image that we denote $u_\text{ps}$. As explained in section \ref{sec:ps}, the statistics to impose are learnt on the input $u$. What follows is a synthesis step where the output image is projected on the subspaces of constraints. There exist several local solutions to this projection step. When initializing PS with the result of MSLG, the initialization image is generally quite close to the images living in the sub-space of the whole set of constraints. Thus the result obtained is improved compared to PS images starting from a random noise image. Naturally fixing the initialization of the PS algorithm removes the randomness of the generated texture. But this is not the case since the initialization is itself random as it is generated from another random process. This combination is illustrated in Figure~\ref{fig:mslg-ps}.

\begin{figure}[t]
  \centering
  \begin{tikzpicture}
    \node[anchor=south, inner sep=0] (input1) at (0,0) {\includegraphics[width = .18\textwidth]{figures/Fabric_0008}};
    \node [anchor=north east, rounded corners, fill=white, opacity=.5, text opacity=1] at (input1.north east) {\footnotesize input};
    \node[anchor=south, inner sep=0] (output1) at (4,0) {\includegraphics[width = .36\textwidth]{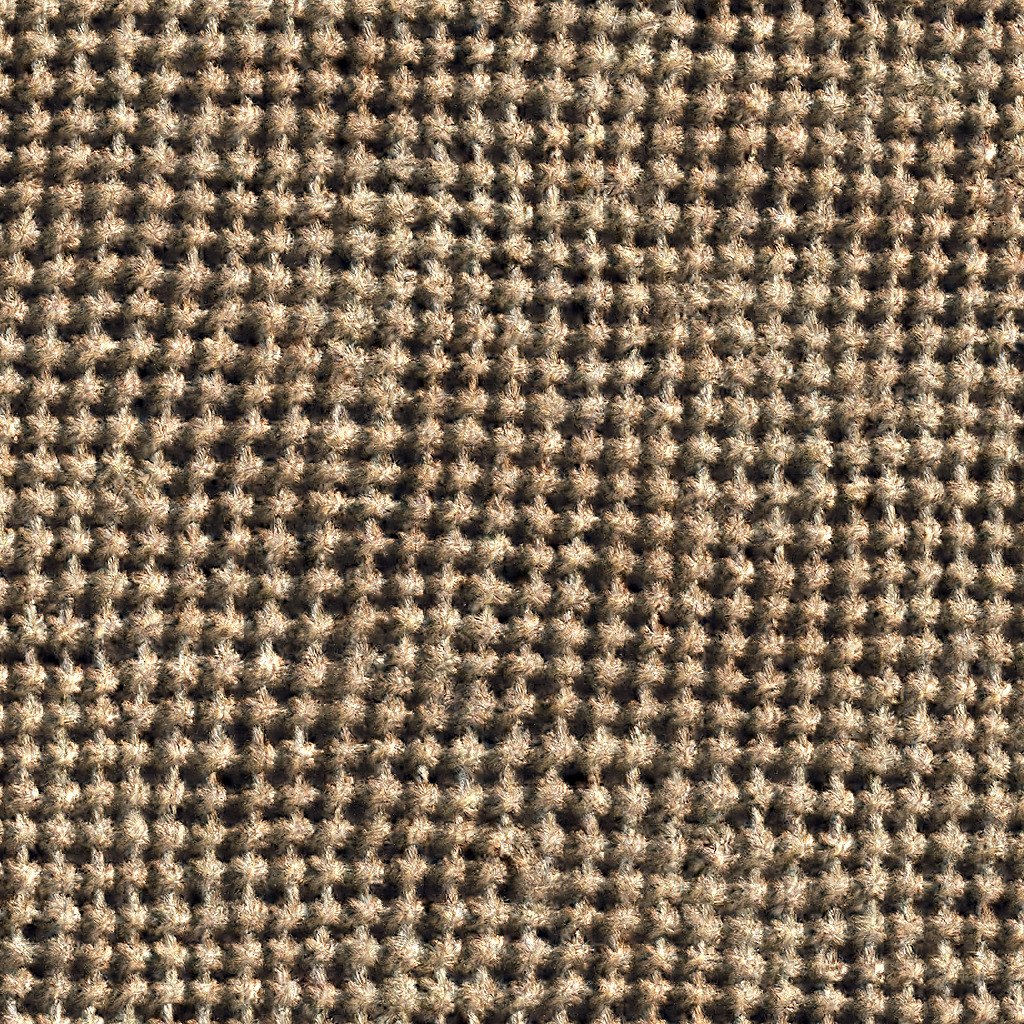}};
    \node [anchor=north east, rounded corners, fill=white, opacity=.5, text opacity=1] at (output1.north east) {\footnotesize MSLG};
    \node[anchor=south, inner sep=0] (output2) at (9,0) {\includegraphics[width = .36\textwidth]{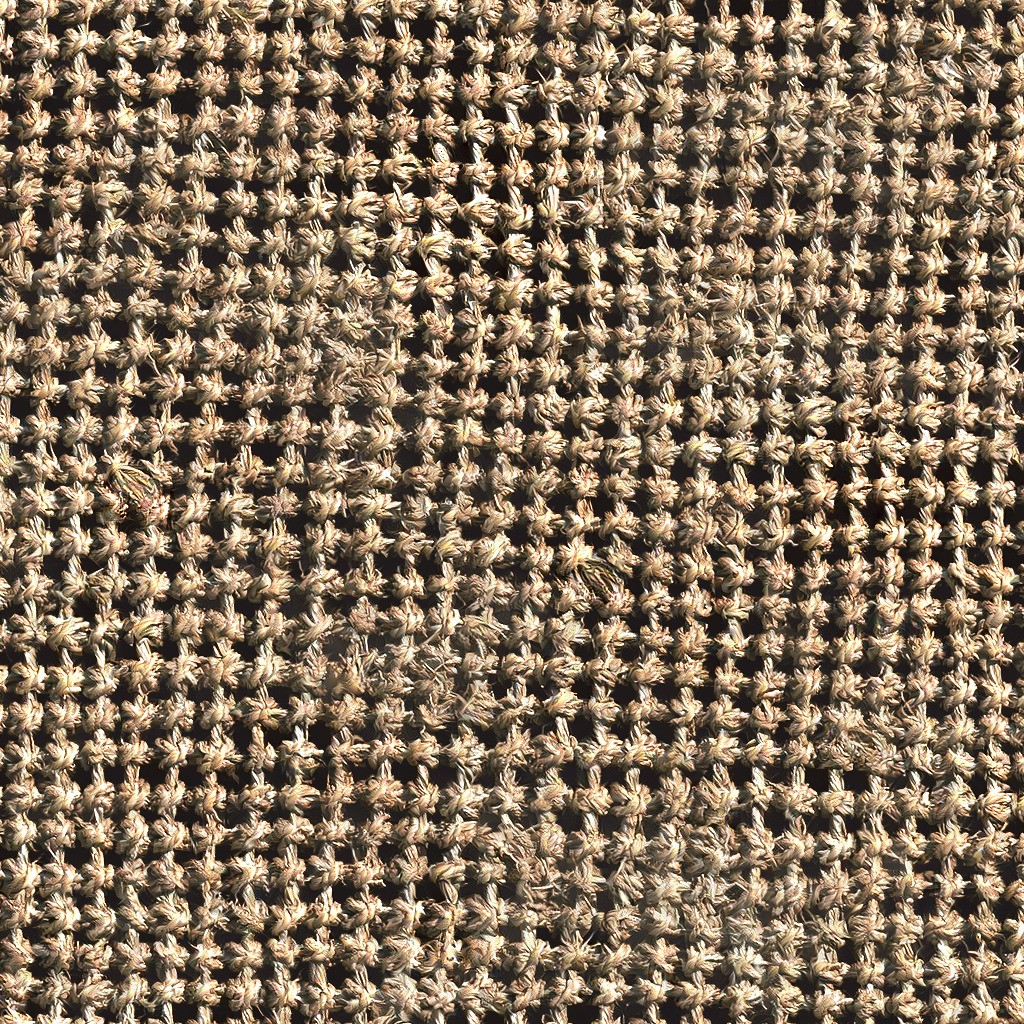}};
    \node [anchor=north east, rounded corners, fill=white, opacity=.5, text opacity=1] at (output2.north east) {\footnotesize MSLG+Gatys};
  \end{tikzpicture}
  \caption{Synthesis results of the combination of the Multi-Scale Locally Gaussian method with Gatys' texture generator (MSLG+Gatys).}
  \label{fig:mslg-gatys}
\end{figure}

The combination of the Multi-Scale Locally Gaussian method with Gatys' texture generator (MSLG+Gatys) is very similar to its combination with the Portilla and Simoncelli method.
The texture generator is initialized with the result of MSLG $u_\text{mslg}$, and the statistics of the target image are enforced via several iterations of backpropagation generating the output image denoted as $u\text{gatys}$. This combination is illustrated in Figure~\ref{fig:mslg-gatys}.

\section{Experiments}\label{sec:exp}

The first part of this section compares the exemplar-based texture synthesis methods described before on a set of standard textures. These results illustrate the advantages and limitations of each one. Then, the second part attempts at the synthesis of real life and more complex textured images, revealing the shortcomings still present in all the methods when confronted with such a demanding task.

\subsection{Comparative evaluation}\label{sec:comparative-evaluation}

We will compare the results of the following texture synthesis methods: Random Phase Noise (RPN)~\cite{Wijk_spot_noise_texture_synthesis_1991,GalerneRPN}, Heeger and Bergen (HB)~\cite{HeegerBergen}, Portilla and Simoncelli (PS)~\cite{PS}, Gatys (Gatys)~\cite{gatys}, SGAN~\cite{jetchev2016texture}, Efros and Leung (EL)~\cite{EfrosLeung}, Efros and Freeman (EF)~\cite{EfrosFreeman}{, CNNMRF~\cite{li2016combining}} and MSLG~\cite{raad2016}. Figures~\ref{fig:algComp1-1} to~\ref{fig:algComp2-2} show results for various  texture samples, one per column; in each figure, the first row shows the sample image and the following rows correspond, as indicated, to one of the algorithms. We focus on these original texture synthesis algorithms, and do not show the numerous variants. For several of our sample textures, these variants could get better results, but we think that showing the results of the original algorithms better underlines their intrinsic strengths and weaknesses. Similarly we won't present the results of all the combinations of the different methods.

\newlength{\widthfigureresultsTwo}
\setlength{\widthfigureresultsTwo}{0.09\textwidth}


\begin{figure}[p]
  \centering
  \begin{tikzpicture}[scale=.9]

    \node[inner sep=0, anchor=south east] (im8) at (0,15.6) {\includegraphics[width=\widthfigureresultsTwo]{figures/tissus27}};
    \node[inner sep=0, anchor=south east]  at (3,15.6) {\includegraphics[width=\widthfigureresultsTwo]{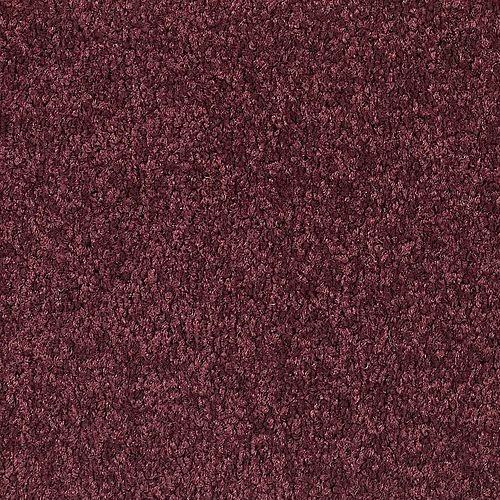}};
    \node[inner sep=0, anchor=south east]  at (6,15.6) {\includegraphics[width=\widthfigureresultsTwo]{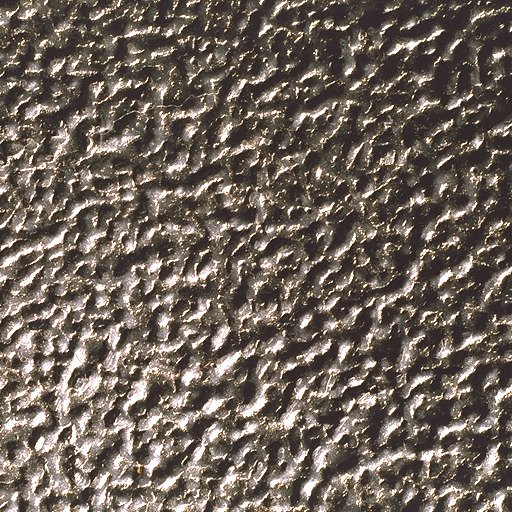}};
    \node[inner sep=0, anchor=south east]  at (9,15.6) {\includegraphics[width=\widthfigureresultsTwo]{figures/Fabric_0008}};

    \node[inner sep=0, anchor=south] (im7) at (0,13) {\includegraphics[width=2\widthfigureresultsTwo]{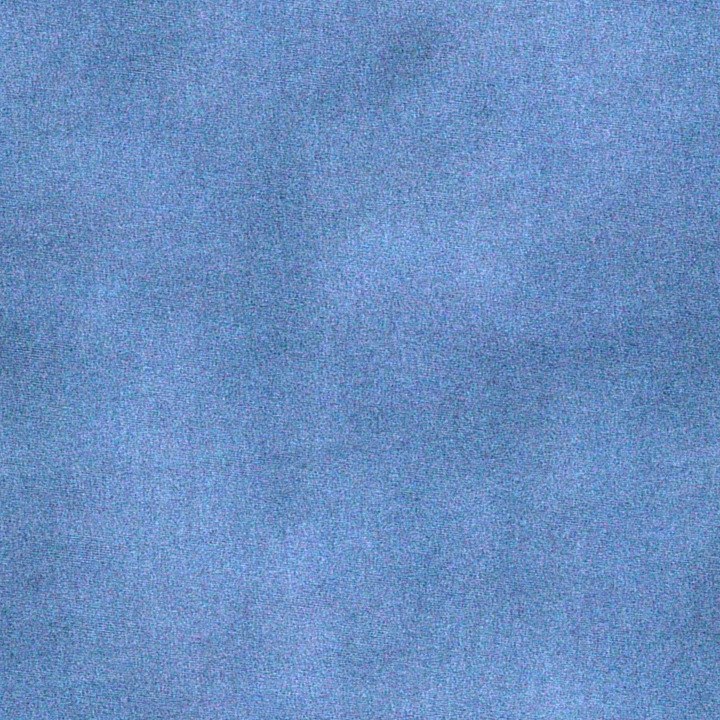}};
    \node[inner sep=0, anchor=south]  at (3,13) {\includegraphics[width=2\widthfigureresultsTwo]{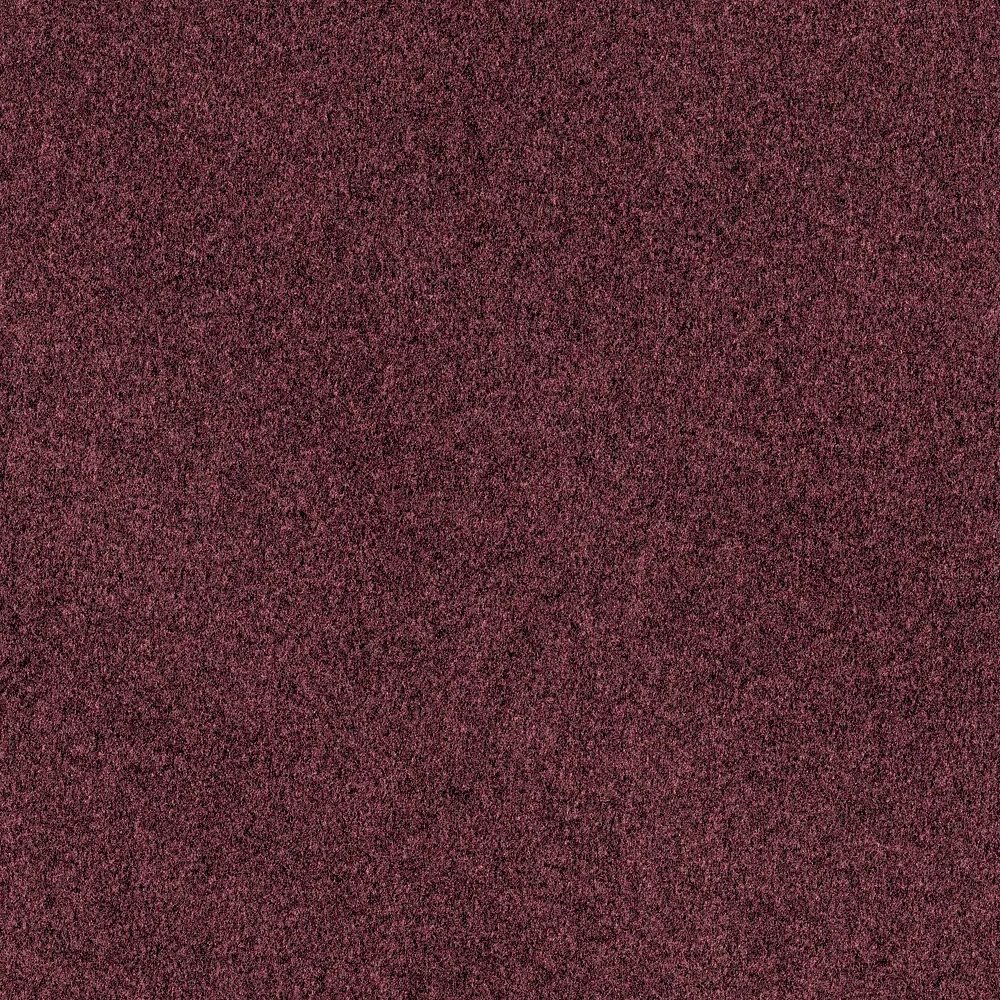}};
    \node[inner sep=0, anchor=south]  at (6,13) {\includegraphics[width=2\widthfigureresultsTwo]{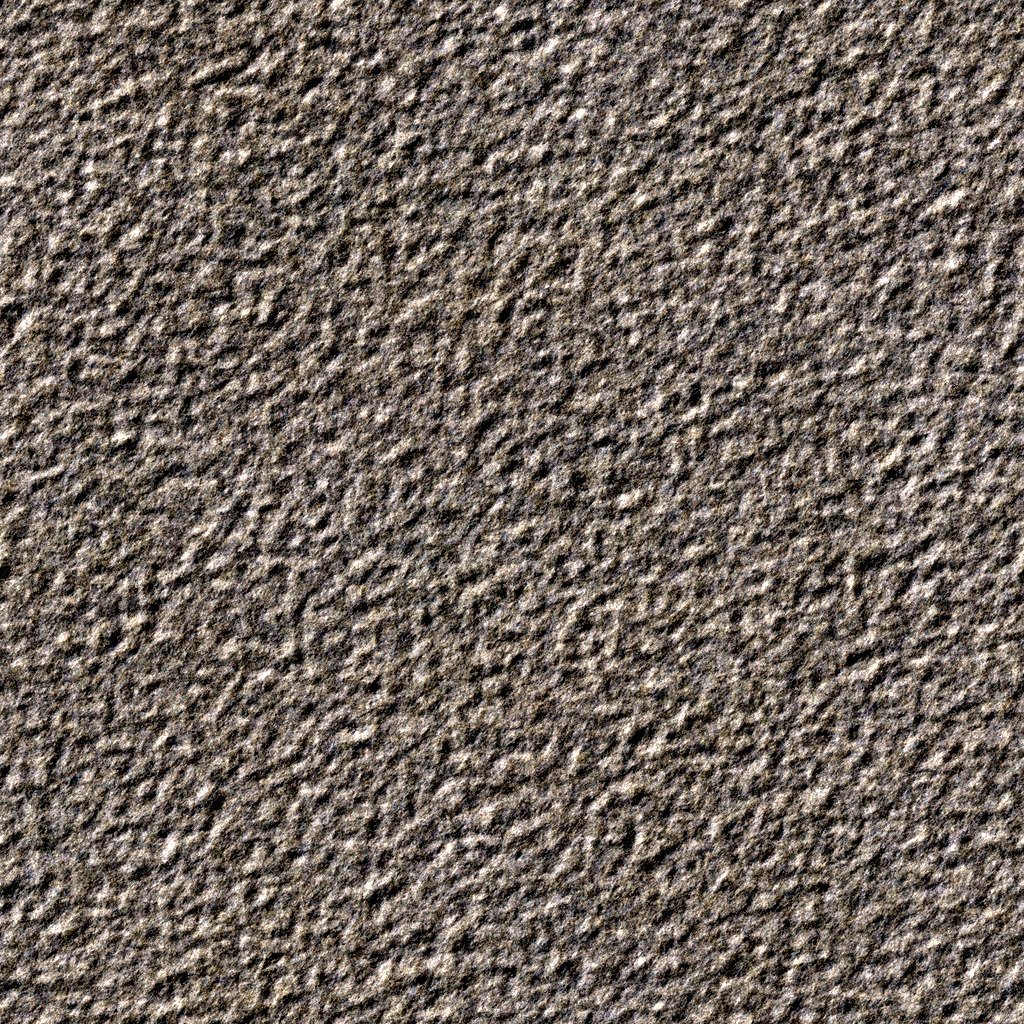}};
    \node[inner sep=0, anchor=south]  at (9,13) {\includegraphics[width=2\widthfigureresultsTwo]{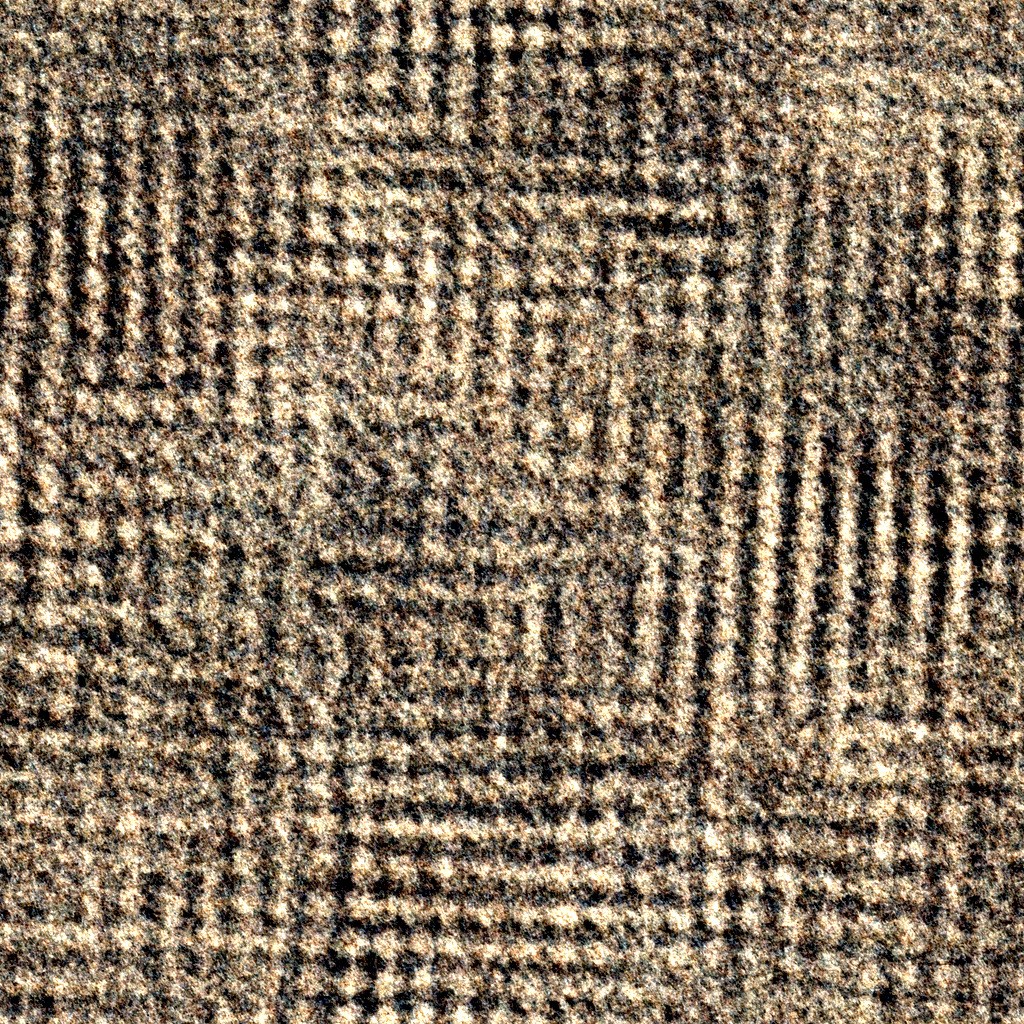}};

    \node[inner sep=0, anchor=south] (im6) at (0,10.4) {\includegraphics[width=2\widthfigureresultsTwo]{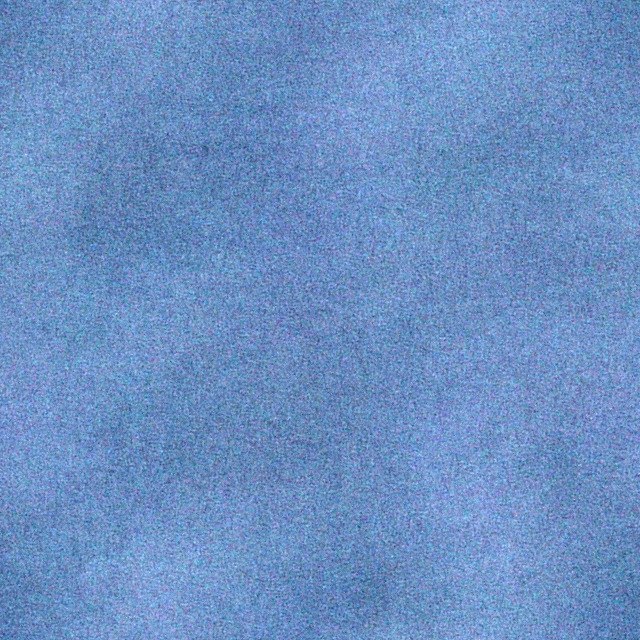}};
    \node[inner sep=0, anchor=south]  at (3,10.4) {\includegraphics[width=2\widthfigureresultsTwo]{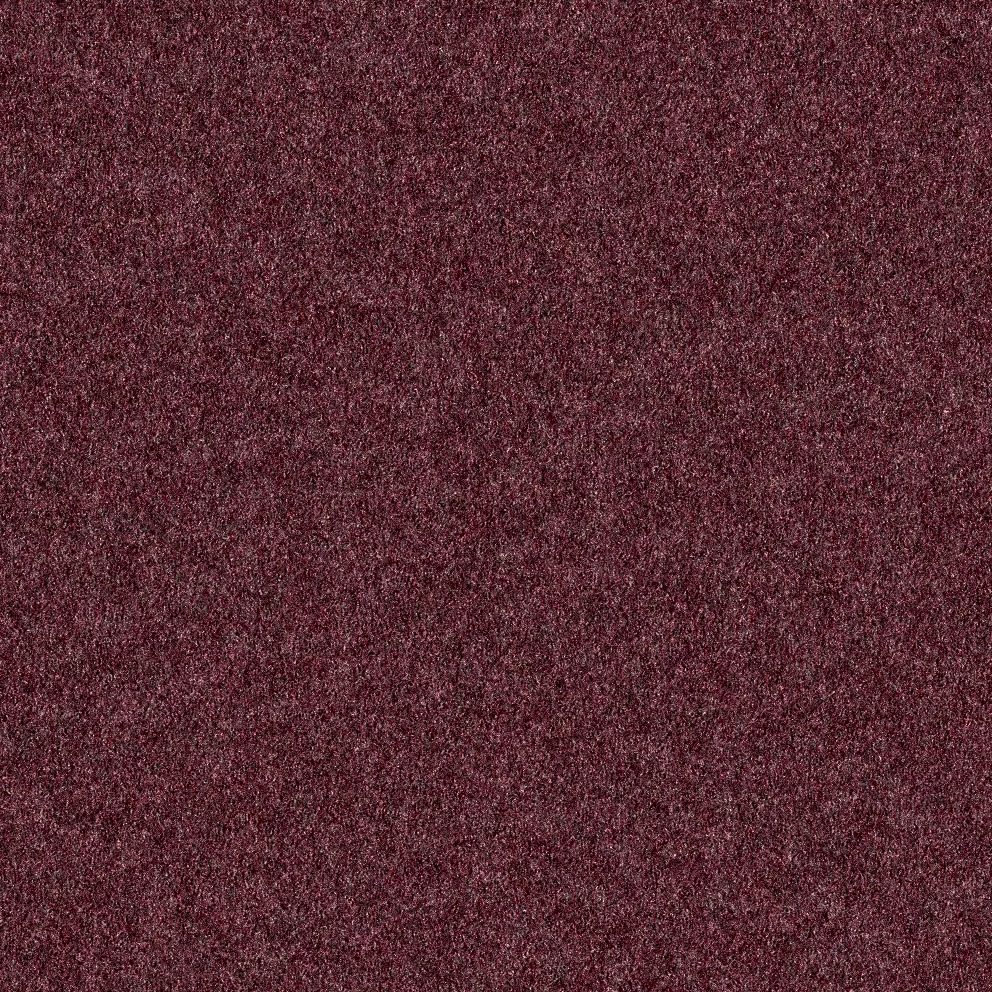}};
    \node[inner sep=0, anchor=south]  at (6,10.4) {\includegraphics[width=2\widthfigureresultsTwo]{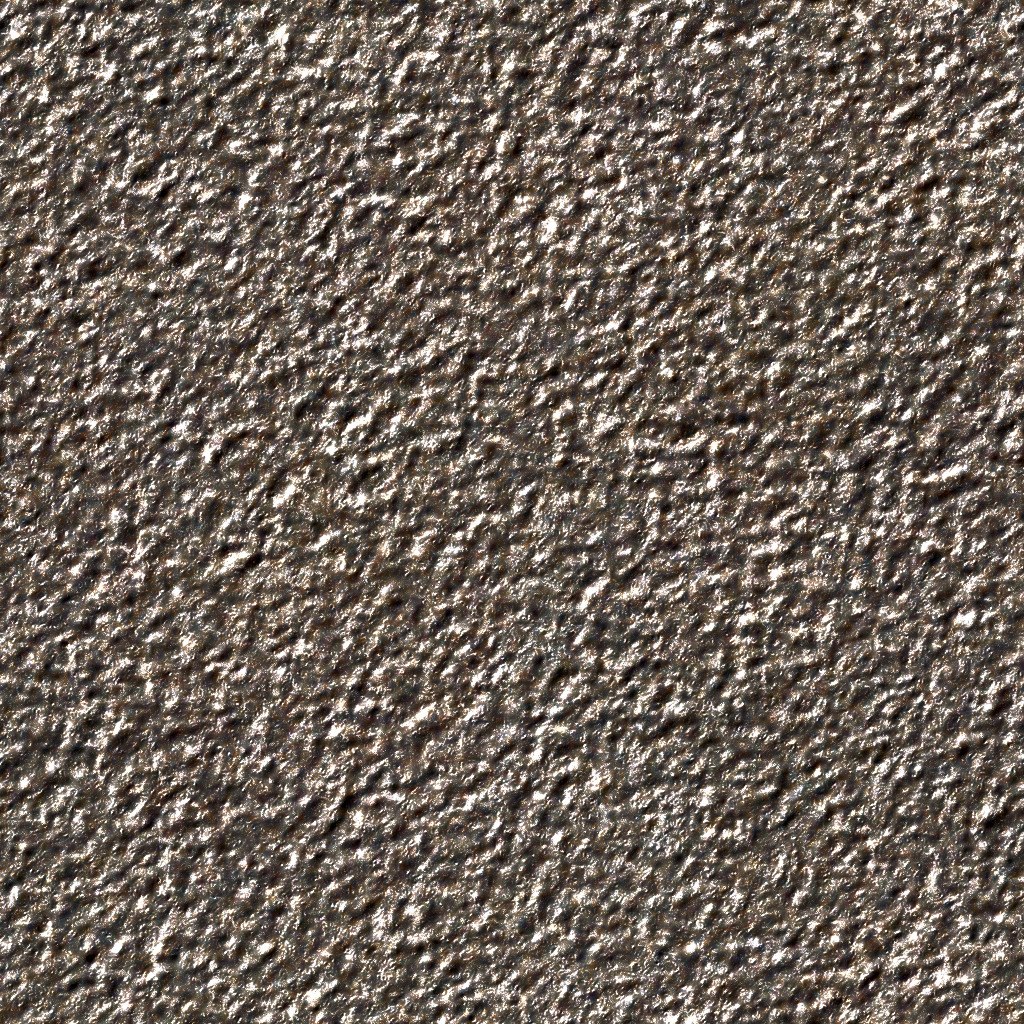}};
    \node[inner sep=0, anchor=south]  at (9,10.4) {\includegraphics[width=2\widthfigureresultsTwo]{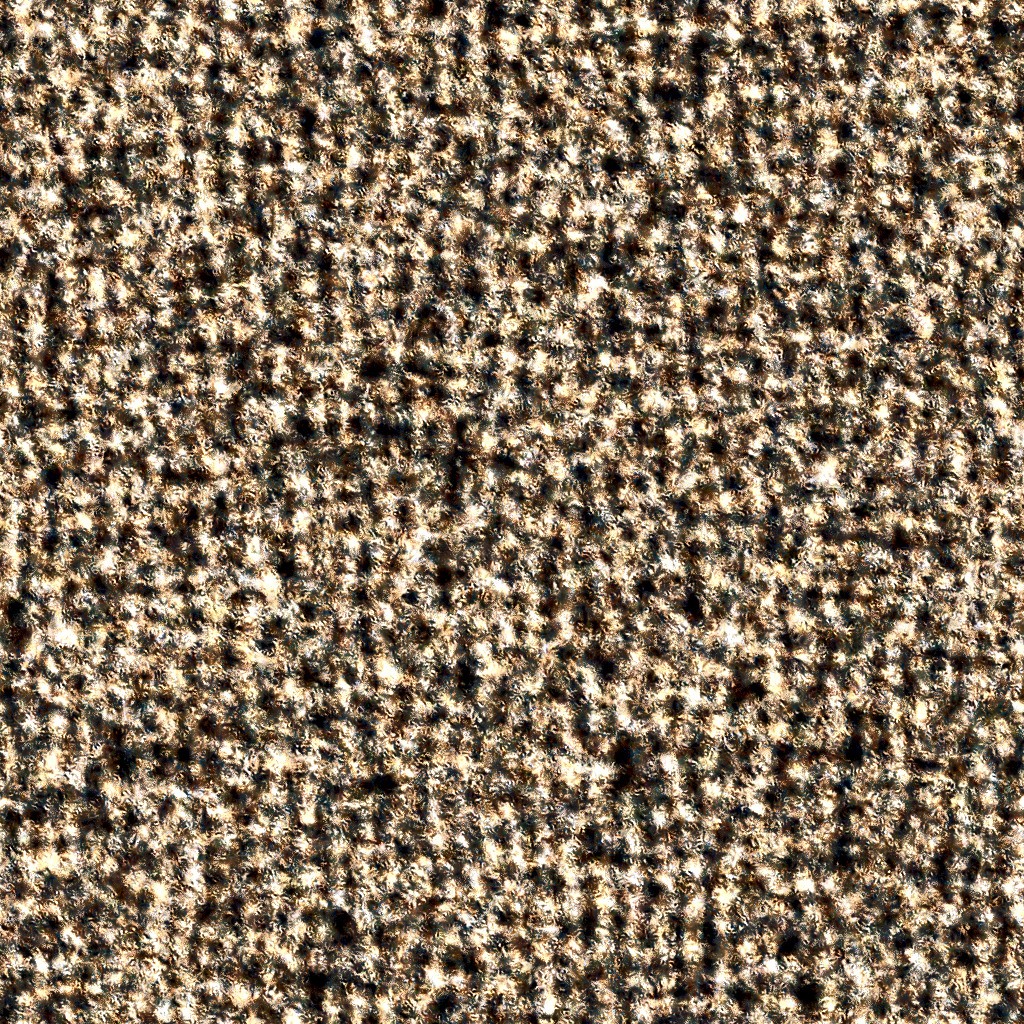}};

    \node[inner sep=0, anchor=south] (im5) at (0,7.8) {\includegraphics[width=2\widthfigureresultsTwo]{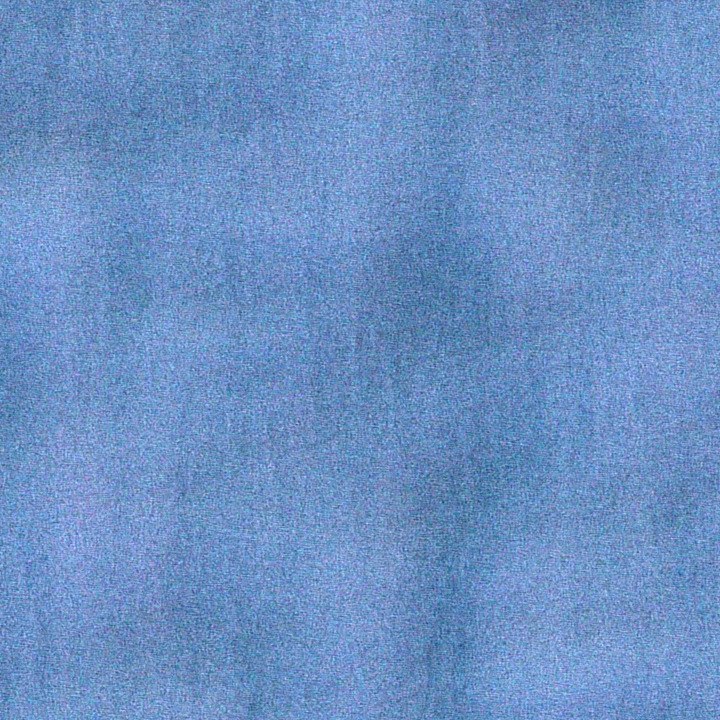}};
    \node[inner sep=0, anchor=south]  at (3,7.8) {\includegraphics[width=2\widthfigureresultsTwo]{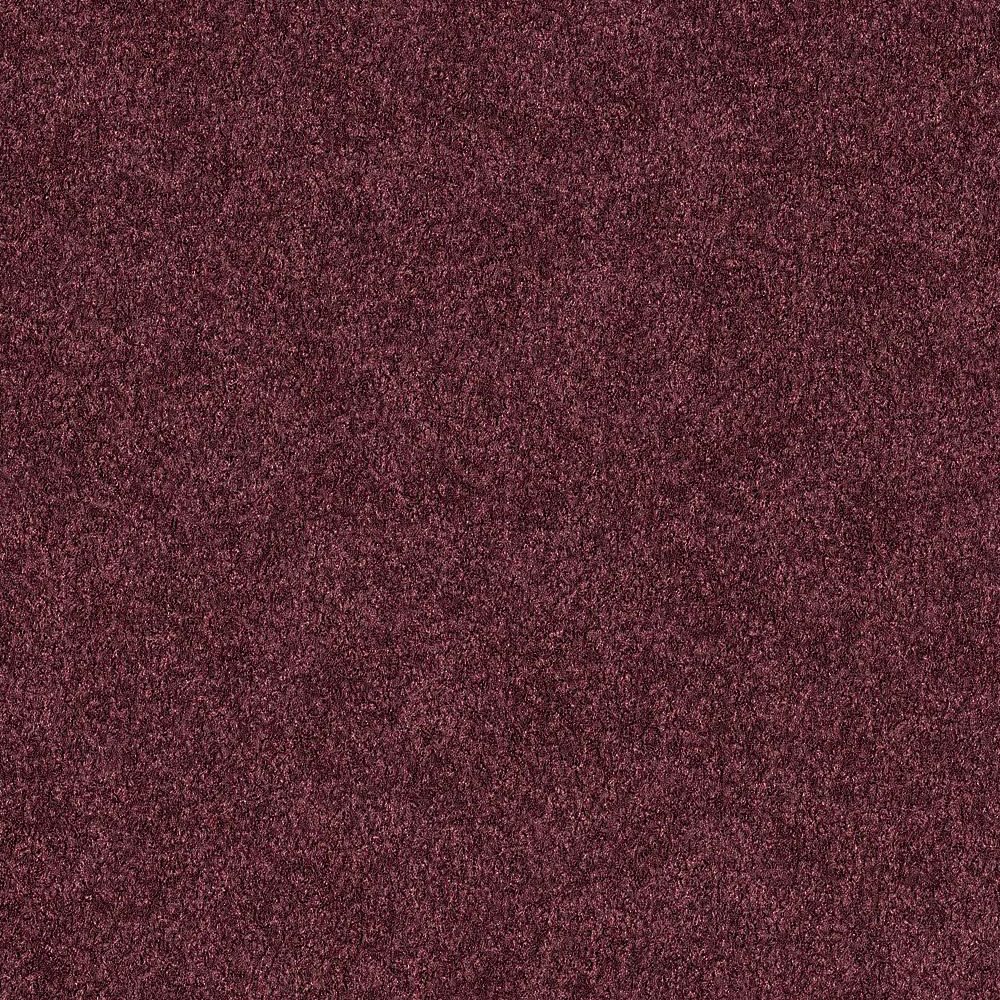}};
    \node[inner sep=0, anchor=south]  at (6,7.8) {\includegraphics[width=2\widthfigureresultsTwo]{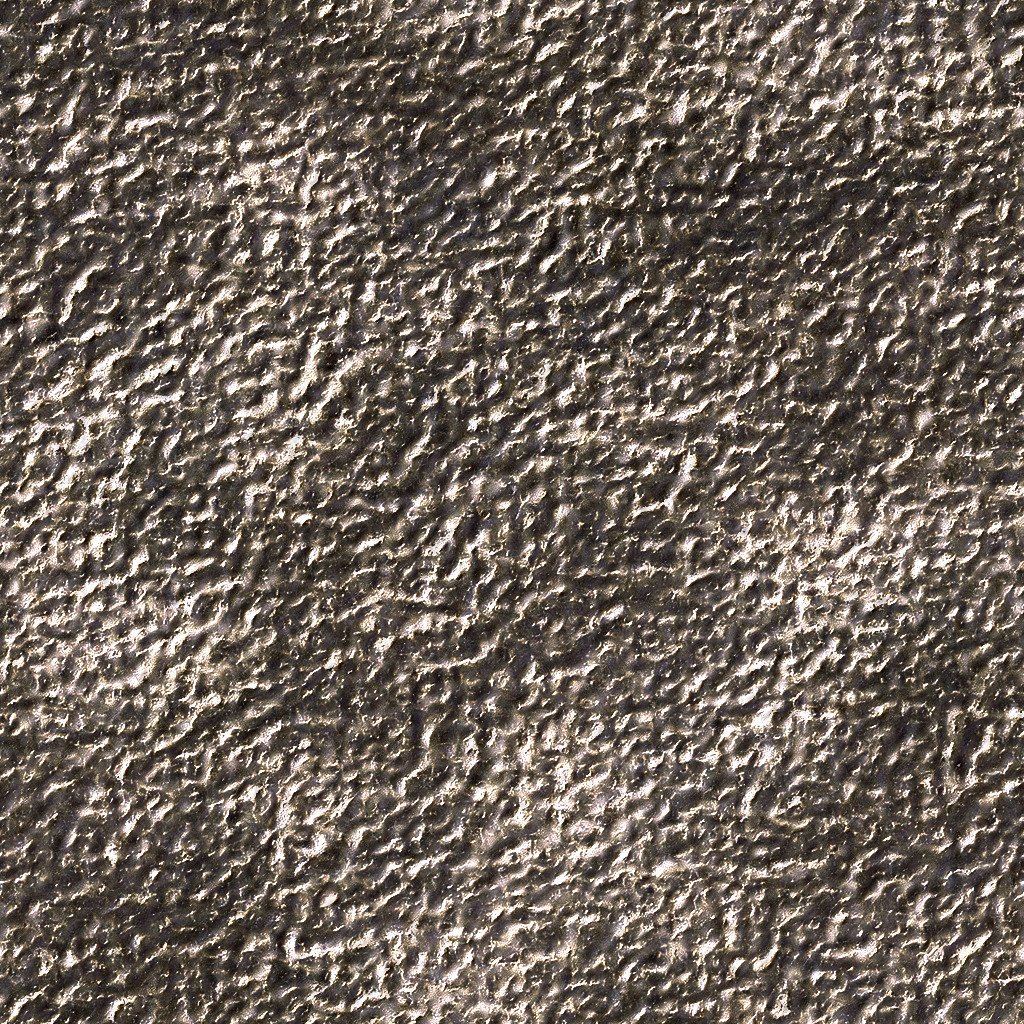}};
    \node[inner sep=0, anchor=south]  at (9,7.8) {\includegraphics[width=2\widthfigureresultsTwo]{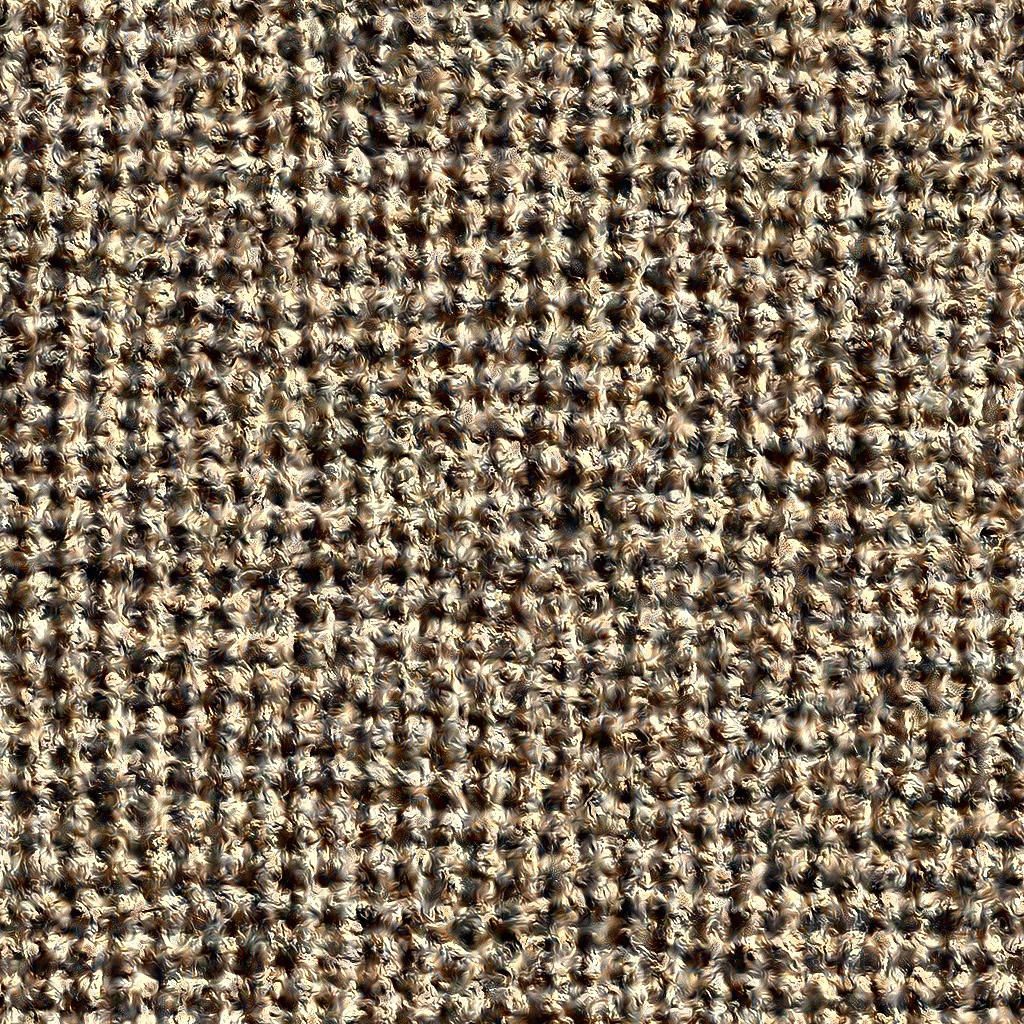}};

    \node[inner sep=0, anchor=south] (im4) at (0,5.2) {\includegraphics[width=2\widthfigureresultsTwo]{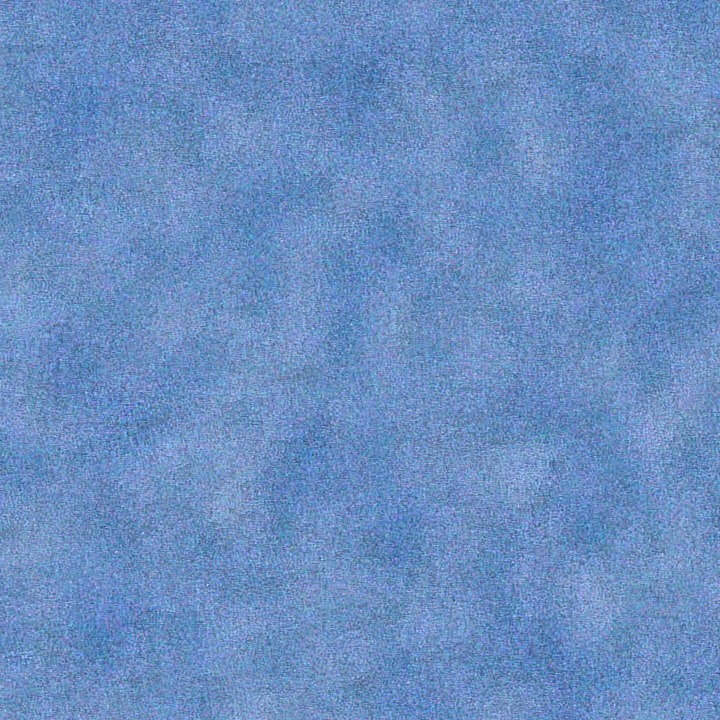}};
    \node[inner sep=0, anchor=south]  at (3,5.2) {\includegraphics[width=2\widthfigureresultsTwo]{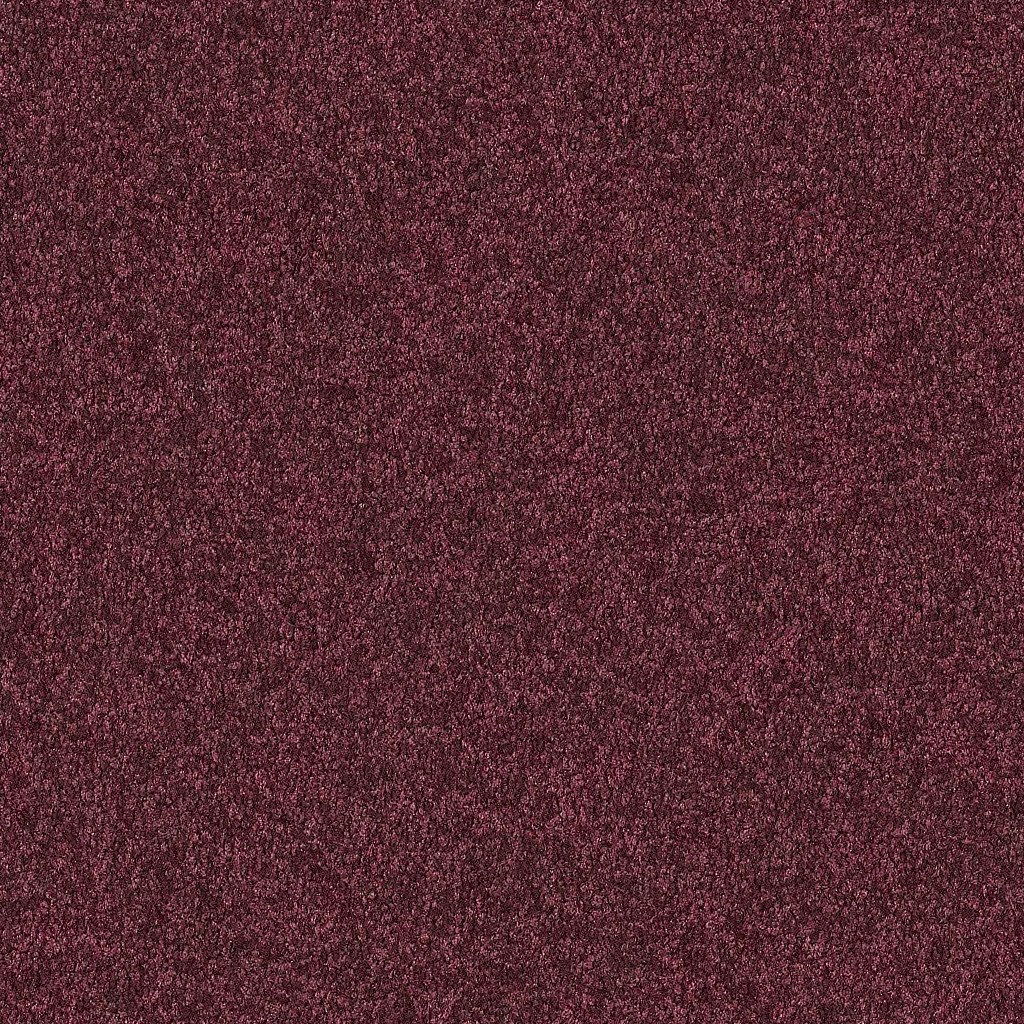}};
    \node[inner sep=0, anchor=south]  at (6,5.2) {\includegraphics[width=2\widthfigureresultsTwo]{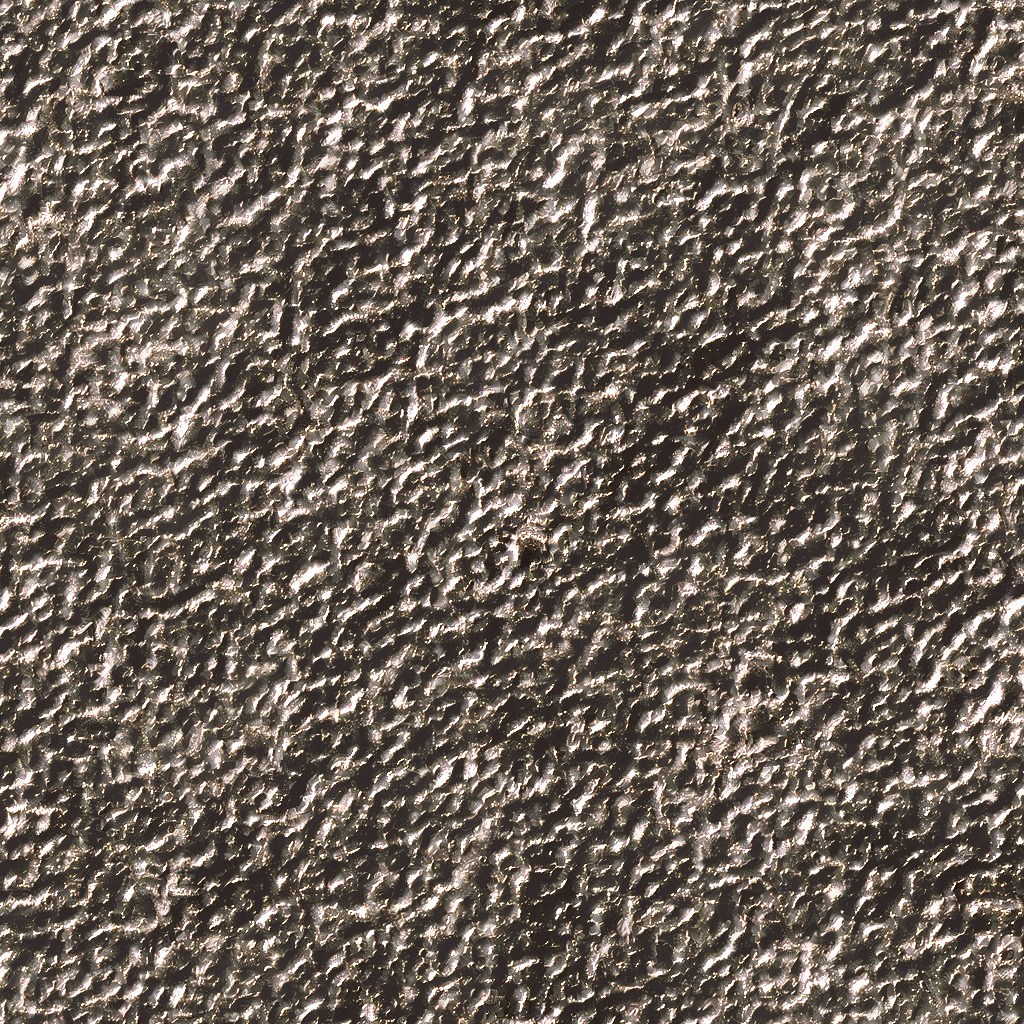}};
    \node[inner sep=0, anchor=south]  at (9,5.2) {\includegraphics[width=2\widthfigureresultsTwo]{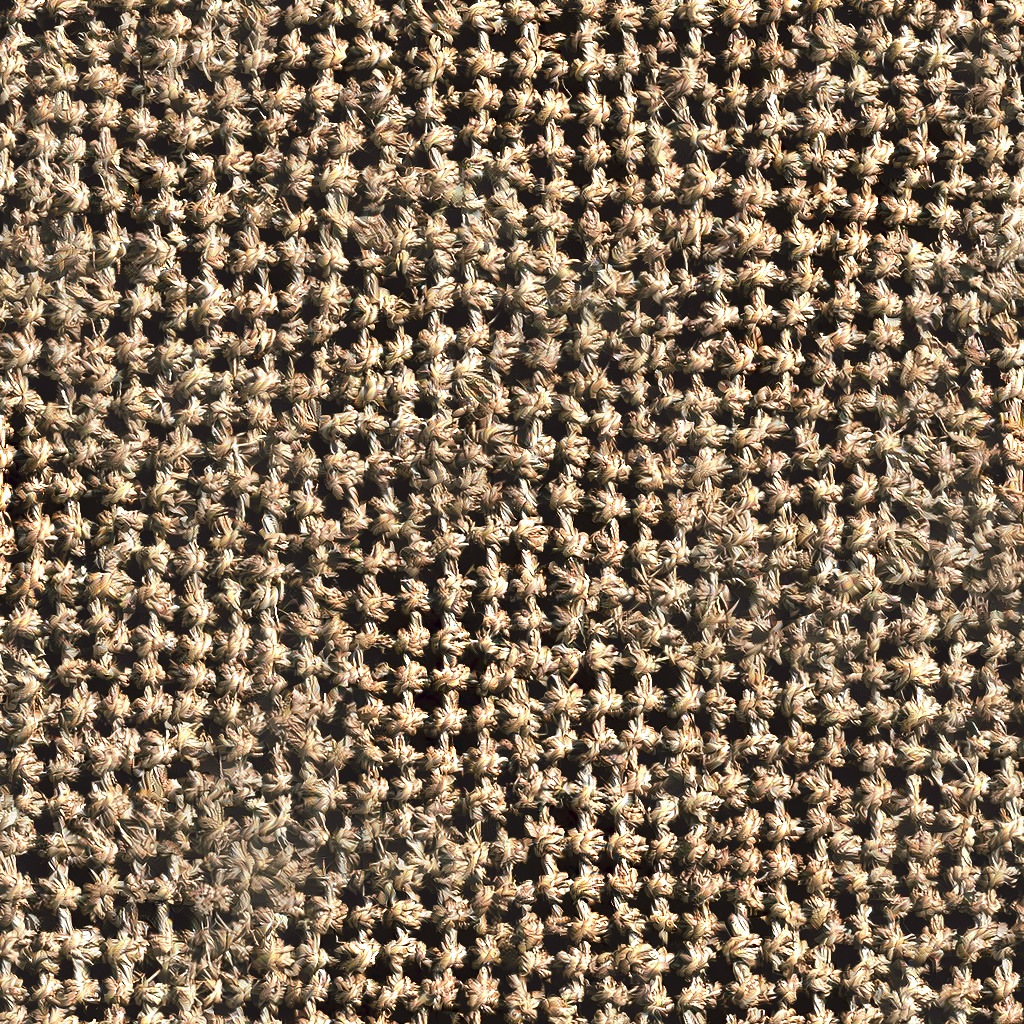}};

    \node[inner sep=0, anchor=south] (im3) at (0,2.6) {\includegraphics[width=2\widthfigureresultsTwo]{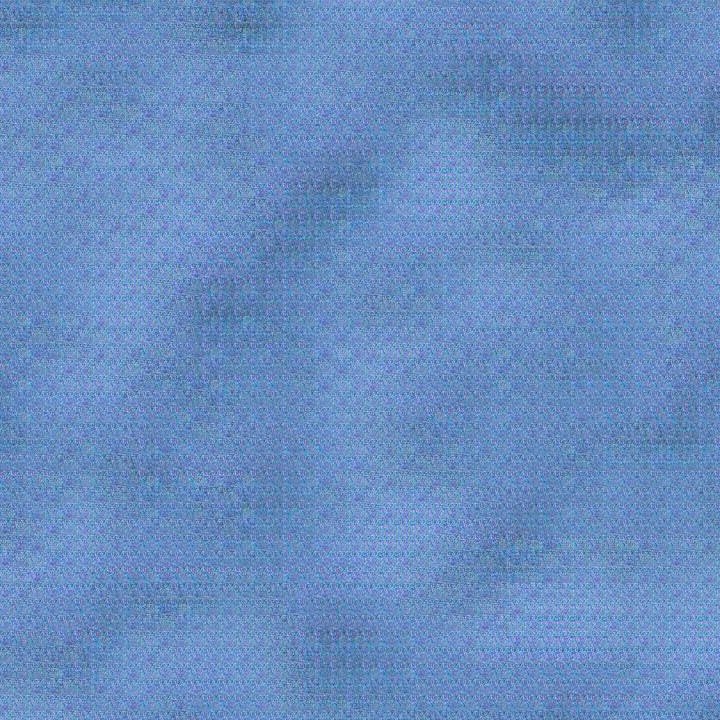}};
    \node[inner sep=0, anchor=south]  at (3,2.6) {\includegraphics[width=2\widthfigureresultsTwo]{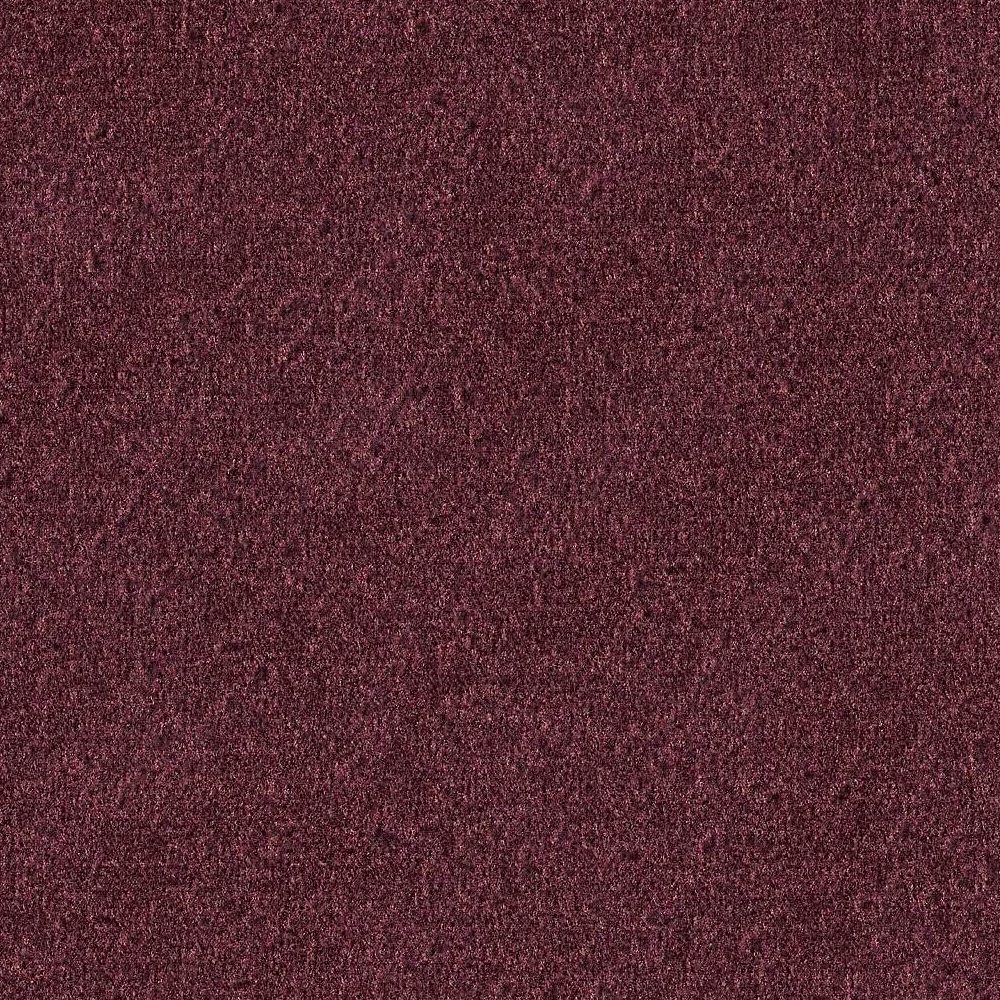}};
    \node[inner sep=0, anchor=south]  at (6,2.6) {\includegraphics[width=2\widthfigureresultsTwo]{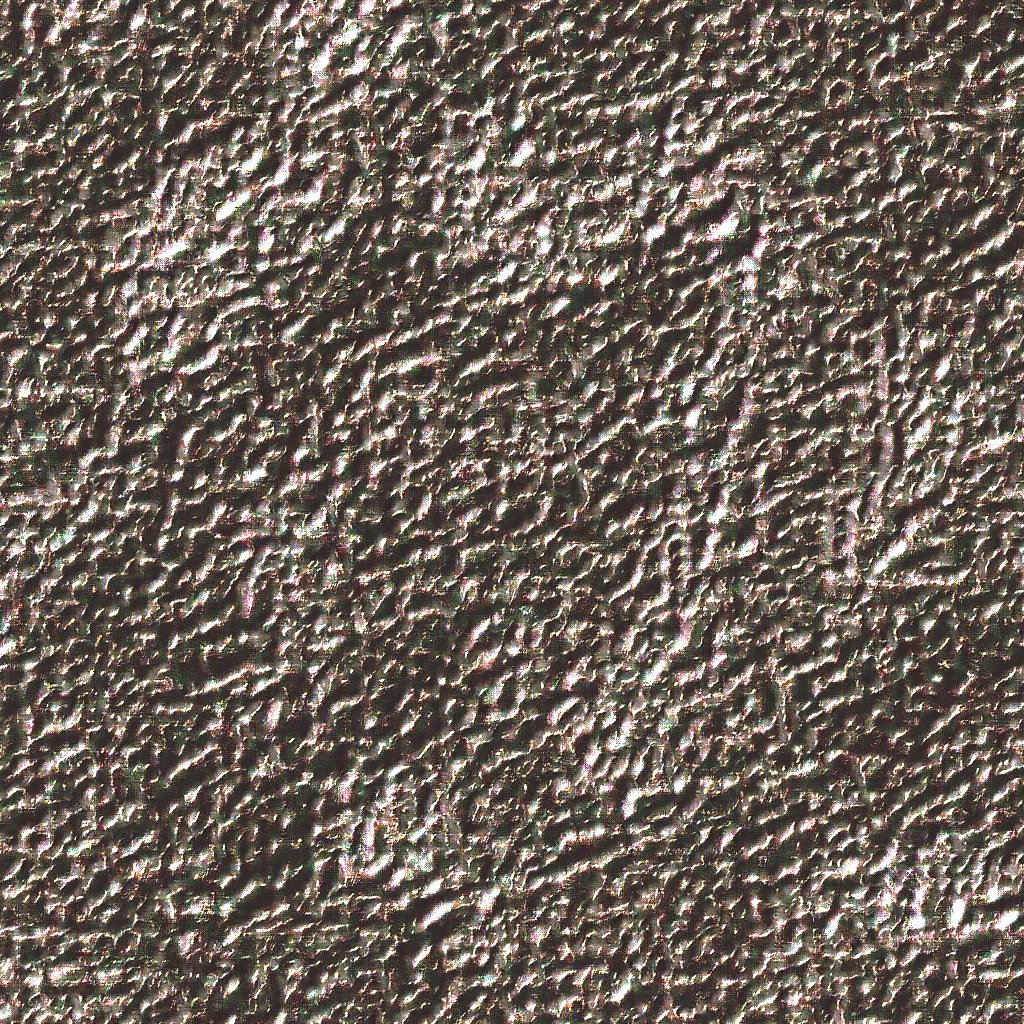}};
    \node[inner sep=0, anchor=south]  at (9,2.6) {\includegraphics[width=2\widthfigureresultsTwo]{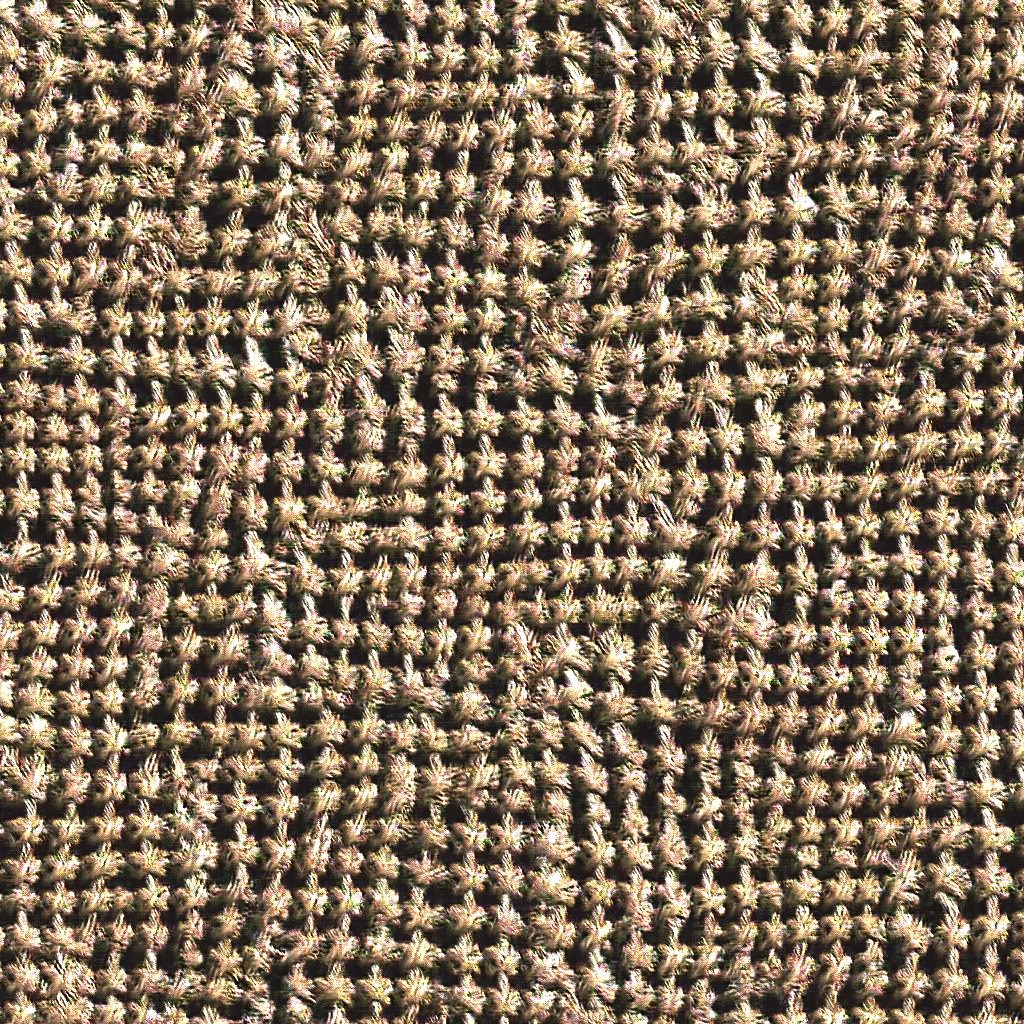}};

    \node [anchor=east] at (im3.west) {SGAN \cite{jetchev2016texture}};
    \node [anchor=east] at (im4.west) {Gatys \cite{gatys}};
    \node [anchor=east] at (im5.west) {PS \cite{PS}};
    \node [anchor=east] at (im6.west) {HB \cite{HeegerBergen}};
    \node [anchor=east] at (im7.west) {RPN \cite{GalerneRPN}};
    \node [anchor=east] at (im8.west) {input};
  \end{tikzpicture}
  \caption{\emph{Comparison of texture synthesis methods.} From top to bottom: input sample, Random Phase Noise (RPN)~\cite{GalerneRPN}, Heeger and Bergen (HB)~\cite{HeegerBergen}, Portilla and Simoncelli (PS)~\cite{PS}, Gatys (Gatys)~\cite{gatys} and SGAN \cite{jetchev2016texture}.}
  \label{fig:algComp1-1}
\end{figure}


\begin{figure}[p]
  \centering
  \begin{tikzpicture}[scale=.9]

    \node[inner sep=0, anchor=south east] (im4) at (0,2.6) {\includegraphics[width=\widthfigureresultsTwo]{figures/tissus27}};
    \node[inner sep=0, anchor=south east]  at (3,2.6) {\includegraphics[width=\widthfigureresultsTwo]{figures/moquette}};
    \node[inner sep=0, anchor=south east]  at (6,2.6) {\includegraphics[width=\widthfigureresultsTwo]{figures/Metal_0005}};
    \node[inner sep=0, anchor=south east]  at (9,2.6) {\includegraphics[width=\widthfigureresultsTwo]{figures/Fabric_0008}};

    \node[inner sep=0, anchor=south] (im3) at (0,0) {\includegraphics[width=2\widthfigureresultsTwo]{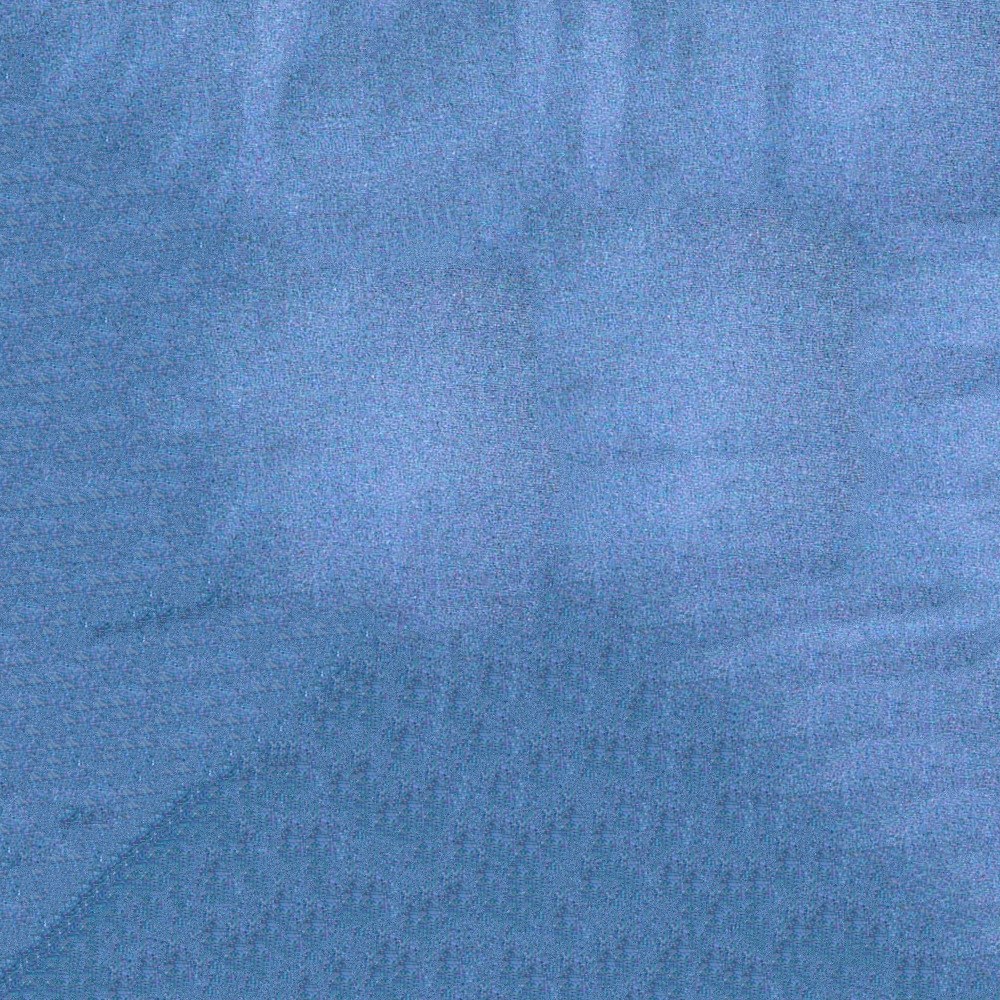}};
    \node[inner sep=0, anchor=south]  at (3,0) {\includegraphics[width=2\widthfigureresultsTwo]{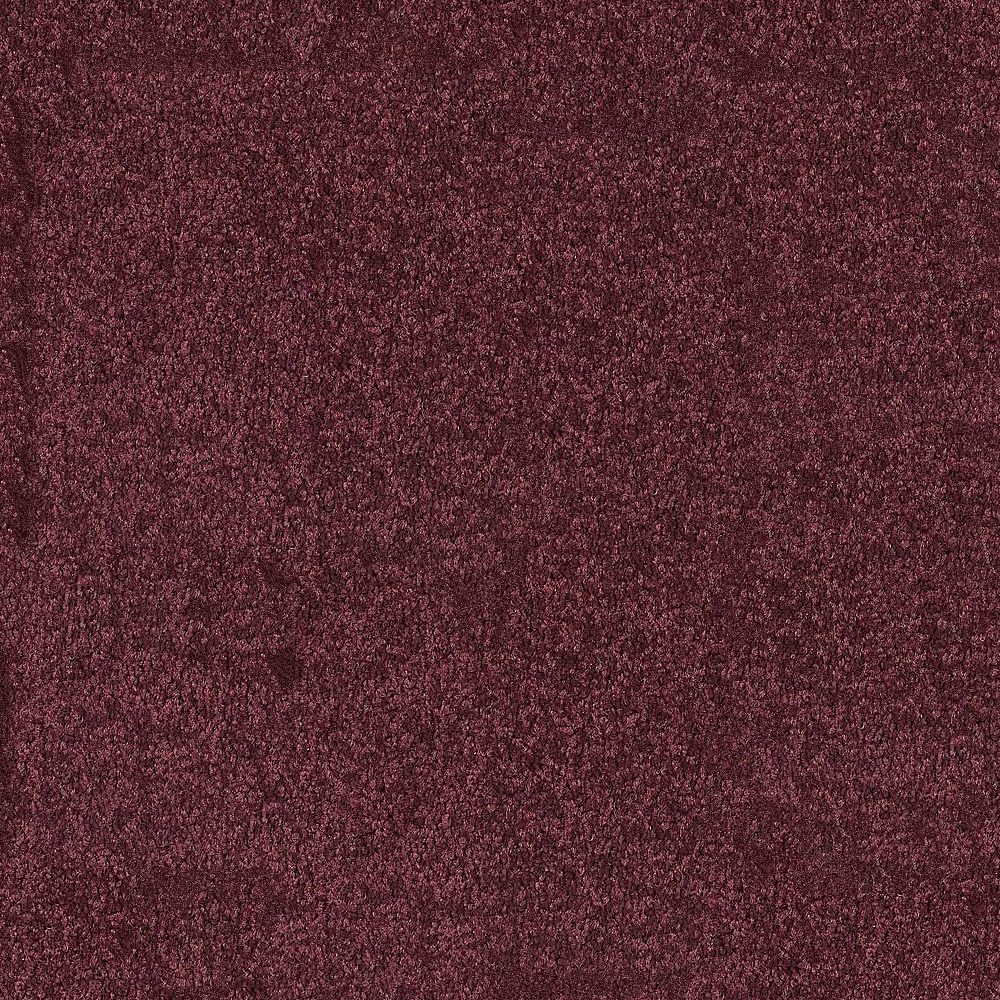}};
    \node[inner sep=0, anchor=south]  at (6,0) {\includegraphics[width=2\widthfigureresultsTwo]{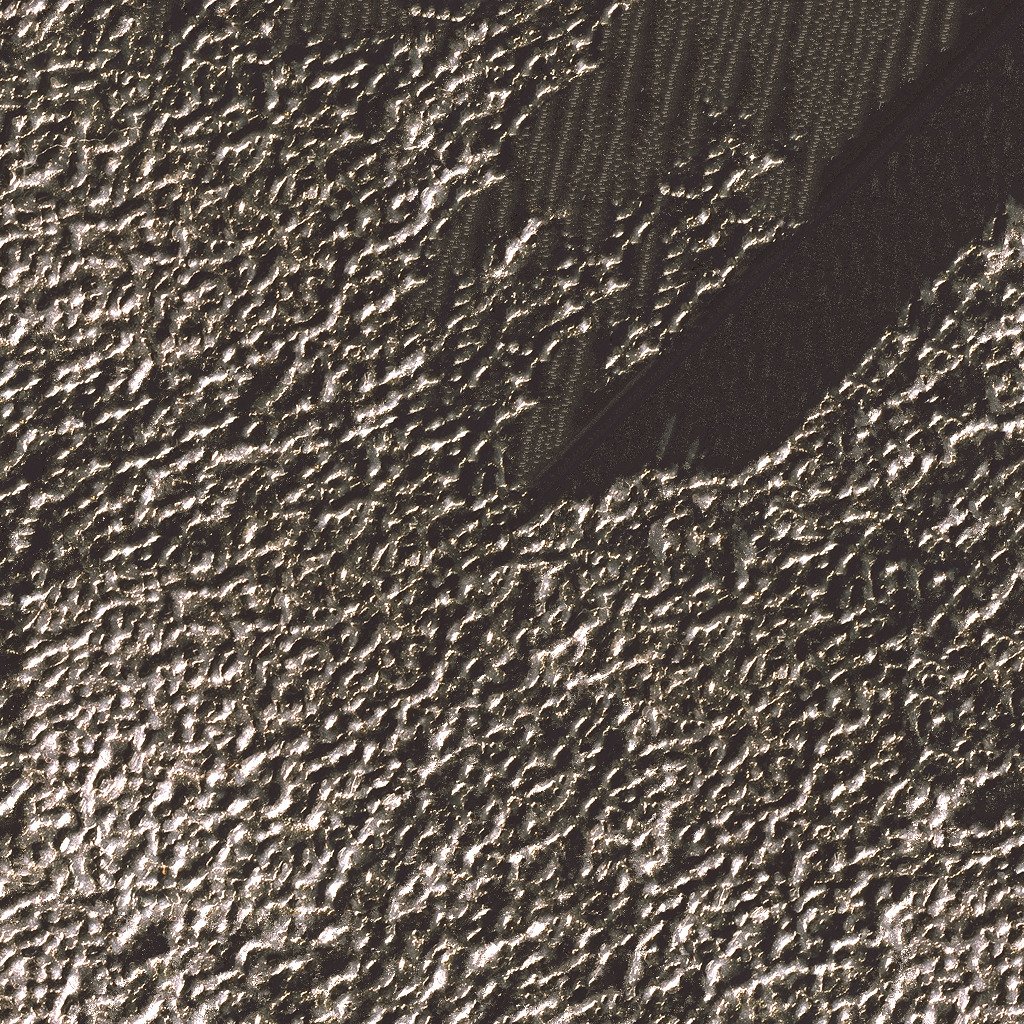}};
    \node[inner sep=0, anchor=south]  at (9,0) {\includegraphics[width=2\widthfigureresultsTwo]{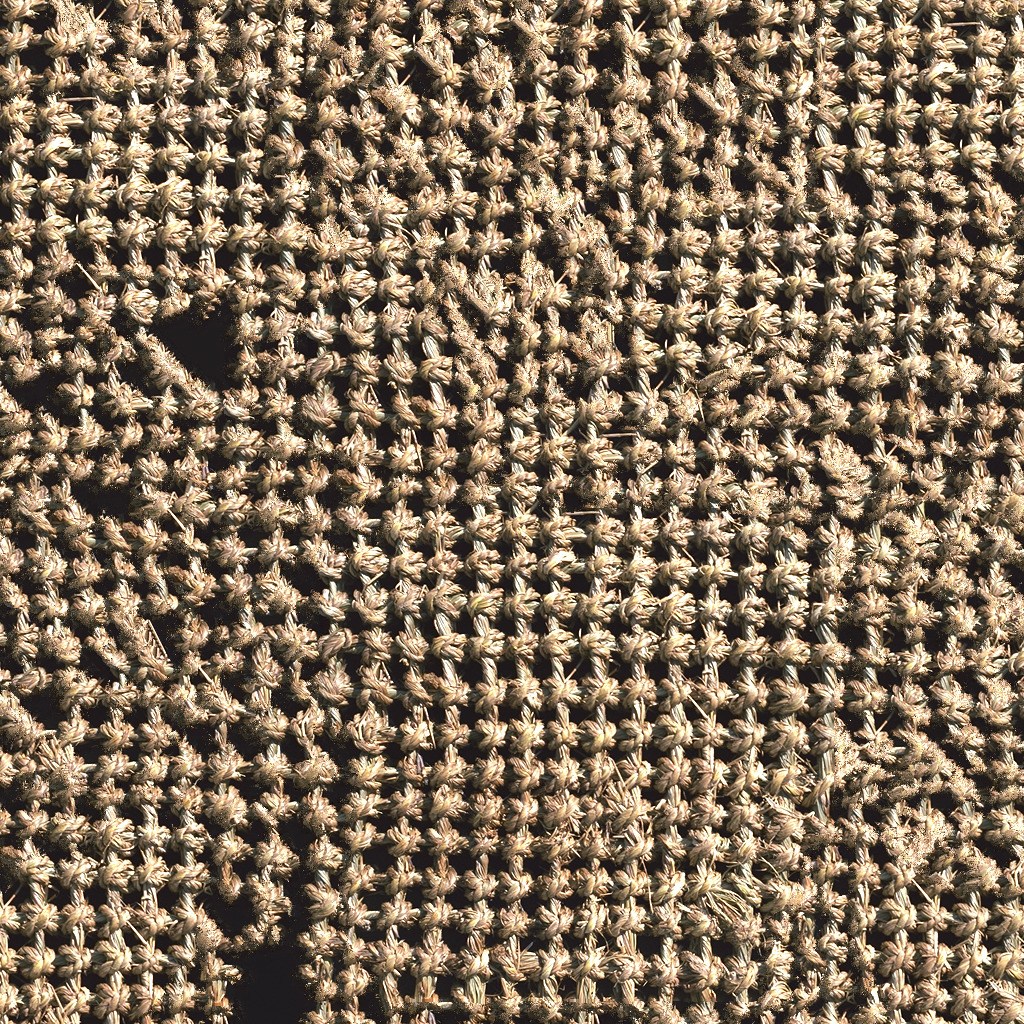}};

    \node[inner sep=0, anchor=south] (im2) at (0,-2.6) {\includegraphics[width=2\widthfigureresultsTwo]{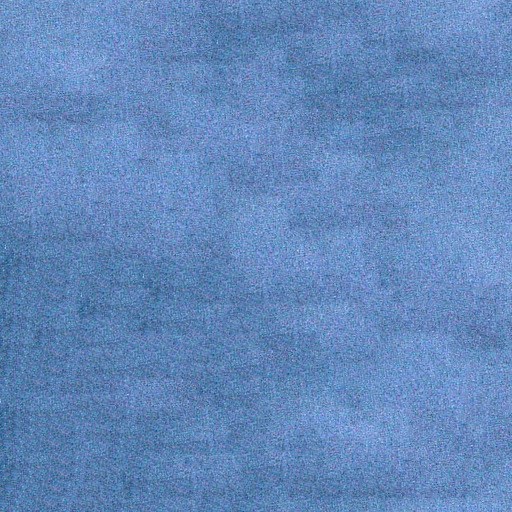}};
    \node[inner sep=0, anchor=south]  at (3,-2.6) {\includegraphics[width=2\widthfigureresultsTwo]{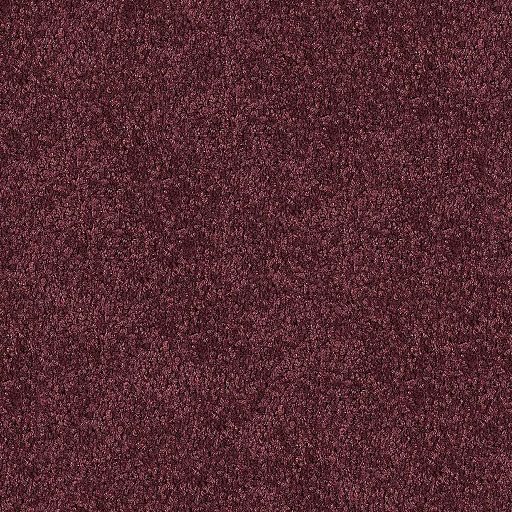}};
    \node[inner sep=0, anchor=south]  at (6,-2.6) {\includegraphics[width=2\widthfigureresultsTwo]{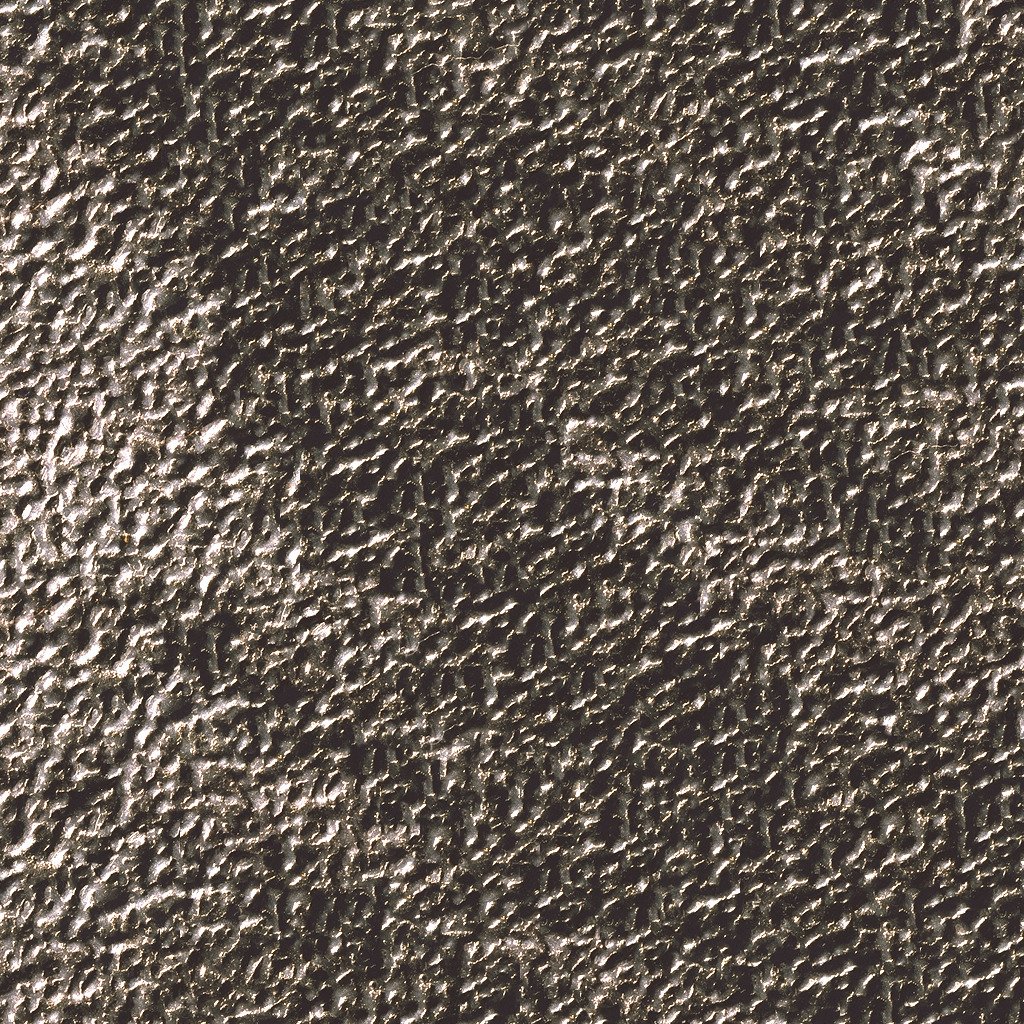}};
    \node[inner sep=0, anchor=south]  at (9,-2.6) {\includegraphics[width=2\widthfigureresultsTwo]{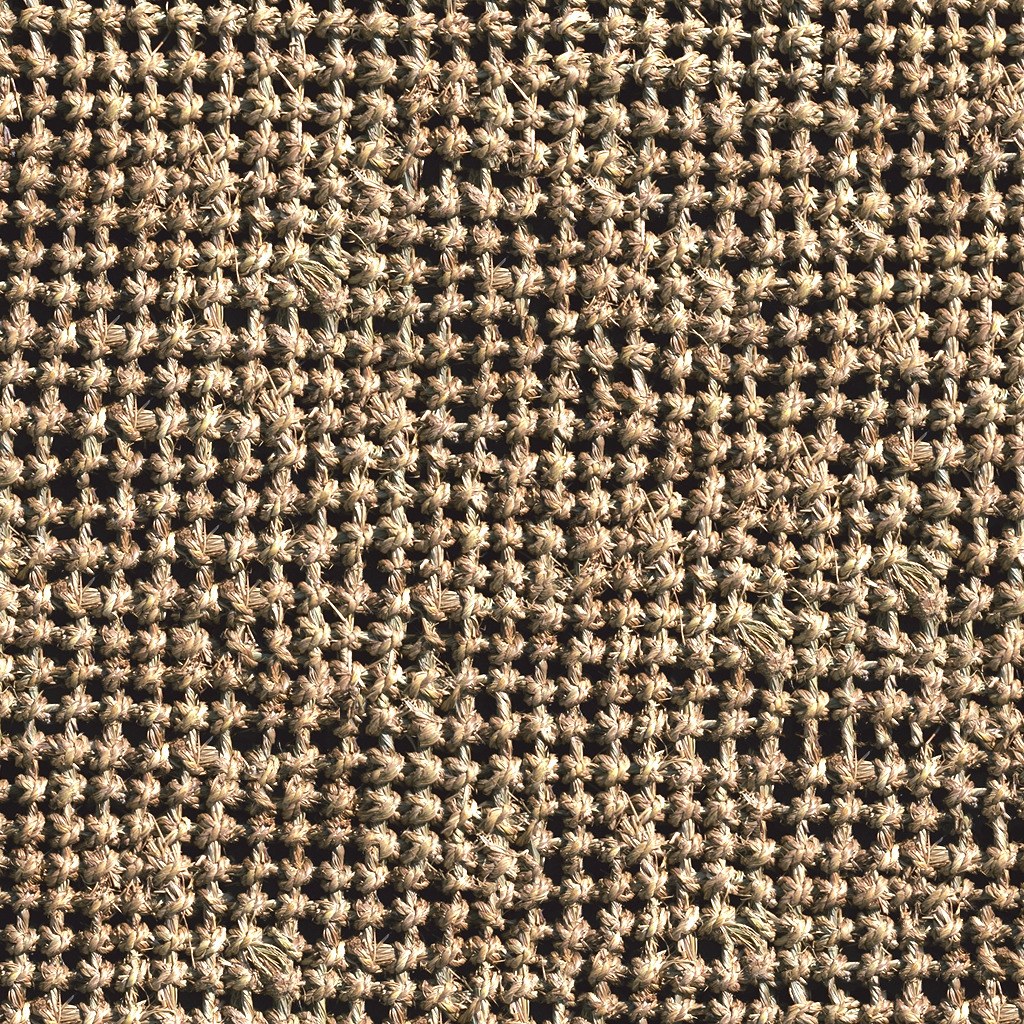}};

    \node[inner sep=0, anchor=south] (im1) at (0,-5.2) {\includegraphics[width=2\widthfigureresultsTwo]{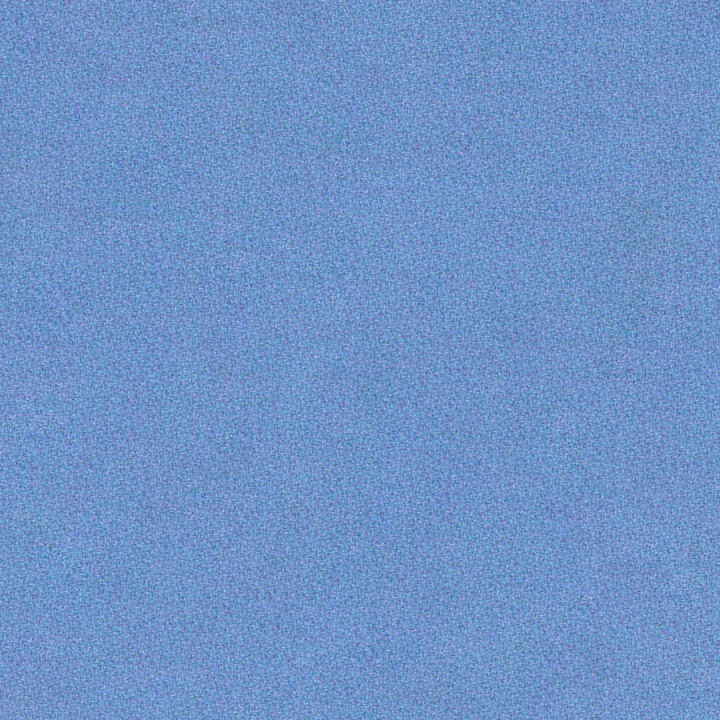}};
    \node[inner sep=0, anchor=south]  at (3,-5.2) {\includegraphics[width=2\widthfigureresultsTwo]{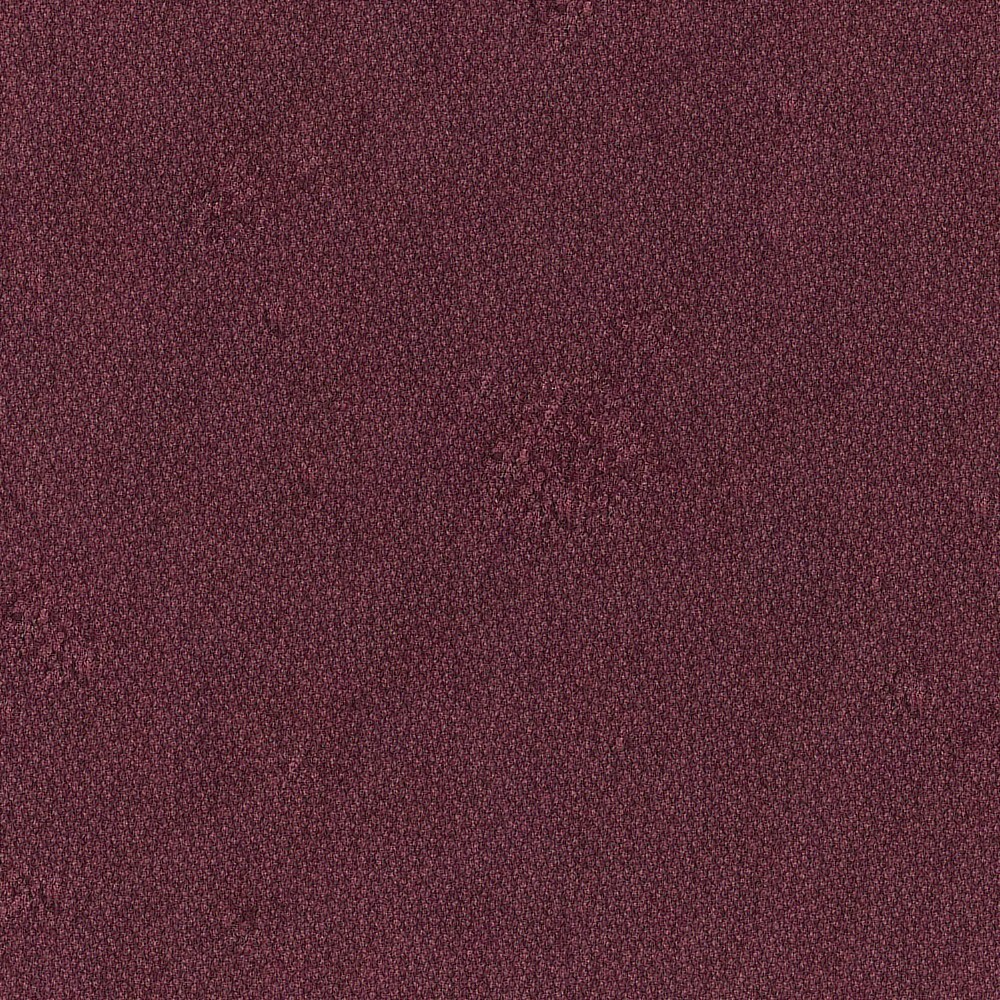}};
    \node[inner sep=0, anchor=south]  at (6,-5.2) {\includegraphics[width=2\widthfigureresultsTwo]{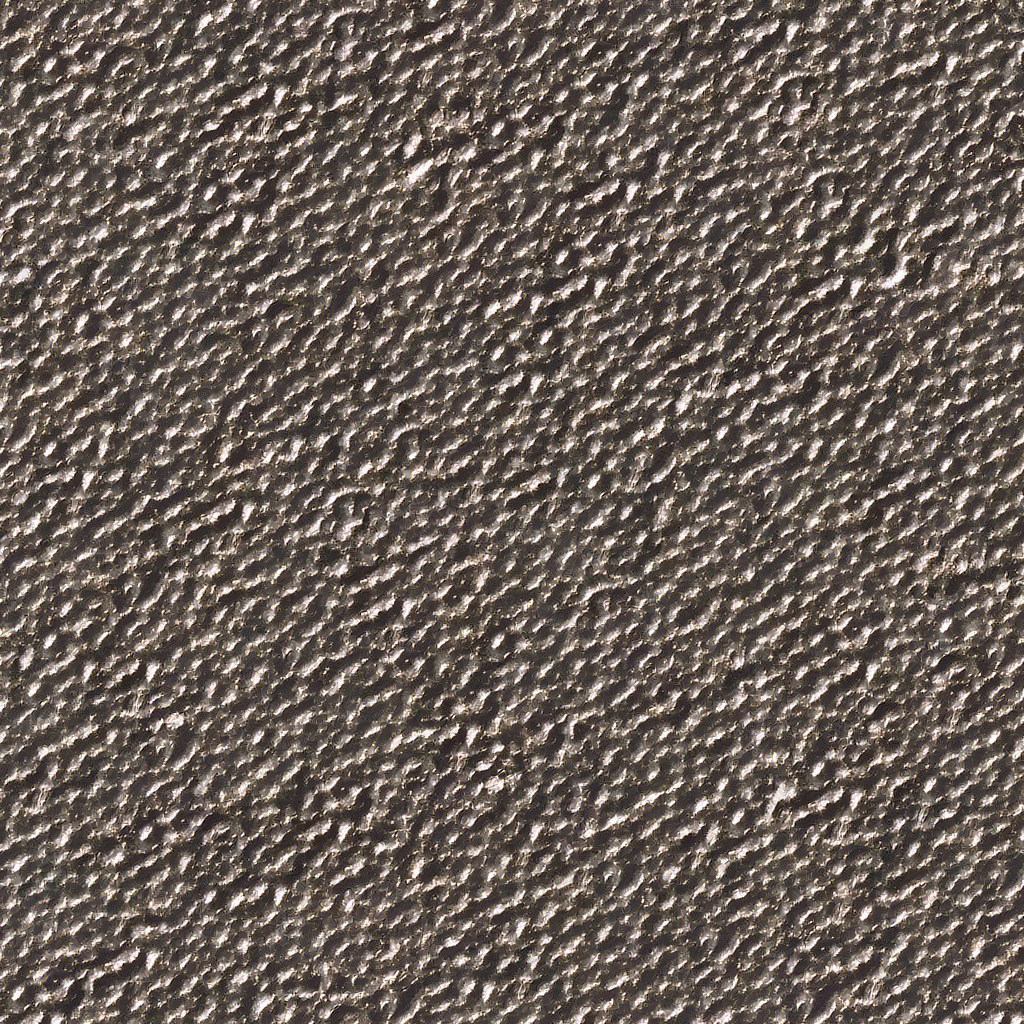}};
    \node[inner sep=0, anchor=south]  at (9,-5.2) {\includegraphics[width=2\widthfigureresultsTwo]{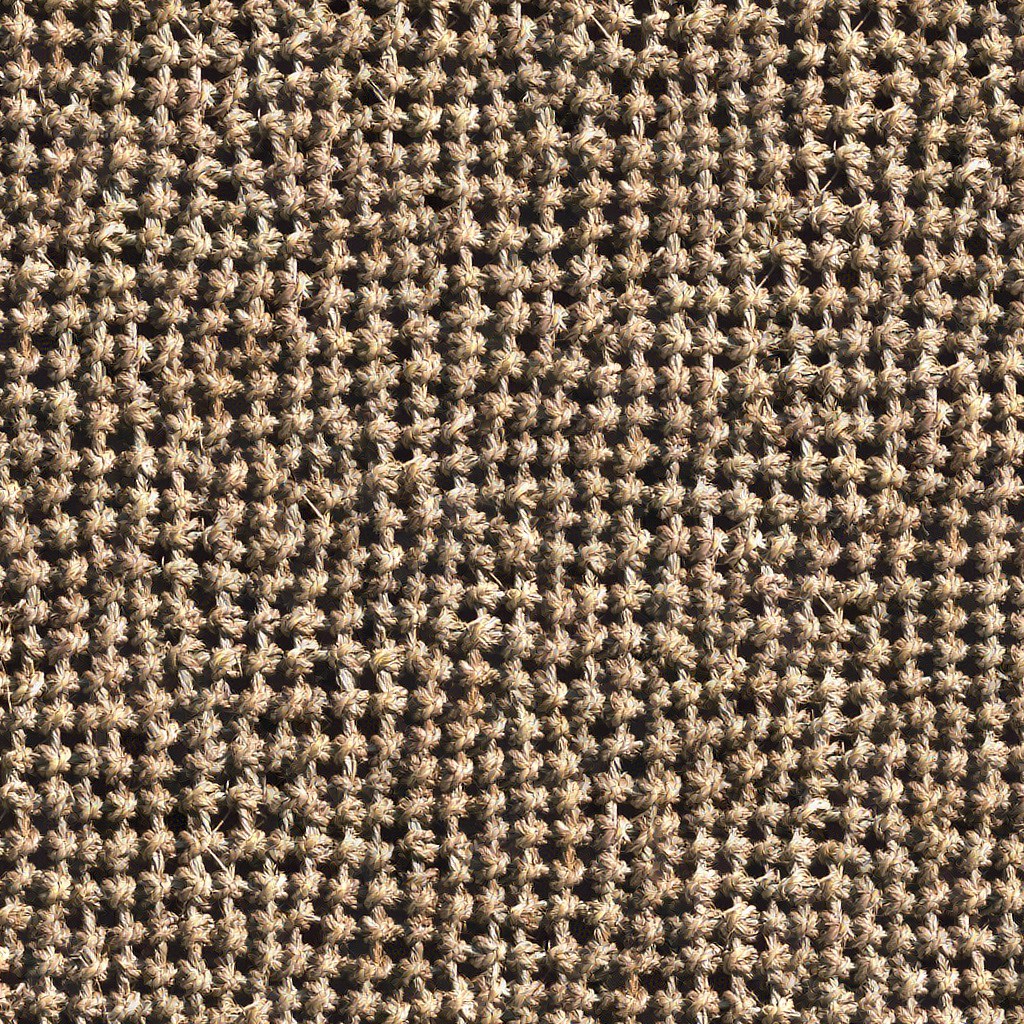}};

    \node[inner sep=0, anchor=south] (im0) at (0,-7.8) {\includegraphics[width=2\widthfigureresultsTwo]{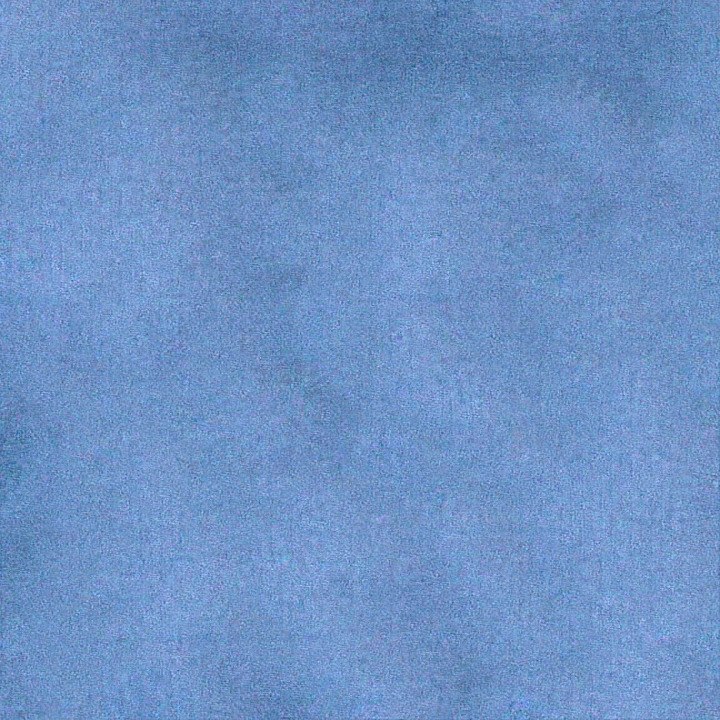}};
    \node[inner sep=0, anchor=south]  at (3,-7.8) {\includegraphics[width=2\widthfigureresultsTwo]{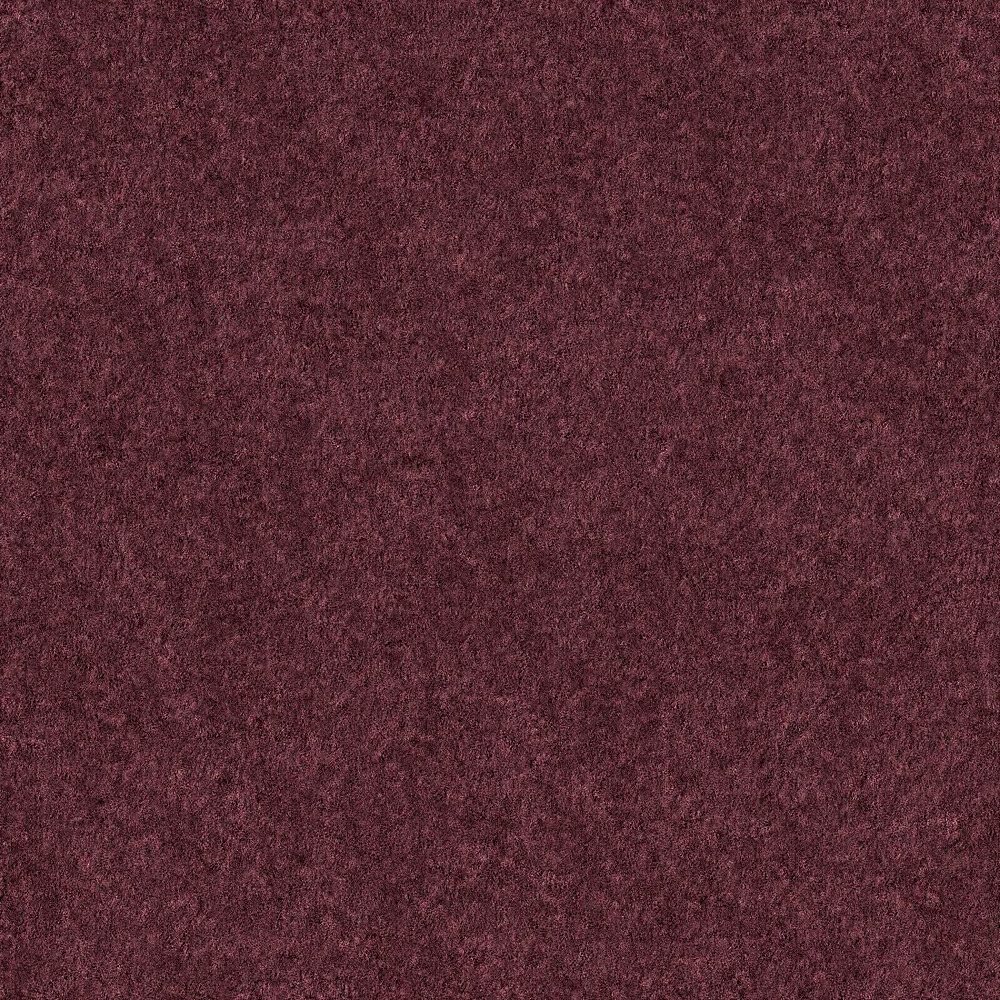}};
    \node[inner sep=0, anchor=south]  at (6,-7.8) {\includegraphics[width=2\widthfigureresultsTwo]{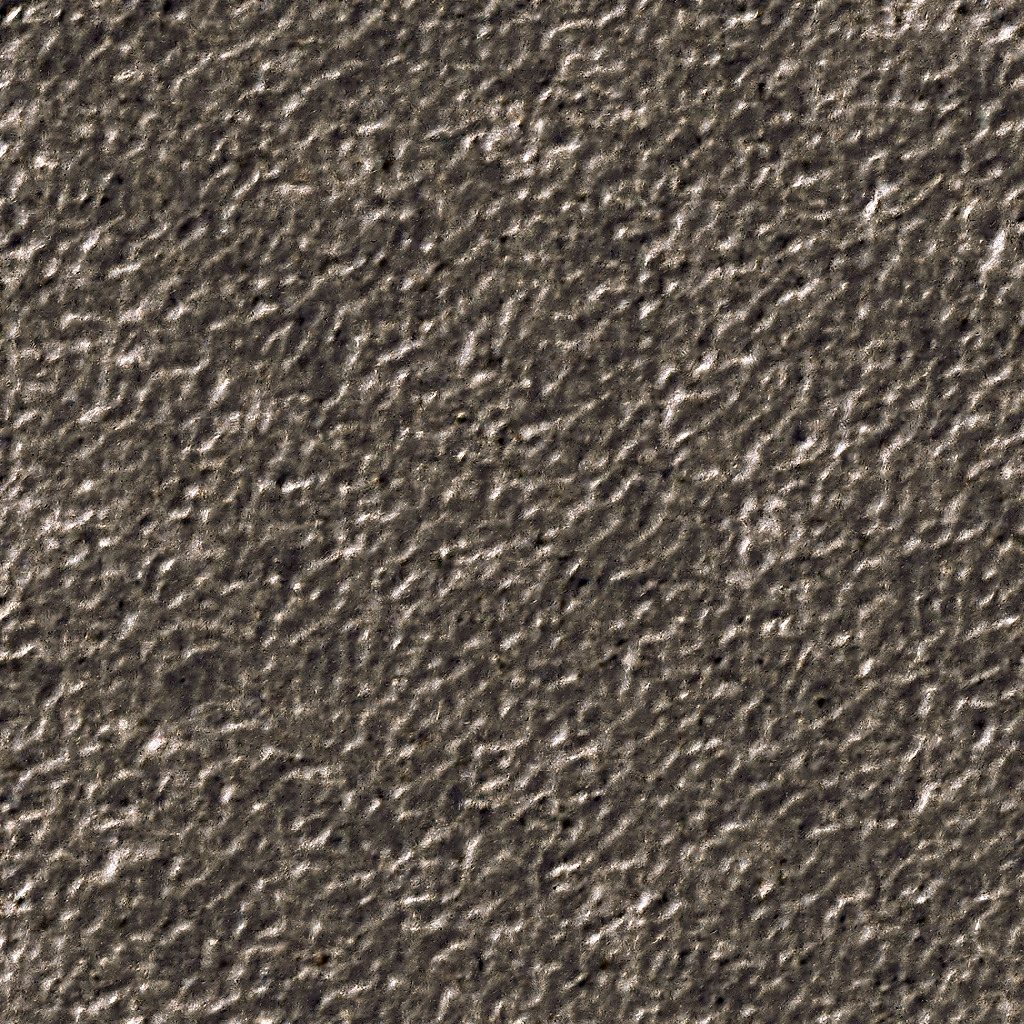}};
    \node[inner sep=0, anchor=south]  at (9,-7.8) {\includegraphics[width=2\widthfigureresultsTwo]{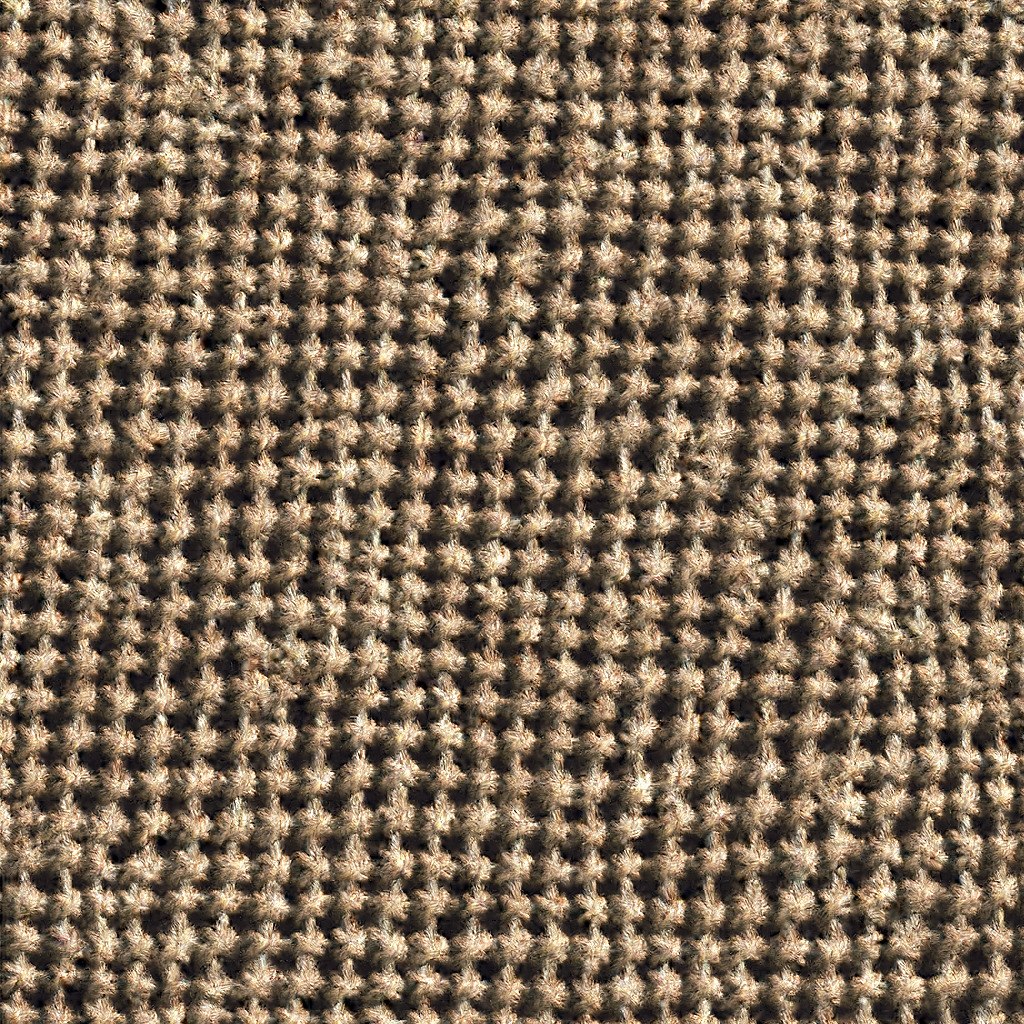}};

    \node [anchor=east] at (im0.west) {MSLG \cite{raad2016}};
    \node [anchor=east] at (im1.west) {CNNMRF \cite{li2016combining}};
    \node [anchor=east] at (im2.west) {EF \cite{EfrosFreeman}};
    \node [anchor=east] at (im3.west) {EL \cite{EfrosLeung}};
    \node [anchor=east] at (im4.west) {input};
  \end{tikzpicture}
  \caption{\emph{Comparison of texture synthesis methods.} From top to bottom: input sample, Efros and Leung (EL)~\cite{EfrosLeung}, Efros and Freeman (EF)~\cite{EfrosFreeman}, CNNMRF~\cite{li2016combining} and MSLG~\cite{raad2016}.}
  \label{fig:algComp1-2}
\end{figure}

\begin{figure}[p]
  \centering
  \begin{tikzpicture}[scale=.9]

    \node[inner sep=0, anchor=south east] (im8) at (0,15.6) {\includegraphics[width=\widthfigureresultsTwo]{figures/Fabric_0000}};
    \node[inner sep=0, anchor=south east]  at (3,15.6) {\includegraphics[width=\widthfigureresultsTwo]{figures/Food_0008}};
    \node[inner sep=0, anchor=south east]  at (6,15.6) {\includegraphics[width=\widthfigureresultsTwo]{figures/Flowers_0000}};
    \node[inner sep=0, anchor=south east]  at (9,15.6) {\includegraphics[width=\widthfigureresultsTwo]{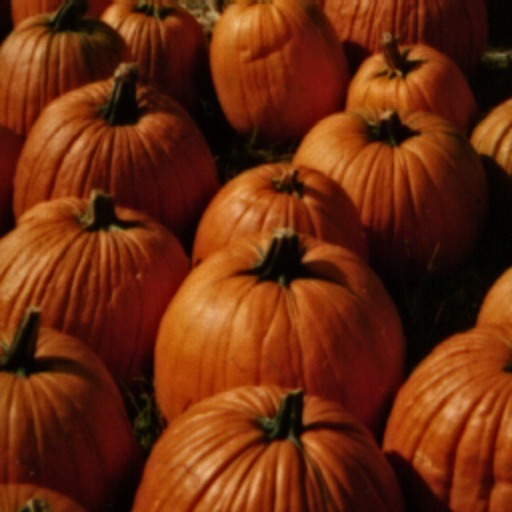}};

    \node[inner sep=0, anchor=south] (im7) at (0,13) {\includegraphics[width=2\widthfigureresultsTwo]{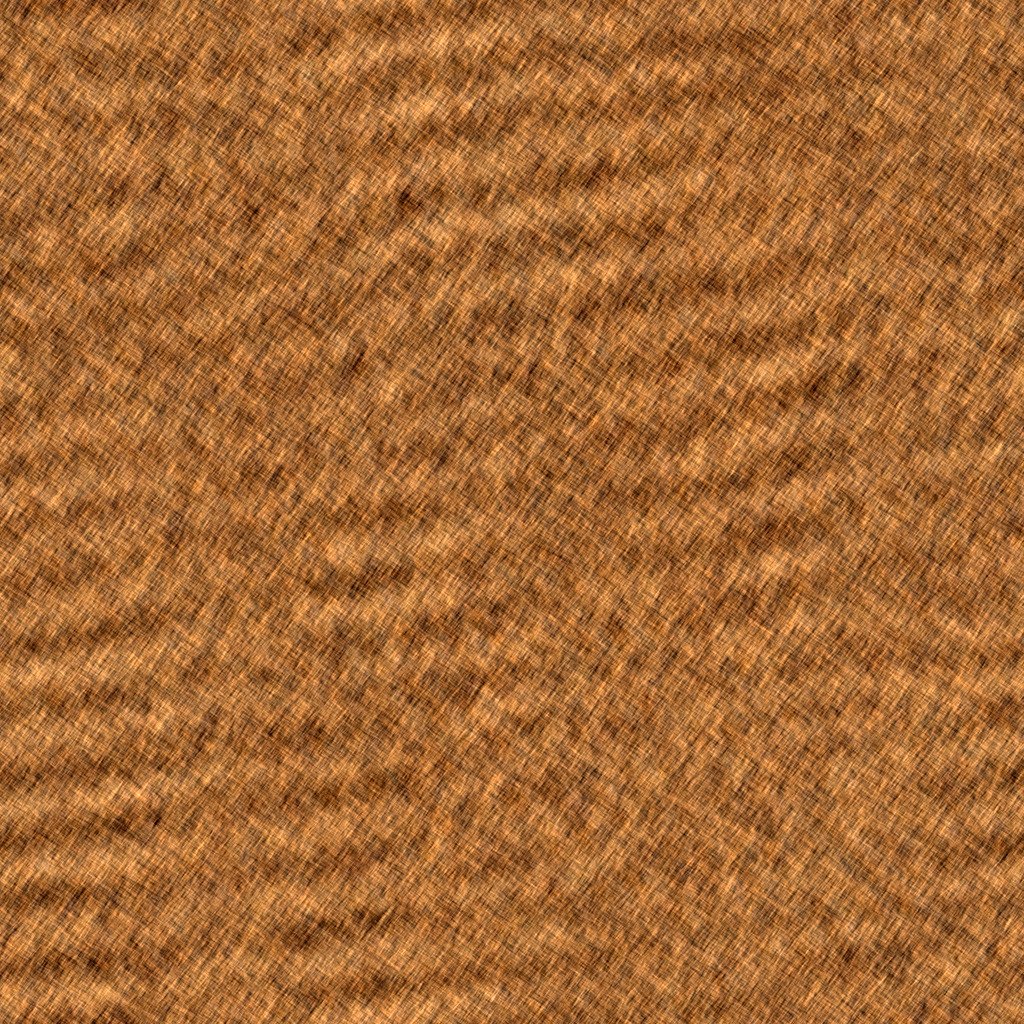}};
    \node[inner sep=0, anchor=south]  at (3,13) {\includegraphics[width=2\widthfigureresultsTwo]{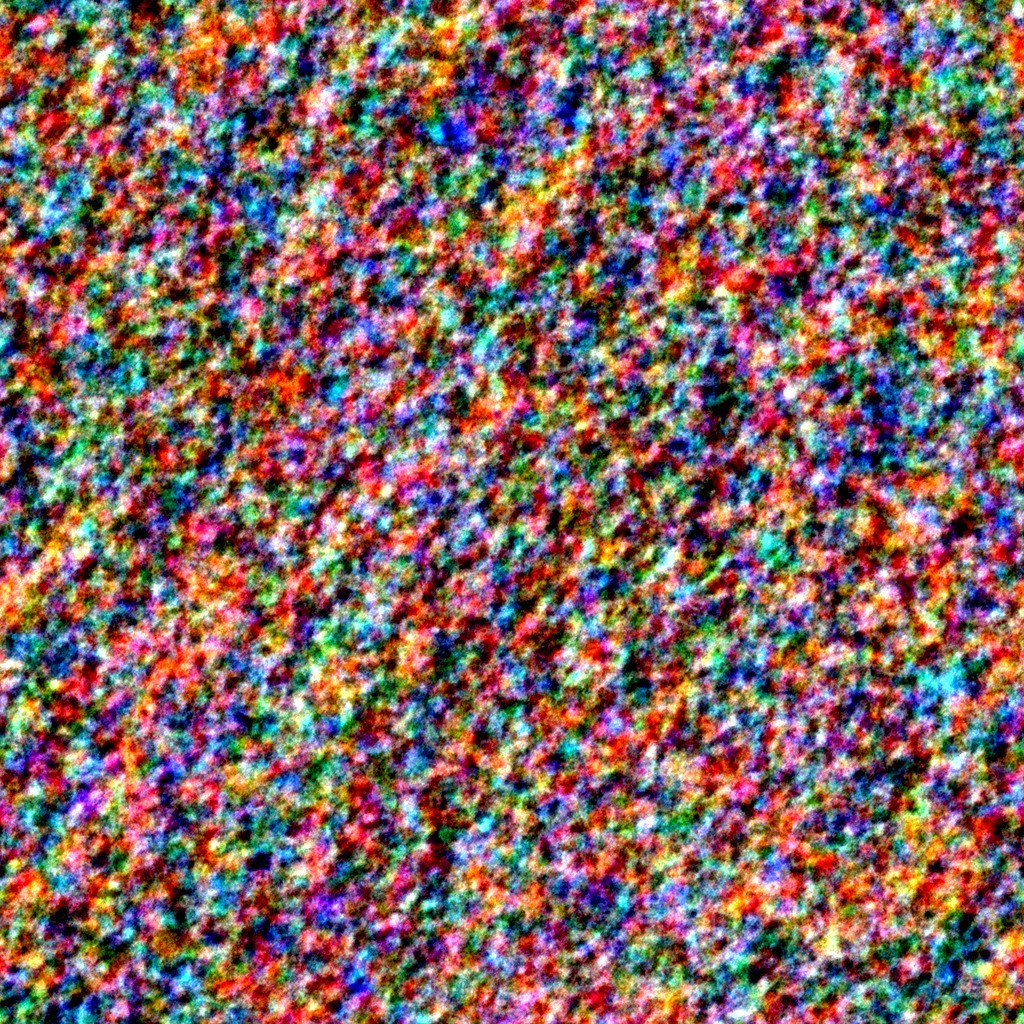}};
    \node[inner sep=0, anchor=south]  at (6,13) {\includegraphics[width=2\widthfigureresultsTwo]{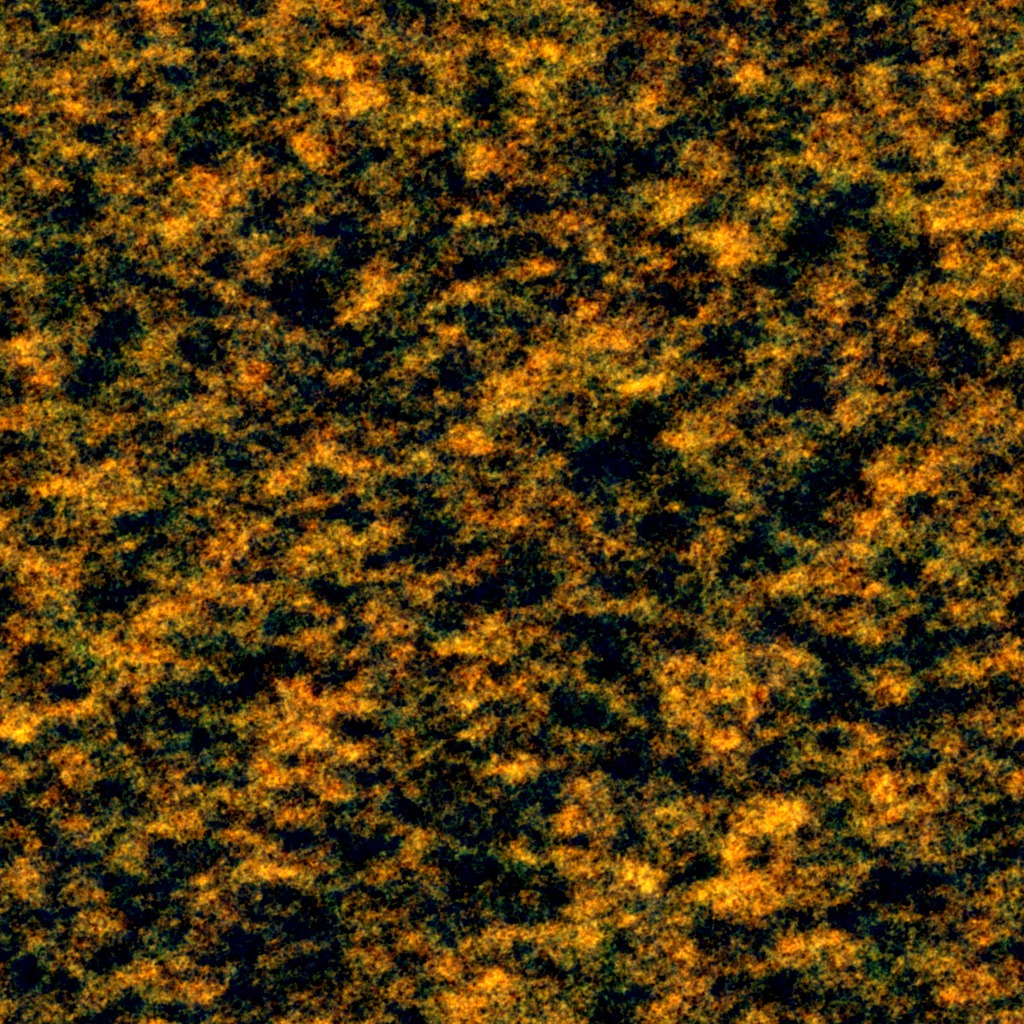}};
    \node[inner sep=0, anchor=south]  at (9,13) {\includegraphics[width=2\widthfigureresultsTwo]{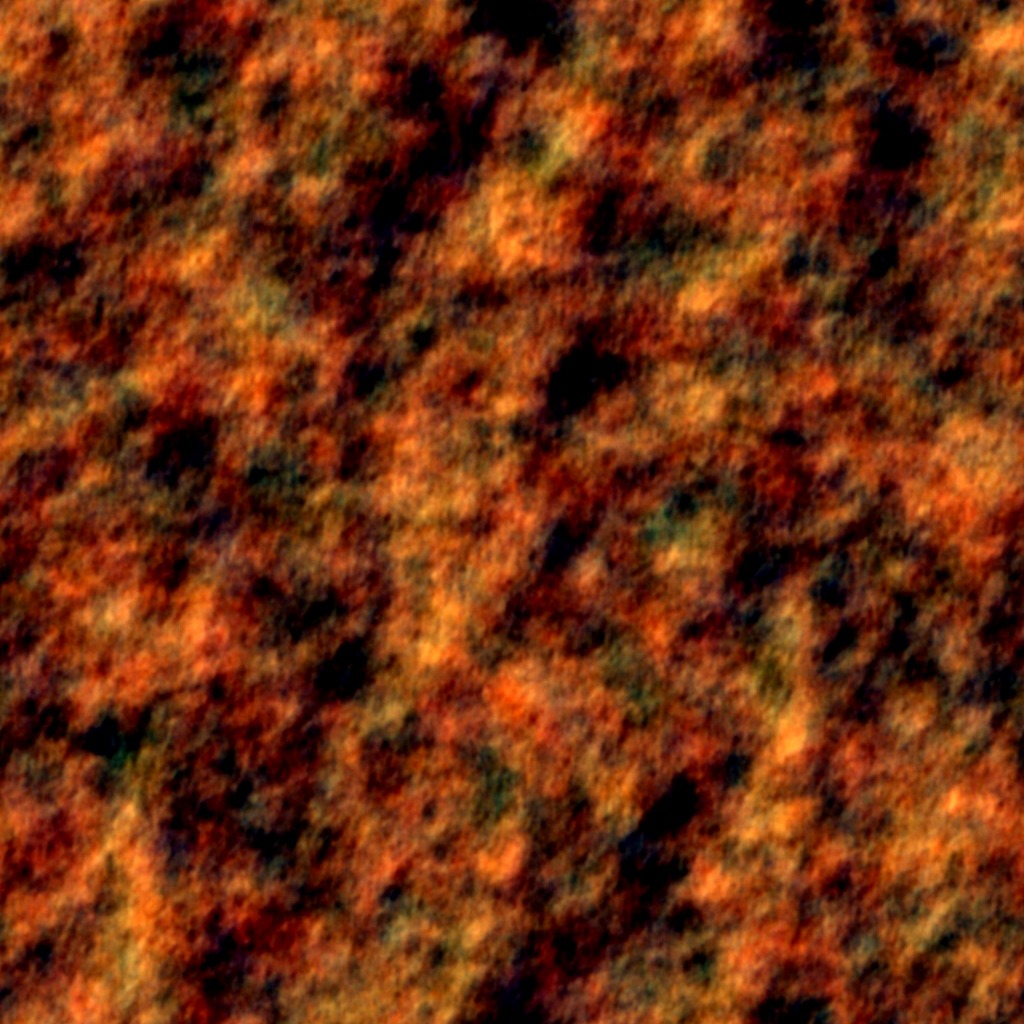}};

    \node[inner sep=0, anchor=south] (im6) at (0,10.4) {\includegraphics[width=2\widthfigureresultsTwo]{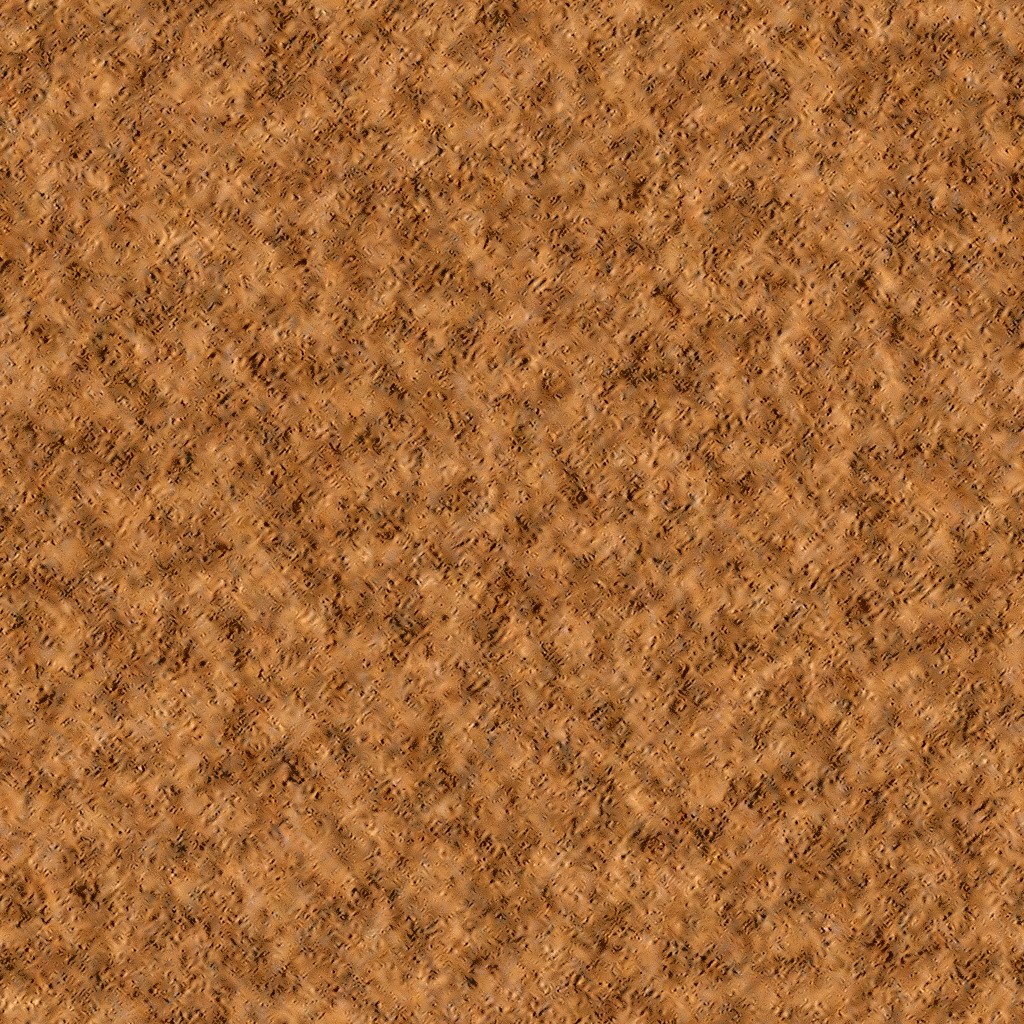}};
    \node[inner sep=0, anchor=south]  at (3,10.4) {\includegraphics[width=2\widthfigureresultsTwo]{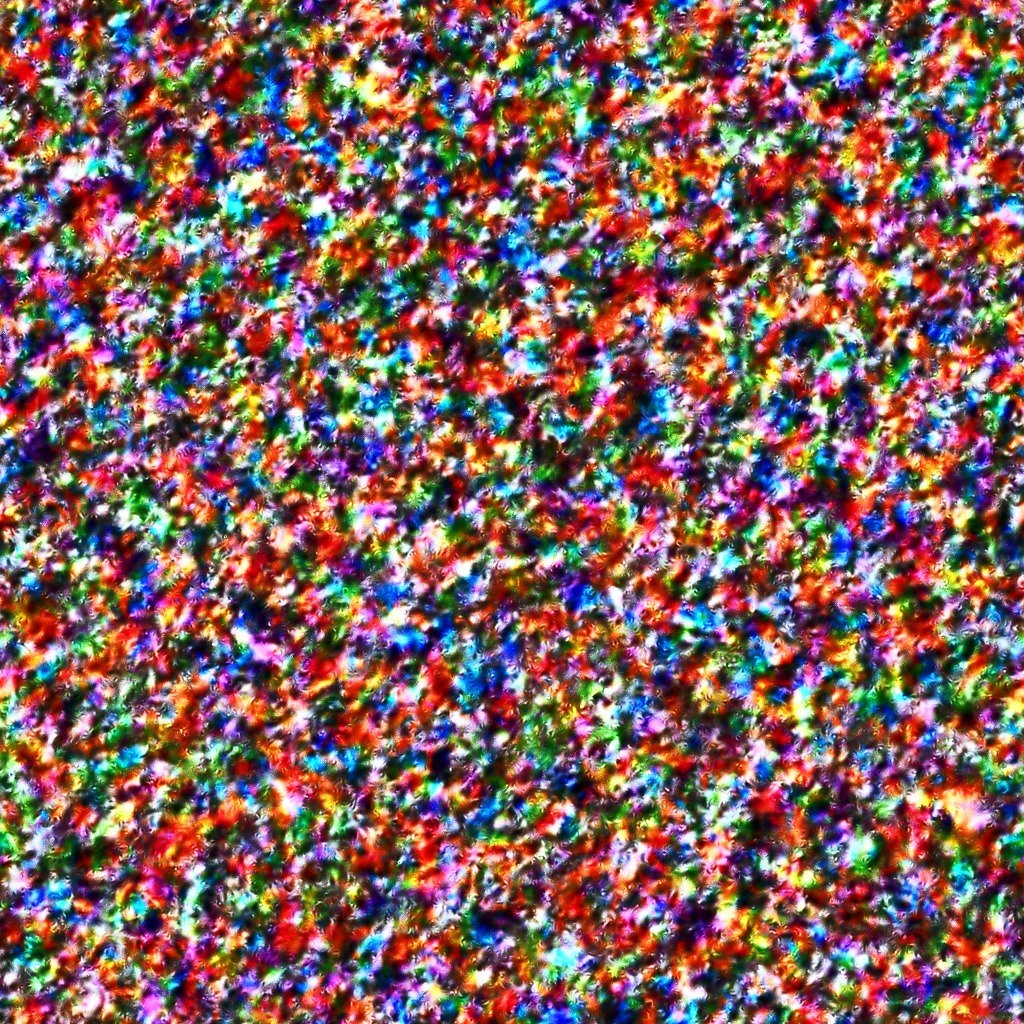}};
    \node[inner sep=0, anchor=south]  at (6,10.4) {\includegraphics[width=2\widthfigureresultsTwo]{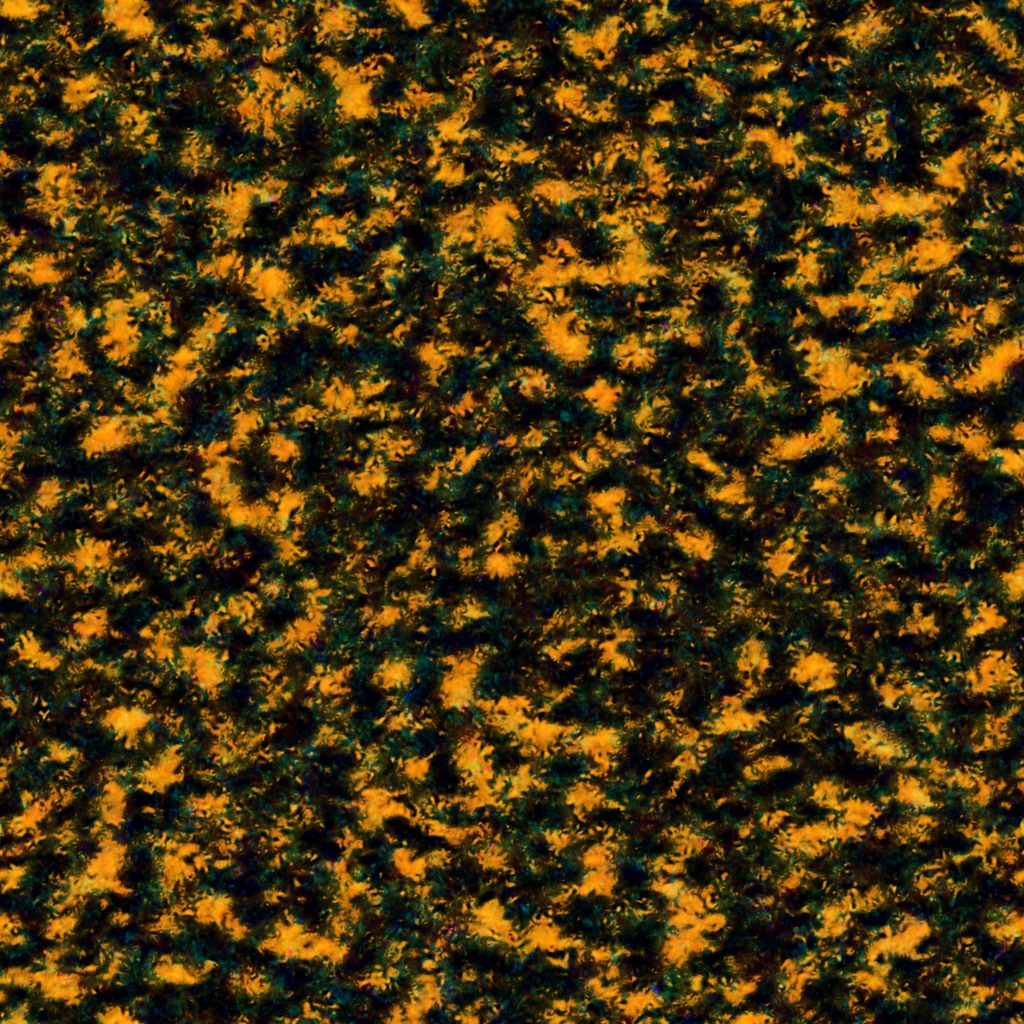}};
    \node[inner sep=0, anchor=south]  at (9,10.4) {\includegraphics[width=2\widthfigureresultsTwo]{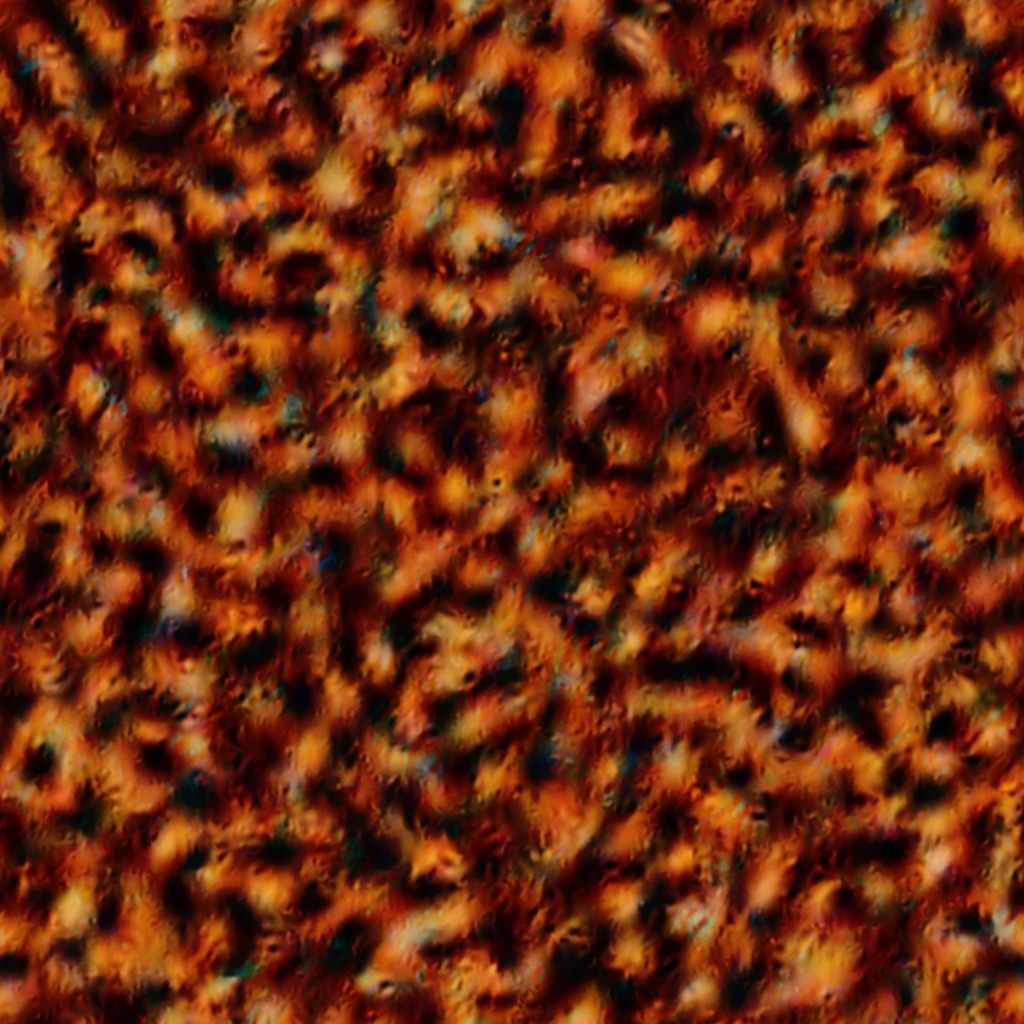}};

    \node[inner sep=0, anchor=south] (im5) at (0,7.8) {\includegraphics[width=2\widthfigureresultsTwo]{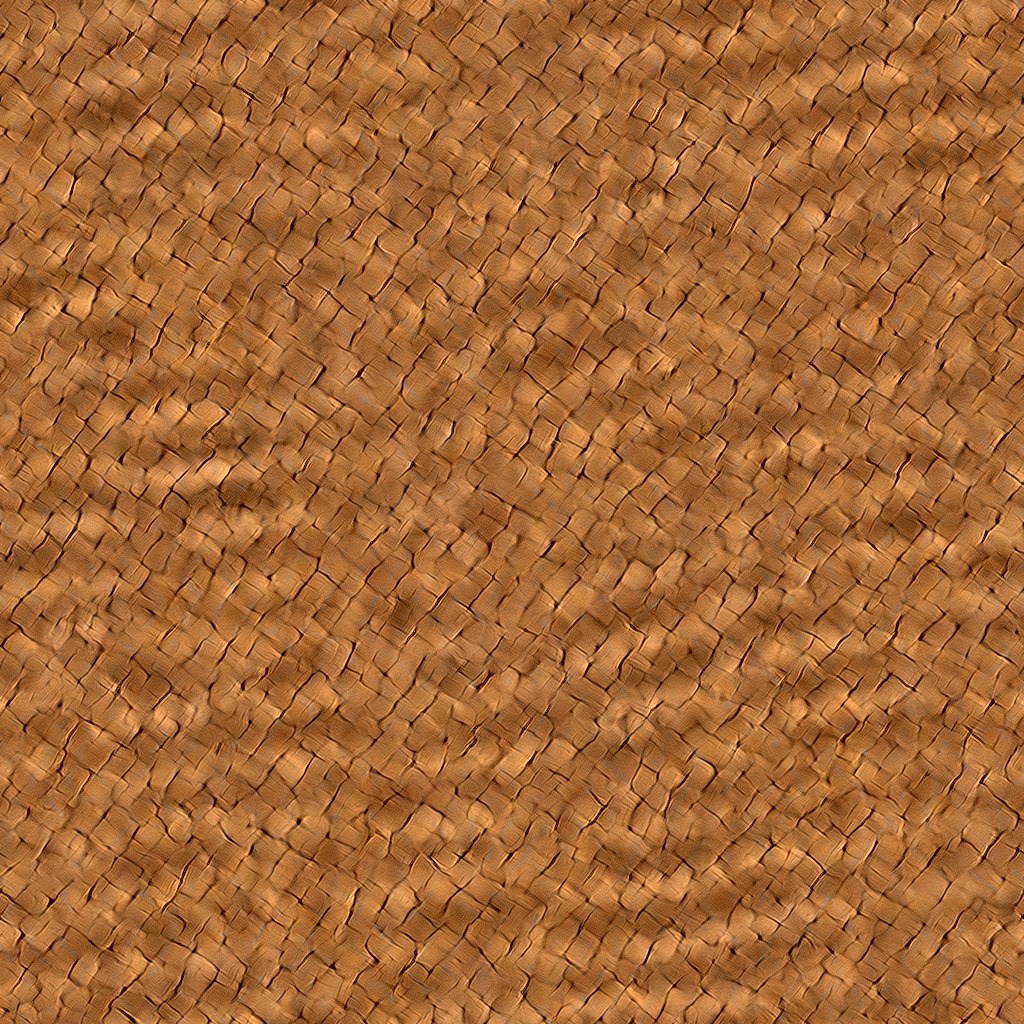}};
    \node[inner sep=0, anchor=south]  at (3,7.8) {\includegraphics[width=2\widthfigureresultsTwo]{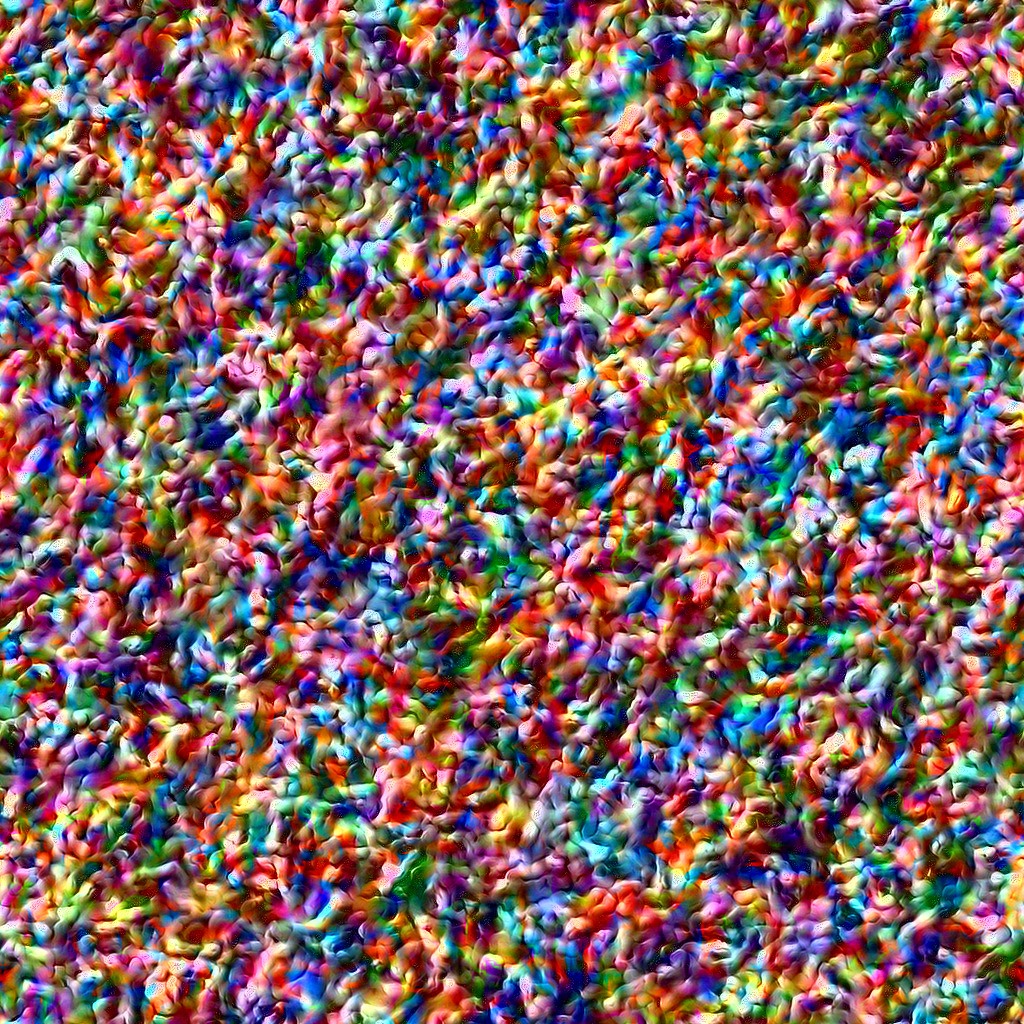}};
    \node[inner sep=0, anchor=south]  at (6,7.8) {\includegraphics[width=2\widthfigureresultsTwo]{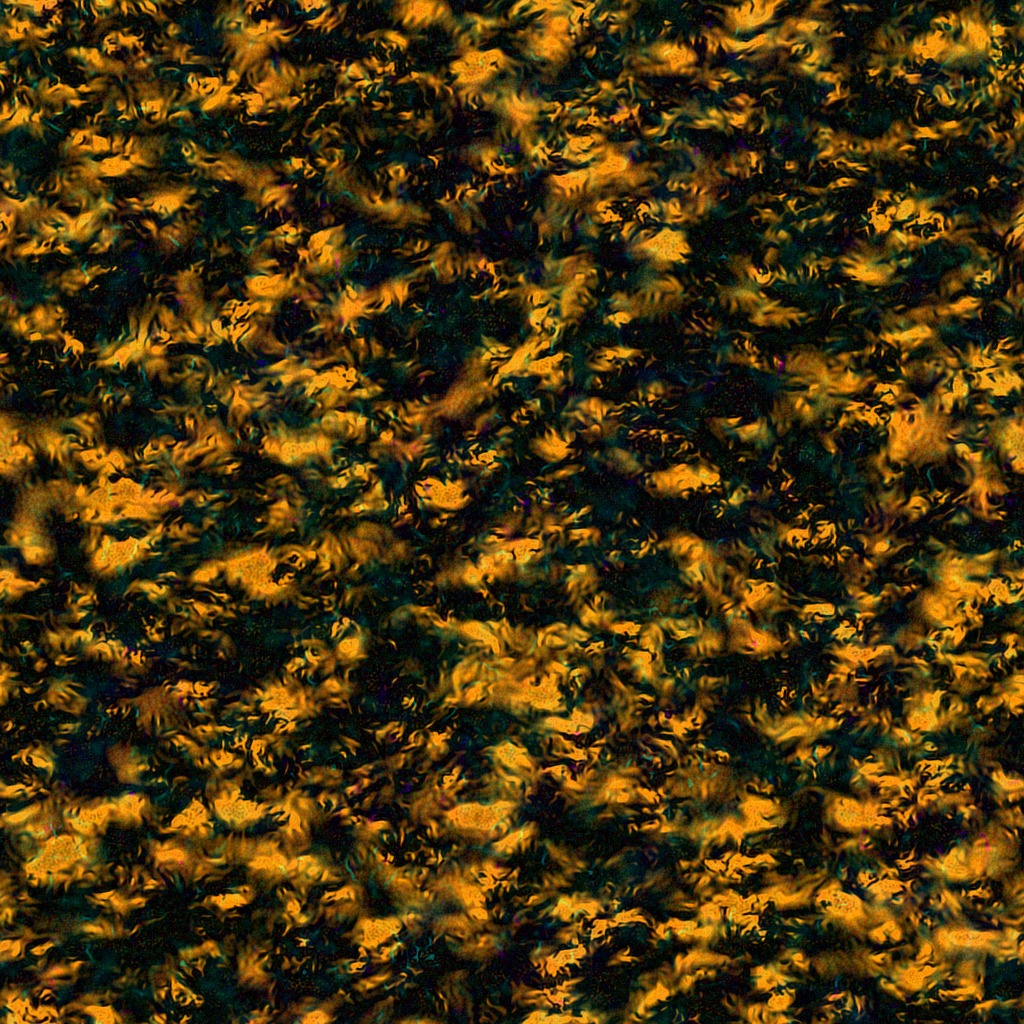}};
    \node[inner sep=0, anchor=south]  at (9,7.8) {\includegraphics[width=2\widthfigureresultsTwo]{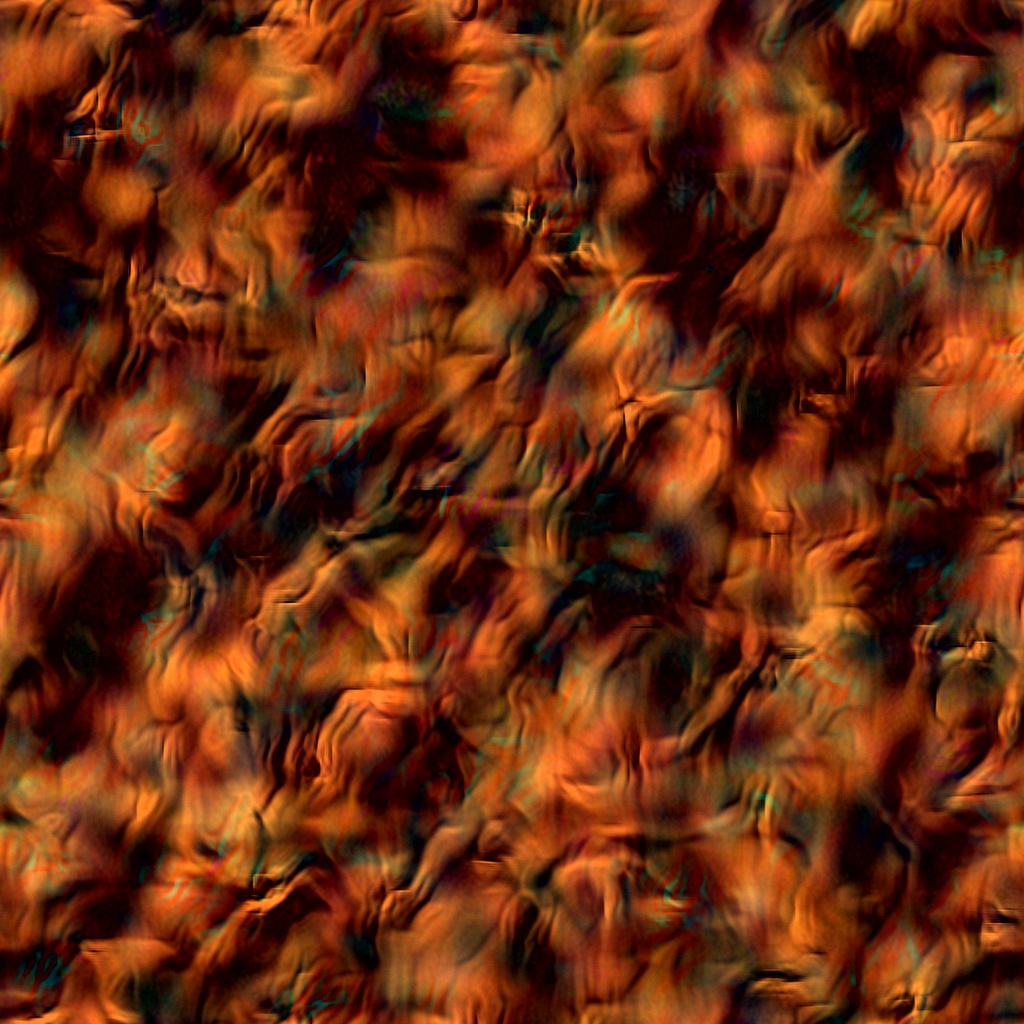}};

    \node[inner sep=0, anchor=south] (im4) at (0,5.2) {\includegraphics[width=2\widthfigureresultsTwo]{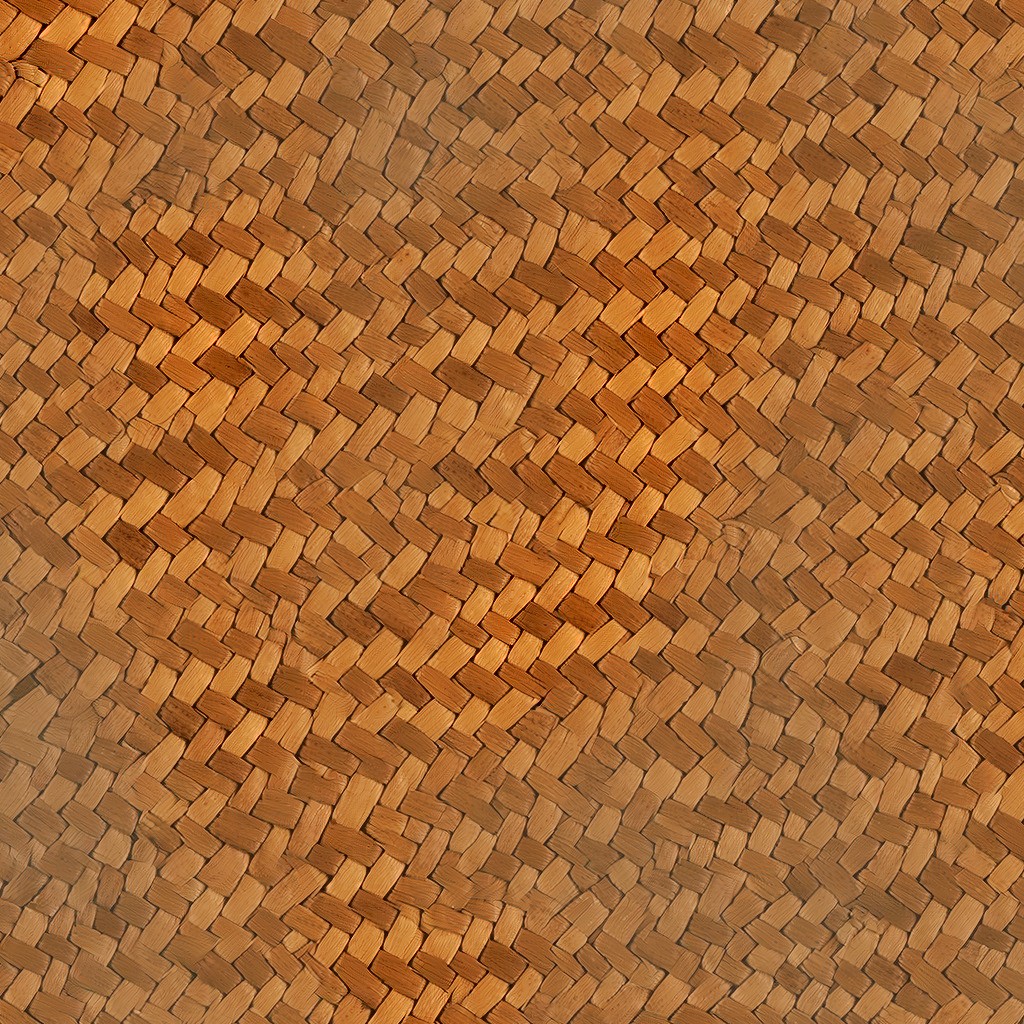}};
    \node[inner sep=0, anchor=south]  at (3,5.2) {\includegraphics[width=2\widthfigureresultsTwo]{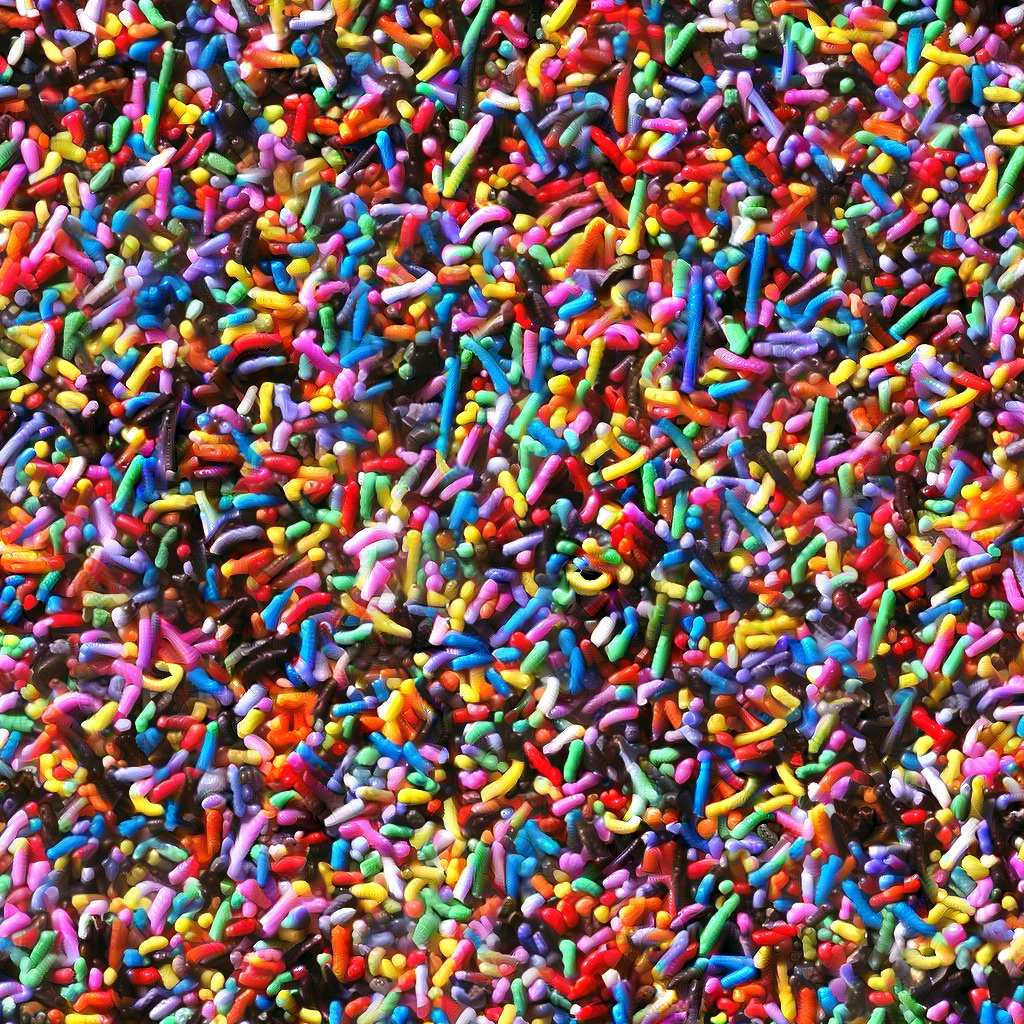}};
    \node[inner sep=0, anchor=south]  at (6,5.2) {\includegraphics[width=2\widthfigureresultsTwo]{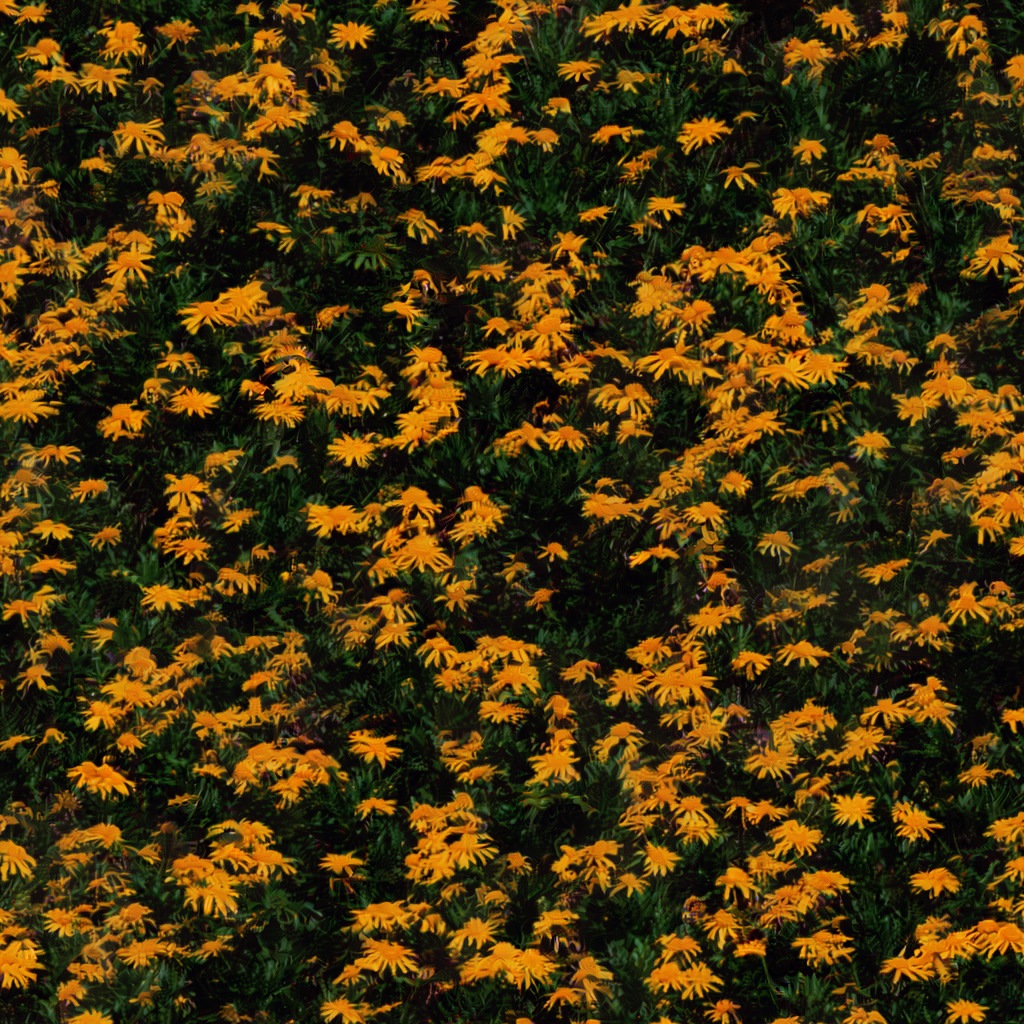}};
    \node[inner sep=0, anchor=south]  at (9,5.2) {\includegraphics[width=2\widthfigureresultsTwo]{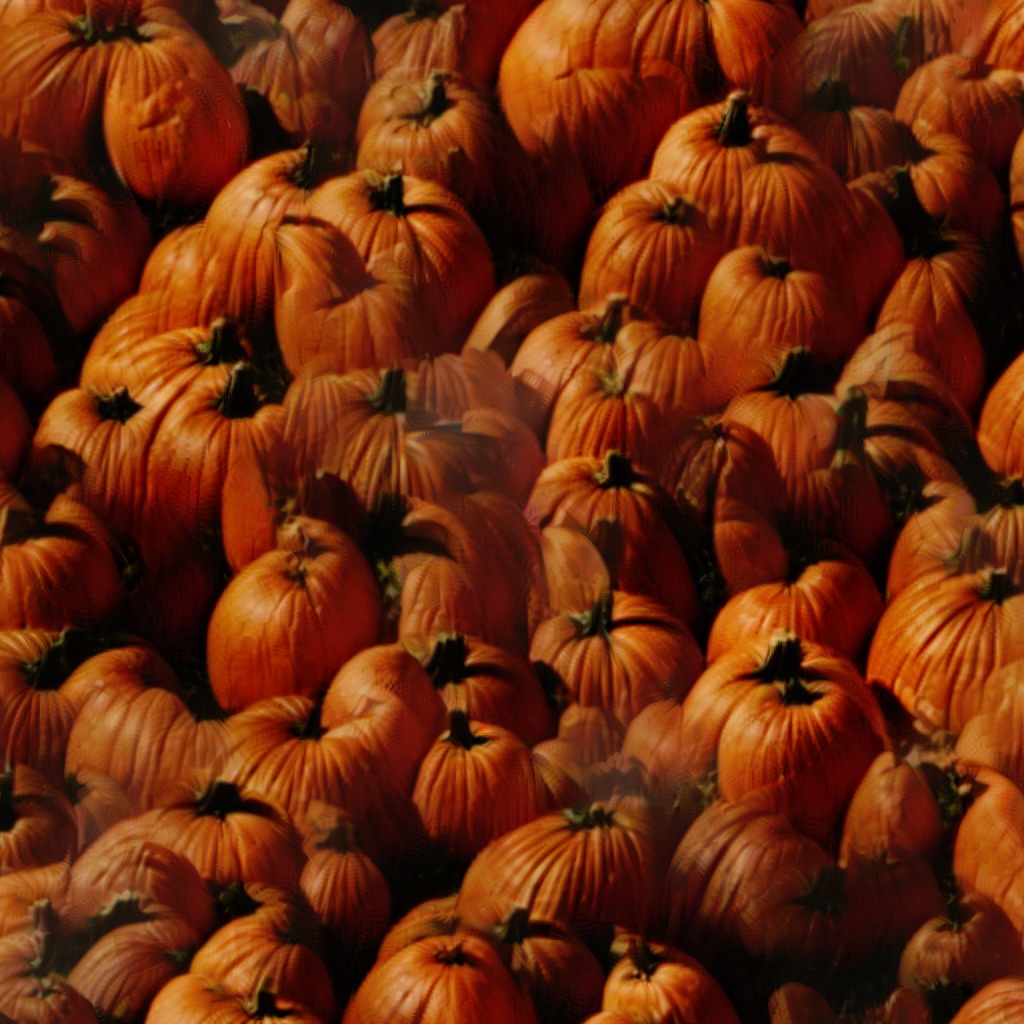}};

    \node[inner sep=0, anchor=south] (im3) at (0,2.6) {\includegraphics[width=2\widthfigureresultsTwo]{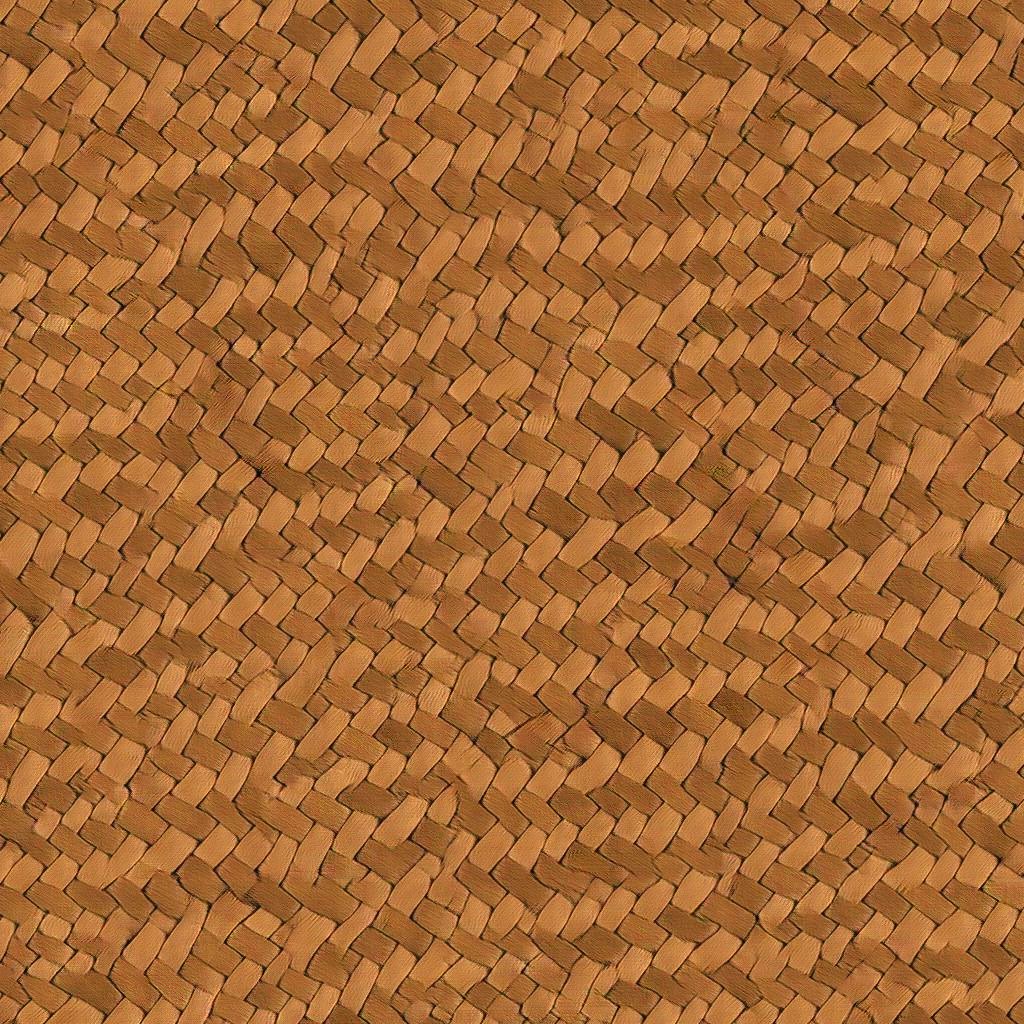}};
    \node[inner sep=0, anchor=south]  at (3,2.6) {\includegraphics[width=2\widthfigureresultsTwo]{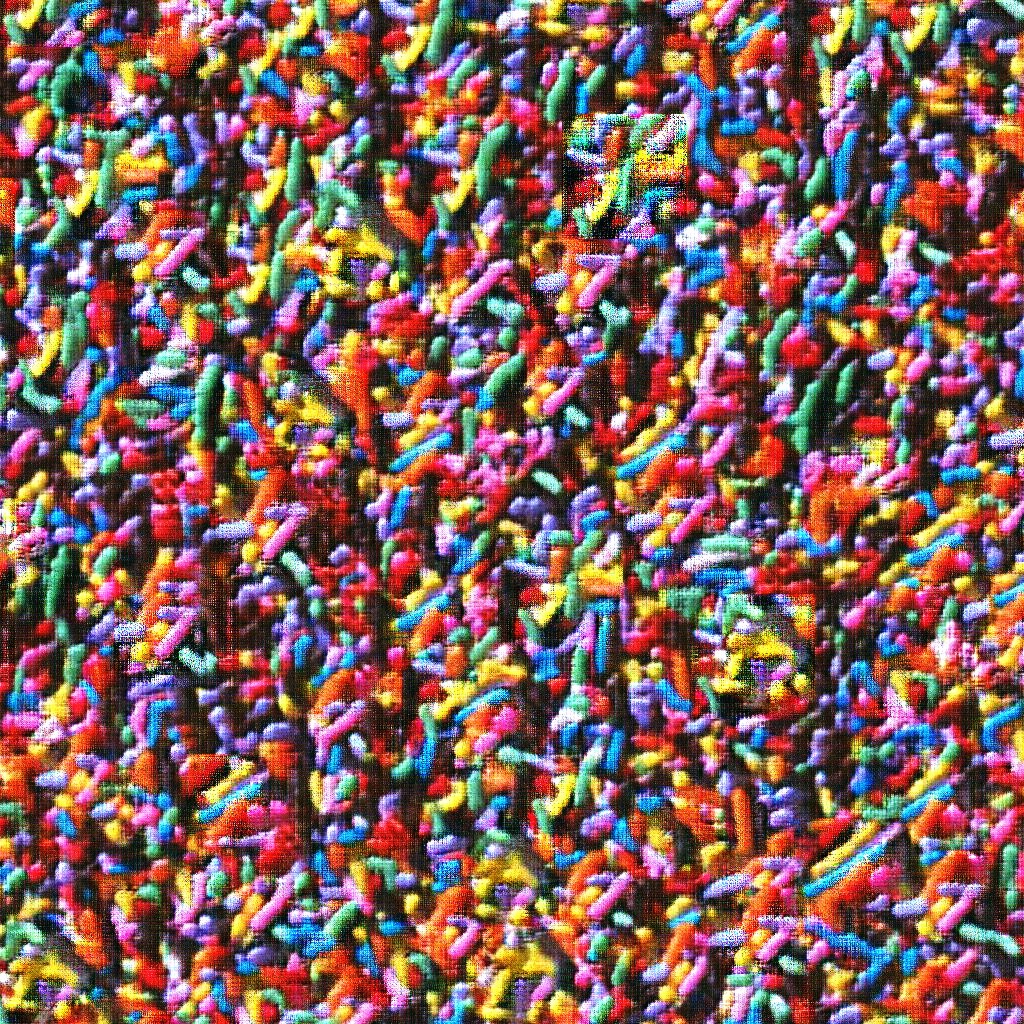}};
    \node[inner sep=0, anchor=south]  at (6,2.6) {\includegraphics[width=2\widthfigureresultsTwo]{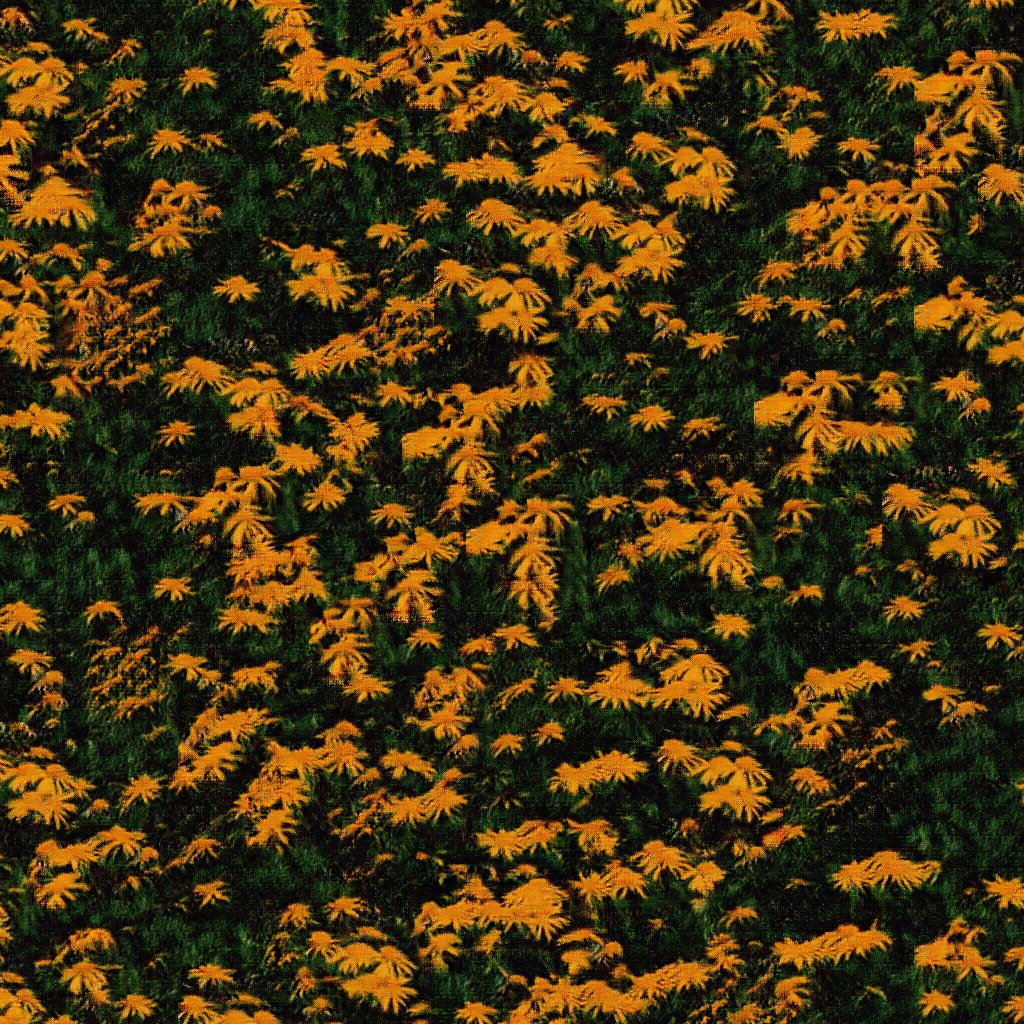}};
    \node[inner sep=0, anchor=south]  at (9,2.6) {\includegraphics[width=2\widthfigureresultsTwo]{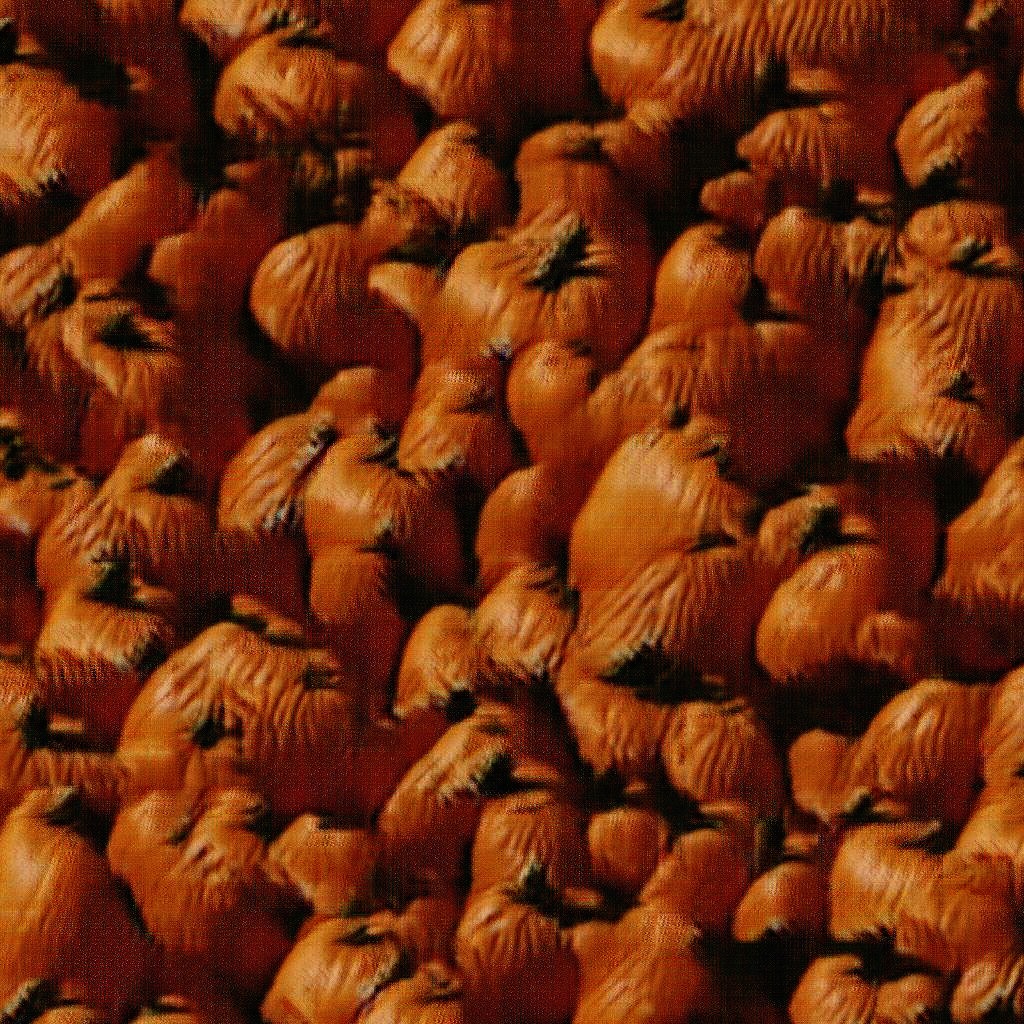}};

    \node [anchor=east] at (im3.west) {SGAN \cite{jetchev2016texture}};
    \node [anchor=east] at (im4.west) {Gatys \cite{gatys}};
    \node [anchor=east] at (im5.west) {PS \cite{PS}};
    \node [anchor=east] at (im6.west) {HB \cite{HeegerBergen}};
    \node [anchor=east] at (im7.west) {RPN \cite{GalerneRPN}};
    \node [anchor=east] at (im8.west) {input};

    \end{tikzpicture}
    \caption{\emph{Comparison of texture synthesis methods.} From top to bottom: input sample, Random Phase Noise (RPN)~\cite{GalerneRPN}, Heeger and Bergen (HB)~\cite{HeegerBergen}, Portilla and Simoncelli (PS)~\cite{PS}, Gatys (Gatys)~\cite{gatys} and SGAN \cite{jetchev2016texture}.}
    \label{fig:algComp2-1}
\end{figure}


\begin{figure}[p]
  \centering
  \begin{tikzpicture}[scale=.9]

    \node[inner sep=0, anchor=south east] (im4) at (0,2.6) {\includegraphics[width=\widthfigureresultsTwo]{figures/Fabric_0000}};
    \node[inner sep=0, anchor=south east]  at (3,2.6) {\includegraphics[width=\widthfigureresultsTwo]{figures/Food_0008}};
    \node[inner sep=0, anchor=south east]  at (6,2.6) {\includegraphics[width=\widthfigureresultsTwo]{figures/Flowers_0000}};
    \node[inner sep=0, anchor=south east]  at (9,2.6) {\includegraphics[width=\widthfigureresultsTwo]{figures/Food_0010}};

    \node[inner sep=0, anchor=south] (im3) at (0,0) {\includegraphics[width=2\widthfigureresultsTwo]{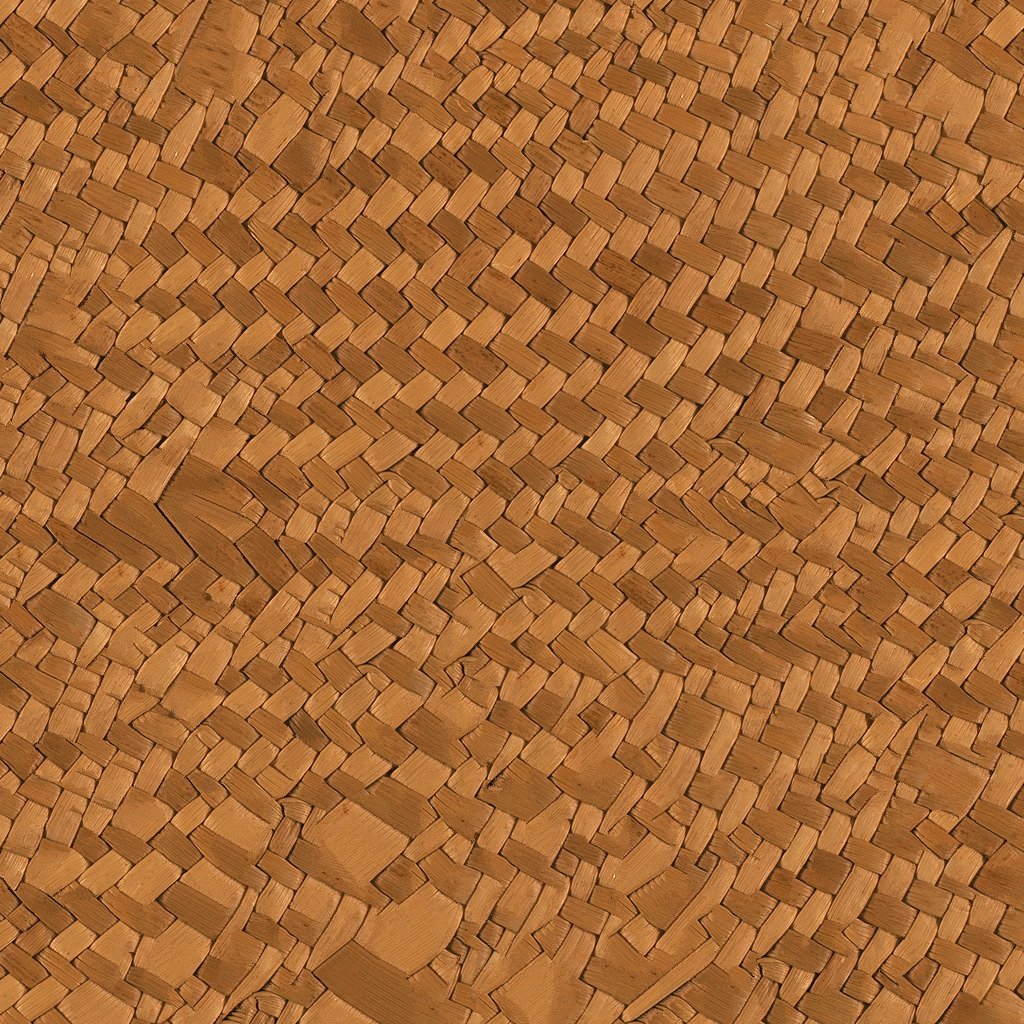}};
    \node[inner sep=0, anchor=south]  at (3,0) {\includegraphics[width=2\widthfigureresultsTwo]{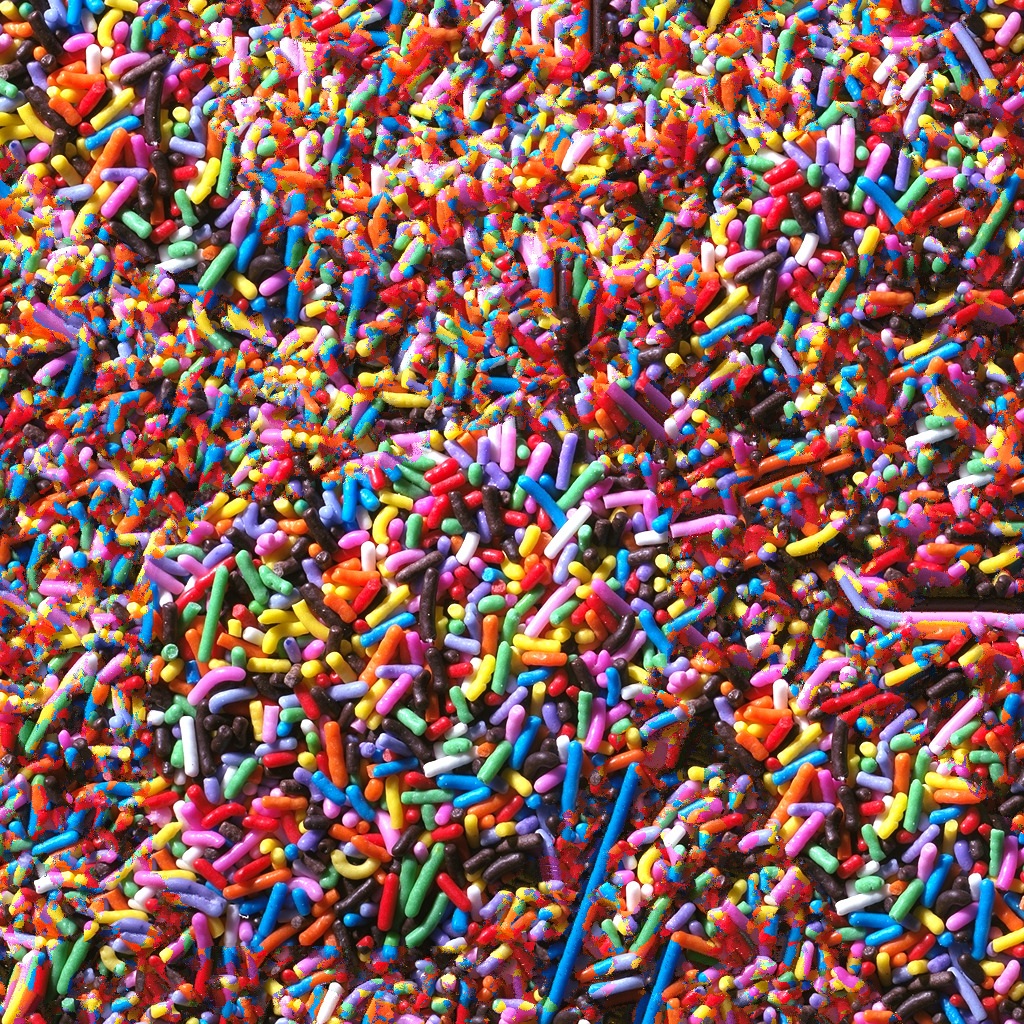}};
    \node[inner sep=0, anchor=south]  at (6,0) {\includegraphics[width=2\widthfigureresultsTwo]{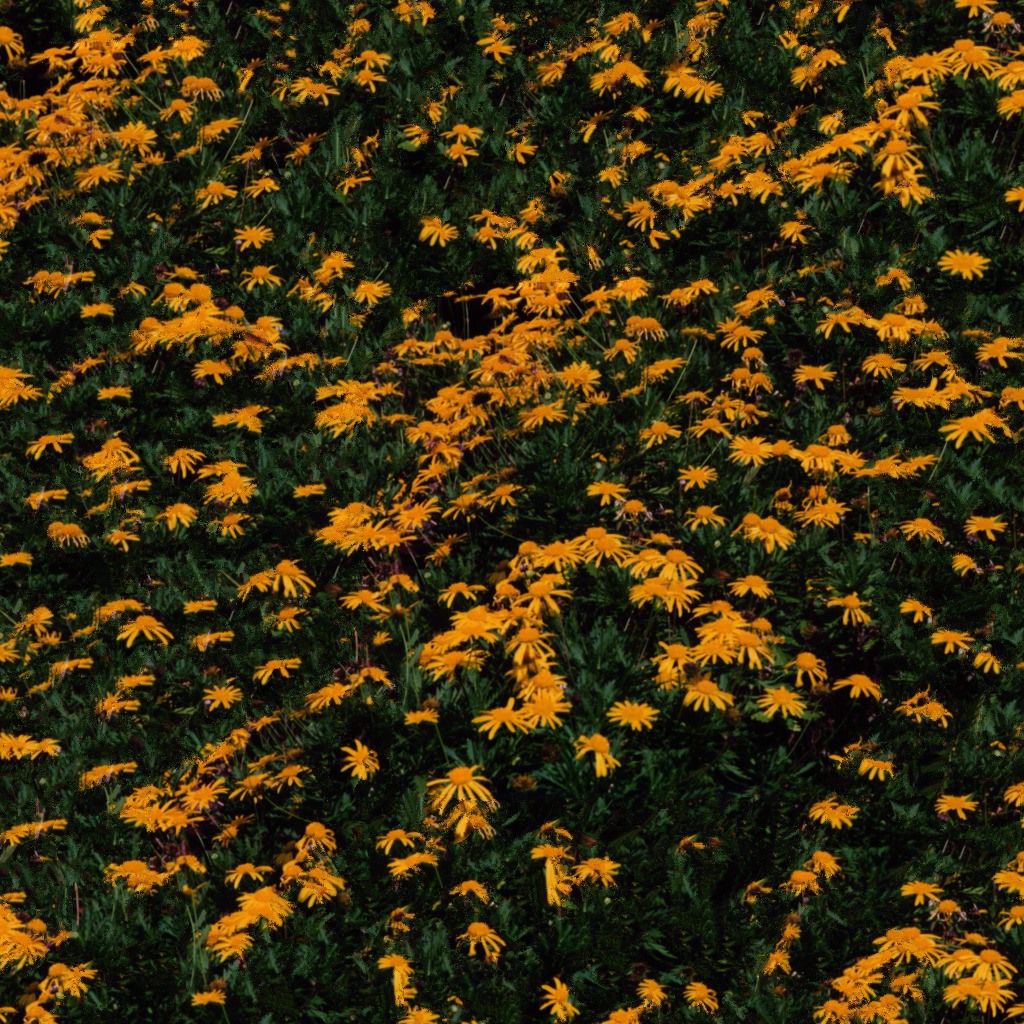}};
    \node[inner sep=0, anchor=south]  at (9,0) {\includegraphics[width=2\widthfigureresultsTwo]{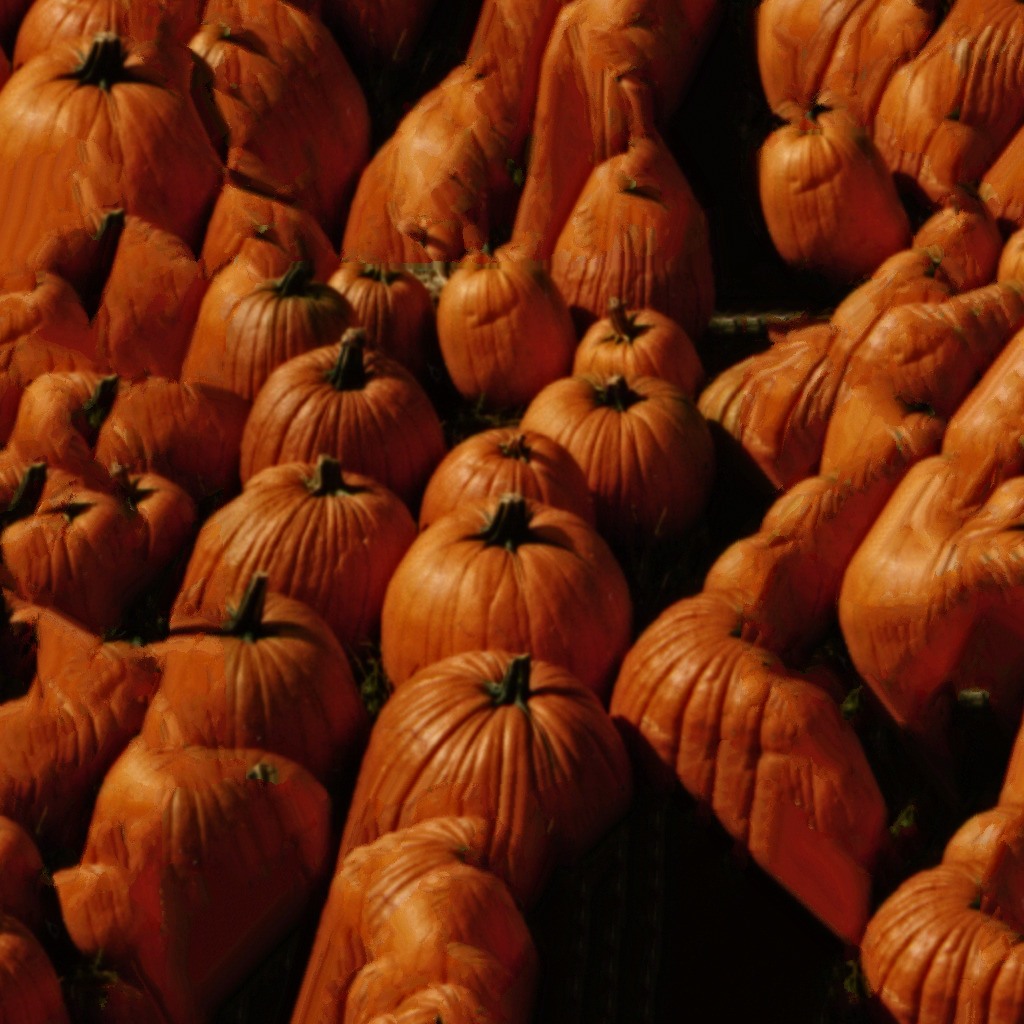}};

    \node[inner sep=0, anchor=south] (im2) at (0,-2.6) {\includegraphics[width=2\widthfigureresultsTwo]{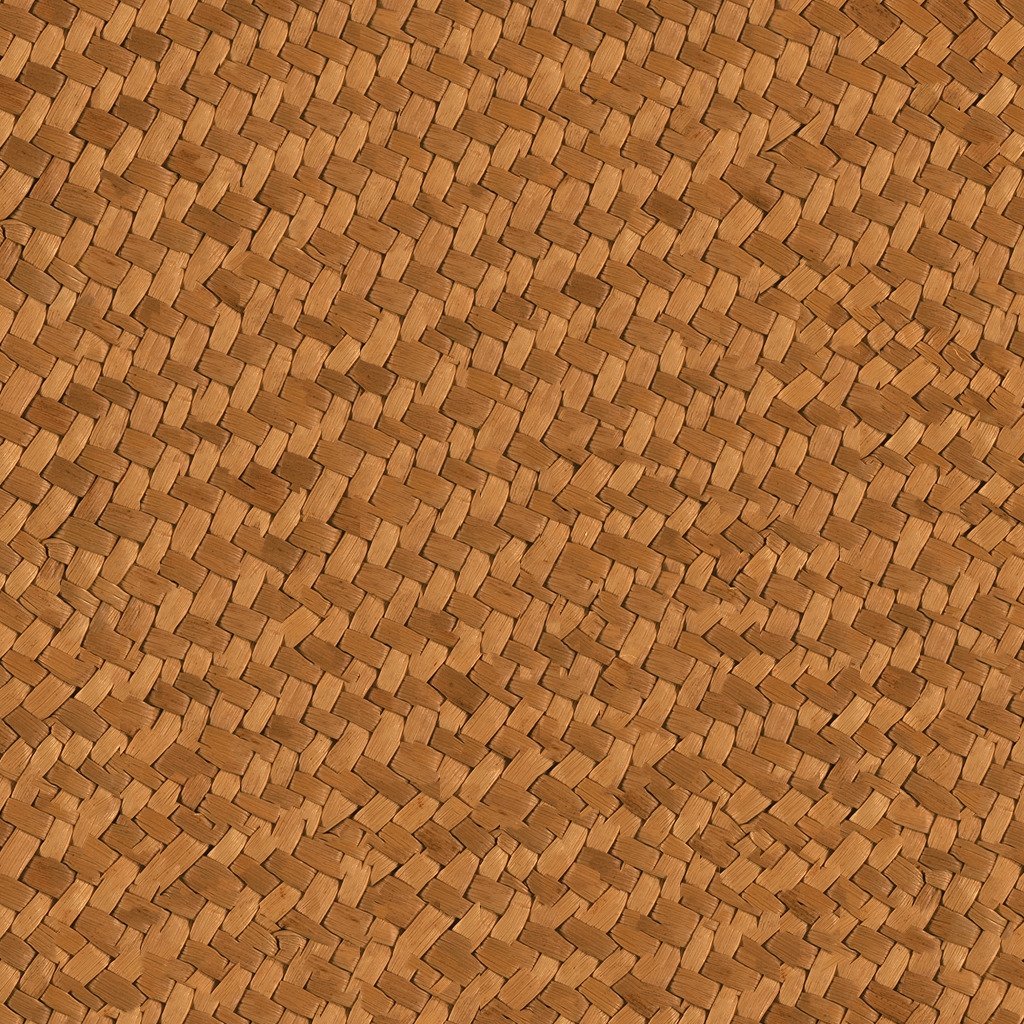}};
    \node[inner sep=0, anchor=south]  at (3,-2.6) {\includegraphics[width=2\widthfigureresultsTwo]{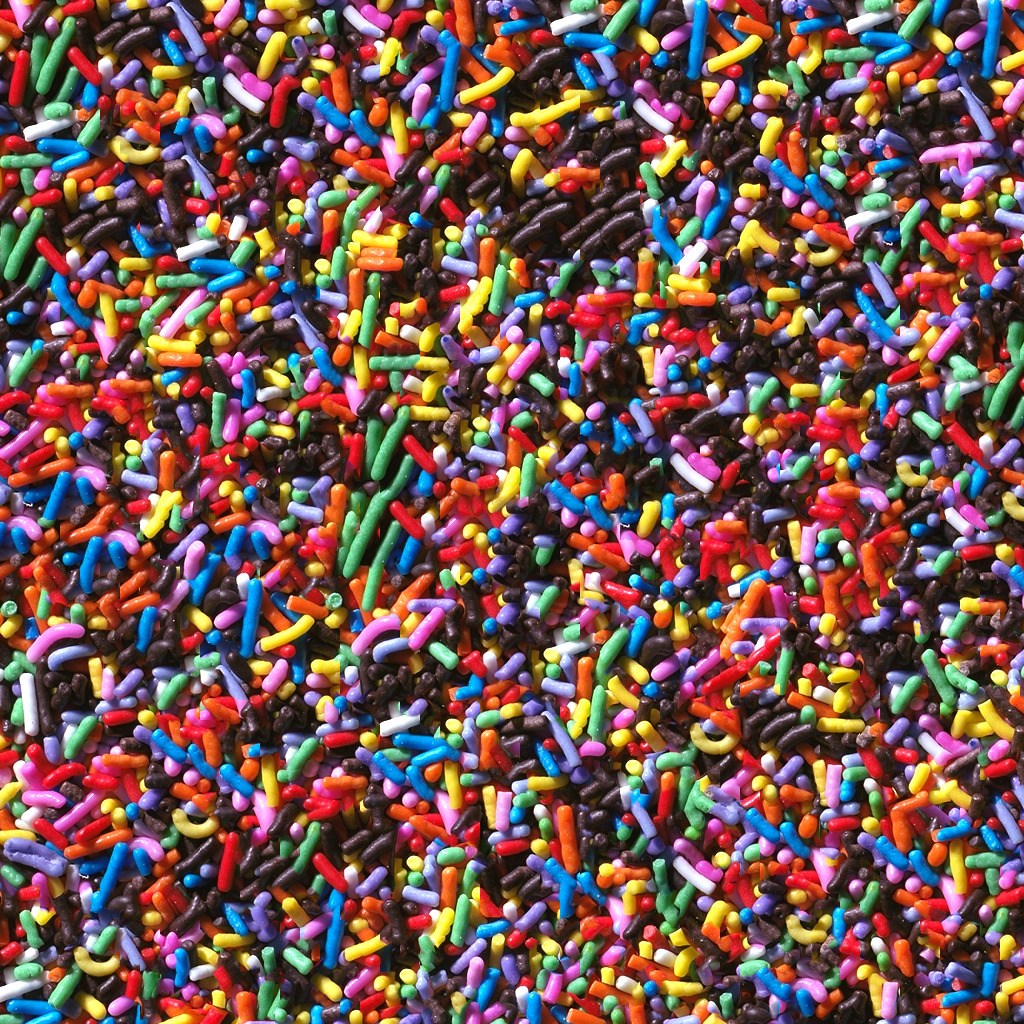}};
    \node[inner sep=0, anchor=south]  at (6,-2.6) {\includegraphics[width=2\widthfigureresultsTwo]{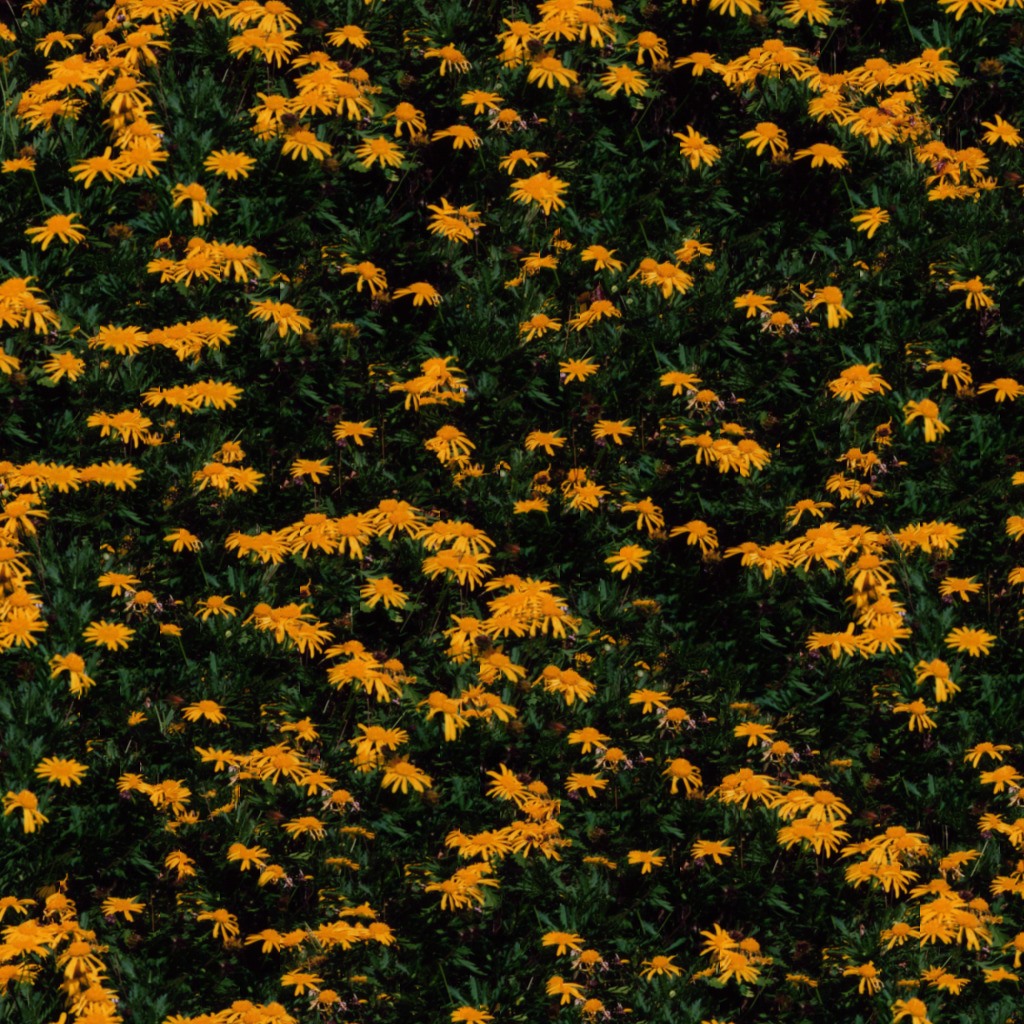}};
    \node[inner sep=0, anchor=south]  at (9,-2.6) {\includegraphics[width=2\widthfigureresultsTwo]{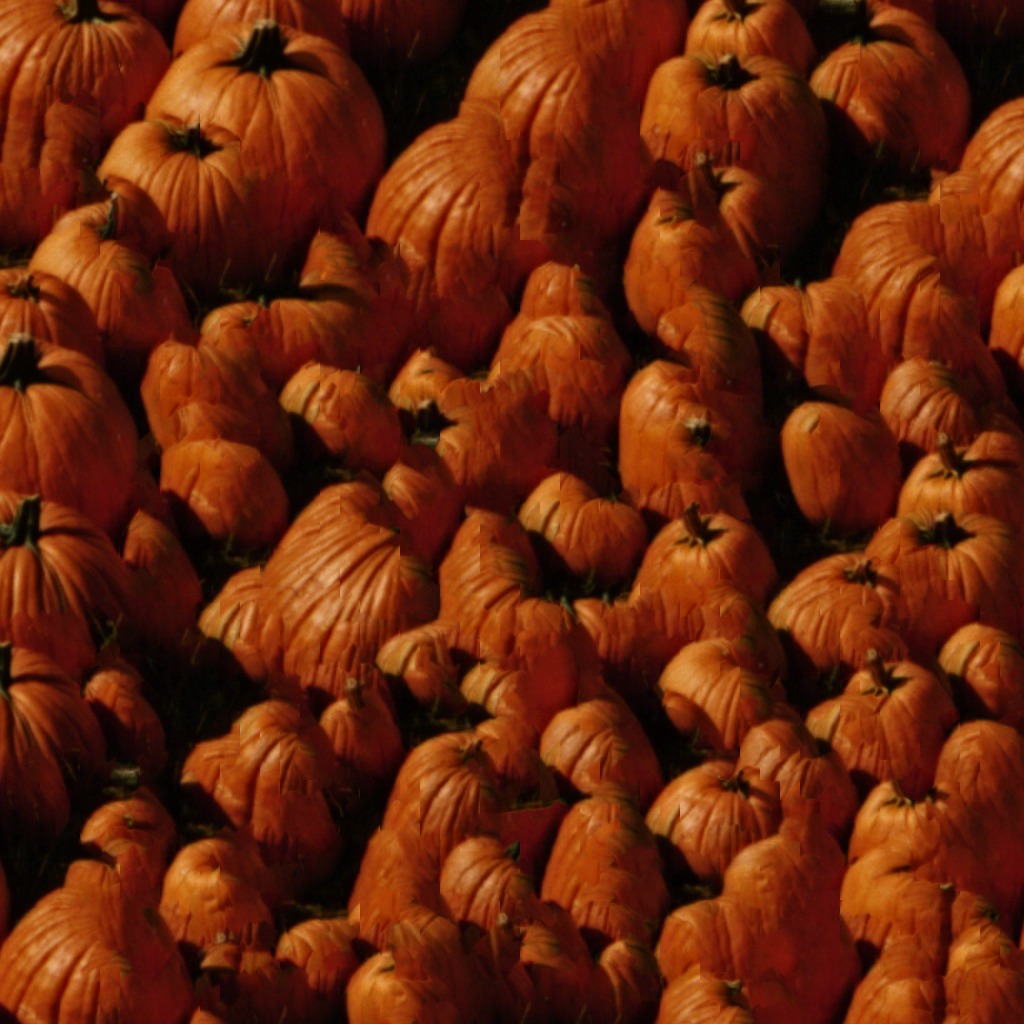}};

    \node[inner sep=0, anchor=south] (im1) at (0,-5.2) {\includegraphics[width=2\widthfigureresultsTwo]{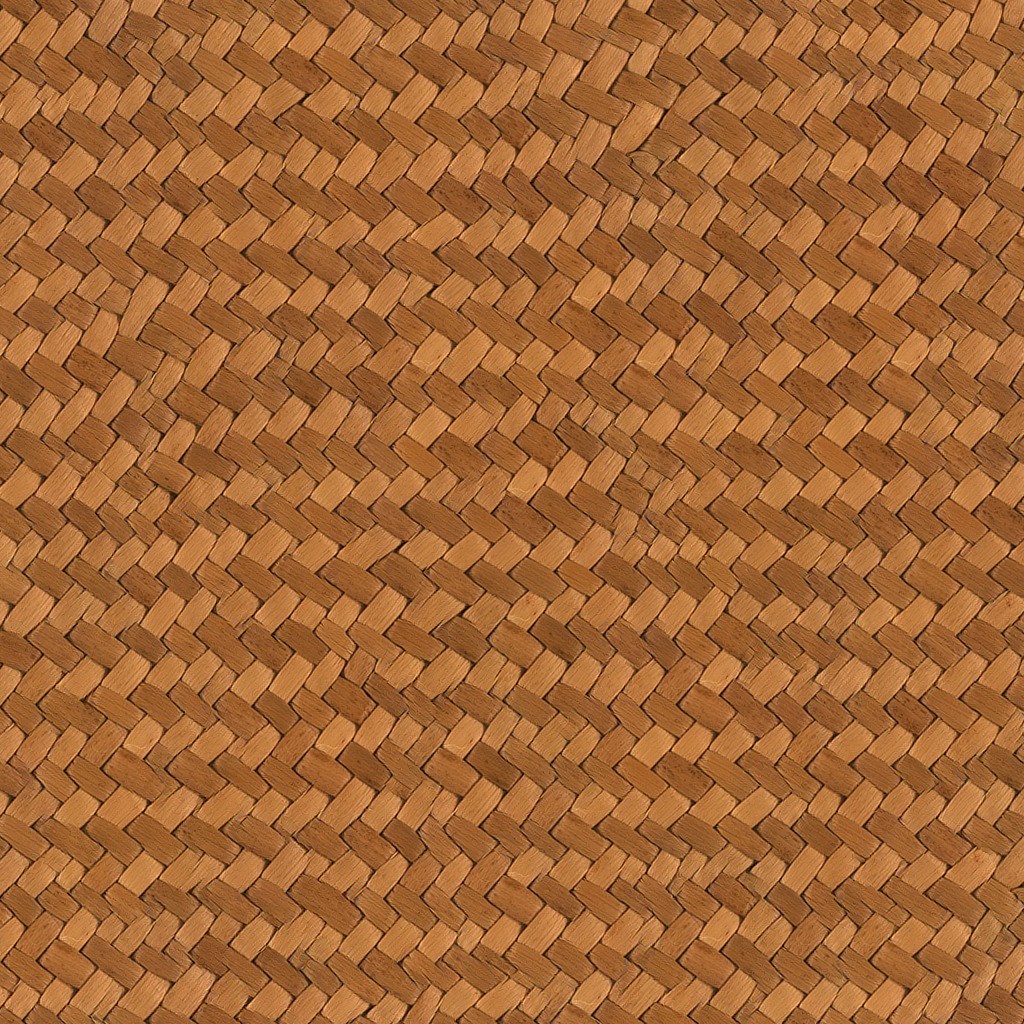}};
    \node[inner sep=0, anchor=south]  at (3,-5.2) {\includegraphics[width=2\widthfigureresultsTwo]{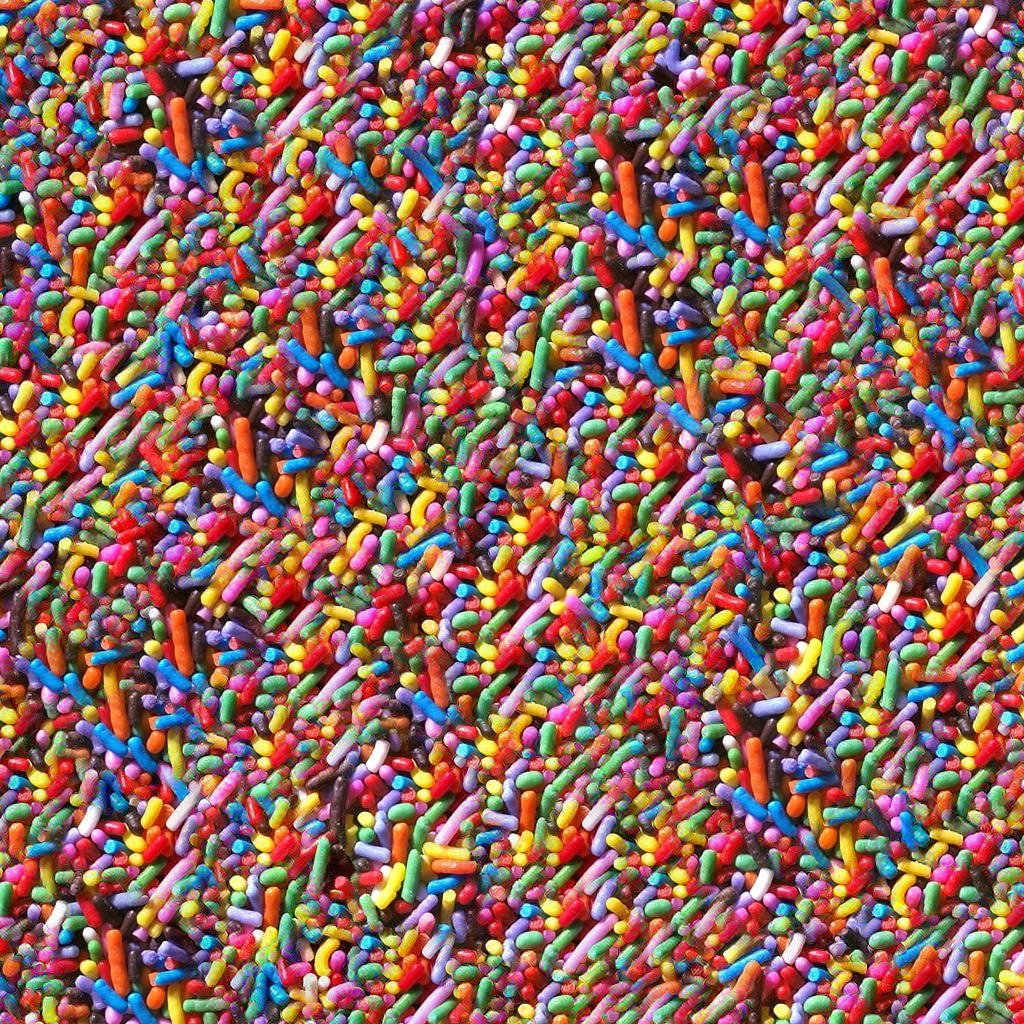}};
    \node[inner sep=0, anchor=south]  at (6,-5.2) {\includegraphics[width=2\widthfigureresultsTwo]{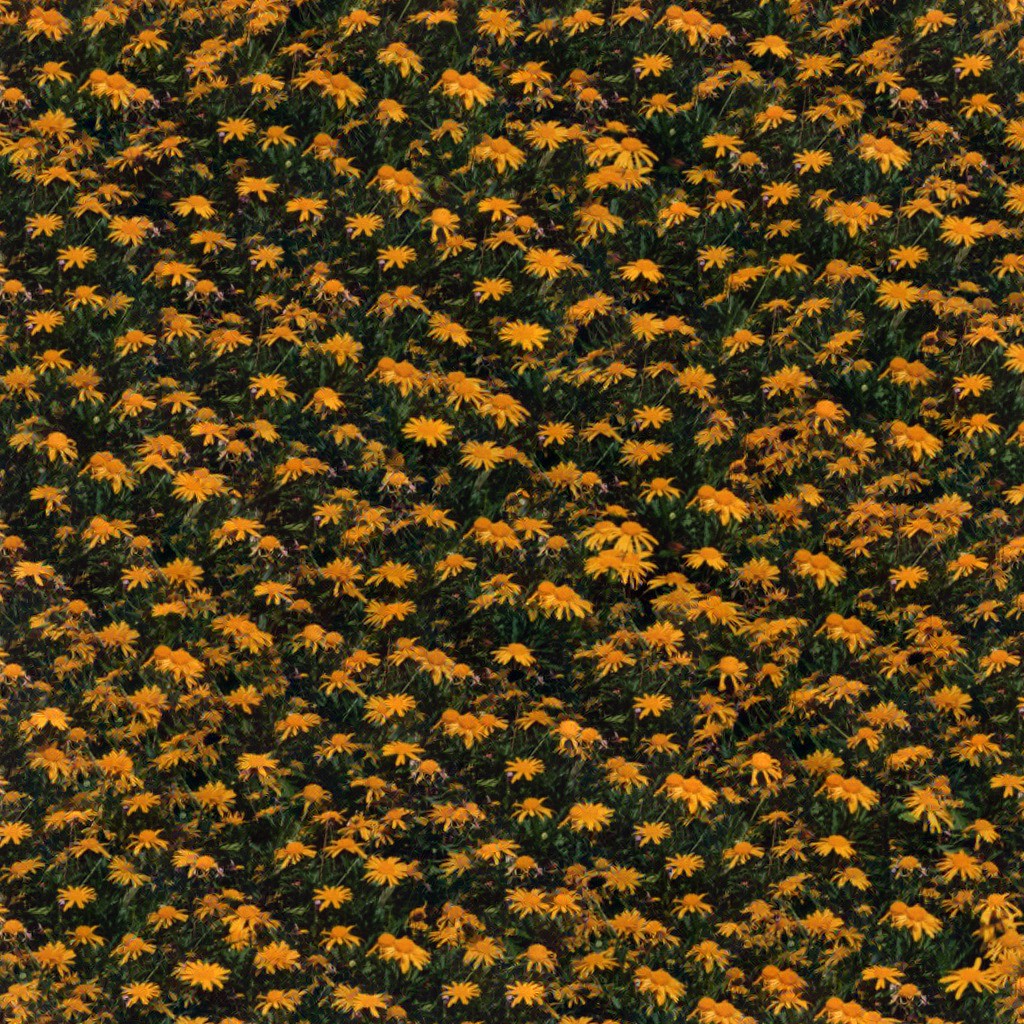}};
    \node[inner sep=0, anchor=south]  at (9,-5.2) {\includegraphics[width=2\widthfigureresultsTwo]{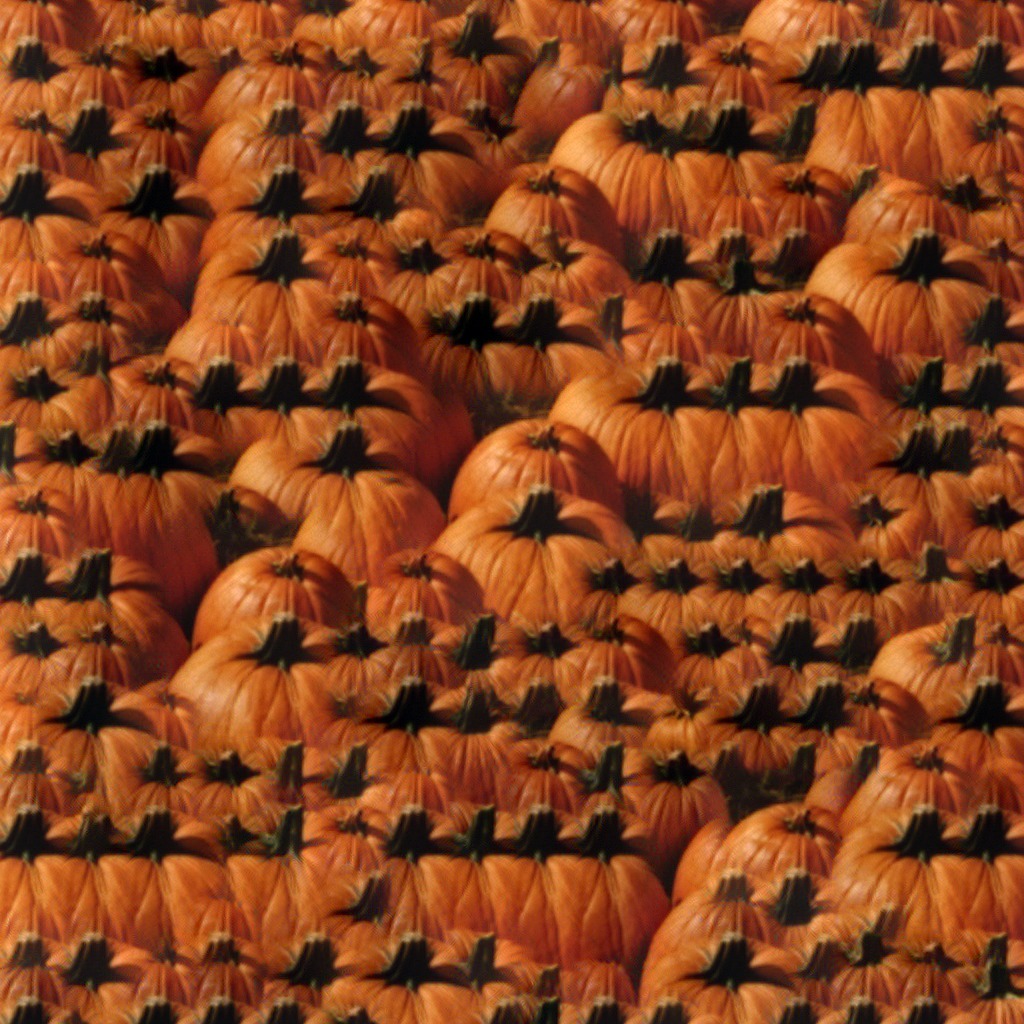}};

    \node[inner sep=0, anchor=south] (im0) at (0,-7.8) {\includegraphics[width=2\widthfigureresultsTwo]{figures/Fabric_0000_MSLG}};
    \node[inner sep=0, anchor=south]  at (3,-7.8) {\includegraphics[width=2\widthfigureresultsTwo]{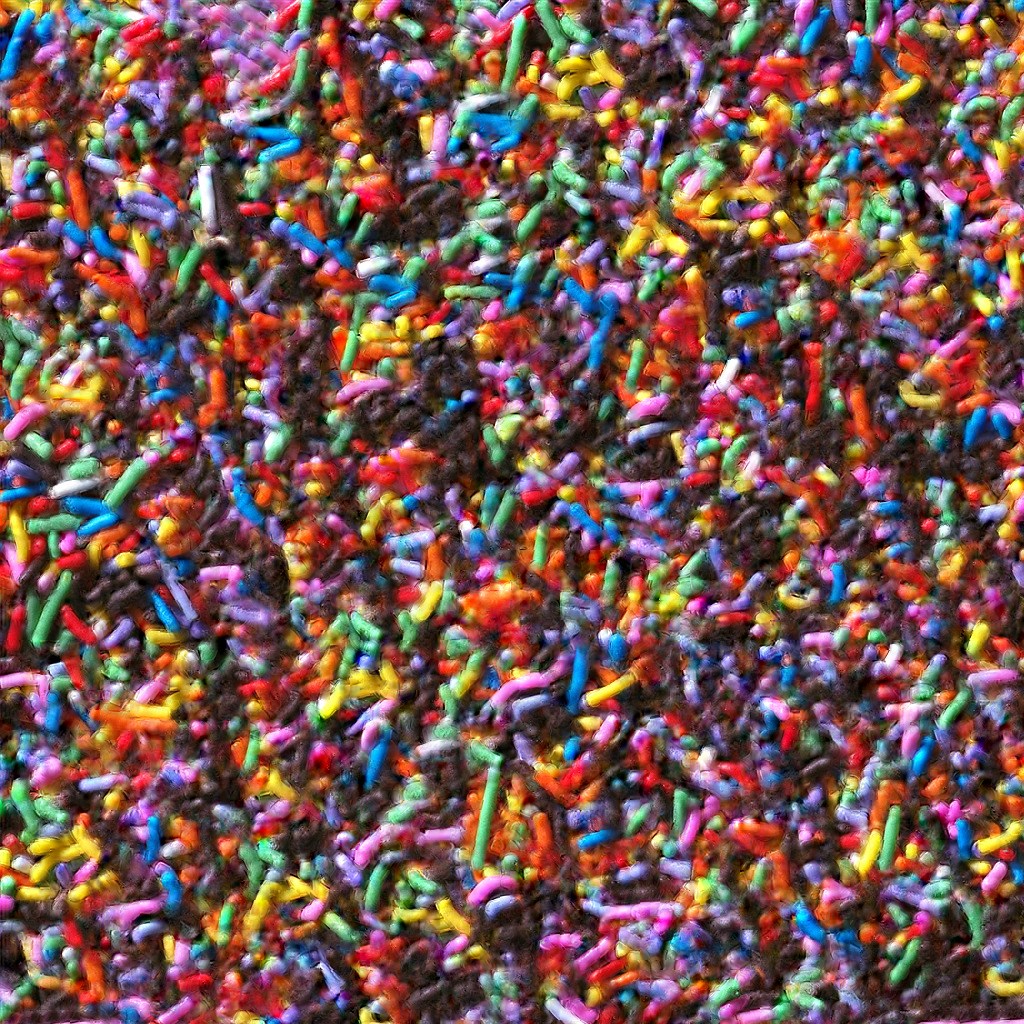}};
    \node[inner sep=0, anchor=south]  at (6,-7.8) {\includegraphics[width=2\widthfigureresultsTwo]{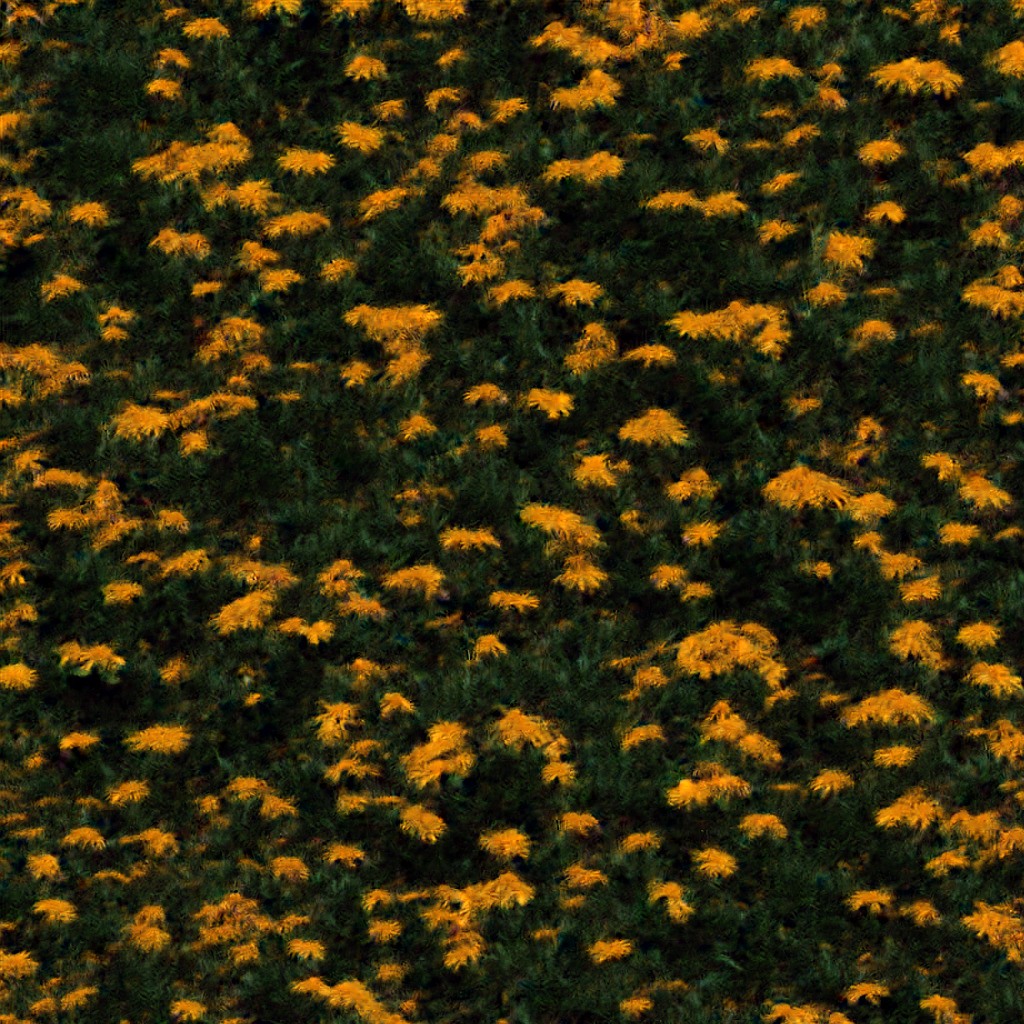}};
    \node[inner sep=0, anchor=south]  at (9,-7.8) {\includegraphics[width=2\widthfigureresultsTwo]{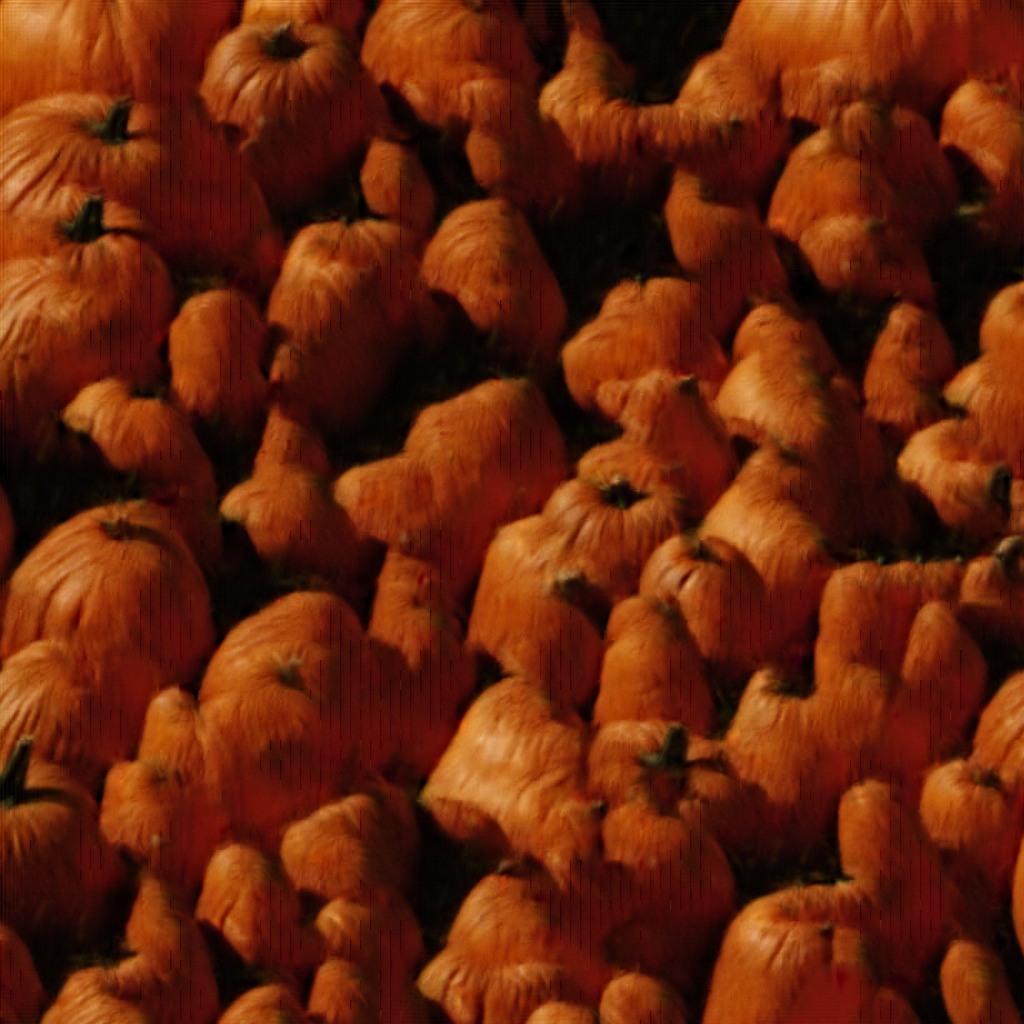}};

    \node [anchor=east] at (im0.west) {MSLG \cite{raad2016}};
    \node [anchor=east] at (im1.west) {CNNMRF \cite{li2016combining}};
    \node [anchor=east] at (im2.west) {EF \cite{EfrosFreeman}};
    \node [anchor=east] at (im3.west) {EL \cite{EfrosLeung}};
    \node [anchor=east] at (im4.west) {input};

    \end{tikzpicture}
    \caption{\emph{Comparison of texture synthesis methods.} From top to bottom: input sample Efros and Leung (EL)~\cite{EfrosLeung}, Efros and Freeman (EF)~\cite{EfrosFreeman}, CNNMRF~\cite{li2016combining} and MSLG~\cite{raad2016}.}
    \label{fig:algComp2-2}
\end{figure}

The second to sixth rows correspond to statistics-based methods described in Section~\ref{sec:parametric}, namely Random Phase Noise, Heeger-Bergen, Portilla-Simoncellli, Gatys and SGAN. Early statistics-based methods: Heeger and Bergen (1995), Portilla and Simoncelli (2000) and Random Phase Noise (1991) yield good results for microtextures, i.e. textures with no conspicuous structures, as can be seen in the first {texture example} of Figure~\ref{fig:algComp1-1} and to a lesser extent {for the second and third example.} {The Heeger and Bergen's method is inspired on a model of the early visual cortex; it provides satisfying results in some cases but there is no theoretical proof of the convergence of the method. On the other hand, the RPN method yields a simple and elegant theory with no convergence issue. The visual results yield by both methods are in general satisfying for microtextures.} However, for textures with local structures, the results are blurry and unsatisfying. Among these three methods, the results obtained by Portilla and Simoncelli are by far the most remarkable. These results contain recognizable configurations from the sample. This can be observed for the last texture example in Figure~\ref{fig:algComp1-1} and the first two examples in Figure~\ref{fig:algComp2-1}. Notice that the Heeger and Bergen and RPN methods yield unsatisfying results for these three examples. Clearly the global statistics considered by these methods are not enough to characterize these highly structured textures.

The second and third rows of the Figures~\ref{fig:algComp1-2} and~\ref{fig:algComp2-2} correspond to the patch re-arrangement methods Efros-Leung (1999) and Efros-Freeman (2001) described in Sections~\ref{sec:non-parametric}.  The first three textures in Figure~\ref{fig:algComp1-2} have no conspicuous structures but are not stationary (for example, there are small changes of illumination). The Efros and Leung method, being too local, fails to recover the global characteristics of these textures. A similar and attenuated behavior is observed in Efros-Freeman's results. The methods are significantly better than their predecessors in the presence of local structure, but have their specific problems. Efros and Leung's results for the third and fourth texture in Figure~\ref{fig:algComp1-2} show two clear examples of garbage growing. The method has repeated a very small part of the input in an inconsistent way creating ``garbage''. In general this phenomenon is more evident in Efros and Leung's results, compared to those of Efros and Freeman. The results of Efros-Leung and Efros-Freeman for the first texture in Figure~\ref{fig:algComp2-2} show that the global organization is sometimes missed, mostly due to the fact that these methods work at a single scale. The second texture example in Figure~\ref{fig:algComp2-2} yields impressive results in the case of Efros-Freeman's method. Nevertheless, looking carefully one can notice the verbatim copies of the piece of chalks in the input image. The hybrid method MSLG (2016) described in Section~\ref{sec:hybrid}, whose results are on the fifth row of Figures~\ref{fig:algComp1-2} and~\ref{fig:algComp2-2}, faces the same issues for the three first examples in Figure~\ref{fig:algComp1-2}.  This is less visible though, since the Gaussian models tend to smooth slightly the result. However, the original granularity of the input sample is lost in MSLG. As mentioned in Section~\ref{sec:non-parametric}, Efros and Leung's and Efros and Freeman's results depend on the patch size, while the multiscale approach (MSLG) is more robust to that parameter. When the former two methods fail to preserve global organization, MSLG, working at multiple scales, manages to preserve this organization. In the second texture example in Figure~\ref{fig:algComp2-2}, MSLG avoids the verbatim copy since the patches are being sampled from their Gaussian model and therefore are different from their original patches. Nevertheless, the Gaussian model strongly smooths the output. The synthesis of the flower texture (Figure~\ref{fig:algComp2-2} third row) is very satisfying for the three methods. Finally, the pumpkin texture shows a clear example of the verbatim copy effect in the Efros-Leung and Efros-Freeman methods.
{The fourth row of the Figures~\ref{fig:algComp1-2} and~\ref{fig:algComp2-2} corresponds to the CNN patch re-arrangement method CNNMRF (2016) described in Section~\ref{sec:non-parametric}. By bringing the patch re-arrangement to CNN features, the blending between the patches is improved. For example on the fourth texture of Figure~\ref{fig:algComp2-2}, the separation between the patches are visible on the results of Efros-Freeman, which is not the case for CNNMRF. However CNNMRF suffers particularly from verbatim copies and fails to recover the global statistics of the image. Efros-Leung and Efros-Freeman do suffer less from these problems because the patch selection step for these methods picks randomly among a selection of patches, whereas CNNMRF takes the most likely patch.
}

{The recent statistics-based CNN methods, discussed in Section~\ref{sec:cnn}, show significant improvement over their predecessors}. Gatys (2015) is the best statistics-based method at respecting the fine details for all the textures of Figures~\ref{fig:algComp1-1} and~\ref{fig:algComp2-1}, which can be well noticed with a zoom-in. However, some low frequencies or structure organizations are missed, as seen on the fourth texture of Figure~\ref{fig:algComp1-1}, and some contrast instabilities can be noticed, for example on the first texture of Figure~\ref{fig:algComp2-1}. As discussed in Section~\ref{sec:cnn}, some variants were proposed to fix these problems. SGAN (2016), on the other hand, better respects the low frequencies, and the results often look better than Gatys when zoomed-out. However on the fine scale, the results are incomplete and noisy, as seen on all the textures of the Figures~\ref{fig:algComp1-1} and~\ref{fig:algComp2-1}. SGAN fails to generate correctly the first texture of Figure~\ref{fig:algComp1-1}, possibly because this texture has no structure and is a microtexture. It is likely that better results can be obtained by tuning the parameters, but as said in Section~\ref{sec:cnn}, the default parameters were used.

Among all these methods, the CNN based methods are the most expensive in computational time. Pixel based methods, like Efros-Leung, are more expensive than patch based methods like Efros-Freeman or MSLG. The speed of statistics-based methods depends on how global the optimization is, and on the number of iterations needed. Portilla-Simoncelli's and Heeger-Bergen's speeds are comparable to patch based methods, while Random Phase Noise is the cheapest of the methods reviewed here.

These comparative evaluations show the strengths and weaknesses of the different original methods described in this survey. As said previously, some variants of these methods can get better results on some pictures. For example, a better result for the fourth texture of the Figure~\ref{fig:algComp1-1} can be seen on Figure~\ref{fig:mslg-gatys}. For this texture, first generating with MSLG, then refining with Gatys' texture generator, enables to combine the best  of both algorithms: The fabric elements are well aligned, and look good at a fine scale. Overall, over the last three decades, tremendous progress was made to generate convincing new texture samples from a small and stationary texture sample. However one could argue that the samples used in this comparison are toy examples. Indeed, except for the third texture of Figure~\ref{fig:algComp1-1}, and fourth texture of Figure~\ref{fig:algComp2-1}, the samples do not suffer much of illumination changes or perspective, and are essentially stationary. Nevertheless, most textures are not  stationary. Think for example of  a wood texture. This leads us to wonder whether the presented algorithms get acceptable results on these complex scenarios.

\subsection{Getting out of toy examples}

\begin{figure}[t]
  \centering
  \begin{tikzpicture}[scale=3.83]
    \node[inner sep=0, anchor=south] (im1) at (0,0) {\includegraphics[width=.4\textwidth]{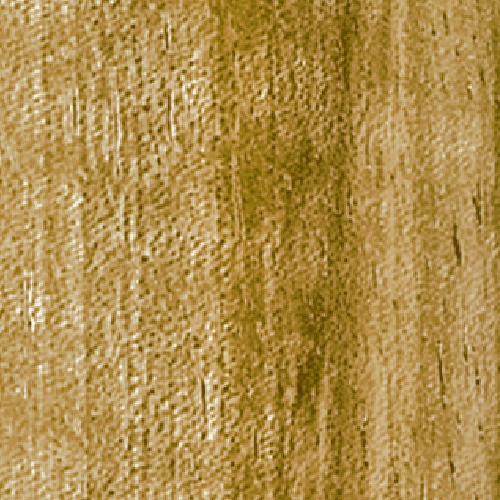}};
    \node[inner sep=0, anchor=south] (im2) at (1.5,0) {\includegraphics[width=.4\textwidth]{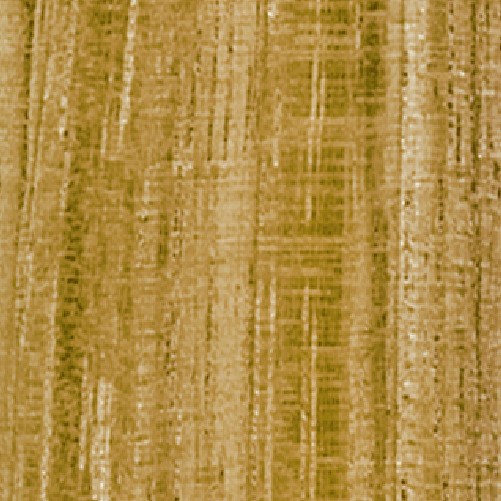}};
    \node[anchor=south east, fill=white, opacity=.4, text opacity=1] at (im1.south east) {\small{crop 1}};
    \node[anchor=south east, fill=white, opacity=.4, text opacity=1] at (im2.south east) {\small{crop 2}};
  \end{tikzpicture}
  \caption{Two crops of different parts of a larger wood texture. The cropped images are of size $500\times 500$ pixels. Each one represents a different texture belonging to a single ``big texture''.}
  \label{fig:wood-crops}
\end{figure}

\begin{figure}[t]
  \centering
  \begin{tikzpicture}[scale=3.83]
    \node[inner sep=0, anchor=south] (im1) at (0,0) {\includegraphics[width=.4\textwidth]{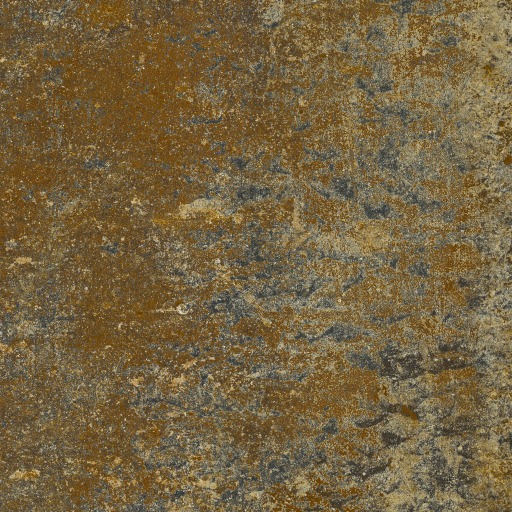}};
    \node[inner sep=0, anchor=south] (im2) at (1.5,0) {\includegraphics[width=.4\textwidth]{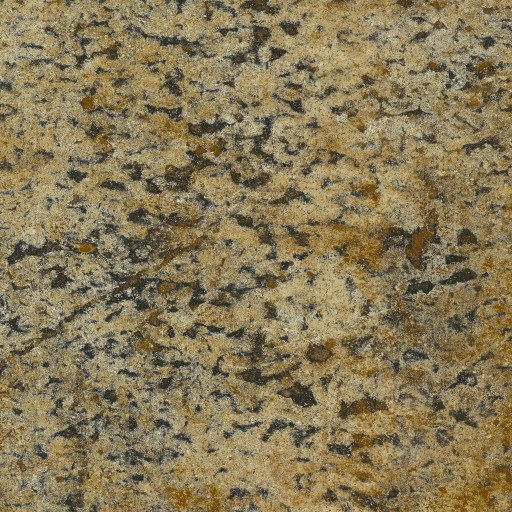}};
    \node[anchor=south east, fill=white, opacity=.4, text opacity=1] at (im1.south east) {\small{crop 1}};
    \node[anchor=south east, fill=white, opacity=.4, text opacity=1] at (im2.south east) {\small{crop 2}};
  \end{tikzpicture}
  \caption{Two crops of different parts of a larger stone texture. The cropped images are of size $512\times 512$ pixels. Each one represents a different texture belonging to a single ``big texture''.}
  \label{fig:stone-crops}
\end{figure}

The previous examples present some quite impressive texture synthesis results by several algorithms. The texture synthesis problem seems to be almost solved for ``academic'' textures. Still, those results were obtained for pictures of relatively small size and taken in almost ideal conditions, in order to get almost stationary textures. In this section, we discuss the situation for more complex textures: When the same methods are applied to sample images of real and non-stationary textures, where long-range structure is present as well as varying detail at every scale. Figures \ref{fig:wood-crops} and \ref{fig:stone-crops} show some realistic examples of real world images that nobody would hesitate calling textures. Nevertheless on second thoughts they do have  a complex, non-stationary structures, because every large enough image has it. But these are precisely the examples that need being emulated!  In this endeavor, we can relax the requirement that the synthesis must make  a larger image. Let's just ask if a method is able to reproduce a perceptually similar texture  at the very same size.

Each of textures in Figures \ref{fig:wood-crops} and \ref{fig:stone-crops} show different salient sub-textures within the same image. Since the methods in Section~\ref{sec:comparative-evaluation} usually assume that the texture is stationary,  it is not completely fair to use these methods on these samples. Several works have investigated ways to handle these complex cases \cite{rosenberger2009layered,atto2014non,kaspar2015self,lockerman2016multi}. In this section we show the results of the state of the art algorithms presented in this paper, and will show that they are still far from emulating  to real world textures, even without the requirement of building a larger texture patch from the sample.

\begin{figure}[p]
  \centering
  \begin{tikzpicture}[scale=.9]

    \node[inner sep=0, anchor=south] (im8) at (0,16.8) {\includegraphics[width=2\widthfigureresultsTwo]{figures/Stone_2_z3_c1_750_750_256}};
    \node[inner sep=0, anchor=south]  at (3,16.8) {\includegraphics[width=2\widthfigureresultsTwo]{figures/Stone_2_z3_c7_2000_300_256}};
    \node[inner sep=0, anchor=south]  at (6,16.8) {\includegraphics[width=2\widthfigureresultsTwo]{figures/wood_h_1_4}};
    \node[inner sep=0, anchor=south]  at (9,16.8) {\includegraphics[width=2\widthfigureresultsTwo]{figures/wood_h_2_1}};

    \node[inner sep=0, anchor=south] (im7) at (0,14) {\includegraphics[width=2\widthfigureresultsTwo]{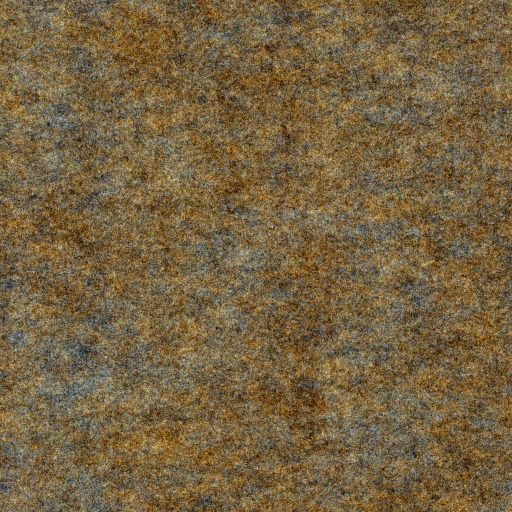}};
    \node[inner sep=0, anchor=south]  at (3,14) {\includegraphics[width=2\widthfigureresultsTwo]{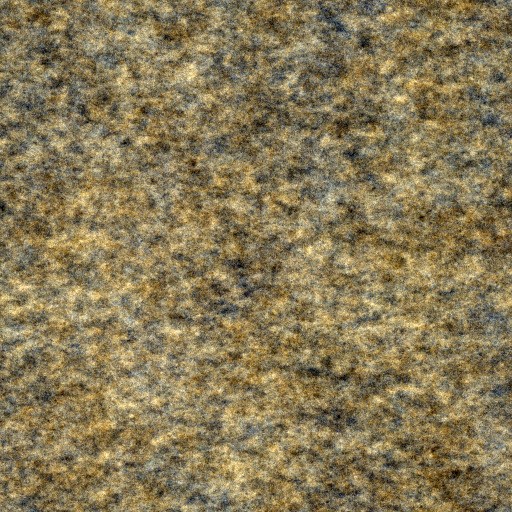}};
    \node[inner sep=0, anchor=south]  at (6,14) {\includegraphics[width=2\widthfigureresultsTwo]{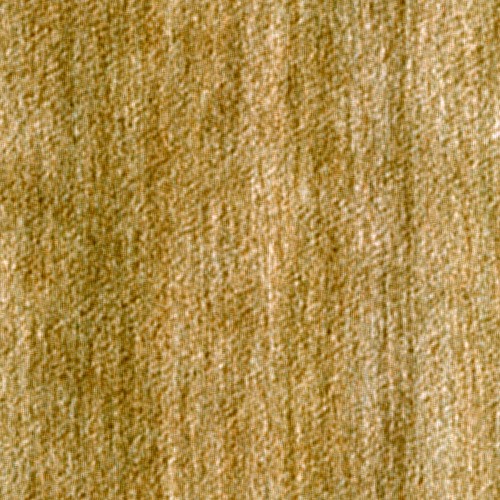}};
    \node[inner sep=0, anchor=south]  at (9,14) {\includegraphics[width=2\widthfigureresultsTwo]{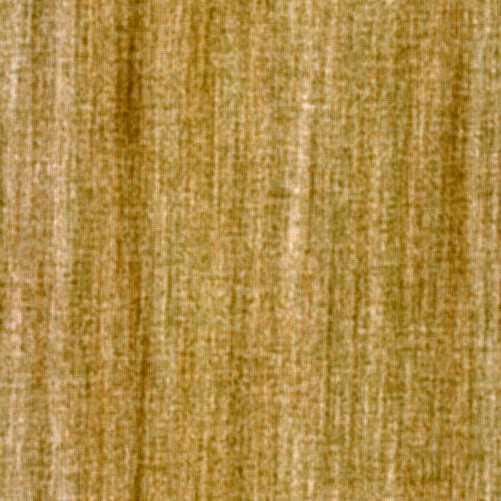}};

    \node[inner sep=0, anchor=south] (im6) at (0,11.2) {\includegraphics[width=2\widthfigureresultsTwo]{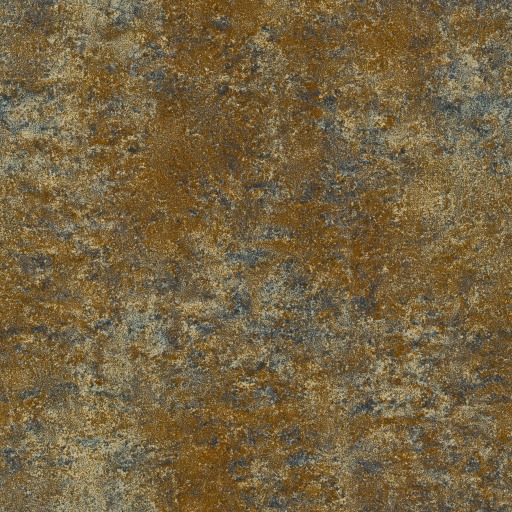}};
    \node[inner sep=0, anchor=south]  at (3,11.2) {\includegraphics[width=2\widthfigureresultsTwo]{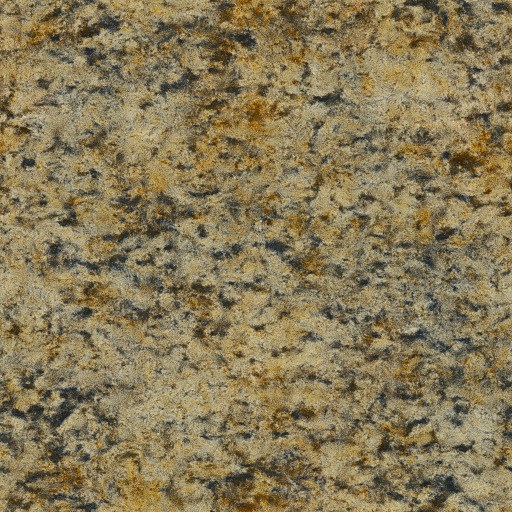}};
    \node[inner sep=0, anchor=south]  at (6,11.2) {\includegraphics[width=2\widthfigureresultsTwo]{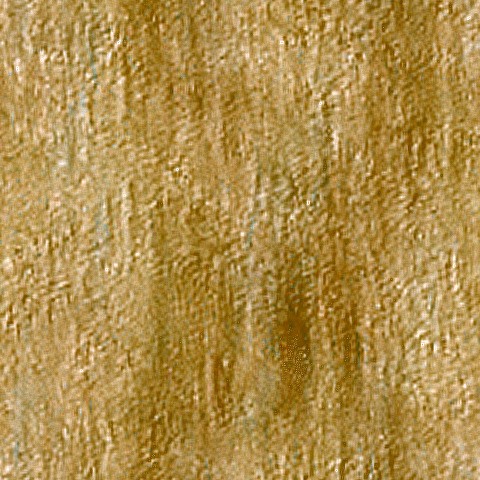}};
    \node[inner sep=0, anchor=south]  at (9,11.2) {\includegraphics[width=2\widthfigureresultsTwo]{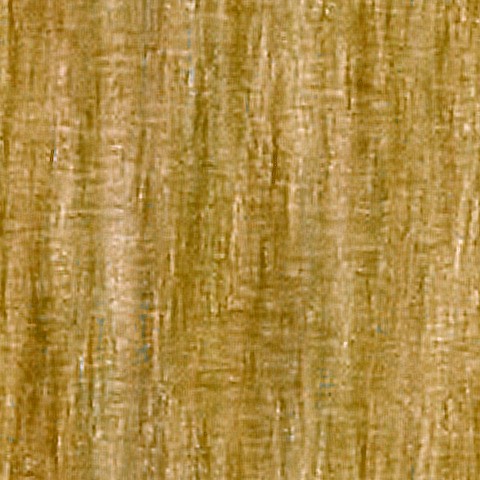}};

    \node[inner sep=0, anchor=south] (im5) at (0,8.4) {\includegraphics[width=2\widthfigureresultsTwo]{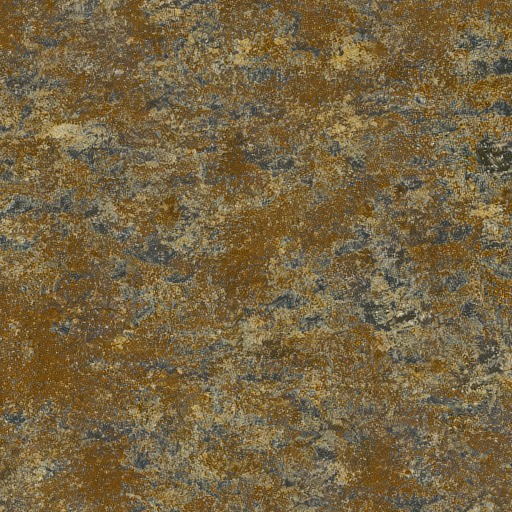}};
    \node[inner sep=0, anchor=south]  at (3,8.4) {\includegraphics[width=2\widthfigureresultsTwo]{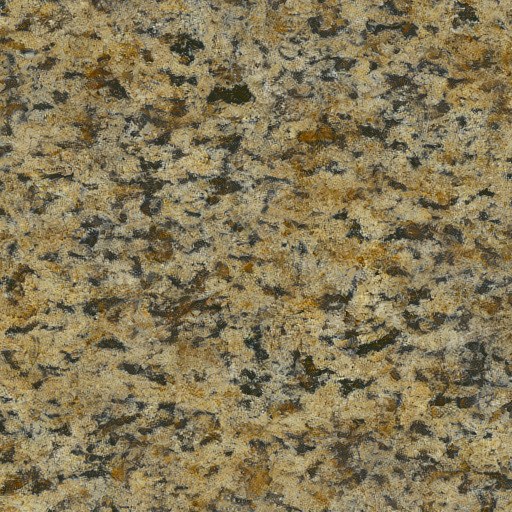}};
    \node[inner sep=0, anchor=south]  at (6,8.4) {\includegraphics[width=2\widthfigureresultsTwo]{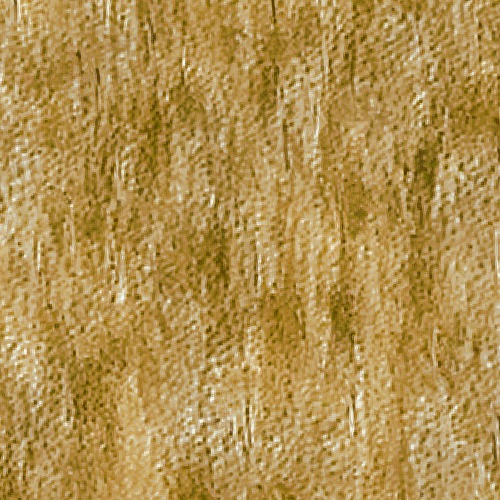}};
    \node[inner sep=0, anchor=south]  at (9,8.4) {\includegraphics[width=2\widthfigureresultsTwo]{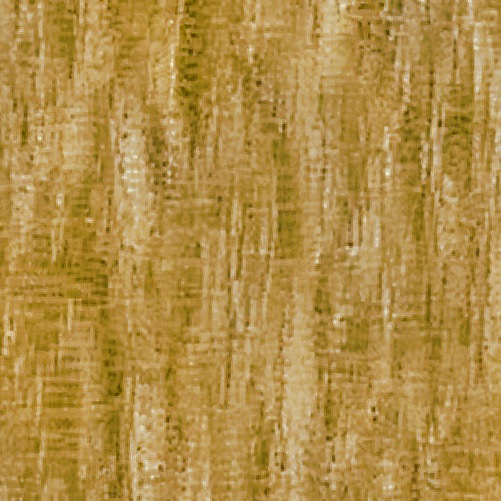}};

    \node[inner sep=0, anchor=south] (im4) at (0,5.6) {\includegraphics[width=2\widthfigureresultsTwo]{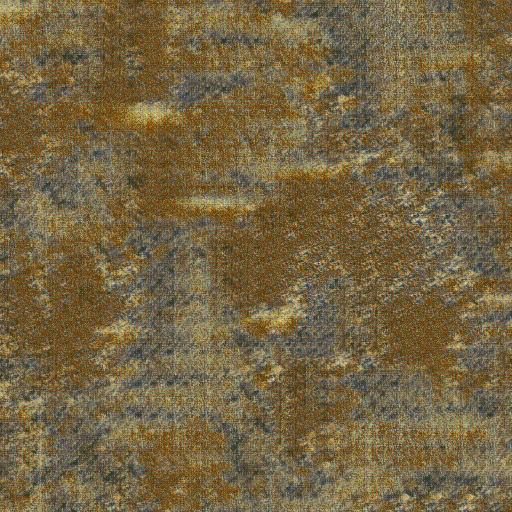}};
    \node[inner sep=0, anchor=south]  at (3,5.6) {\includegraphics[width=2\widthfigureresultsTwo]{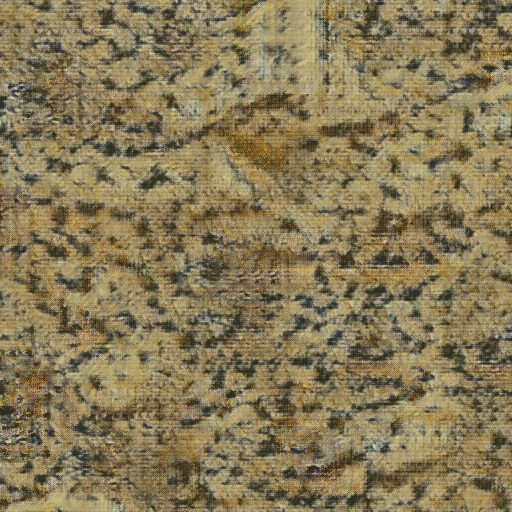}};
    \node[inner sep=0, anchor=south]  at (6,5.6) {\includegraphics[width=2\widthfigureresultsTwo]{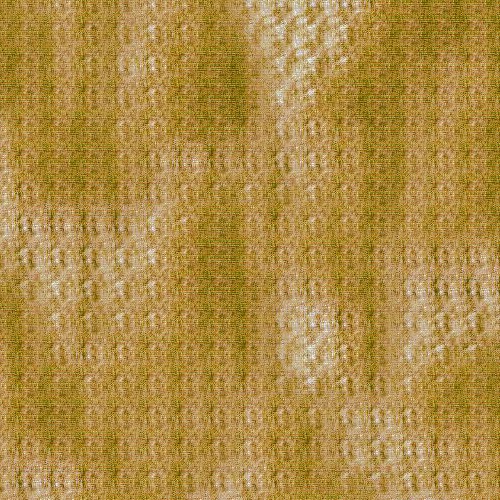}};
    \node[inner sep=0, anchor=south]  at (9,5.6) {\includegraphics[width=2\widthfigureresultsTwo]{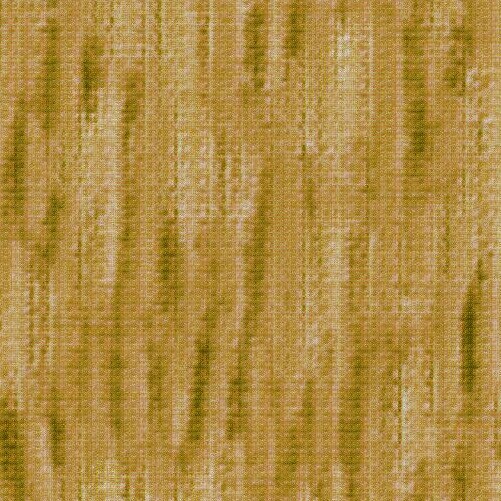}};

    \node[inner sep=0, anchor=south] (im3) at (0,2.8) {\includegraphics[width=2\widthfigureresultsTwo]{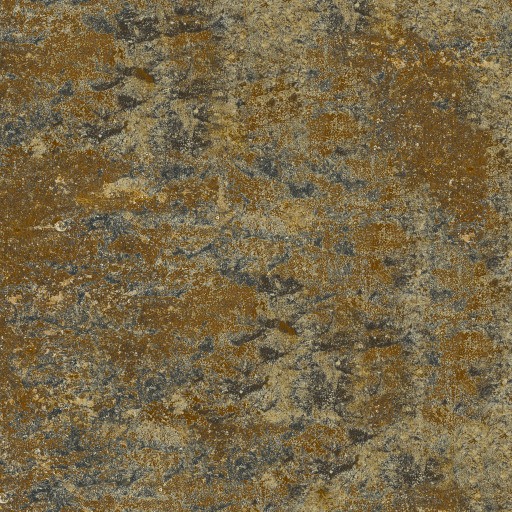}};
    \node[inner sep=0, anchor=south]  at (3,2.8) {\includegraphics[width=2\widthfigureresultsTwo]{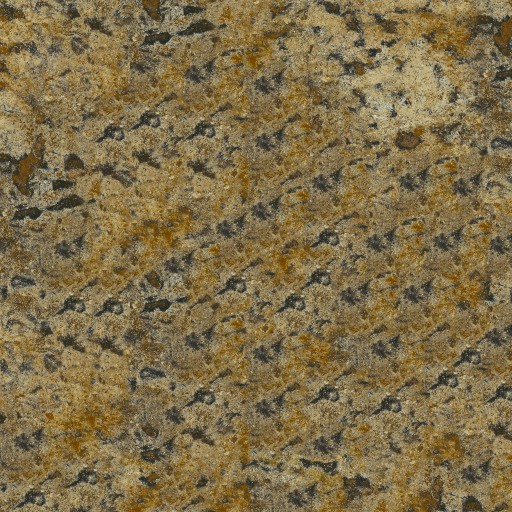}};
    \node[inner sep=0, anchor=south]  at (6,2.8) {\includegraphics[width=2\widthfigureresultsTwo]{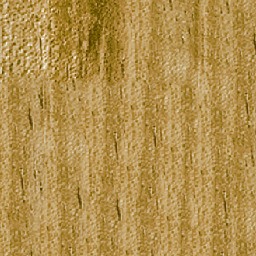}};
    \node[inner sep=0, anchor=south]  at (9,2.8) {\includegraphics[width=2\widthfigureresultsTwo]{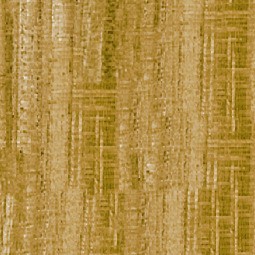}};

    \node[inner sep=0, anchor=south] (im2) at (0,0) {\includegraphics[width=2\widthfigureresultsTwo]{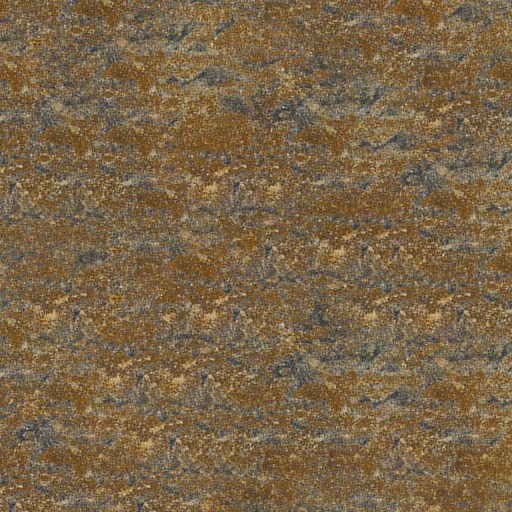}};
    \node[inner sep=0, anchor=south]  at (3,0) {\includegraphics[width=2\widthfigureresultsTwo]{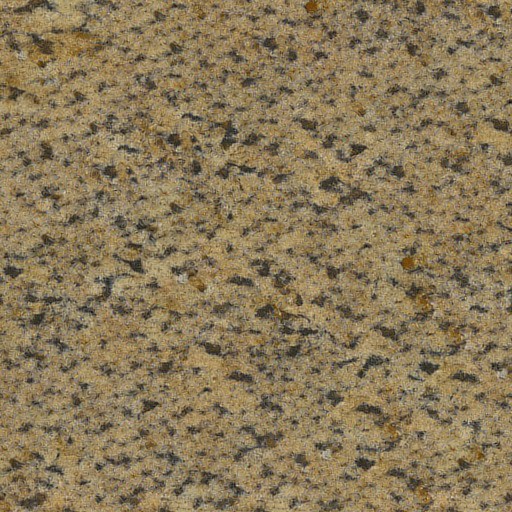}};
    \node[inner sep=0, anchor=south]  at (6,0) {\includegraphics[width=2\widthfigureresultsTwo]{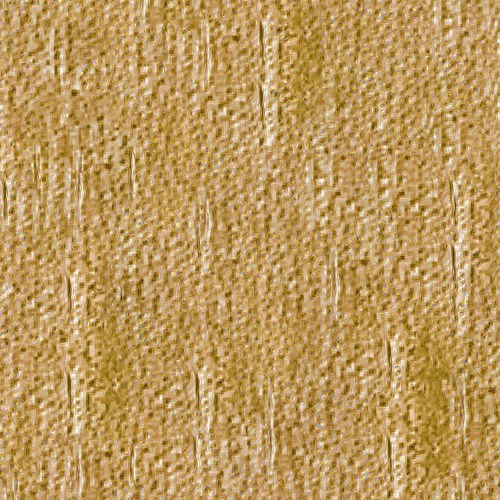}};
    \node[inner sep=0, anchor=south]  at (9,0) {\includegraphics[width=2\widthfigureresultsTwo]{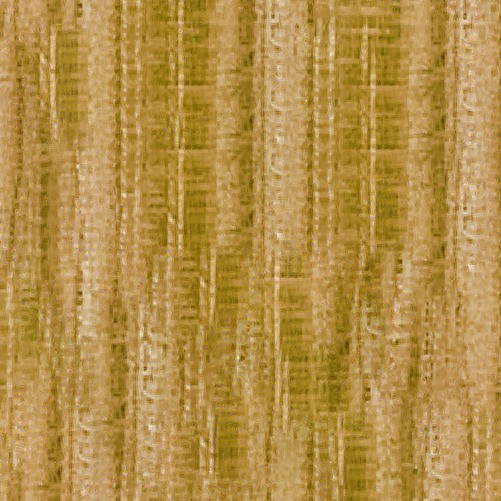}};

    \node [anchor=east] at (im2.west) {CNNMRF \cite{li2016combining}};
    \node [anchor=east] at (im3.west) {EF \cite{EfrosFreeman}};
    \node [anchor=east] at (im4.west) {SGAN \cite{jetchev2016texture}};
    \node [anchor=east] at (im5.west) {Gatys \cite{gatys}};
    \node [anchor=east] at (im6.west) {PS \cite{PS}};
    \node [anchor=east] at (im7.west) {RPN \cite{GalerneRPN}};
    \node [anchor=east] at (im8.west) {input};

  \end{tikzpicture}

  \caption{Synthesis results for statistical based and patch re-arrangement methods on complex texture. They show the current limitations of all best methods.  RPN scrambles the textures. PS loses long range coherence of the wood veins. EF and CNNMRF's copy paste is quite visible for all textures and incurs in garbage growing.   PS and Gatys have satisfying results on the left hand two textures,  but miss to emulate long range  interactions on the wood textures. SGAN  grows periodic noise patterns.}
  \label{fig:wood-crops-synthesis}
\end{figure}

\begin{figure}[p]
  \centering
  \begin{tikzpicture}[scale=0.9]

    \node[inner sep=0, anchor=south] (im8) at (0,16.8) {\includegraphics[width=2\widthfigureresultsTwo]{figures/Stone_2_z3_c1_750_750_256}};
    \node[inner sep=0, anchor=south]  at (2.9,16.8) {\includegraphics[width=2\widthfigureresultsTwo]{figures/Stone_2_z3_c7_2000_300_256}};
    \node[inner sep=0, anchor=south]  at (5.8,16.8) {\includegraphics[width=2\widthfigureresultsTwo]{figures/wood_h_1_4}};
    \node[inner sep=0, anchor=south]  at (8.7,16.8) {\includegraphics[width=2\widthfigureresultsTwo]{figures/wood_h_2_1}};

    \node[inner sep=0, anchor=south] (im7) at (0,14) {\includegraphics[width=2\widthfigureresultsTwo]{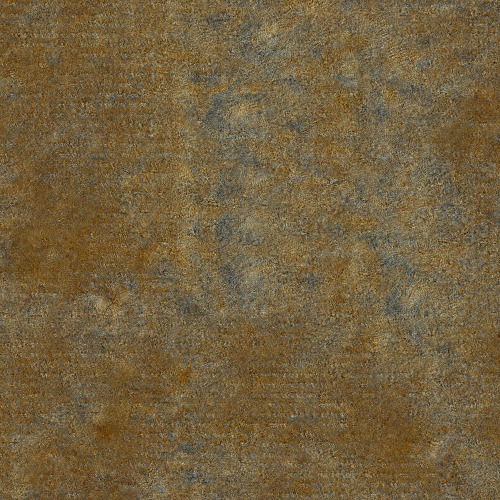}};
    \node[inner sep=0, anchor=south]  at (2.9,14) {\includegraphics[width=2\widthfigureresultsTwo]{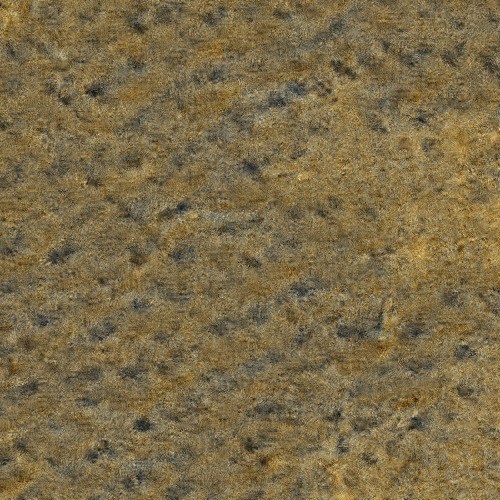}};
    \node[inner sep=0, anchor=south]  at (5.8,14) {\includegraphics[width=2\widthfigureresultsTwo]{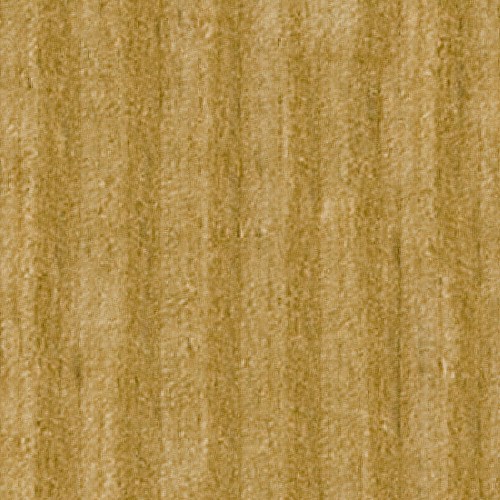}};
    \node[inner sep=0, anchor=south]  at (8.7,14) {\includegraphics[width=2\widthfigureresultsTwo]{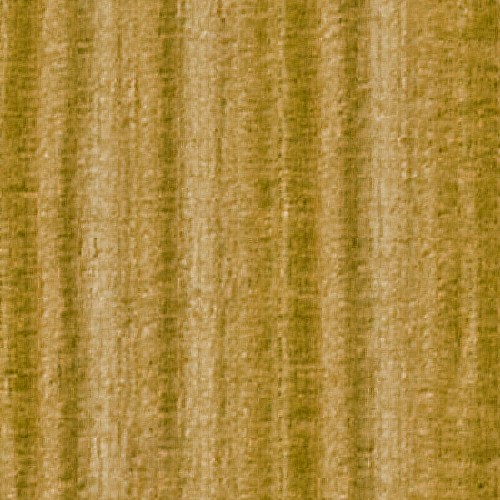}};

    \node[inner sep=0, anchor=south] (im6) at (0,11.2) {\includegraphics[width=2\widthfigureresultsTwo]{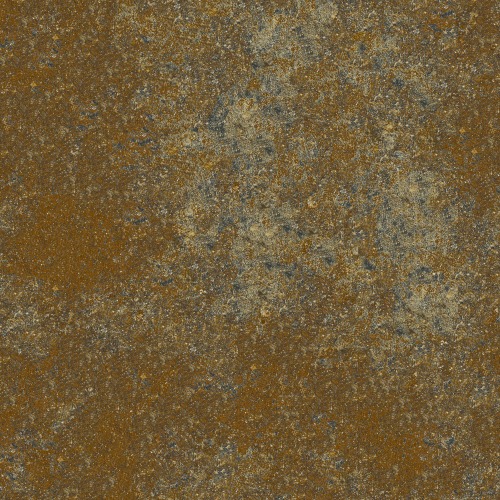}};
    \node[inner sep=0, anchor=south]  at (2.9,11.2) {\includegraphics[width=2\widthfigureresultsTwo]{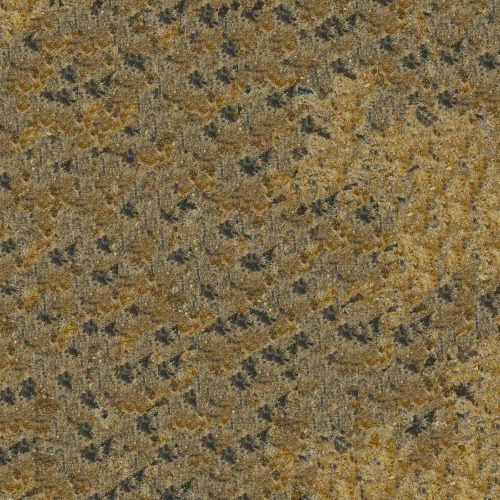}};
    \node[inner sep=0, anchor=south]  at (5.8,11.2) {\includegraphics[width=2\widthfigureresultsTwo]{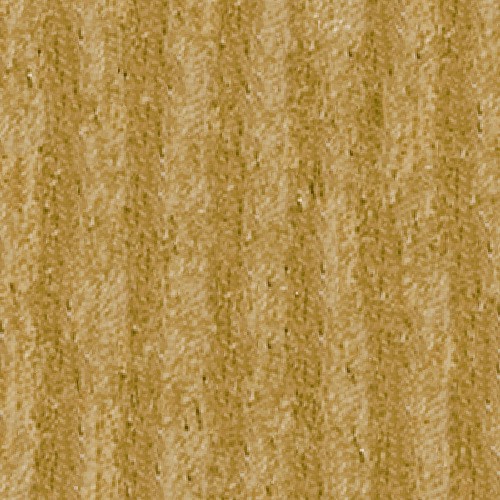}};
    \node[inner sep=0, anchor=south]  at (8.7,11.2) {\includegraphics[width=2\widthfigureresultsTwo]{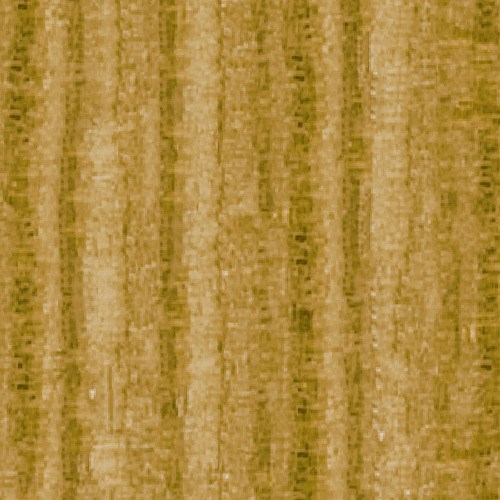}};

    \node[inner sep=0, anchor=south] (im5) at (0,8.4) {\includegraphics[width=2\widthfigureresultsTwo]{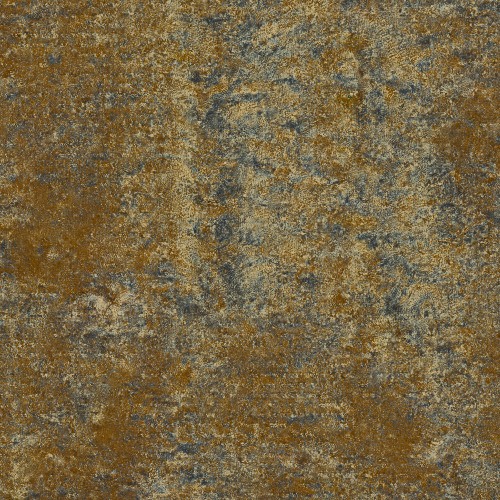}};
    \node[inner sep=0, anchor=south]  at (2.9,8.4) {\includegraphics[width=2\widthfigureresultsTwo]{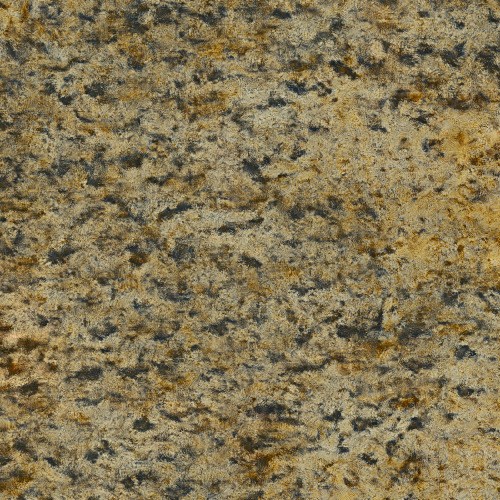}};
    \node[inner sep=0, anchor=south]  at (5.8,8.4) {\includegraphics[width=2\widthfigureresultsTwo]{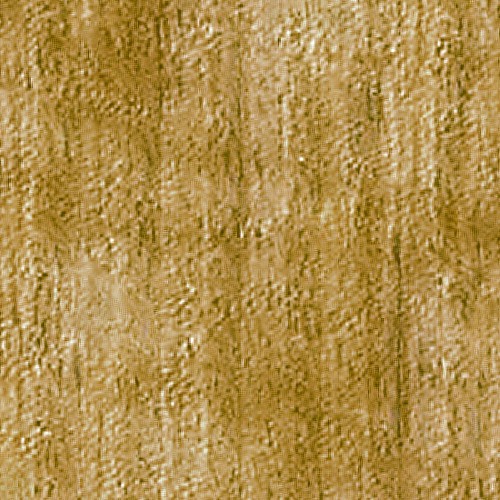}};
    \node[inner sep=0, anchor=south]  at (8.7,8.4) {\includegraphics[width=2\widthfigureresultsTwo]{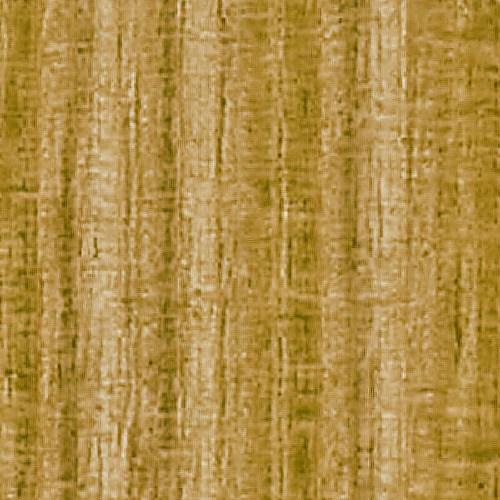}};

    \node[inner sep=0, anchor=south] (im4) at (0,5.6) {\includegraphics[width=2\widthfigureresultsTwo]{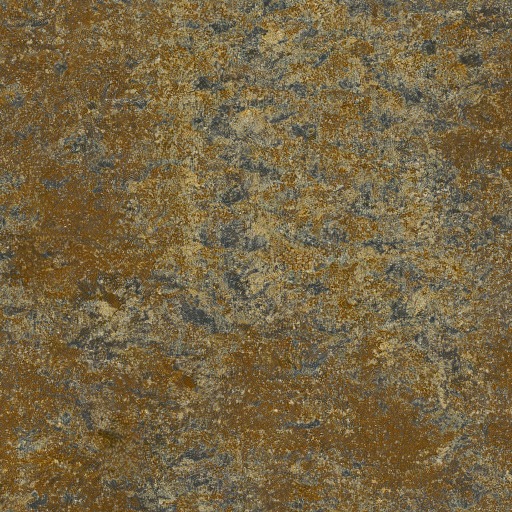}};
    \node[inner sep=0, anchor=south]  at (2.9,5.6) {\includegraphics[width=2\widthfigureresultsTwo]{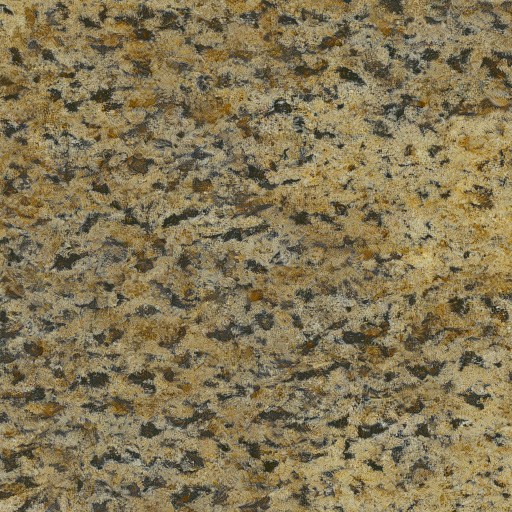}};
    \node[inner sep=0, anchor=south]  at (5.8,5.6) {\includegraphics[width=2\widthfigureresultsTwo]{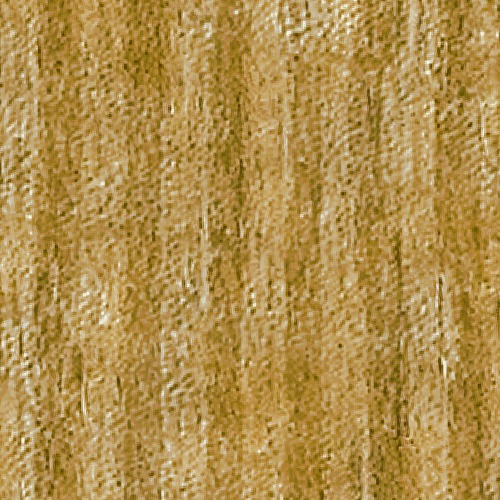}};
    \node[inner sep=0, anchor=south]  at (8.7,5.6) {\includegraphics[width=2\widthfigureresultsTwo]{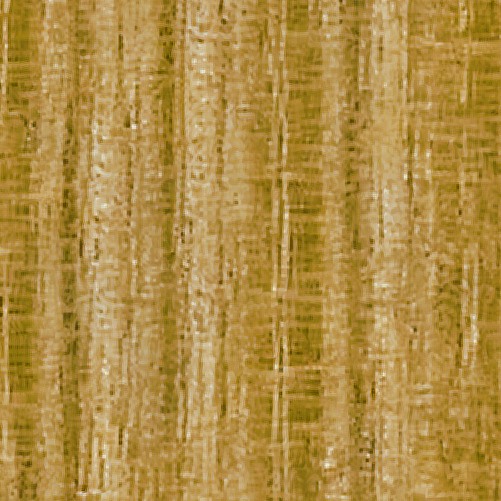}};

    \node [anchor=east] at (im4.west) {MSLG+Gatys};
    \node [anchor=east] at (im5.west) {MSLG+PS};
    \node [anchor=east] at (im6.west) {MSLG+EF};
    \node [anchor=east] at (im7.west) {MSLG};
    \node [anchor=east] at (im8.west) {input};

  \end{tikzpicture}

  \caption{Synthesis results for the hybrid methods. In columns 2) and 3), MSLG has repetitions and garbage growing; thus all the generated results based on the MSLG outputs keep this defect. In columns 1) and 4), MSLG respects well the global statistics of the textures, and the combination with other methods indeed improves the result. MSLG+PS and MSLG+Gatys perform better  on these examples than MSLG+EF.}
\label{fig:wood-crops-synthesis-combined-methods}
\end{figure}

Figures~\ref{fig:wood-crops-synthesis} and~\ref{fig:wood-crops-synthesis-combined-methods} show the results of the presented statistics-based, patch re-arrangement and hybrid methods on some of these more complex examples. The best results are Gatys' texture generator on the second texture and MSLG+PS and MSLG+Gatys on the first texture and fourth texture. When applying RPN or PS to them the results obtained are often too blurry. Gatys' texture generator fails to catch the low frequency structures for the last two textures. EF, {CNNMRF} and MSLG suffer from garbage growing and verbatim copies on the first three textures. This is true especially when the input is {not stationary.} SGAN fails to generate properly on the first two textures, and while the global organization of the third and fourth pictures is good, it suffers from the noise at small scale mentioned previously. As noticed in the previous section, the MSLG results are slightly blurry.

These results show that while some methods can get good results on some of these challenging texture samples, no method manages to get satisfying results for all four textures.

\section{Conclusion}\label{sec:conclusion}

With the multiplication of applications in computer graphics to the entertainment industry, the interest in the generation of synthetic objects with realistic texture has grown rapidly. High budget film sets, computer games, and in some cases digital art, spend intensive human efforts to imitate the appearance and feel of real world items. For this reason, exemplar-based texture synthesis has been the focus of intensive work for three decades. And as the available computational power increased, so has the sophistication of these methods.

In the end of the last millennium, statistics-based methods, such as RPN, Heeger-Bergen and Portilla-Simoncelli focused on a reduced set of statistics. The results were quite satisfactory on micro-textures, but could be blurry and far from the originals for more complex structures. Patch re-arrangement methods, such as Efros-Leung and Efros-Freeman, managed to respect significantly better the feel and the low level structures of these textures, but could have issues, such as discontinuities, verbatim copy, garbage growing or simply not respecting some essential statistics of the textures, such as the average intensity. Hybrid methods, such as MSLG, fix some of the issues of patch re-arrangement methods, but still share some of their issues. Very recently, statistics-based methods have been revisited with Convolutional Neural Networks (CNNs). {CNN based methods significantly increase the number of texture statistics involved in their model, for example by a factor of 25 approximately in the case of Gatys' texture generator compared to Portilla-Simoncelli. The results show a spectacular progress over their predecessors, but no method is perfect yet. Patch re-arrangement methods were revisited as well by CNNs. CNNMRF improves the blending between the image patches, but the results still suffer from the problems mentioned above for patch re-arrangement methods.} In this review, we presented three statistics-based neural methods with different models: Gatys' texture generator, DeepFrame and SGAN. When zoomed-in, the outputs of Gatys' texture generator are the best among the statistics-based methods, but miss some important low frequency constraints of the texture when zoomed-out. Some variants aim at fixing this shortcoming. SGAN succeeds better on several examples to respect the global structure of the texture, but the details of the texture are poor. While all the other statistics-based methods have an explicit texture model, the SGAN model is more implicit.

Our experimental results look no doubt sometimes worse than in the original papers, but  precisely we did not select the best examples. Our examination of the history of the method leads to the following conclusions.
\newline\noindent - The \textit{exemplar-based texture synthesis} problem is implicitly ill-posed, as it requires to extrapolate a Fourier spectrum by enlarging the image given a very small  sample of it.  Having very small samples may have been historically interesting  in computer  graphics, but is no longer a technical issue, given the available memories and computational power in all computers.
\newline\noindent - By working on small  texture examples the literature has somehow unrealistically restricted the problem. Indeed it is simply not true that textures are as stationary as those examples suggest.
\newline\noindent - When trying to work on larger examples, we have seen that no texture sample is really stationary.  A realistically large texture sample in fact contains smaller patches of very different textures.
\newline\noindent - This explains first why patch based copy-paste methods are doomed in spite of some apparent success in some quasi-periodic texture with no conspicuous detail. On more involved samples, they cannot but reproduce recognizable details.
\newline\noindent - This also explains why progress in this topic is linked to the design of methods enforcing more and more statistical parameters. The number of statistics enforced by statistical models is growing fast: 710 for Portilla-Simoncelli, 176640 for the default model in Gatys' texture generator. With some results showing that the filters can be chosen with random weights \cite{gatysrecent}, one can wonder if the solution is not to just use the highest number of statistics possible to emulate a texture.  One may also wonder where to draw a reasonable limit between synthesizing complex textures and rendering scenes containing textured objects.
\newline\noindent - Thus, the Portilla-Simoncelli method, of enforcing a high number of statistics, wins, but it is somewhat a Pyrrhic victory. Indeed, the more random statistics we pile up, the better the exemplar-based results. But it remains to find numerical  tools applying automatically an Occam's razor as  Portilla and Simoncelli did manually. This is still needed to realize the goal of Julesz' program, which was  to find the minimal  sufficient set of statistics rendering two textures indistinguishable.

\section*{Acknowledgements}
We thank  Rafael Grompone von Gioi for valuable corrections and suggestions,  Arthur Leclaire and Yang Lu for their feedback and for helping produce some of the experiments, and the anonymous referees for valuable advice.
\bibliographystyle{plain}
\bibliography{all}

\end{document}